\documentclass{article} 
\usepackage{authblk}
\usepackage[table]{xcolor}
\usepackage{times}
\usepackage{eso-pic} 
\usepackage[margin=1.1in]{geometry}
\usepackage{authblk}

\usepackage[square,numbers]{natbib}
\usepackage[utf8]{inputenc} 
\usepackage[T1]{fontenc}    
\usepackage{hyperref}       
\usepackage{url}            
\usepackage{booktabs}       
\usepackage{amsfonts}       
\usepackage{nicefrac}       
\usepackage{microtype}      
\usepackage{xcolor}         
\usepackage{url}            
\usepackage{booktabs}       
\usepackage{amsfonts}       
\usepackage{nicefrac}       
\usepackage{microtype}      
\usepackage{xcolor}         
\usepackage{placeins}       
\usepackage[normalem]{ulem}

\usepackage{amsmath,amssymb,amsthm,bbm,color,nccmath}
\usepackage{graphicx}
\usepackage[capitalize,noabbrev]{cleveref}
\usepackage{algorithm}
\usepackage{algorithmic}
\usepackage{caption}
\usepackage{subcaption}
\usepackage[export]{adjustbox}
\usepackage{afterpage}
\usepackage[hang,flushmargin]{footmisc}
\usepackage{float}
\usepackage{enumitem,kantlipsum}
\usepackage{multirow}

\usepackage{array}
\theoremstyle{plain}
\newtheorem{theorem}{Theorem}[section]
\newtheorem{proposition}[theorem]{Proposition}
\newtheorem{lemma}[theorem]{Lemma}
\newtheorem{corollary}[theorem]{Corollary}
\theoremstyle{definition}

\theoremstyle{remark}
\newtheorem{remark}[theorem]{Remark}
\newtheorem{example}{Example}
\usepackage[export]{adjustbox}
\usepackage{amsmath, amssymb} 

\newcommand{\ep}{\varepsilon}
\newcommand{\EE}[1]{\mathbb{E}\left[{#1}\right]}

\newcommand{\PP}[1]{\mathbb{P}\left\{{#1}\right\}}
\newcommand{\Pp}[2]{\mathbb{P}_{{#1}}\left\{{#2}\right\}}
\newcommand{\PPst}[2]{\mathbb{P}\left\{{#1}\  \middle| \ {#2}\right\}}

\newcommand{\X}{\mathcal{X}}
\newcommand{\Y}{\mathcal{Y}}
\newcommand{\R}{\mathbb{R}}
\newcommand{\ch}{\widehat{C}}
\newcommand{\iidsim}{\stackrel{\textnormal{iid}}{\sim}}

\newcommand{\dtv}{\textnormal{d}_{\textnormal{TV}}}
\DeclareMathOperator*{\argmin}{argmin}
\newcommand{\scp}{\texttt{SPI}}
\newcommand{\scph}{\texttt{SPI-Whole}}
\newcommand{\scpc}{\texttt{SPI-Subset}}
\newcommand{\cp}{\texttt{OnlyReal}}
\newcommand{\cpsynt}{\texttt{OnlySynth}}
\newcommand{\One}[1]{{\mathbbm{1}}\left\{{#1}\right\}}

\definecolor{Green}{RGB}{119,221,119}
\usepackage{pdflscape}
\usepackage{placeins}

\usepackage{enumitem}
\allowdisplaybreaks

\newcommand{\printfnsymbol}[1]{%
  \textsuperscript{\@fnsymbol{#1}}%
}
\title{Synthetic-Powered Predictive Inference}

\author[1]{Meshi Bashari$^*$}
\author[1]{Roy Maor Lotan$^*$}
\author[2]{Yonghoon Lee$^*$}
\author[2]{Edgar Dobriban}
\author[1,3]{Yaniv Romano}
\affil[1]{Department of Electrical and Computer Engineering, Technion IIT, Israel}
\affil[2]{Department of Statistics and Data Science, The Wharton School, University of Pennsylvania, USA}
\affil[3]{Department of Computer Science, Technion IIT, Israel}
\date{}

\begin{document}
\renewcommand{\thefootnote}{\fnsymbol{footnote}}
\footnotetext[1]{Equal contribution.}
\renewcommand{\thefootnote}{\arabic{footnote}}
\maketitle
\begin{abstract}
Conformal prediction is a framework for predictive inference with a distribution-free, finite-sample guarantee. However, it tends to provide uninformative prediction sets when calibration data are scarce. This paper introduces Synthetic-powered predictive inference ($\scp$), a novel framework that incorporates synthetic data---e.g., from a generative model---to improve sample efficiency. At the core of our method is a score transporter: an empirical quantile mapping that aligns nonconformity scores from trusted, real data with those from synthetic data. By carefully integrating the score transporter into the calibration process, $\scp$ provably achieves finite-sample coverage guarantees without making any assumptions about the real and synthetic data distributions. When the score distributions are well aligned, $\scp$ yields substantially tighter and more informative prediction sets than standard conformal prediction. Experiments on image classification---augmenting data with synthetic diffusion-model generated images---and on tabular regression demonstrate notable improvements in predictive efficiency in data-scarce settings.
\end{abstract}

\section{Introduction}

\subsection{Background and motivation}

Conformal prediction~\citep{vovk1999machine,saunders1999transduction,vovk2005algorithmic} is a general framework for quantifying predictive uncertainty, providing finite-sample statistical guarantees for any machine learning model. Given a test instance with an unknown label (e.g., an image), conformal prediction constructs a prediction set---a collection of plausible labels guaranteed to include the true label with a user-specified coverage probability (e.g., 95\%). To do so, it relies on a labeled holdout calibration set to compute nonconformity scores, which measure how well a model's prediction aligns with the true labeled outcome. These scores are then used to assess uncertainty in future predictions. Crucially, the coverage guarantee holds whenever the calibration and test data are exchangeable (e.g., i.i.d.), without any assumption on the sampling distribution.

While conformal prediction offers a powerful coverage guarantee, its reliance on a holdout set limits its effectiveness when labeled data is scarce---becoming
unstable and highly variable in coverage, 
or overly conservative and uninformative. 
As a result, it offers limited value in applications where labeled data is inherently limited,
such as those requiring personalization or subgroup-specific guarantees.
Importantly,  this is not merely an abstract concern 
\citep{banerji2023clinical}---for example, in medical settings, it is natural to seek valid inference tailored to specific patient characteristics such as age, health condition, and/or other group identifiers of interest, see e.g., \citep{liu2016there,chernozhukov2023toward}. Similarly, in image classification tasks, one may wish to ensure that coverage holds for the true class label, see e.g.,  \citep{vovk2003mondrian}. 
In these cases and many others, we often have only a few representative holdout examples for each group or class, which severely restricts the applicability of standard conformal prediction.

\newpage

Meanwhile, we are witnessing rapid progress in the ability to train accurate machine learning models even under data-scarce settings, 
driven by the rising quality of synthetic data produced by modern generative models
and by advances in domain adaptation, see e.g.,  \citep{bommasani2021opportunities}. 
These developments inspire the question we pursue in this work:
\emph{Can we rigorously enhance the sample efficiency of conformal prediction by leveraging a large pool of synthetic data---such as labeled datapoints
from related subpopulations, or even data sampled from generative models?}

At first glance, 
it may appear hard to 
use synthetic data to boost sample efficiency in a statistically valid way.
After all, the distribution of synthetic data can be 
completely different from that of the data of interest.\footnote{We refer to the limited dataset of interest as the \emph{real calibration set}, to distinguish it from the (potentially synthetic) calibration data.}
Overcoming this challenge, we propose a principled framework that unlocks conformal prediction with the ability to incorporate synthetic data while preserving rigorous, model-agnostic, non-asymptotic coverage guarantees. Crucially, our method---\scp---provides a coverage bound that requires no assumptions about the similarity between the real and synthetic data distributions. Still, when the distribution of synthetic and real scores is close, our approach yields a substantial boost in sample efficiency---resulting in more informative prediction sets than those produced by standard conformal prediction.
A discussion of related literature is deferred to~\Cref{sec:literature}.

\subsection{Preview of the proposed method and our key contributions}

\begin{figure}[!h]
    \centering
    \includegraphics[width=\linewidth]{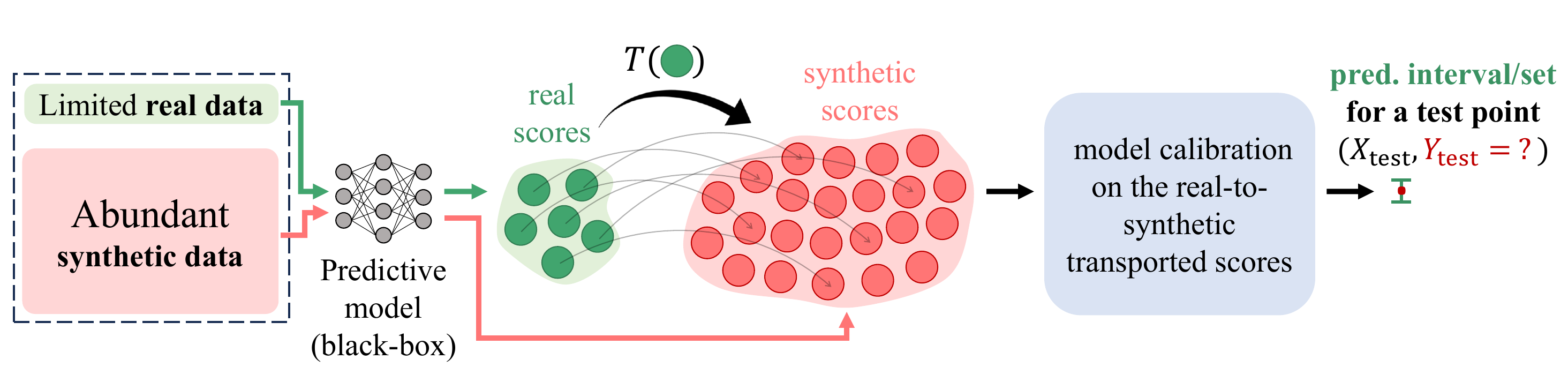}
    \caption{\textbf{A high-level overview of the proposed method}. The approach leverages a small labeled real dataset alongside a large labeled synthetic dataset. The \emph{score transporter} maps scores from the real domain to the synthetic one. Calibration is then performed using the transported real scores and the synthetic scores.}
    \label{app-fig:overview}
\end{figure}

Our key innovation is the introduction of the \emph{score transporter}: a data-driven empirical quantile mapping function that transports the real calibration scores to resemble the synthetic scores. This mapping enables the construction of prediction sets for new test datapoints, leveraging the abundance of synthetic data. Crucially, the score transporter does not require data splitting, allowing full use of the real and synthetic calibration data. Furthermore, 
we develop a computationally efficient algorithm 
with a runtime complexity similar to that of standard conformal prediction. A pictorial illustration of our proposed calibration framework is provided in~\Cref{app-fig:overview}.

We support our proposed $\scp$ framework with two theoretical guarantees. The first shows that when the synthetic and real score distributions are close, 
the achieved coverage closely matches the desired level.
More generally, it characterizes how 
the distributional shift between the real and synthetic scores, and the construction of the score transporter affect the realized coverage.  

The second theoretical result establishes worst-case 
bounds on the coverage probability,
which can be directly controlled by the user. 
This bound allows the user to set a ``guardrail,'' i.e., a lower bound on the coverage probability (say 80\%) that holds regardless of the distribution of the synthetic data.
Remarkably, this bound holds
even when the synthetic data depend on the real calibration set. 
This flexibility enables users to adapt or filter the synthetic data to improve efficiency, for example by selecting datapoints that resemble the real ones---all without requiring any data splitting.    

We demonstrate the practicality of our method on 
multi-class classification and regression tasks. 
For image classification on ImageNet, we explore two practical strategies for constructing synthetic data. The first leverages a generative model (Stable Diffusion \citep{Rombach_2022_CVPR} or FLUX~\citep{flux2024}) to generate artificial images for each class. 
The second uses another set of real data, drawn from a different distribution, as the synthetic data.
In the regression setting, we consider tabular panel data, 
using past panels as synthetic data and a recent panel as the real calibration set. Across all experiments, our method shows improvements in statistical efficiency---even when the real calibration set is very small, with as few as 15 datapoints. Software for reproducing the experiments is available at \hyperlink{https://github.com/Meshiba/spi}{https://github.com/Meshiba/spi}.

\section{Problem setup}\label{sec:problem-setup}
Consider the standard setting of a prediction problem where we have $m$ i.i.d.~(real) calibration datapoints\footnote{Some of our results rely on a weaker assumption than i.i.d.---namely, exchangeability of the real calibration datapoints.} $\smash{{(X_i, Y_i)}_{i} \iidsim P_{X,Y} = P_X \times P_{Y \mid X}}$, 
for $i \in [m]:=\{1, \ldots, m\}$
on $\mathcal{X} \times \mathcal{Y}$, where each $X_i \in \mathcal{X}$ represents the features and $Y_i \in \mathcal{Y}$ denotes the label or outcome for the $i$-th datapoint. 
Given a new test input $X_{m+1} \sim P_X$, 
the task is
to construct a prediction set $\smash{\ch(X_{m+1})}$ for the unknown label $Y_{m+1}$ with the following distribution-free coverage guarantee:
\begin{equation}\label{eq:coverage}
    \Pp{(X_i,Y_i)_{i \in [m+1]}\iidsim P_{X,Y}}{ Y_{m+1}\in \ch(X_{m+1})} \ge 1-\alpha, \text{ for any distribution $P_{X,Y}$ on $\X \times \Y$},
\end{equation}
where $1-\alpha \in (0,1)$ is a predetermined target level of coverage. 
Here and below, we abbreviate by 
$(a_i)_{i \in [k]}$ vectors $(a_1,a_2,\cdots,a_k)$.
While standard conformal prediction \citep{vovk1999machine,papadopoulos2002inductive,vovk2005algorithmic, angelopoulos2021gentle} 
has this property, 
its efficiency can be limited when the calibration sample size $m$ is small.

Suppose now that we also have access to a set of synthetic datapoints $(\tilde{X}_j, \tilde{Y}_j )_{j \in [N]} \iidsim Q_{X,Y}$.
These could be datapoints
collected from related distributions,  sampled from a generative model, or obtained otherwise. 
We hope that $Q_{X,Y}$ is close to $P_{X,Y}$, but do not assume this.
We are interested in the setting where $m \ll N$, 
aiming to improve inference with a small calibration set by leveraging a large synthetic dataset. 
To make this concrete, we aim to construct 
a prediction set map $\smash{\ch : \X \rightarrow \mathcal{P}(\Y)}$---where $\mathcal{P}(\Y)$ is the set of subsets of $\Y$---as a function of the datasets ${(X_i, Y_i)}_{i \in [m]}$ and $(\tilde{X}_j, \tilde{Y}_j )_{j \in [N]}$, such that the prediction set $\smash{\ch(X_{m+1})}$ satisfies, for any distributions $P_{X,Y}$ and $Q_{X,Y}$ on $\X \times \Y$,
\begin{equation}\label{eqn:coverage_2}
\begin{split}
    &\Pp{(X_i,Y_i)_{i \in [m+1]}\iidsim P_{X,Y}, (\tilde{X}_i,\tilde{Y}_i)_{i \in [N]}\iidsim Q_{X,Y}}{ Y_{m+1}\in \ch(X_{m+1})} \ge 1-\alpha.
\end{split}
\end{equation}

For the classification task where $Y$ is discrete, we extend our discussion beyond the marginal coverage guarantee in \eqref{eq:coverage} and consider the following label-conditional coverage guarantee \citep{vovk2003mondrian,vovk2005algorithmic}:
\begin{equation}\label{eq:coverage_class}
    \PPst{ Y_{m+1}\in \ch(X_{m+1})}{Y_{m+1} = y} \ge 1-\alpha, \text{ for all $y \in \Y$},
\end{equation}
where, as before, we aim for a distribution-free guarantee, under any distributions $P_{X,Y}$ and $Q_{X,Y}$---although the inequality in \eqref{eq:coverage_class} is written in a simplified form.

\subsection{Background: split conformal prediction}
Split conformal prediction \citep{papadopoulos2002inductive}
is an approach to attain 
the coverage guarantee~\eqref{eq:coverage}.
The first step 
is to construct a nonconformity score function $s : \mathcal{X} \times \mathcal{Y} \rightarrow \mathbb{R}$
from an independent dataset.\footnote{A typical example for regression problems is $s(x,y) = |y-\hat\mu(x)|$, where $\hat \mu$ is a predictor pre-trained on a separate dataset; see e.g., \citep{vovk2005algorithmic, angelopoulos2021gentle}, etc.}  
Next, we compute the scores on the calibration datapoints: $S_i = s(X_i,Y_i)$ for $i \in [m]$. 
The prediction set is then given as
\begin{equation}\label{eqn:split_conformal}
    \ch(X_{m+1}) := \left\{ y \in \mathcal{Y} : s(X_{m+1}, y) \le \hat{Q}_{1-\alpha} \right\},
\end{equation}
where $\hat{Q}_{1-\alpha}$ denotes the $\lceil (1-\alpha)(m+1)\rceil$-th smallest score from the (multi-)set $(S_i)_{i \in [m]}$.

If the scores $(S_i)_{i \in [m]}$ are distinct almost surely, then the split conformal prediction set~\eqref{eqn:split_conformal}
attains the following coverage bounds~\citep{vovk1999machine,papadopoulos2002inductive,vovk2005algorithmic}:
\begin{equation}\label{eqn:bound_split}
    1-\alpha \leq \PP{Y_{m+1} \in \ch(X_{m+1})} \leq 1-\alpha+1/(m+1).
\end{equation}
If $m$ is very small, the split conformal
set might be conservative.
In particular, if $m+1 < 1/\alpha$, then the only way to achieve
$1-\alpha$ coverage with $m$ datapoints is 
by producing a trivial prediction set that includes all labels.
For a typical value of $\alpha = 0.05$, this is the case when $m<19$.
Since we aim to handle situations with very low sample sizes, 
this motivates us to develop
a procedure capable
of producing more informative prediction sets, by leveraging synthetic data.

\section{Methodology}

\subsection{Synthetic-powered predictive inference}
\label{sec:sp-cp}
    In this section, we introduce our method---\scp---which is designed to leverage the synthetic datapoints to effectively increase the sample size, thereby producing a non-conservative prediction set. We construct a split-conformal-type method that performs inference based on pre-constructed nonconformity scores. Throughout the section, we assume that the score function $s: \mathcal{X} \times \mathcal{Y} \rightarrow \mathbb{R}$ is fixed, and denote the real and synthetic scores as $S_i = s(X_i,Y_i)$ for $i \in [m+1]$ and $\smash{\tilde{S}_j = s(\tilde{X}_j,\tilde{Y}_j)}$ for $j \in [N]$, respectively.
Here $S_{m+1}$ is the unobserved test score.

Our strategy is to construct a score-transporter $T$ that maps a real score to a synthetic score---as a function of the observed scores. 
We then run split conformal prediction on the synthetic scores and apply 
$T$ to obtain a prediction set for the real score $S_{m+1}$.
A carefully constructed map $T$ can generate a prediction set with a theoretically controlled coverage rate, while effectively leveraging the large synthetic dataset.
The procedure has three steps.

\paragraph{Step 1. Construct windows in the space of synthetic scores.} 
 Denote by $S_{(1)}, \ldots, S_{(m+1)}$ the real scores arranged in increasing order.
 We first define a ``window"
 $I_m(r)$
  designed to contain the $r$-th score $S_{(r)}$
 for each $r \in [m+1]$, as follows: 
 \begin{equation}\label{eqn:window_bounds}
 \begin{split}
     &R_r^- = \max\left\{t \in [N+1] : F(t-1) \leq \tfrac{\beta}{2}\right\},\,
     R_r^+ = \min\left\{t \in [N+1] : F(t)\ge 1-\tfrac{\beta}{2}\right\},
     \end{split}
 \end{equation}
 where $\beta \in (0,1)$ is a predefined level, and  $F:=F_{m,N,r}$ is defined as\footnote{Here, for non-negative integers $a\le b$, $\binom{b}{a} = b!/(a!(b-a)!)$ denotes the binomial coefficient, where $x!=x\cdot (x-1) \cdot \ldots\cdot  1$ is the factorial of a non-negative integer $x$.
  Also, \( p_{m,N,r}(k) \) is the \emph{probability mass function of the 
 \(r\)-th order statistic} from a random sample of size \( m +1\) drawn \emph{without replacement} from a finite population of size \( N + m+1 \)
 \citep[e.g.,][p. 243]{wilks1962mathematical}.}
 \[F(t) = \sum_{k=1}^t p_{m,N,r}(k), \text{ with } p_{m,N,r}(k) = \tbinom{k+r-2}{r-1}\tbinom{N+m-k-r+2}{m-r+1}/{\tbinom{N+m+1}{m+1}}.\]
Then, with the 
synthetic scores 
$\smash{\tilde{S}_{(1)}, \ldots, \tilde{S}_{(N)}}$ in increasing order,
we construct the window as
\begin{equation}\label{im}
I_m(r) = [L_{m}(r), U_{m}(r)], \quad \text{where} \quad L_{m}(r) := \tilde{S}_{(R_r^-)} \text{ and } U_{m}(r) := \tilde{S}_{(R_r^+)},
\end{equation}
and where $ \tilde{S}_{(N+1)} = +\infty $. This window is designed to satisfy the following property, 
where we denote the distribution of $s(X,Y)$ under $P_{X,Y}$ and $Q_{X,Y}$ by $P$ and $Q$, respectively:
\begin{lemma}\label{lem:window}
    If $P = Q$ and both are continuous distributions, then $\PP{S_{(r)} \in I_m(r)} \ge 1 - \beta$ for all $r \in [m+1]$.
\end{lemma}

The proof is deferred to Appendix~\ref{app-sec:proofs}. 
Intuitively, the window $I_m(r)$ represents a region in the synthetic score space where $S_{(r)}$ is likely to lie, and the transporter we construct in the next step maps the real score to an element within its corresponding window.

\paragraph{Step 2. Construct the score-transporter.}
We now define the map $T(\cdot) = T(\cdot ; (S_{i})_{i\in [m]}, (\tilde{S}_{j})_{j\in [N]})$
mapping real to synthetic scores
as follows. 
For a scalar $\eta$, let $r_\eta = \sum_{i=1}^m \One{S_i < \eta}+1$ denote the rank of $\eta$ among $(S_1, \dots, S_m, \eta)$ in increasing order, and 
with $L_m,U_m$ from \eqref{im},
define
\begin{equation}\label{eqn:score_transporter}
T(\eta) = 
\begin{cases}
U_{m}(r_\eta), & \text{if } \eta \ge U_{m}(r_\eta), \\
\mathrm{NN}_{m}^-(r_\eta, \eta), & \text{if } L_{m}(r_\eta) \le \eta < U_{m}(r_\eta), \\
L_{m}(r_\eta), & \text{if } \eta < L_{m}(r_\eta),
\end{cases}
\end{equation}
where the \emph{lower nearest neighbor} $\mathrm{NN}_m^-$ is defined as 
\[\mathrm{NN}_{m}^-(r,\eta) := \max_{
R_r^- \le j \le  R_r^+} 
\left\{ \tilde{S}_{(j)} : \tilde{S}_{(j)} \le \eta \right\}.\]
Roughly speaking, the score-transporter $T$ maps $\eta$ to a synthetic score in the corresponding window $I_m(r_\eta)$ that is closest to $\eta$.
The lower nearest neighbor $\mathrm{NN}_{m}^-$ is chosen carefully to act as a lower bound on the score, ensuring that the coverage
can be tightly controlled.

\begin{figure}[!h]
    \begin{subfigure}[t]{\linewidth}
    \centering
    \includegraphics[width=0.7\linewidth]{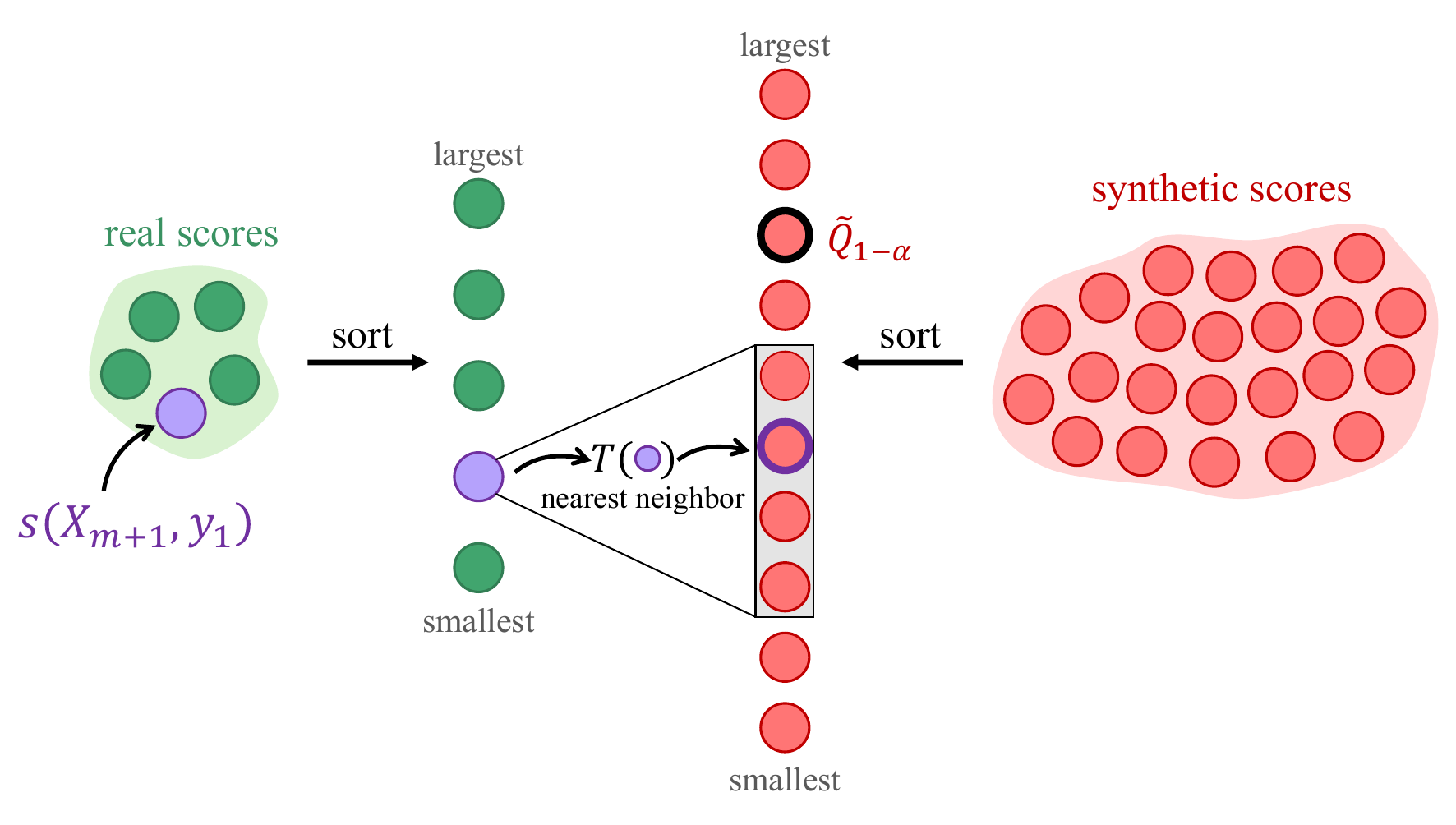}
    \caption{Candidate label $y_1$: the test score (purple) ranks second among the real scores. Its mapped synthetic neighbor---computed via~\eqref{eqn:score_transporter} and outlined in purple---falls below the empirical quantile $\tilde{Q}_{1-\alpha}$, hence $y_1\in \ch(X_{m+1})$.}
    \label{fig:score-transporter-1}
     \end{subfigure}
     \begin{subfigure}[t]{\linewidth}
    \centering
    \includegraphics[width=0.7\linewidth]{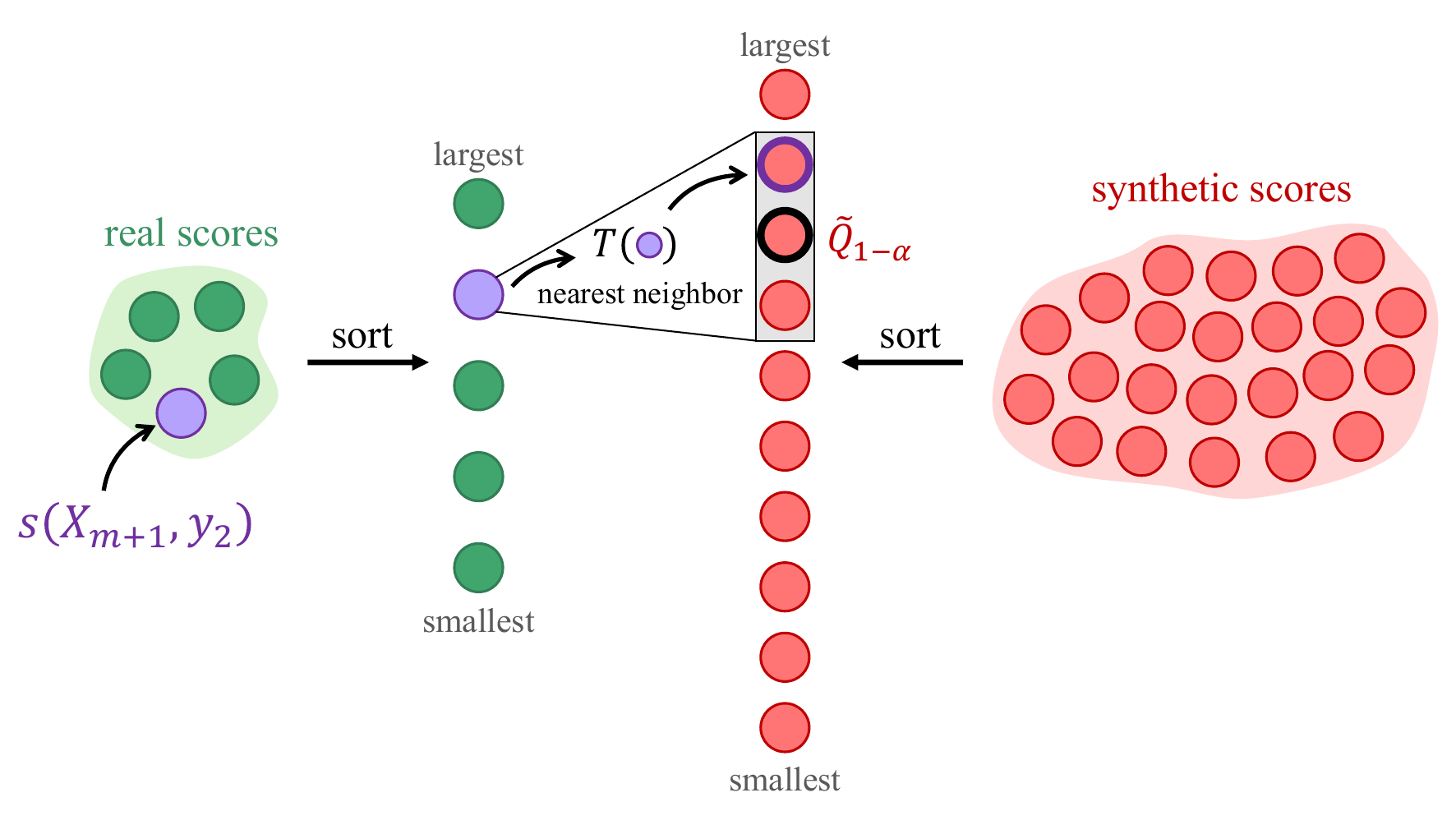}
         \caption{Candidate label $y_2$: the test score (purple) ranks fourth among the real scores. Its mapped synthetic neighbor---computed via~\eqref{eqn:score_transporter} and outlined in purple---exceeds the empirical quantile $\tilde{Q}_{1-\alpha}$, thus $y_2\notin \ch(X_{m+1})$.}
    \label{fig:score-transporter-2}
     \end{subfigure}
    \caption{\textbf{Illustration of the synthetic-powered predictive inference for two candidate labels.} 
Each panel displays sorted nonconformity scores: real scores on the left and synthetic scores on the right. The rectangle indicates the window in the synthetic space to which the test score can be mapped (as defined in~\eqref{im}). The black-outlined circle indicates the $(1-\alpha)(1+\frac{1}{N})$th empirical quantile of the synthetic scores, $\tilde{Q}_{1-\alpha}$.}
    \label{fig:score-transporter}
\end{figure}

\paragraph{Step 3. Conformal prediction after transport-mapping.}
Applying the score-transporter $T$ to a hypothetical score $s(X_{m+1}, y)$, 
we construct the prediction set as those $y$ values for which this mapped value lies in the conformal prediction region constructed from the synthetic data:
\begin{equation}\label{eqn:pred_set}
    \ch(X_{m+1}) = \left\{ y \in \mathcal{Y} : T\left(s(X_{m+1}, y)\right) \le \tilde{Q}_{1-\alpha}\right\}.
\end{equation}
Here
\(\tilde{Q}_{1-\alpha}\)
is the 
$\lceil (1-\alpha)(N+1)\rceil$-th smallest score in
$(\tilde{S}_j)_{j \in [N]}$. 
We term this procedure \textit{Synthetic-powered predictive inference} (\scp).
\Cref{fig:score-transporter} presents a schematic overview of $\scp$ with two candidate labels, illustrating each of the steps discussed above.
Building on the ideas of~\citep{vovk2003mondrian,vovk2005algorithmic}, we 
extend our proposed method to achieve label-conditional coverage guarantees in~\cref{app-sec:label-cond-calib}.

\subsection{Simplifying the computation of \texorpdfstring{$\scp$}{}}
\label{simp}
Since 
$T(\cdot) = T(\cdot ; (S_{i})_{i\in [m]}, (\tilde{S}_{j})_{j\in [N]})$
depends on $(S_{i})_{i\in [m]}$ and $(\tilde{S}_{j})_{j\in [N]}$,
the prediction set $\smash{\ch(X_{m+1})}$ in~\eqref{eqn:pred_set} 
has an \emph{a priori} potentially complex dependence on $y$.
Fortunately, the prediction set simplifies to the following formula, which is fast to compute:

\begin{equation}\label{eqn:pred_set_alt}
    \ch^{\mathrm{fast}}(x) = \left\{y \in \Y : s(x,y) \le \max\{\min\{\tilde{Q}_{1-\alpha}', S_{(\tilde{R}^-)}\}, S_{(\tilde{R}^+)}\}\right\},\, S_{(m+1)} = +\infty. 
\end{equation}
Here 
$\tilde{Q}_{1 - \alpha}'$ 
is the 
$(\lceil {(1 - \alpha)(N+1)}\rceil+1)$-th smallest score among 
$(\tilde{S}_j)_{j \in [N]}$,
and
\begin{equation}\label{eqn:tilde-R}
\begin{split}
    &\tilde{R}^\pm = \max\{r \in [m+1] : R_r^\pm \le \lceil {(1 - \alpha)(N+1)}\rceil\}.
\end{split}
\end{equation}
The following result shows that the prediction set $\ch^{\mathrm{fast}}$ is equivalent to the prediction set~\eqref{eqn:pred_set}---here, for two sets $A$ and $B$, $A \triangle B$ denotes the symmetric set difference $(A \cap B^c) \cup (A^c \cup B)$.
\begin{proposition}\label{prop:pred_set}
    Recall the prediction sets $\ch$ from~\eqref{eqn:pred_set} and
    $\ch^{\mathrm{fast}}$ from~\eqref{eqn:pred_set_alt}. If $Q$ is continuous, then \[\PP{\{Y_{m+1} \in \ch(X_{m+1})\} \,\triangle\, \{Y_{m+1} \in \ch^{\mathrm{fast}}(X_{m+1})\}} = 0.\]
\end{proposition}
\noindent Based on this simplification, we present the complete $\scp$ procedure in Algorithm~\ref{algo:ot-calibration}.

\subsection{Theoretical guarantees}\label{sec:thr-guarantees}

We now derive bounds on the coverage rate of the $\scp$ prediction set~\eqref{eqn:pred_set}.
The first bound shows that when the real and synthetic scores are similar (as measured by total variation distance), our method has a tight coverage around the desired level.
\begin{theorem}[Coverage depending on the closeness of real and synthetic distributions]\label{thm:coverage_bound}
Suppose the real calibration set $(X_i,Y_i)_{i\in [m]}$ is exchangeable with the test point $(X_{m+1},Y_{m+1})$
and the synthetic calibration datapoints $(\tilde{X}_j,\tilde{Y}_j)_{j\in[N]}$ are drawn i.i.d., where the distribution $Q$ of their scores is continuous.
Let $P_{(r)}^{m+1}$ and $Q_{(r)}^{m+1}$ denote the distribution of the $r$-th order statistic among $m+1$ i.i.d.~draws from $P$ and $Q$, respectively. 
Then the prediction set $\smash{\ch(X_{m+1})}$ from~\eqref{eqn:pred_set} satisfies
\[1-\alpha-\beta - \ep_{P,Q}^{m+1} \le \PP{Y_{m+1} \in \ch(X_{m+1})} \le 1-\alpha+\beta + \ep_{P,Q}^{m+1} + 1/(N+1),\]
where $\ep_{P,Q}^{m+1} = \frac{1}{m+1}\sum_{i=1}^{m+1} \dtv(P_{(i)}^{m+1}, Q_{(i)}^{m+1})$
and $\dtv$ denotes the total variation distance.
\end{theorem}

When $P = Q$, we have $\ep_{P,Q}^{m+1} = 0$, 
and thus
our procedure provides a tighter upper bound than split conformal prediction using only the real calibration data~\eqref{eqn:bound_split}  when $\beta + 1/(N+1) \le {1}/({m+1})$. When $N\gg m$ and $\beta \ll 1/m$, our method offers a tighter coverage. In practice, however, we often observe tight coverage even for relatively large $\beta$---in the proof, the $\pm \beta$ error in the bounds corresponds to an extreme case in which all the true scores fall outside their respective windows, a scenario that is typically unlikely when $P \approx Q$.

\begin{remark}
The continuity of the score distribution required in Theorem~\ref{thm:coverage_bound} can generally be attained conveniently. 
For example, in settings where the originally constructed score outputs discrete values, one can simply add a negligible amount of i.i.d.~Uniform$[-\delta,\delta]$ noise to the scores, so that the perturbed scores---which are nearly identical to the original ones---have a continuous distribution.
\end{remark}

When the distributions $P$ and $Q$ differ greatly, 
the bounds in Theorem~\ref{thm:coverage_bound} may be loose,
as they do not sufficiently account for the adjustment introduced by the map $T$ under distribution shift. 
Below, we provide alternative worst-case bounds for the coverage rate of the $\scp$ prediction set, 
which depend 
only on the sample sizes, and hold regardless of the relationship between $Q$ and $P$.

\begin{theorem}[Worst-case coverage]\label{thm:worst_case}
Suppose that the real calibration set $(X_1,Y_1),\dots,(X_m,Y_m)$ is exchangeable with the test point $(X_{m+1},Y_{m+1})$, and that the synthetic score distribution $Q$ is continuous. Then the prediction set $\smash{\ch(X_{m+1})}$ in~\eqref{eqn:pred_set} satisfies
\begin{multline*}
\frac{|\{ j \in[m+1] : R_j^+ \le \left\lceil (1 - \alpha)(N + 1) \right\rceil \}|}{m+1} \le \PP{Y_{m+1} \in \ch(X_{m+1})}\\
\le \frac{|\{ j \in[m+1] : R_j^- \le \left\lceil (1 - \alpha)(N + 1) \right\rceil \}|}{m+1}.
\end{multline*}  
\end{theorem}
While this result is somewhat non-explicit,
the bounds can be computed fast, and remain close to the target level $1 - \alpha$, as illustrated in Section~\ref{sec:bound_plots} through a set of plots.
We emphasize that the bounds hold due to the careful construction of the score transport map from \eqref{eqn:score_transporter}, and would not hold if we were to simply mix together the real and synthetic data.
Moreover, the bounds impose no condition on the distribution of the synthetic scores---it is even allowed for the synthetic scores 
to depend
on the real calibration set. 
This provides significant flexibility in the choice of synthetic data even with a separate score function; see Section~\ref{sec:score_trans}.

The bounds in Theorem~\ref{thm:worst_case} 
depend solely on $m$, $N$, $\alpha$, and $\beta$. 
This allows us to adjust the levels $\alpha$ and $\beta$ to achieve a specific lower bound, say $1-\alpha'$, for a predetermined value of $\alpha'$---which implies that the guarantee~\eqref{eqn:coverage_2} can be achieved.~However, we recommend the procedure without level adjustment, 
in the spirit of Theorem~\ref{thm:coverage_bound}, 
since Theorem~\ref{thm:worst_case} provides worst-case bounds.

\subsection{Improving the quality of synthetic scores}\label{sec:score_trans}

The quality of the $\scp$ prediction set depends on how well the distribution of the synthetic score $Q$ approximates the true score distribution $P$, as supported by Theorem~\ref{thm:coverage_bound}. 
Thus, we can improve the prediction set by carefully constructing the synthetic scores. For example, we can seek a map $g$ such that the distribution of the adjusted synthetic score $\smash{\tilde{S}_j' = g(\tilde{S}_j)}$ better approximates the true distribution $P$.

More generally, we may construct
a separate score function $\tilde{s} : \X \times \Y \rightarrow \R$ for the synthetic data, so that the distribution of the synthetic score $\tilde{s}(\tilde{X},\tilde{Y})$ better approximates that of the real score $s(X,Y)$. 
Or, we may select a subset of the synthetic scores that is expected to provide better approximation. 
Below, we present two approaches to improve the quality of synthetic scores.

\subsubsection{Constructing a separate synthetic score function}

We first discuss the approach of constructing a separate synthetic score function $\tilde{s}$. For example, one might choose to construct $\tilde{s}$ using a split of the real data and a split of the synthetic data. However, if the original data sample size $m$ is small---which is the main focus of this work---we may prefer to reuse the data both for constructing the adjustment function or synthetic scores and for performing inference. Therefore, in this section, we focus on data-dependent score construction, while the details of the data-splitting-based approach are deferred to Section~\ref{sec:score_data_split}.

For example, one might consider constructing an adjustment function $g$ as $s\mapsto g(s) := \hat{\theta}_1 s + \hat{\theta}_2$, where the parameters $(\hat{\theta}_1, \hat{\theta}_2)$ are fitted using the calibration scores $(S_i)_{i \in [m]}$ and $(\tilde{S}_j)_{j \in [N]}$ via least squares:
\[
   \smash{ (\hat{\theta}_1, \hat{\theta}_2) = \argmin_{a,b} \sum_{i=1}^{m} |a \cdot \tilde{S}_{\left(\lfloor {iN}/{m} \rfloor \right)} + b - S_{(i)}|^2,}
\]
and then constructing the adjusted synthetic scores $(g(\tilde{S}_j))_{j \in [N]}$, setting $\tilde{s} = g \circ s$; see~\citep{xie2024boosted} for a more sophisticated approach to learning the score function $\tilde{s}$.

For such a synthetic score $\tilde{s}$ constructed in a data-dependent manner, can we still expect a provable coverage bound?
The answer is yes, since the bounds in Theorem~\ref{thm:worst_case} hold for synthetic scores with arbitrary dependence on the real calibration set.

\begin{corollary}\label{cor:data_dependent_score}
   Suppose the synthetic score function $\tilde{s}$ is constructed using both the real data $(X_i,Y_i)_{i \in [m]}$ and the calibration data $(\tilde{X}_j,\tilde{Y}_j)_{j \in [N]}$. 
   Then the prediction set $\ch$ from~\eqref{eqn:pred_set}, constructed using $\tilde{S}_j = \tilde{s}(\tilde{X}_j, \tilde{Y}_j)$ for $j \in [N]$, attains the bounds stated in Theorem~\ref{thm:worst_case}.
\end{corollary}

\subsubsection{Constructing a subset of synthetic data}

Now, we shift to a different approach for improving the quality of synthetic scores: constructing a subset of the synthetic data that is more relevant for inference on the real data. 
This approach is particularly useful when the synthetic data comes from different sources, 
rather than sampled from a generative model.
The idea is to select synthetic datapoints based on how well they approximate the real data. 
Then, we form a subset consisting of points with high approximation quality.
Again, since Theorem~\ref{thm:worst_case} imposes no condition on the joint distribution of the synthetic scores, the bounds also hold for the $\scp$ prediction set constructed with this subset of synthetic scores.

\begin{corollary}\label{cor:sunthetic_subset}
Let $\smash{I_\text{subset} = \{j_1, \cdots, j_{\tilde{N}}\} \subset [N]}$ denote the indices of a subset of synthetic data points, and suppose that $\tilde{N} = |I_\text{subset}|$ is fixed. Then the prediction set $\smash{\ch}$ from~\eqref{eqn:pred_set}, constructed using $\smash{(\tilde{S}_{j_l})_{l \in [\tilde{N}]}}$ as the synthetic scores, satisfies the bounds stated in Theorem~\ref{thm:worst_case}, with $N$ replaced by $\tilde{N}$.
\end{corollary}

Note that this result requires the number of selected points $\tilde{N}$ to be fixed.\footnote{More generally, if $\tilde{N}$ is random but independent of the real scores, the same bounds hold for the conditional probability $\mathbb{P}\{Y_{m+1} \in \ch(X_{m+1}) \mid \tilde{N}\}$.}
For example, one can use a nearest-neighbor procedure,
in which we partition the synthetic data into subsets of a fixed size $n$, and then select $k$ subsets whose score distributions most closely resemble that of the real data, resulting in $\tilde{N} = nk$ synthetic data points (see Algorithm~\ref{algo:ot-calibration-clustering}).

\section{Experiments}\label{sec:exp}
\vspace{-0.5em}
In this section, we compare the performance of the proposed $\scp$ procedure to that of standard conformal prediction in a setting where a small real calibration set and a large synthetic calibration set are available, each drawn from distinct and unknown distributions.

\vspace{-1em}
\paragraph{Setup and performance metrics} We randomly sample two disjoint subsets from the real data, assigning one as the real calibration set and the other as the test set. 
Additionally, we sample a synthetic calibration set from the synthetic data, which is intentionally larger than the real calibration set, aligning with the focus of this paper.
The test set is used to evaluate the procedure based on two metrics: the coverage rate, and the prediction set size (for classification problems) or prediction interval width (for regression problems). We report the results from 100 repeated trials, each with different random calibration, test, and synthetic datasets.

\vspace{-1em}
\paragraph{Methods}
We compare the following methods: $\cp$—standard split conformal prediction \citep{papadopoulos2002inductive} using the real calibration set; $\cpsynt$---conformal prediction applied to the synthetic calibration set as if it were real, which does not provide coverage guarantees; and $\scp$ (ours)---the proposed procedure outlined in \Cref{algo:ot-calibration}, applied with $\beta=0.4$.

\vspace{-1em}
\subsection{Multi-class classification on the ImageNet data}
\label{sec:imagenet-exp}
We begin by evaluating our method on a multi-class classification task using the ImageNet dataset~\citep{deng2009imagenet}. In particular, we aim for marginal \eqref{eq:coverage} and label-conditional coverage guarantees \eqref{eq:coverage_class}; the latter requires hold-out data for each class. Since our experiments involve generating thousands of images per class, we restrict our study to a subset of 30 classes (listed in~\Cref{app-tab:acc_per_class}) that form the real population.

We consider two scenarios for constructing the synthetic data. In the first scenario, we apply a generative model to produce synthetic images. In the second scenario, the synthetic set is formed using real images drawn from classes not included in the real population. 

Across all experiments and methods, we use a CLIP model \citep{radford2021learning} as the predictive model, along with the adaptive prediction sets (APS) score function~\citep{romano2020classification}. Importantly, CLIP is not trained on ImageNet images. Additional details on the score function and pre-trained model are provided in \cref{app-sec:score-funcs,app-sec:models}, respectively.

\vspace{-1em}
\subsubsection{\texorpdfstring{$\scp$}{} with generated synthetic data}
\label{sec:exp-imagenet-gen}
\begin{figure}[h!]
    \centering
    \setlength{\tabcolsep}{2pt} 
    \renewcommand{\arraystretch}{0.8} 

    \begin{tabular}{c!{\vrule width 2pt}cccccc}
			
            \includegraphics[height=0.1\linewidth]{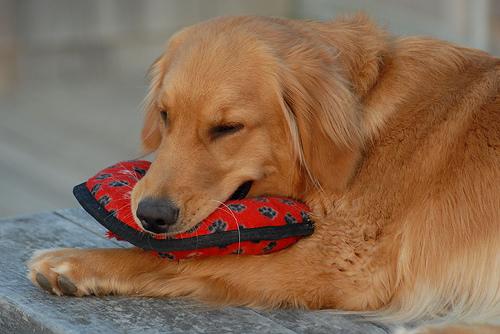} &
            \includegraphics[height=0.1\linewidth]{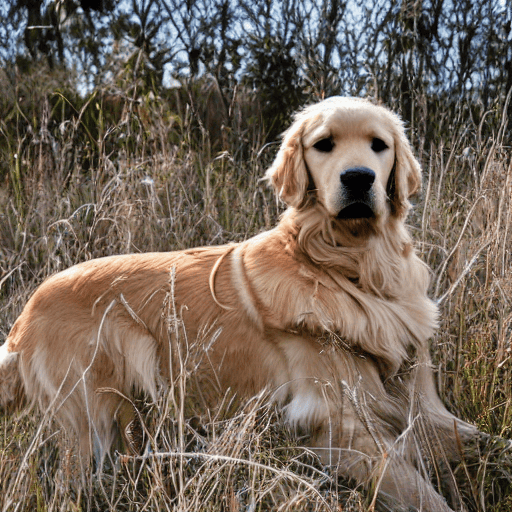} &
            \includegraphics[height=0.1\linewidth]{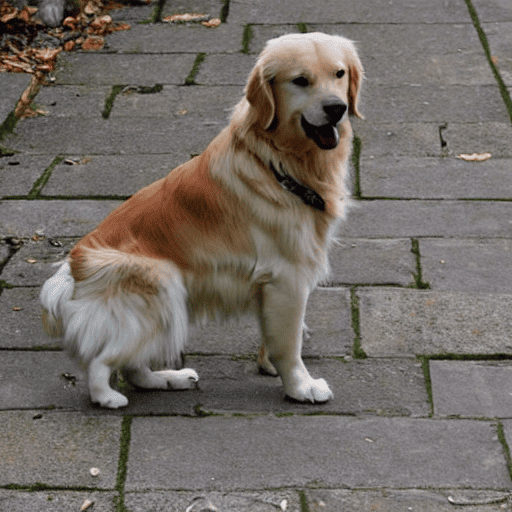} &
            \includegraphics[height=0.1\linewidth]{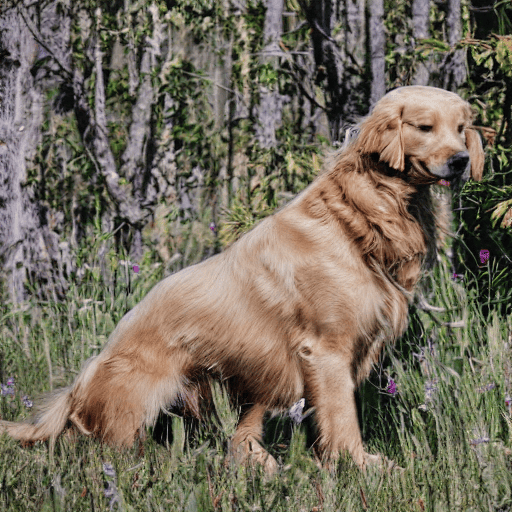} &
            \includegraphics[height=0.1\linewidth]{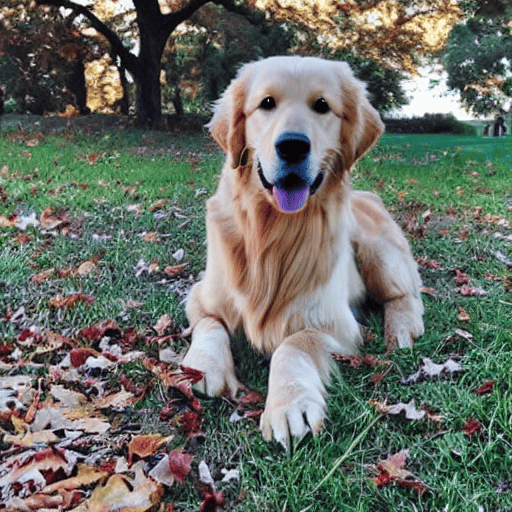} &
            \includegraphics[height=0.1\linewidth]{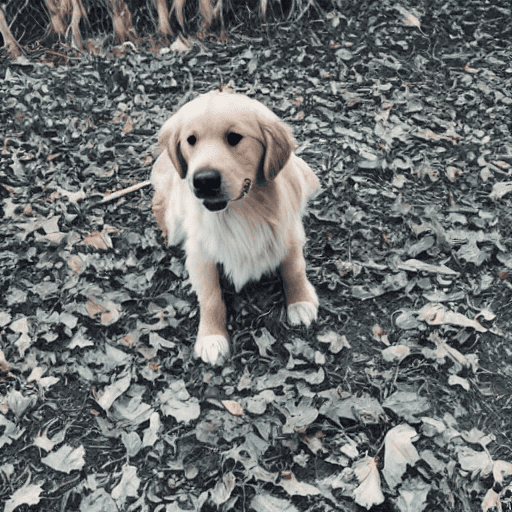} &
            \includegraphics[height=0.1\linewidth]{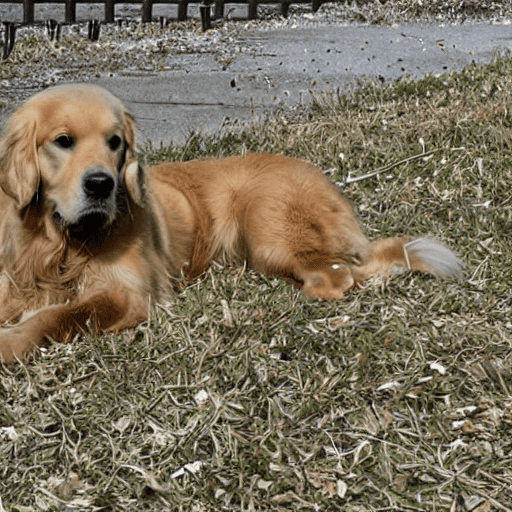} \\
			
    \includegraphics[height=0.1\linewidth]{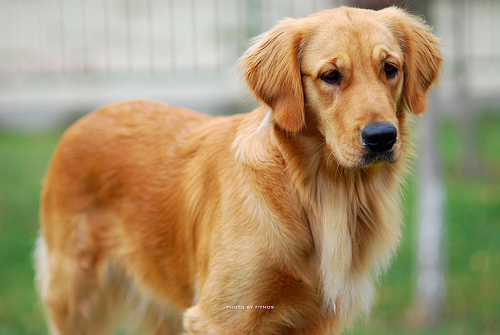} &
    \includegraphics[height=0.1\linewidth]{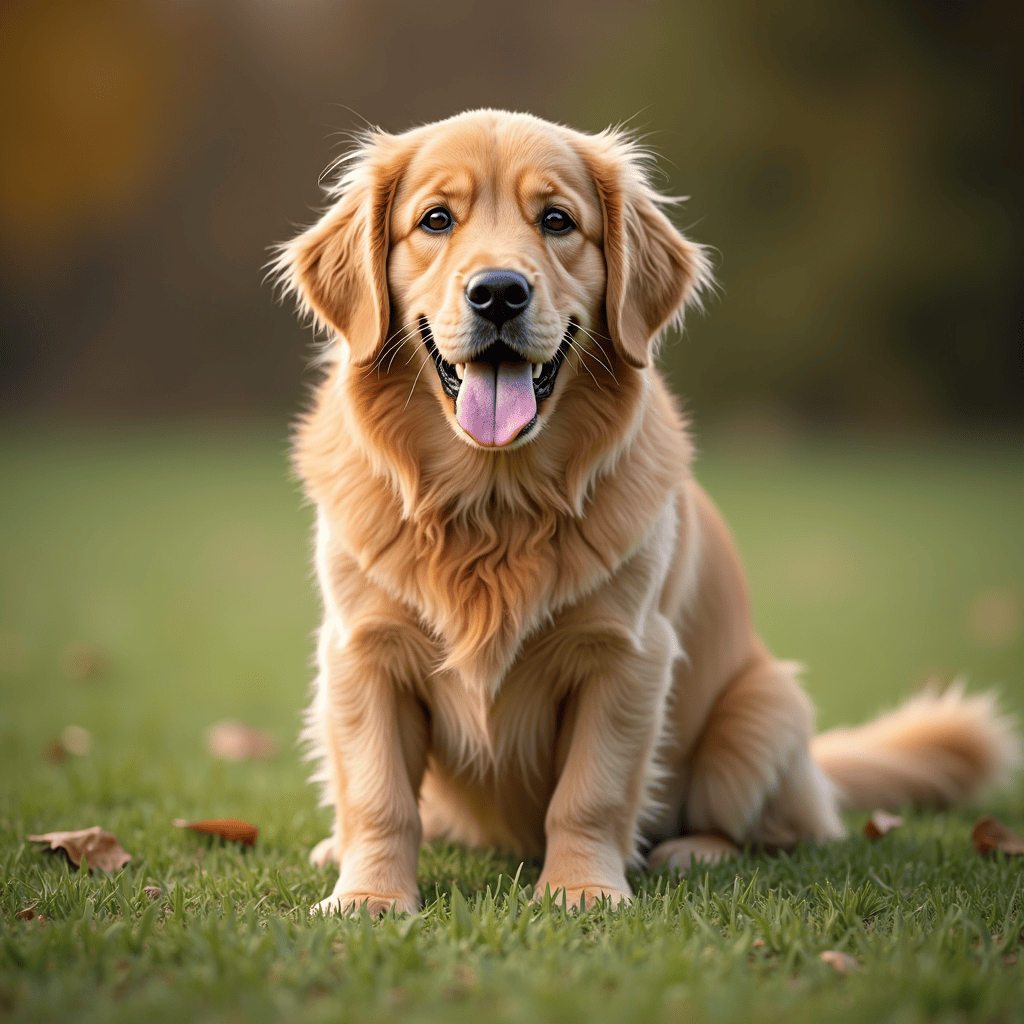} &
    \includegraphics[height=0.1\linewidth]{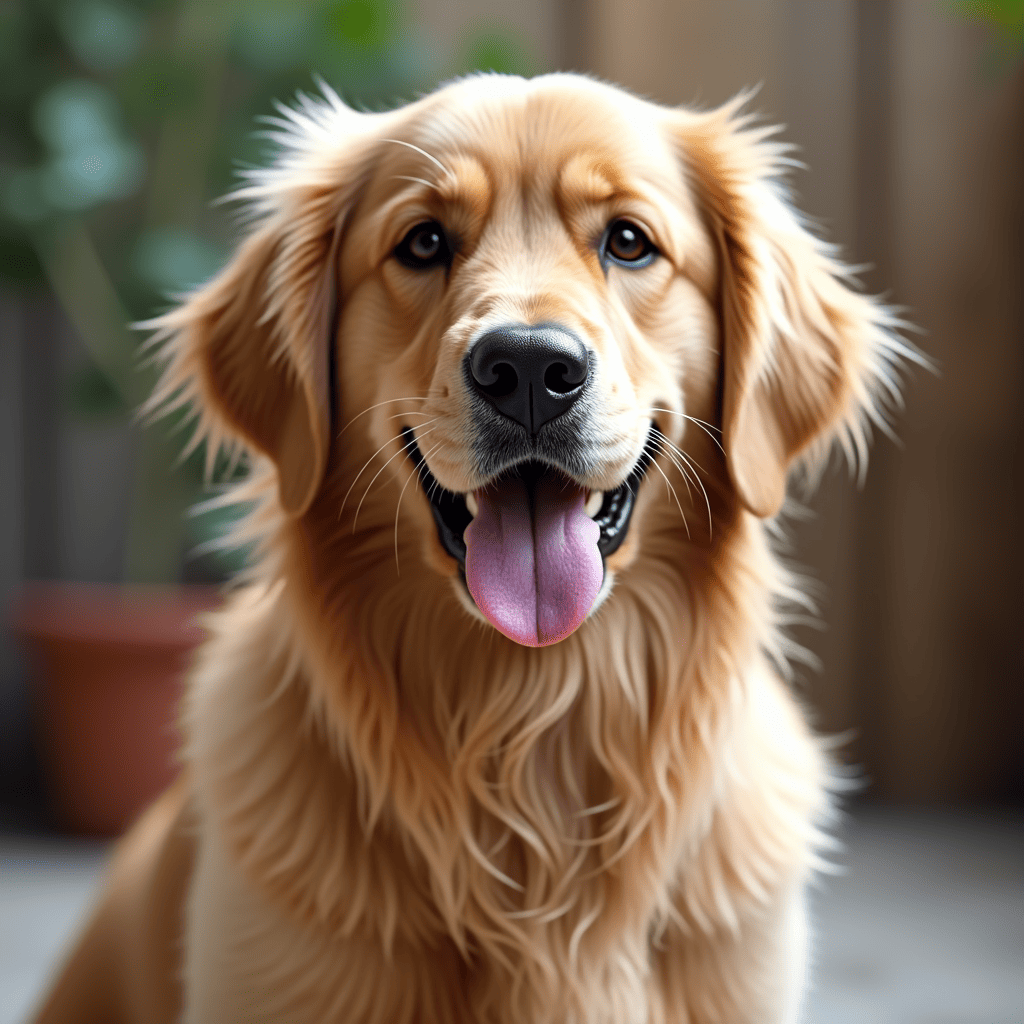} &
    \includegraphics[height=0.1\linewidth]{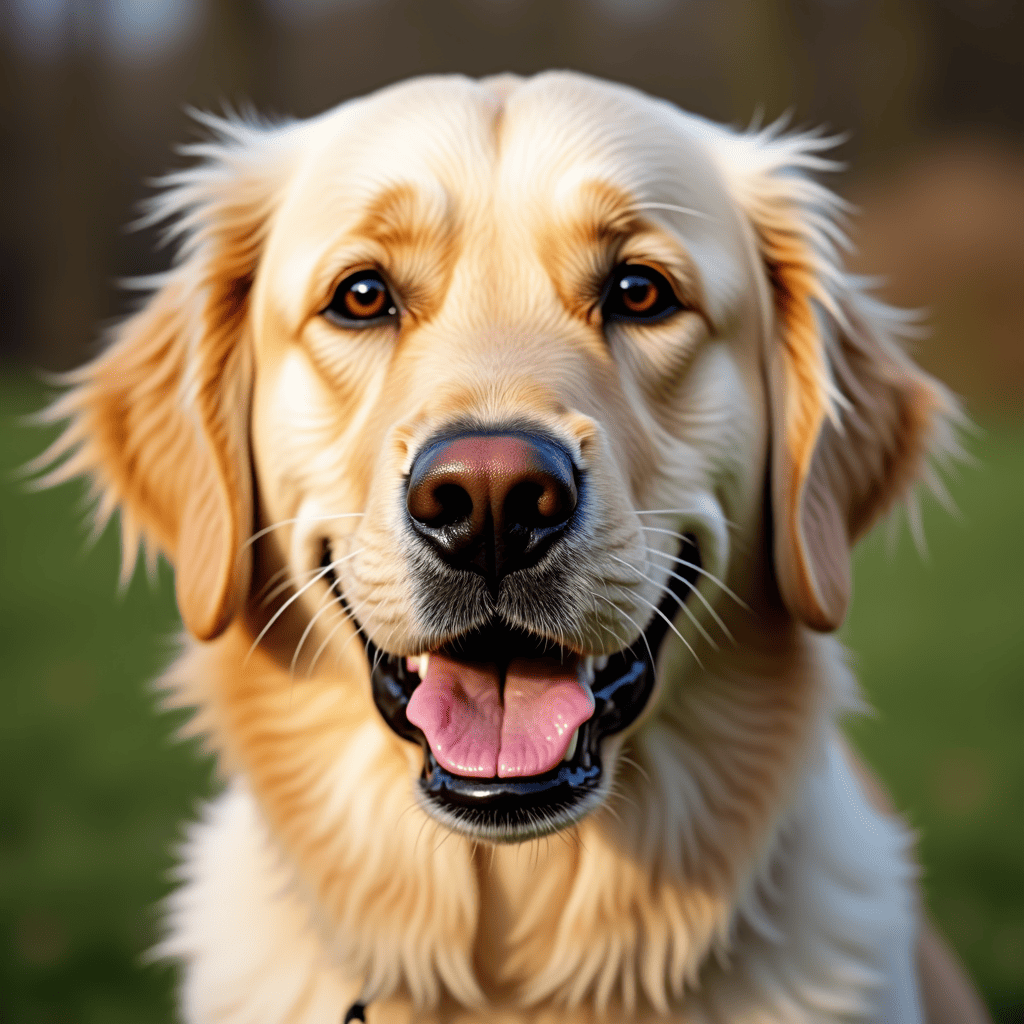} &
    \includegraphics[height=0.1\linewidth]{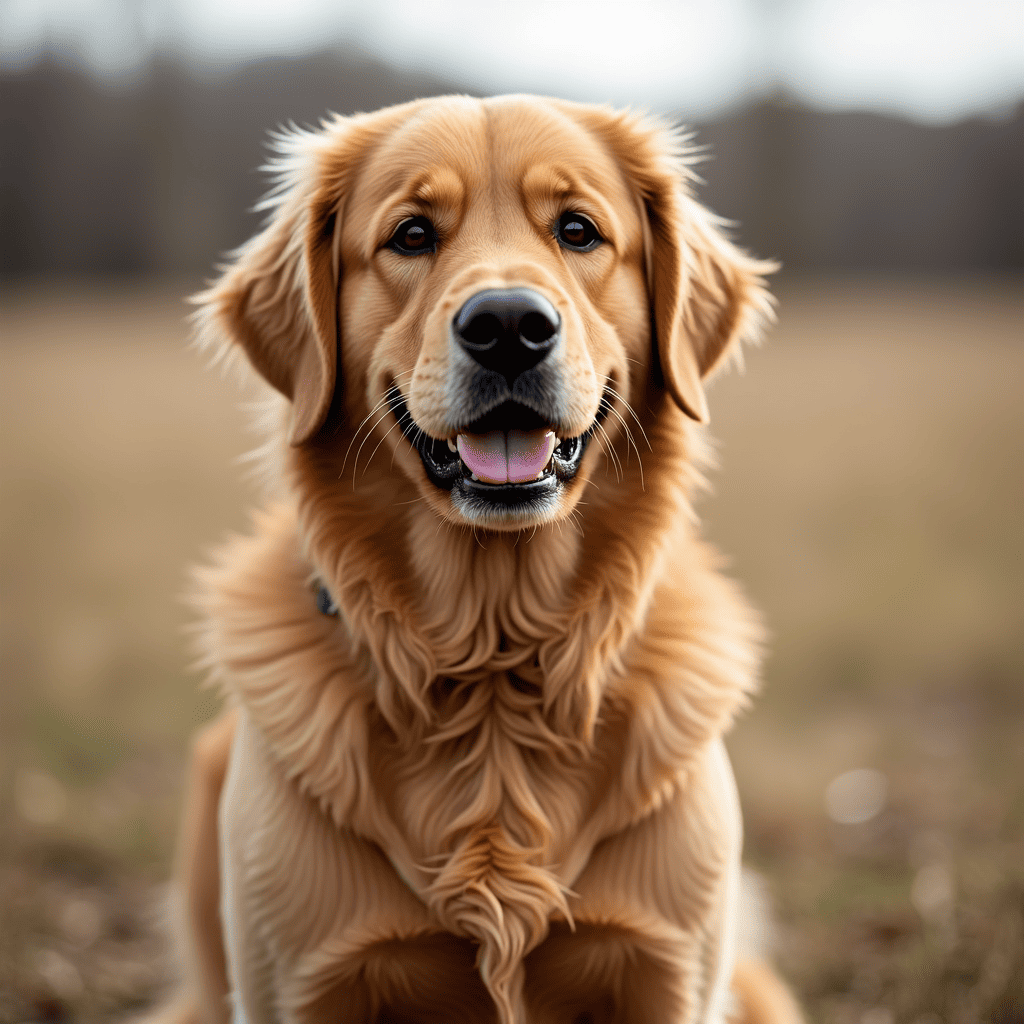} &
    \includegraphics[height=0.1\linewidth]{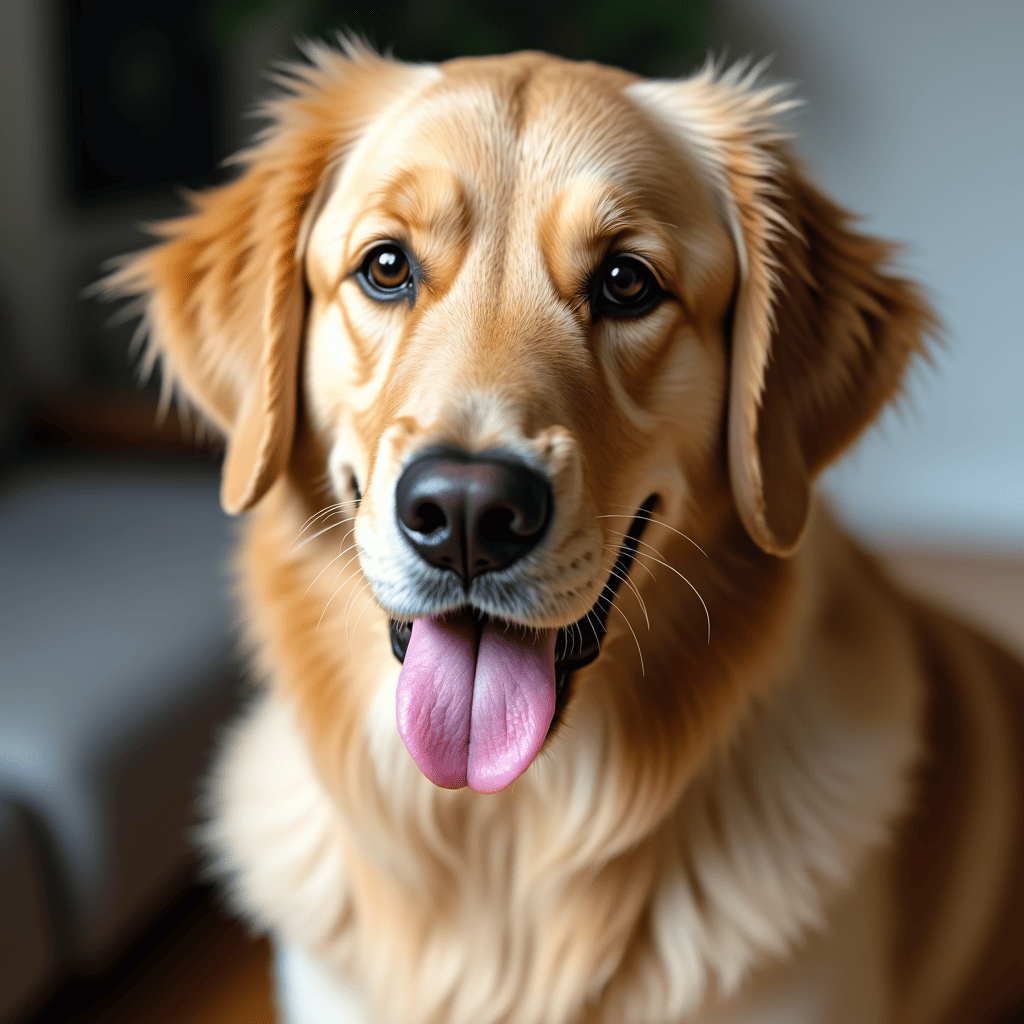} &
    \includegraphics[height=0.1\linewidth]{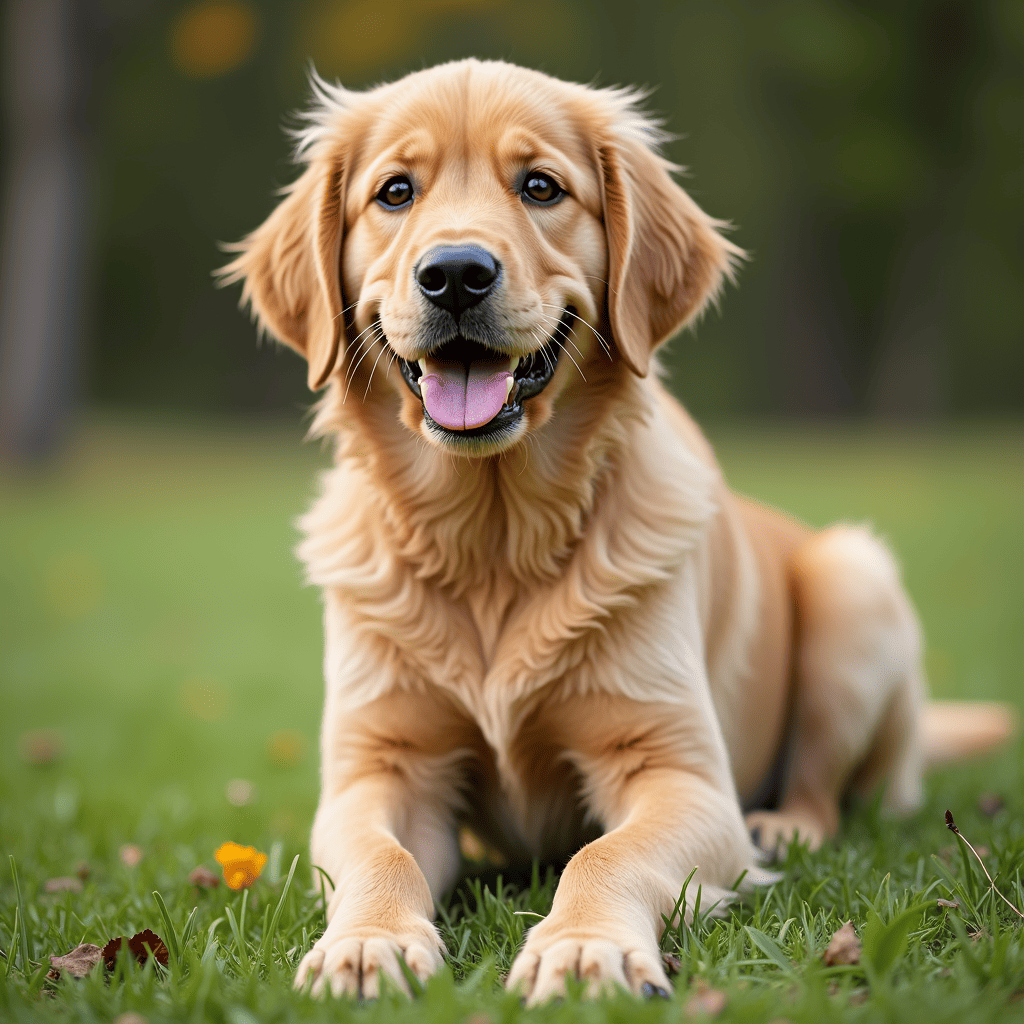} \\
			
    \end{tabular}

    \caption{Examples of real and generated images for the {\em golden retriever} class.
The first column displays a real ImageNet images, while the remaining columns show generated samples. The top row contains images generated by Stable Diffusion~\citep{Rombach_2022_CVPR}, and the bottom row by FLUX~\citep{flux2024}.}
    \label{fig:imagenet-images}
\end{figure}

We use Stable Diffusion~\citep{Rombach_2022_CVPR} to generate synthetic images resembling those in ImageNet. \Cref{fig:imagenet-images} shows representative examples, including images from an additional generative model discussed later. Additional examples and further details are provided in \cref{app-fig:imagenet-images} and \cref{app-sec:data_generation}.

For the marginal coverage experiments, we randomly select $m=15$ ImageNet images from the real data, 
chosen from among $30$ classes, 
to construct the real calibration set. 
The test set consists of 15,000 real images, and the synthetic calibration set includes $N=1,000$ generated images, sampled uniformly across all classes.
For the label-conditional experiments, we randomly select $m=15$ real images for each of the $k=30$ classes to form the real calibration set (resulting in $mk = 450$ real data points), 
500 real images per class to form the test set, and $n=1,000$ generated images per class to form the synthetic calibration set (resulting in $N = nk = 30,000$ synthetic data points).

\begin{figure}[!h]
    \includegraphics[height=0.22\textwidth, valign=t]{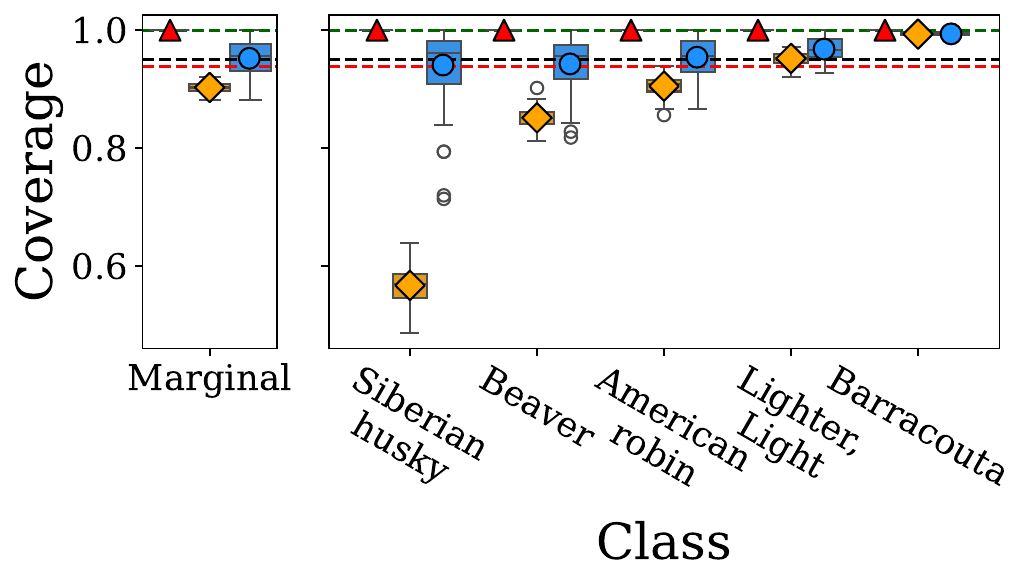}
    \includegraphics[height=0.22\textwidth, valign=t]{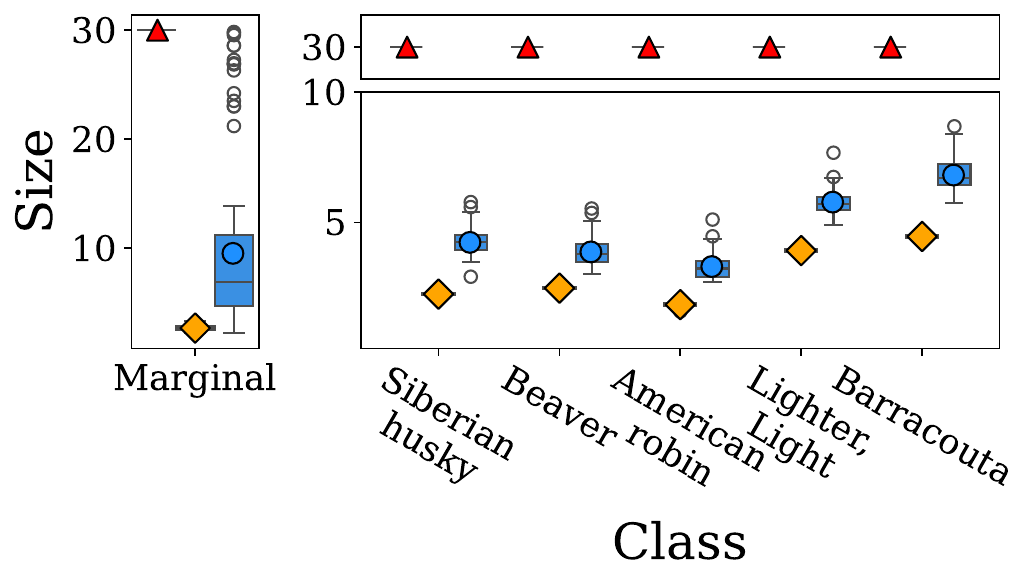}
    \includegraphics[height=0.22\textwidth, valign=t]{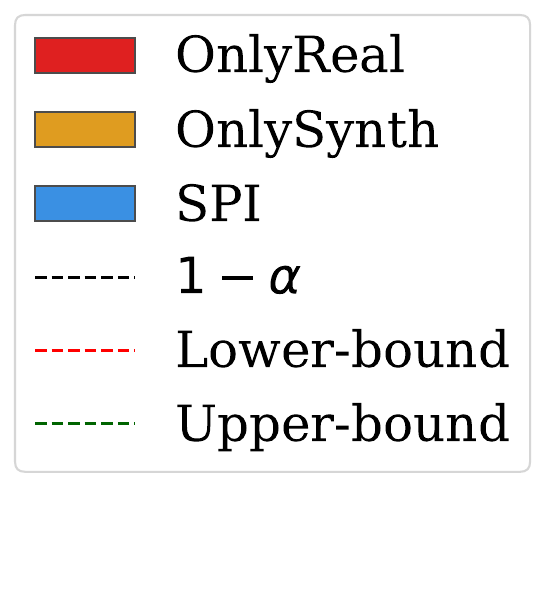}
    \caption{\small Results for the ImageNet data:
     Coverage rates of $\cp$, $\cpsynt$, and $\scp$ at target level $1-\alpha = 0.95$, averaged over 100 trials. Left: Average coverage. Right: Average prediction set size, both under marginal (leftmost box in each group) and label-conditional coverage settings.  Label-conditional results are shown for selected classes; see~\Cref{app-tab:imagenet-gen-all-classes_0.05} for results across all classes.}
    \label{fig:imagenet-gen_0.05}
\end{figure}

\Cref{fig:imagenet-gen_0.05} presents the performance of various methods under both marginal and label-conditional guarantees at target coverage level $1-\alpha=0.95$.
The label-conditional results are shown for five representative classes. 
The observations below apply to both the marginal and label-conditional settings.
We observe that $\cp$ controls the coverage at level $1 - \alpha = 0.95$. However, it remains conservative due to the small size of the real calibration set, which results in trivial prediction sets. The $\cpsynt$ approach fails to achieve the target coverage level of $1 - \alpha$, exhibiting undercoverage for some classes. This violation arises from the distribution shift between the real and synthetic data.

In contrast, the proposed method, $\scp$, achieves coverage within the theoretical bounds established in \cref{thm:worst_case}. For example, for the ``Siberian husky'' class, where the synthetic images differ significantly from the real ones, $\scp$ still produces informative prediction sets. 
For classes where the synthetic and real data are more aligned, such as the ``lighter'' class, $\scp$ shows low variance in coverage with smaller prediction set sizes.

We provide results for additional $\alpha$ levels in~\Cref{app-sec:exp-imagenet-gen}.
Further experiments regarding the effect of the size of the real calibration set on the performance of $\scp$ are shown in \Cref{app-sec:imagenet-gen-calib-size}. In addition, we provide experiments using the FLUX generative model~\citep{flux2024}, which exhibit similar trends to those observed with Stable Diffusion; see \Cref{app-sec:exp-imagenet-flux}. Examples of generated images and details are provided in~\Cref{fig:imagenet-images,app-fig:flux-imagenet-images,app-sec:data_generation}, respectively.

\subsubsection{\texorpdfstring{$\scp$}{} with synthetic data from \texorpdfstring{$k$}{k}-nearest subset selection}
\label{sec:exp-imagenet-clusters}

We now explore the performance of the subset-based variant of our approach, referred to as $\scpc$ and described in~\Cref{algo:ot-calibration-clustering}. The experiments in this section reflect scenarios where a generative model is unavailable.

As before, we aim to control both marginal and label-conditional coverage. In the marginal setting, we randomly select $m=15$ real images, across 30 classes, to form the real calibration set, and 15,000 real images from the same classes to form the test set. In the label-conditional setting, we randomly sample $m=15$ real images per class to form the real calibration set and 500 real images per class for testing. In both cases, the synthetic calibration set consists of $N=1,500$ annotated ImageNet images, drawn from 100 classes that are disjoint from the real classes, with $n=15$ images per class.

We apply the subset selection approach to improve the quality of the synthetic data, using a $k = 20$ nearest-subset selection strategy, leading to $\tilde N = nk = 300$ selected synthetic datapoints (see \Cref{algo:ot-calibration-clustering}). We compare this $\scpc$ variant of our method to $\scph$, where the latter denotes the $\scp$ procedure run with the entire synthetic set.
Additionally, as a baseline, we include standard conformal prediction applied to the real set, $\cp$. 

\begin{figure}[!h]
    \centering
    \includegraphics[height=0.22\linewidth]{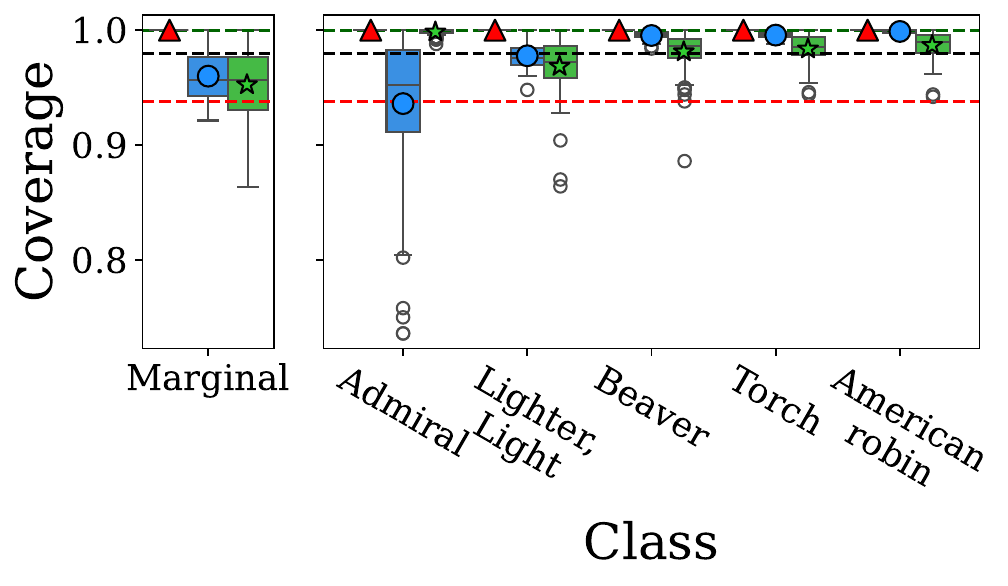}
    \includegraphics[height=0.22\linewidth]{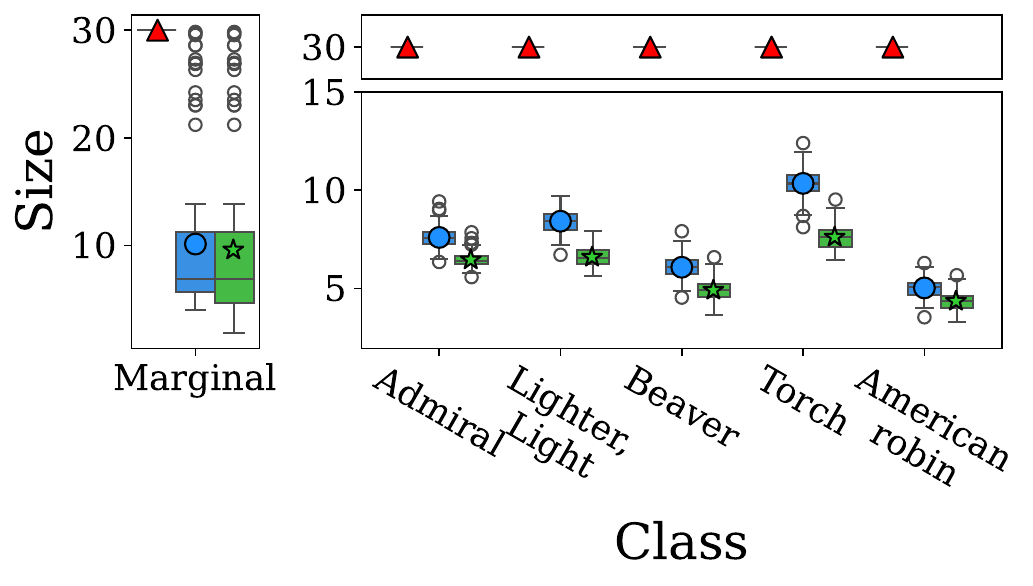}
    \includegraphics[height=0.21\linewidth]{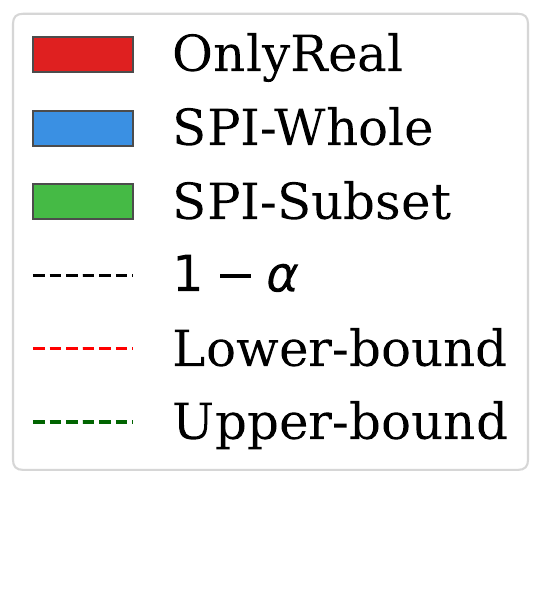}
    \caption{\small Results for the ImageNet data:
    Coverage rates of $\cp$, $\scph$, and $\scpc$ at target level $1-\alpha = 0.98$, averaged over 100 trials. Left: Average coverage. Right: Average prediction set size, both under marginal (leftmost box in each group) and label-conditional coverage settings.  Label-conditional results are shown for selected classes; see~\Cref{app-tab:imagenet-clusters-all-classes_0.02} for results across all classes.
    }
    \label{fig:imagenet-clusters_0.02}
\end{figure}

\Cref{fig:imagenet-clusters_0.02} shows the performance of different methods with marginal and label-conditional guarantees at target coverage level $1-\alpha=0.98$. The label-conditional results are shown for five representative classes. We see that $\cp$ controls the coverage at the $1 - \alpha$ level as expected, but it produces overly conservative---in fact, trivial---prediction sets that contain all 30 possible labels. 
This is not surprising as split conformal prediction needs at least 50 datapoints to produce a nontrivial prediction set at level $\alpha = 0.02$.

Both $\scph$ and $\scpc$ achieve coverage within the theoretical bounds, generating smaller prediction sets compared to $\cp$. 
In the label-conditional setting, $\scpc$ achieves coverage that more closely aligns with the target $1-\alpha$ and produces smaller prediction sets, outperforming $\scph$. This highlights the benefit of aligning the synthetic set more closely with the real distribution through $k$-nearest subset selection. 

However, in the marginal setting, where the real calibration set includes images from 30 different classes, selecting a small subset of $k$ synthetic classes does not necessarily improve alignment with the real distribution. Consequently, $\scpc$, which uses only a subset of the synthetic data (300 images), exhibits higher variance in coverage compared to $\scph$, which leverages the entire synthetic calibration set of 1,500 images.

We provide results for all classes appearing in the real data at additional values of the target level $\alpha$ in~\Cref{app-sec:exp-imagenet-clusters}. In~\cref{app-sec:imagenet-clusters-k}, we further illustrate the performance of the $\scpc$ procedure for different values of the hyperparameter $k$.

\subsection{Regression on the MEPS dataset}
\label{sec:exp-meps}
In this experiment, we evaluate our method on a regression task using the Medical Expenditure Panel Survey (MEPS) datasets~\citep{MEPS19}. We first fit a regression model on MEPS panel survey number 19. MEPS panel survey number 20 is then used as the synthetic data, and panel survey number 21 serves as the real data. This setup reflects a scenario in which large historical panels are leveraged as synthetic data to improve calibration on a recent, smaller real-world population. For all methods, we use the conformalized quantile regression (CQR) score function~\citep{romano2019conformalized}. Further details on the score function and the regression model are provided in \cref{app-sec:score-funcs,app-sec:models}, respectively. For each age group, we construct a real calibration set with $m=15$ examples and a synthetic calibration set with $N=1,000$ examples.

\begin{figure}[!h]
    \centering
    \includegraphics[height=0.2\textwidth, valign=t]{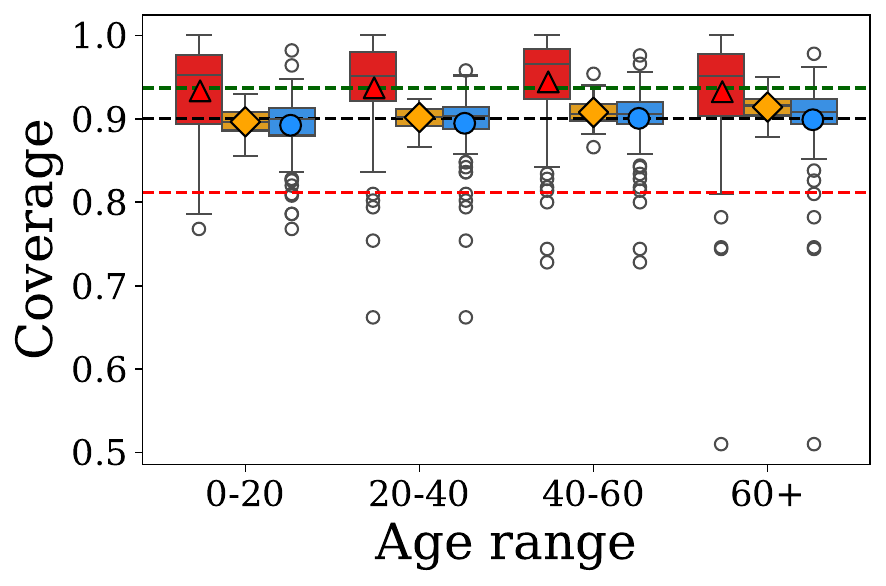}
    \includegraphics[height=0.2\textwidth, valign=t]{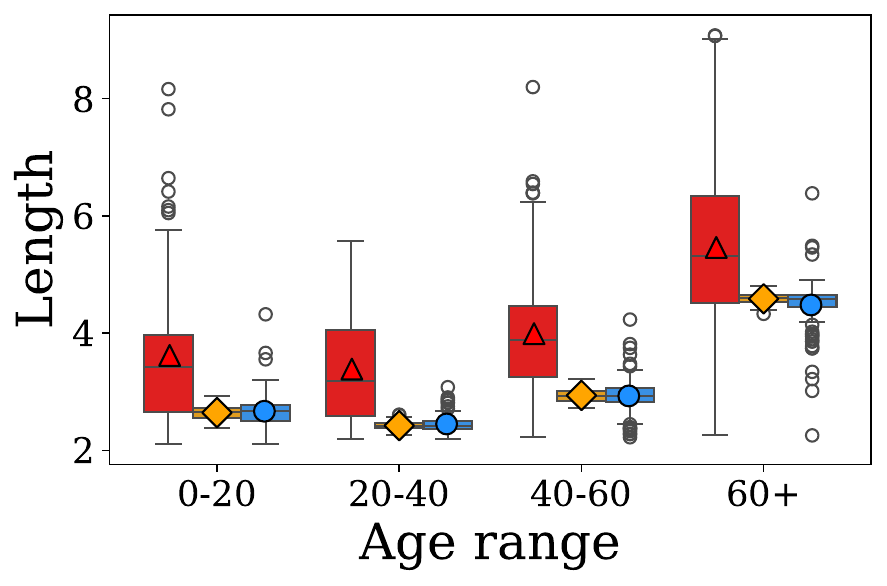}
    \includegraphics[height=0.22\textwidth, valign=t]{figures/results/legend_spi.pdf}
    \caption{\small Results for the MEPS dataset: Marginal coverage and interval length for each age-group, obtained by $\cp$, $\cpsynt$, and $\scp$. Target coverage is $1-\alpha=0.9$; experiments are repeated for 100 trials. 
    }
    \label{fig:meps}
\end{figure}

\Cref{fig:meps} presents the coverage and interval length results for $\cp$, $\cpsynt$, and $\scp$ at target coverage level $1-\alpha=0.9$, across different age-groups. Similarly to the classification experiments,  $\cp$ attains valid coverage but has a higher variance due to the small size of the real calibration set. 
Following that figure, we see that the synthetic and real data are well aligned. However, 
$\cpsynt$, which relies solely on synthetic data, lacks formal coverage guarantees. In contrast, $\scp$ achieves coverage close to the nominal level of 0.9, as predicted by~\Cref{thm:coverage_bound}.
We provide results for additional $\alpha$ levels and further experiments in~\Cref{app-sec:exp-meps}.

\section{Discussion}
\label{sec:dicsussion}
In this work, we presented a novel framework that enhances the sample efficiency of conformal prediction by leveraging synthetic data in a theoretically grounded manner. 
While we focused on marginal 
and label-conditional coverage, 
many applications require feature-conditional guarantees. Extending our approach to such settings---e.g., by drawing on ideas from \citet{gibbs2025conformal}---is an important direction for future work. 
Another limitation is the assumption that the real calibration data and the test point are i.i.d., which may not hold in practice. We believe our results can be extended beyond the i.i.d.~setting by building on techniques developed in \citep{tibshirani2019conformal,barber2023conformal,sesia2024adaptive,podkopaev2021distribution,joshi2025likelihood}. Lastly, our data-dependent $k$-nearest subset approach requires the size of the adapted synthetic dataset to be specified in advance. Developing theory that allows this size to vary and depend on the real data is of interest in future work.

\section*{Acknowledgments}
M.~B., R.~M-L., and Y.~R. were supported by the European Union (ERC, SafetyBounds, 101163414). Views and opinions expressed are however those of the authors only and do not necessarily reflect those of the European Union or the European Research Council Executive Agency. Neither the European Union nor the granting authority can be held responsible for them. This research was also partially supported by the Israel Science Foundation (ISF grant 729/21). 
E.~D. and Y.~L. were partially supported by the US NSF, NIH, ARO, AFOSR, ONR, and the Sloan Foundation.
Y.~R. acknowledges additional support from the Career Advancement Fellowship at the Technion, and is deeply grateful to Shai Feldman and Jeremias Sulam for their insightful discussions and valuable feedback.
\clearpage
\bibliography{bib}
\clearpage

\setcounter{figure}{0}
\renewcommand{\thefigure}{S\arabic{figure}}
\setcounter{table}{0}
\renewcommand{\thetable}{S\arabic{table}}
\appendix
\section{Algorithmic details}

\begin{algorithm}[!ht]
\caption{Synthetic-powered predictive inference ($\scp$)}
\label{algo:ot-calibration}
\begin{algorithmic}[1]
\STATE \textbf{Input:} Real calibration set $(X_i,Y_i)_{i\in [m]}$; synthetic calibration set $(\tilde{X}_i,\tilde{Y}_i)_{i\in [N]}$; test input $X_{m+1}$; score function $s$; target coverage level $1-\alpha$; parameter for window construction $\beta$.
\STATE Compute the real scores $S_i=s(X_i,Y_i)$, for $i\in [m]$.
\STATE Compute the synthetic scores $\tilde{S}_j=s(\tilde{X}_j,\tilde{Y}_j)$, for $j\in [N]$, and let $\tilde{Q}'_{1-\alpha} = \tilde{S}_{(\lceil(N+1)(1 - \alpha)\rceil+1)}$.
\STATE Compute $R_r^-$ and $R_r^+$ for $r\in [m+1]$, according to~\eqref{eqn:window_bounds}. 
\STATE Compute $\tilde{R}^-$ and $\tilde{R}^+$, according to \eqref{eqn:tilde-R}.
\STATE Compute the bound $Q = \max\{\min\{\tilde{Q}_{1-\alpha}', S_{(\tilde{R}^-)}\}, S_{(\tilde{R}_+)}\}$.
\STATE Compute $\ch(X_{m+1}) = \{y \in \Y : s(X_{m+1},y) \leq Q\}$.
\STATE \textbf{Output:} Prediction set ${\ch(X_{m+1})}$.

\end{algorithmic}
\end{algorithm}

\begin{algorithm}[!ht]
\caption{$\scp$ with data-dependent $k$-nearest subset selection}
\label{algo:ot-calibration-clustering}
\begin{algorithmic}[1]
\STATE \textbf{Input:} Real calibration set $(X_i,Y_i)_{i\in [m]}$; subsets of synthetic calibration set $(\tilde{X}_j^l,\tilde{Y}_j^l)_{j\in [n]}$, $l=1,2,\cdots,L$; test input $X_{m+1}$; score function $s$; target coverage level $1-\alpha$; parameter for window construction $\beta$; parameter for selection $k$.
\STATE Compute the real scores $S_i=s(X_i,Y_i)$, for $i\in [m]$.
\STATE Compute the synthetic scores $\tilde{S}_j^l=s(\tilde{X}_j^l,\tilde{Y}_j^l)$, for $j\in [N]$ and $l \in [L]$.
\FOR{$l$ in $[L]$}
\STATE $\textit{Distances} [l] \leftarrow \text{Cramer-von-Mises-Statistic}(\{S_i : i\in[m]\}, \{\tilde{S}_j^l: j\in[n]\})$ \{\Cref{algo:cramer-von-mises-test}\}
\ENDFOR
\STATE Let $\mathcal{L}$ be the set of $k$ subsets in $[L]$ with the smallest values in \textit{Distances}.
\STATE Apply \Cref{algo:ot-calibration} with $\{(\tilde{X}_j^l,\tilde{Y}_j^l) : j\in [n], l \in \mathcal{L}\}$ as the synthetic calibration data.

\STATE \textbf{Output:} Prediction set $\smash{\ch(X_{m+1})}$.

\end{algorithmic}
\end{algorithm}

\begin{algorithm}[!ht]
\caption{Cramer-von Mises two-sample test statistic}
\label{algo:cramer-von-mises-test}
\begin{algorithmic}[1]
\STATE \textbf{Input:} $(X_i)_{i\in [N]}$; $(Y_i)_{i\in [M]}$ (all distinct)
\STATE Let $W = \{X_1, \dots, X_N\} \cup \{Y_1, \dots, Y_M\}$ be the set of all datapoints.
\STATE Compute the ranks:\\
$r_i = \text{(the rank of $X_i$ in $W$)}$ for $i \in [N]$, and $s_i = \text{(the rank of $Y_i$ in $W$)}$ for $i \in [M]$.
\STATE Let $r_{(1)} < \cdots < r_{(N)}$ and $s_{(1)} < \cdots < s_{(M)}$ be the order statistics of $(r_i)_{i \in [N]}$ and $(s_i)_{i \in [M]}$, respectively.
\STATE Compute
\[U = N \sum_{i=1}^{N} (r_{(i)} - i)^2 + M \sum_{j=1}^{M} (s_{(j)} - j)^2.\]
\STATE Compute the Cramer–von Mises test statistic \( T \) as:
\[
T = \frac{U}{NM(N+M)} - \frac{4MN-1}{6(M+N)}.
\]
\STATE \textbf{Output:} \( T \).

\end{algorithmic}
\end{algorithm}
\clearpage

\begin{algorithm}[!ht]
\caption{$\beta$-selection}
\label{algo:beta-selection}
\begin{algorithmic}[1]
\STATE \textbf{Input:} Real calibration set size $m$; synthetic calibration set size $N$; target coverage level $1-\alpha$; desired worst-case lower bound $L$; step size $\epsilon$.
\STATE Set $\beta \gets \epsilon$.
\STATE Compute $R^+_{r}$ with $\beta$ for $r\in [m+1]$, according to~\eqref{eqn:window_bounds}.
\STATE Compute $\tilde{L} \gets |\{ i \in[m+1] : R_i^+ \le \left\lceil (1 - \alpha)(N + 1) \right\rceil \}|\ /\ (m+1)$.
\WHILE{$\tilde{L} < L$}
    \STATE $\beta += \epsilon$
    \STATE Compute $R^+_{r}$ with $\beta$ for $r\in [m+1]$, according to~\eqref{eqn:window_bounds}.
    \STATE Compute $\tilde{L} \gets |\{ i \in[m+1] : R_i^+ \le \left\lceil (1 - \alpha)(N + 1) \right\rceil \}|\ /\ (m+1)$.
\ENDWHILE
\STATE \textbf{Output:} \( \beta \).
\end{algorithmic}
\end{algorithm}

\section{Related work}\label{sec:literature}
\vspace{-1em}
The concept of prediction sets dates back to foundational works such as \citet{wilks1941determination}, \citet{wald1943extension}, \citet{scheffe1945non}, and \citet{tukey1947non,tukey1948nonparametric}. 
The initial ideas behind conformal prediction were introduced by \citet{saunders1999transduction} and \citet{vovk1999machine}. 
Since then, with the rise of machine learning, conformal prediction has emerged as a widely used framework for constructing distribution-free prediction sets \citep[e.g.,][]{papadopoulos2002inductive,vovk2005algorithmic,vovk2012conditional,Chernozhukov2018,dunn2022distribution,lei2013distribution,lei2014distribution,lei2015conformal,lei2018distribution,Sadinle2019,gibbs2021adaptive,park2021pac,angelopoulos2021gentle,guan2022prediction,guan2023localized,guan2024conformal,romano2020classification,bates2023testing,einbinder2022training,liang2022integrative,liang2023conformal}.

More recently, there has been growing interest in extending conformal prediction to offer more refined guarantees beyond standard marginal coverage. 
In particular, several works aim to offer approximate local coverage guarantees in the feature space \citep{guan2023localized,zhang2024posterior,hore2025conformal}; group-conditional coverage, which aims to guarantee valid coverage across pre-defined groups based on features and/or labels \citep{vovk2003mondrian,jung2022batch,gibbs2025conformal,bairaktari2025kandinsky}; and cluster-conditional coverage, which focuses on label-conditioned subgroups \citep{ding2023class}. However, these approaches still face the inherent limitations of conformal inference in settings where labeled data for the group-of-interest is limited, as previously discussed.

In contrast, we are interested in obtaining exact label- or group-of-interest conditional coverage guarantees even when the dataset from our distribution of interest is small. 
To this end, we take a different approach,
aiming to enhance sample efficiency by incorporating synthetic data.

A related line of work explores the use of unlabeled data to improve sample efficiency \citep{einbinder2024semi,angelopoulos2023prediction}. These methods assume that the unlabeled data is drawn from the same distribution as the labeled calibration set. 
In contrast, we consider settings where this assumption is violated and develop methods that remain valid under such unknown distributional shifts. Moreover, the above methods cannot be applied in the label-conditional setting, as they require knowing the labels of the unlabeled data.

Another related line of work is few-shot conformal prediction~\citep{fisch2021few,park2023few}, which addresses settings where only limited data is available for the target task, along with additional auxiliary tasks. These approaches leverage related but distinct tasks to improve sample efficiency. \citet{fisch2021few} provides asymptotic task-conditional coverage guarantees, whereas our focus is on finite-sample guarantees. \citet{park2023few} mitigate the small-sample challenge using cross-validation, but their methods remain constrained by the overall number of available datapoints---which we assume to be small in our setting.
\citet{dutta2024estimating} propose to retrieve web images to enable conformal prediction in zero-shot settings, by leveraging conformal prediction with ambiguous ground truth \citep{stutz2024conformal}, but do not provide coverage guarantees for their method.

\vspace{-1.5em}
\section{Technical background}
\vspace{-1em}
\subsection{Score functions}\label{app-sec:score-funcs}
\vspace{-1em}
\paragraph{Adaptive prediction sets  \citep{romano2020classification}} 
For classification tasks, we assume that the pre-trained model outputs an estimated probability vector $\hat{\pi}\in [0,1]^K$, where $K$ is the number of classes and each entry represents the estimated probability of the corresponding class.
We consider the APS score function that is defined for a given pair $(X,Y)$ as follows:
Let $\hat{\pi}_{(1)}(X) \ge \hat{\pi}_{(2)}(X) \ge \dots \ge \hat{\pi}_{(K)}(X)$ be the sorted values of the probability vector $\hat{\pi}(X)$, and let $r(Y,\hat{\pi}(X))$ denote the rank of the label $Y$ within this sorted vector. The nonconformity score is then given by:
\begin{align}
\label{eq:aps-score}
    s(X,Y) = \hat{\pi}_{(1)}(X) + \hat{\pi}_{(2)}(X) + \dots + \hat{\pi}_{(r(Y,\hat{\pi}(X)))}(X) - U\cdot \hat{\pi}_{(r(Y,\hat{\pi}(X)))}(X),
\end{align}
where $U$ is a uniform random variable on $[0,1]$, independent of everything else.

\vspace{-2em}
\paragraph{Conformalized quantile regression \citep{romano2019conformalized}} For the regression task, suppose we have a pre-trained quantile regression model that estimates the $\gamma$-th quantile of the distribution $Y\mid X$, denoted as $\hat{q}(X;\gamma)$. The conformalized quantile regression (CQR) score is then defined as 
\begin{align}
\label{eq:cqr-score}
    s(X,Y) = \max \{ \hat{q}(X;\alpha / 2) - Y, Y - \hat{q}(X;1 - \alpha / 2)\}.
\end{align}
Applying conformal prediction with this score, the prediction set takes the form
\begin{align*}
    \ch(X_{n+1}) = \left[\hat{q}(X_{n+1}; \alpha / 2) - \hat{Q}_{1-\alpha}, \hat{q}(X_{n+1}; 1-\alpha / 2) + \hat{Q}_{1-\alpha}\right].
\end{align*}

\section{Explaining the \emph{score transporter}}\label{sec:transporter}

In this section, we provide further intuition about 
the theoretical bounds established in~\Cref{thm:worst_case}.

We illustrate these bounds using the same schematic from \Cref{fig:score-transporter}. Real and synthetic nonconformity scores are shown as circles in sorted order, with each real score connected to a window in the synthetic score space (depicted as rectangles). The $(1 - \alpha)$th empirical quantile of the synthetic scores, $\tilde{Q}_{1 - \alpha}$, is outlined in black.

\begin{figure}[!h]
    \centering
    \begin{subfigure}[t]{0.45\linewidth}
            \centering
            \includegraphics[width=0.7\linewidth]{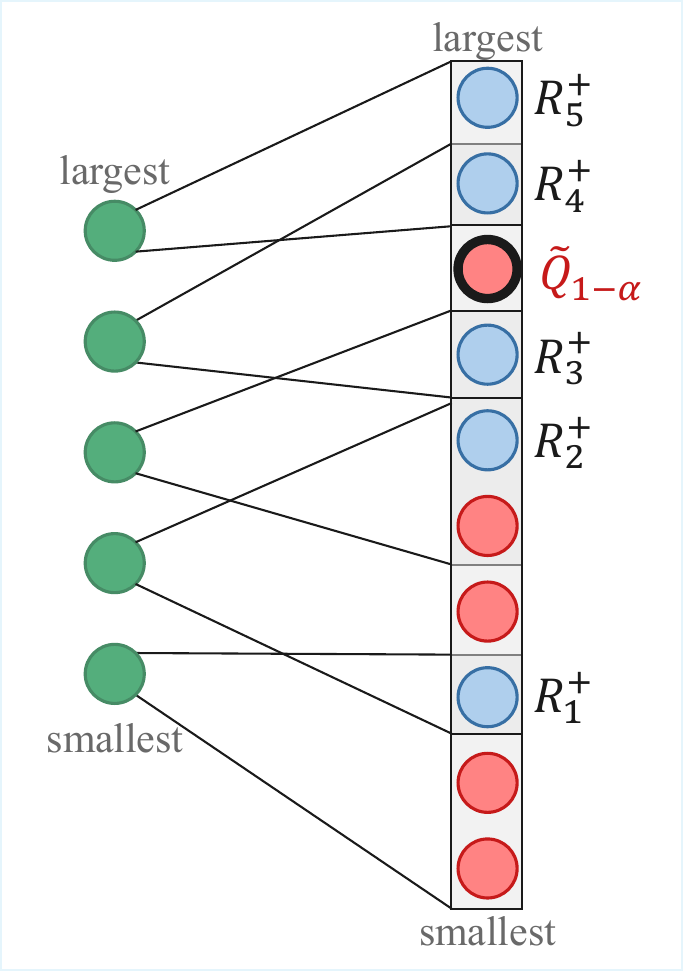}
            \caption{Values of $R^+_{r}$ (in blue), representing the upper endpoints of the synthetic windows. The smallest three real scores satisfy $R^+_{r} \leq \tilde{Q}_{1 - \alpha}$ and are thus guaranteed to be mapped to synthetic scores below the threshold $\tilde{Q}_{1-\alpha}$.}
    \label{app-fig:illus-wc-bounds-lower}
    \end{subfigure}
    \hfill
    \begin{subfigure}[t]{0.45\linewidth}
    \centering
    \includegraphics[width=0.7\linewidth]{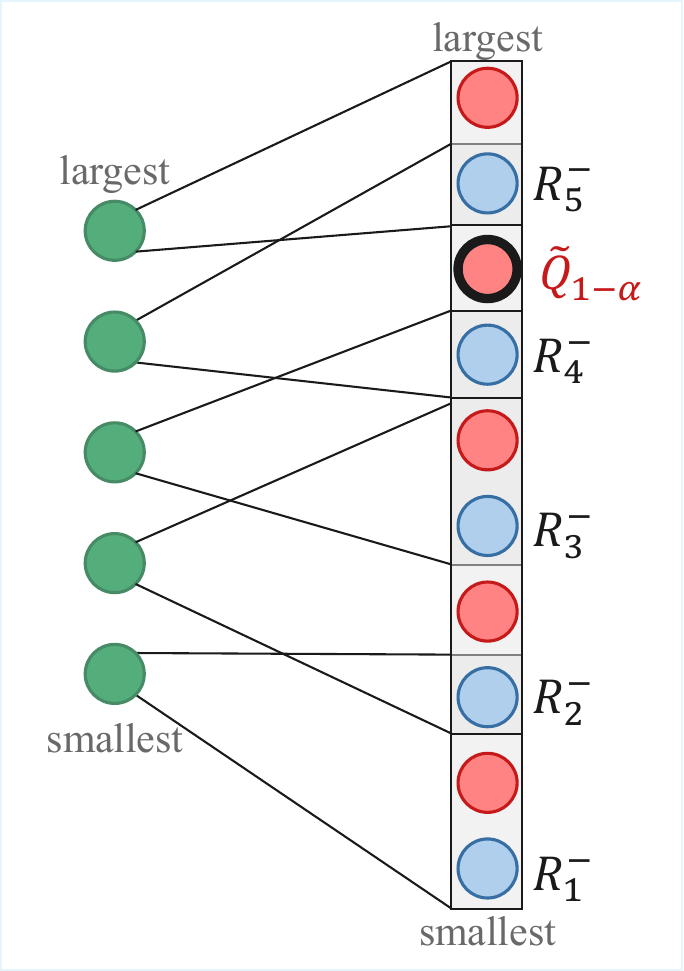}
            \caption{Values of $R^-_{r}$ (in blue), representing the lower endpoints of the synthetic windows. The fifth-ranked real score satisfies $R^-_{5} > \tilde{Q}_{1 - \alpha}$, meaning it is necessarily mapped above the threshold $\tilde{Q}_{1-\alpha}$.}
    \label{app-fig:illus-wc-bounds-upper}
    \end{subfigure}
    \caption{\textbf{Illustration of the quantities used in the worst-case coverage bounds from \Cref{thm:worst_case}}.
Real and synthetic nonconformity scores are shown as circles in sorted order. Each real score is connected to a window in the synthetic score space (depicted as rectangles). The synthetic $(1 - \alpha)$th empirical quantile $\tilde{Q}_{1 - \alpha}$ is marked in blue.
}
    \label{app-fig:illus-wc-bounds}
\end{figure}

\Cref{app-fig:illus-wc-bounds} visualizes the quantities used to derive the worst-case coverage bounds. For each real score $S_{(r)}$ for $r\in[m+1]$, we denote the endpoints of its associated window by $R^-_{r}$ and $R^+_{r}$, as introduced in~\eqref{eqn:window_bounds}. These correspond to the smallest and largest ranks, respectively, of the synthetic scores that $S_{(r)}$ can be mapped to. For convenience, we refer to $R^-_{r}$ and $R^+_{r}$ as the synthetic scores at those ranks.

\Cref{app-fig:illus-wc-bounds-lower} shows the values $R^+_{r}$ (in blue), which are used to compute the lower bound. Since the transported score $T(S_{(r)})$---defined as the nearest synthetic score within the window among those that are smaller than $S_{(r)}$---is always less than or equal to $R^+_{r}$, any real score satisfying $R^+_{r} \leq \tilde{Q}_{1-\alpha}$ must necessarily satisfy $T(S_{(r)}) \leq \tilde{Q}_{1 - \alpha}$. Consequently, the corresponding label will always be included in the prediction set. 

\Cref{app-fig:illus-wc-bounds-upper} shows the values $R^-_{r}$ (in blue), which are used to compute the upper bound. Real scores for which $R^-_{r} \leq \tilde{Q}_{1 - \alpha}$ may be mapped to a synthetic score below the threshold and thus may be included in the prediction set---for example, the bottom four real scores in the figure. In contrast, if $R^-_{r} > \tilde{Q}_{1 - \alpha}$ (as for the fifth-ranked score), then the transported score must exceed the threshold, and the corresponding label is guaranteed to be excluded.

By exchangeability, the test score is equally likely to take any of the $m+1$ possible ranks among the real calibration scores. Therefore, the coverage probability is bounded between the fraction of real scores whose $R^+_{r} \leq \tilde{Q}_{1 - \alpha}$ and the fraction whose $R^-_{r} \leq \tilde{Q}_{1 - \alpha}$, as formalized in~\Cref{thm:worst_case}. In our example, these correspond to $3/5$ and $4/5$, respectively.

\FloatBarrier

\subsection{Coverage guarantee bounds}\label{sec:bound_plots}

\vspace{-0.5em}
To illustrate the distribution-free bounds in~\Cref{thm:worst_case}, we present several visualizations. These bounds are determined solely by the sample sizes $m$ and $N$, the parameter $\beta$, and the target coverage level $1-\alpha$.

\begin{figure}[!h]
    \centering
    \begin{subfigure}[t]{0.32\linewidth}
    \includegraphics[width=\linewidth]{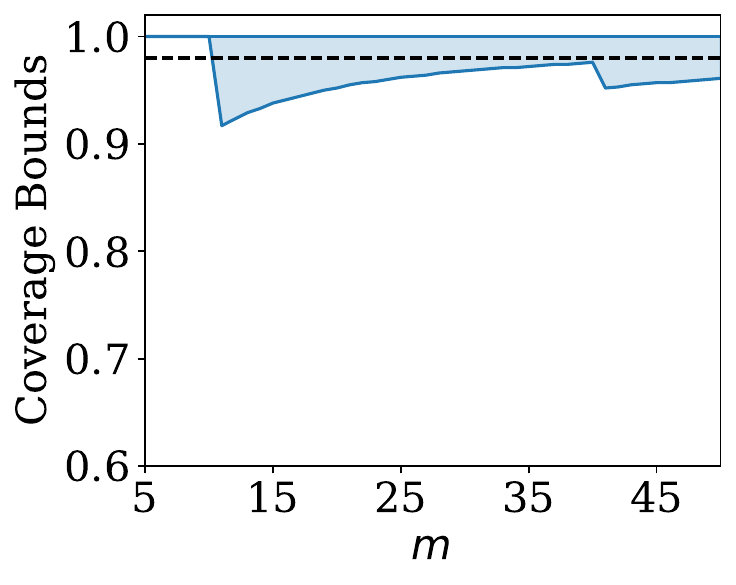}
    \caption{$\alpha=0.02$}
    \end{subfigure}
    \begin{subfigure}[t]{0.32\linewidth}
    \includegraphics[width=\linewidth]{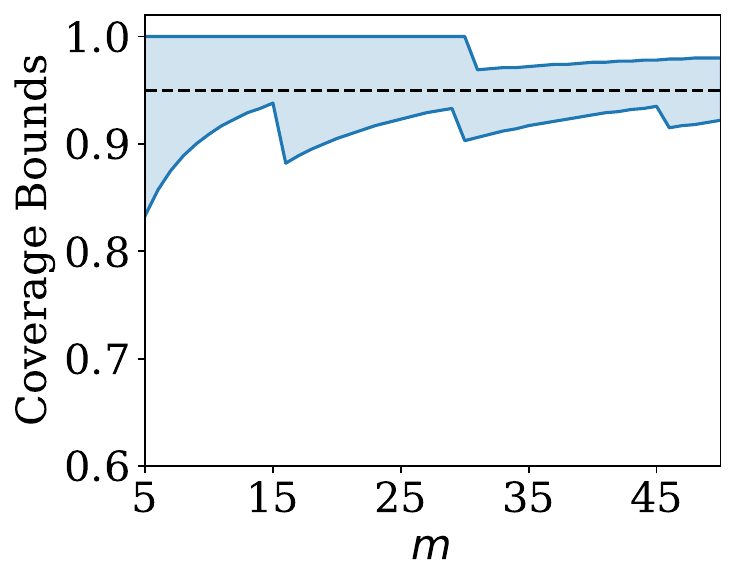}
    \caption{$\alpha=0.05$}
    \end{subfigure}
    \begin{subfigure}[t]{0.32\linewidth}
    \includegraphics[width=\linewidth]{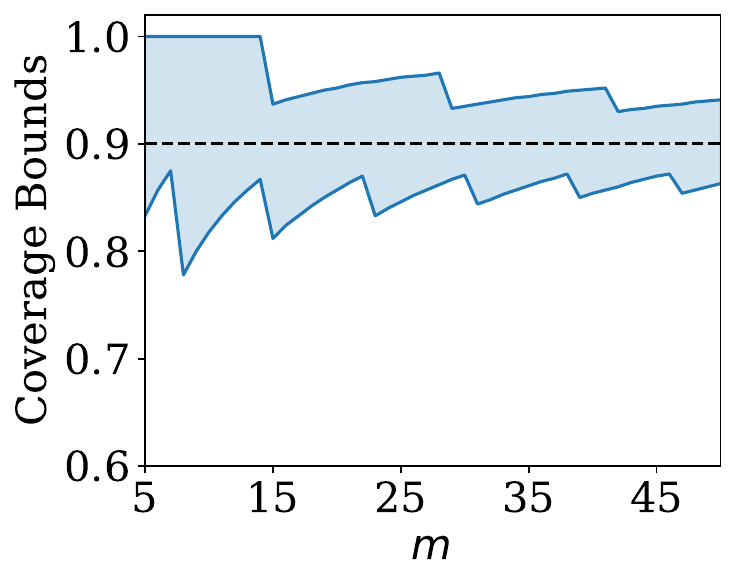}
    \caption{$\alpha=0.1$}
    \end{subfigure}
    \caption{Illustration of the coverage bounds in~\Cref{thm:worst_case} as a function of the real calibration set size $m$. The synthetic calibration size is $N=1000$, and we set $\beta = 0.4$. Results are presented for $\alpha = 0.02$ (a), $0.05$ (b), and $0.1$ (c).
    The shaded regions represent the area between the lower and upper bounds for each $\alpha$ level, with the dashed black lines indicating the target coverage level $1 - \alpha$.}
    \label{app-fig:bounds-m}
\end{figure}

\Cref{app-fig:bounds-m} presents the upper and lower bounds in~\Cref{thm:worst_case} as functions of the calibration set size $m$ and the level $\alpha$, with fixed $N = 1000$ and $\beta = 0.4$. As $m$ increases, the bounds become tighter around the target level $1 - \alpha$.

\vspace{-0.5em}

\begin{figure}[!h]
    \centering
    \begin{subfigure}[t]{0.32\linewidth}
    \includegraphics[width=\linewidth]{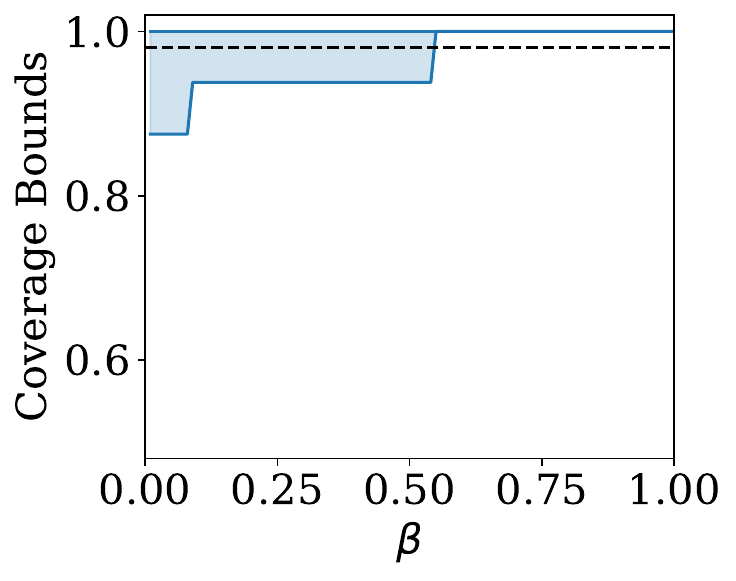}
    \caption{$\alpha=0.02$}
    \end{subfigure}
    \begin{subfigure}[t]{0.32\linewidth}
    \includegraphics[width=\linewidth]{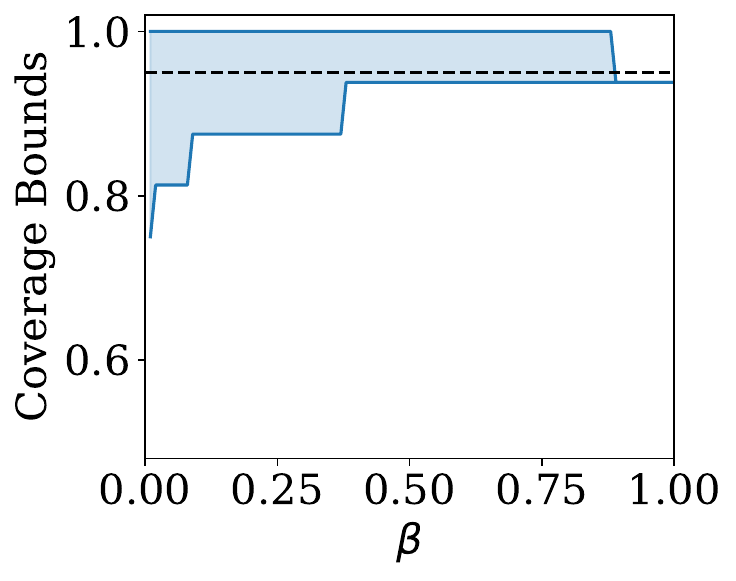}
    \caption{$\alpha=0.05$}
    \end{subfigure}
    \begin{subfigure}[t]{0.32\linewidth}
    \includegraphics[width=\linewidth]{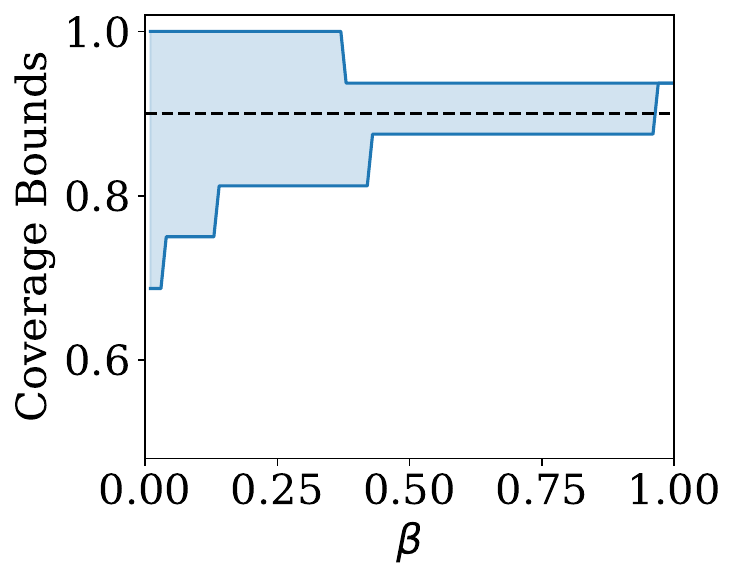}
    \caption{$\alpha=0.1$}
    \end{subfigure}
    \caption{Illustration of the coverage guarantee bounds from \Cref{thm:worst_case} as a function of $\beta$.
    The real calibration set contains $m = 15$ datapoints. Other details are as in~\Cref{app-fig:bounds-m}.}
    \label{app-fig:bounds-beta}
\end{figure}

\vspace{-0.5em}
Next, \Cref{app-fig:bounds-beta} illustrates how the bounds vary with the parameter $\beta$---under $m = 15$, $N = 1000$, and different values of $\alpha$. As $\beta$ decreases, the bounds become looser. This trend can be explained as follows: by the construction of the windows in~\eqref{eqn:window_bounds}, smaller values of $\beta$ lead to wider windows. 
As a result, fewer $R^+_{r}$ values (for $r\in[m+1]$) fall below the $(1-\alpha)$th empirical quantile, loosening the lower bound. 
At the same time, more $R^-_{r}$ values fall below this quantile, resulting in a looser upper bound. This trend is consistent across various $\alpha$, as shown in the figure. 
Further, the bounds exhibit a stepwise pattern due to their discrete nature---their values change in increments of $1/(m+1)$.

In practice, one may be interested in using our method while ensuring that the lower bound guaranteed by~\Cref{thm:worst_case} is no smaller than a user-specified level $L$. To this end, we provide an algorithm (see Algorithm~\ref{algo:beta-selection}) for selecting $\beta$ based on the sample sizes $m$ and $N$, the target miscoverage level $\alpha$, step size $\epsilon$ (e.g., 0.01), and the desired lower bound $L$. As shown in~\Cref{app-fig:bounds-beta}, multiple $\beta$ values may yield the same lower bound. In such cases, the algorithm selects the smallest $\beta$ that results in a lower bound greater than or equal to $L$, inspired by the result in~\Cref{thm:coverage_bound}.
\FloatBarrier

\section{Constructing a separate synthetic score function with data splitting}\label{sec:score_data_split}

In this section, we provide details
constructing the synthetic
score function independently of the calibration data, adding to the discussion in Section~\ref{sec:score_trans}.
For instance, suppose we apply data splitting
(to both the real and the synthetic data) 
and use one split as training data to construct the synthetic score function $\tilde{s}$. If the data has already been split to construct $s$, the same split can be used for constructing $\tilde{s}$. 
Then we use both real and synthetic (training) data to construct $\tilde{s}$, 
to ensure that its distribution 
better approximates that of the real score. 
Throughout this section, 
we condition everything on the training datasets.

We begin with the method of constructing an adjustment function $g$ and using the transformed score function $\tilde{s} = g \circ s$ for the synthetic data. In this case, the prediction set $\ch^g(X_{m+1})$, constructed according to~\eqref{eqn:pred_set}, is given by
\begin{equation*}
    \ch^g(X_{m+1}) = \{y \in \Y : T(s(X_{m+1},y) ; (S_i)_{i \in [m]}, (g(\tilde{S}_j))_{j \in N}) \le \tilde{Q}_{1-\alpha}^g\},
\end{equation*}
where $\tilde{Q}_{1-\alpha}^g$ denotes the $\lceil(N+1)(1-\alpha)\rceil$-th smallest value among $\{g(\tilde{S}_j) : j \in [N]\}$.

\begin{example}\label{ex:tr}
One option is to construct an affine transformation $g(s) = \theta_1 s + \theta_2$ to adjust the scale and bias of $s(\tilde{X},\tilde{Y})$. 
Denote the training sets by $(X_1', Y_1'), \dots, (X_{m_\text{train}}', Y_{m_\text{train}}')$ and $(\tilde{X}_1', \tilde{Y}_1'), \dots, (\tilde{X}_{N_\text{train}}', \tilde{Y}_{N_\text{train}}')$, 
and denote the corresponding real and synthetic (training) scores by $(S_i')_{i \in [m_\text{train}]}$ and $(\tilde{S}_j')_{j \in [N_\text{train}]}$, respectively. 
Then, we can set $\theta_1$ and $\theta_2$ via least squares: 
\[(\theta_1, \theta_2) = \argmin_{a,b} \sum_{i=1}^{m_\text{train}} \left|a \cdot \tilde{S}_{\left(\lfloor i N_\text{train}/{m_\text{train}} \rfloor \right)}' + b - S_{(i)}'\right|^2,\]
where $S_{(i)}'$ and $\tilde{S}_{(j)}'$ denote the order statistics of the real and synthetic training scores, respectively.
\end{example}

More generally, suppose we construct a new score function $\tilde{s}$ using the training data, such that the distribution of $\tilde{s}(\tilde{X}, \tilde{Y})$ approximates that of $s(X, Y)$, where $(\tilde{X}, \tilde{Y}) \sim Q_{X,Y}$ and $(X, Y) \sim P$. 
The prediction set is then constructed according to~\eqref{eqn:pred_set}, using the synthetic scores $\tilde{S}_j = \tilde{s}(\tilde{X}_j, \tilde{Y}_j)$:
\begin{equation}\label{eqn:pred_set_2}
    \ch(X_{m+1}) = \{y \in \Y : T(s(X_{m+1},y) ; (S_i)_{i \in [m]}, (\tilde{S}_j)_{j \in N}) \le \tilde{Q}_{1-\alpha}\}.
\end{equation}

Denoting the distribution of $\tilde{s}(\tilde{X}, \tilde{Y})$ as $\tilde{Q}$, we have the following result as a direct consequence of Theorem~\ref{thm:coverage_bound} and~\ref{thm:worst_case}\footnote{The proofs of these theorems build upon the setting $S_1, \dots, S_{m+1} \iidsim P$ and $\tilde{S}_1, \dots, \tilde{S}_N \iidsim Q$, and do not depend directly on the datasets or the score function. Therefore, the results in Corollary~\ref{cor:transformation} follow directly by applying the same arguments with $\tilde{S}_1, \dots, \tilde{S}_N \iidsim \tilde{Q}$.}.

\begin{corollary}\label{cor:transformation}
    Suppose the distribution $\tilde{Q}$ is continuous. Then the prediction set $\smash{\ch(X_{m+1})}$ from~\eqref{eqn:pred_set_2}, constructed using $\tilde{S}_j = \tilde{s}(\tilde{X}_j, \tilde{Y}_j)$ for $j \in [N]$, satisfies
    \[1-\alpha-\beta - \ep_{P,\tilde{Q}}^{m+1} \le \PP{Y_{m+1} \in \ch(X_{m+1})} \le 1-\alpha+\beta + \ep_{P,\tilde{Q}}^{m+1} + 1/(N+1).\]
    Moreover, the bounds stated in Theorem~\ref{thm:worst_case} also hold for $\smash{\ch(X_{m+1})}$ from~\eqref{eqn:pred_set_2}.
\end{corollary}

\section{Predictive inference with label-conditional coverage control}
\label{app-sec:label-cond-calib}
Here, we review the standard approach \citep{vovk2005algorithmic} for achieving the label-conditional coverage guarantee~\eqref{eq:coverage_class}, and then discuss a variant of this approach based on $\scp$.

The basic idea is to partition the calibration set by classes,
run conformal prediction within each class, and then combine the results to construct a prediction set. Specifically, the prediction set is constructed as follows:
\begin{align*}
    \ch(X_{m+1})=\left\{y\in \Y : s(X_{m+1}, y) \leq Q_{1-\alpha}^y\right\},
\end{align*}
where $Q_{1-\alpha}^y$ denotes the $\lceil(1-\alpha)(n_y + 1)\rceil$-th smallest element among $\{S_i : i \in [m], Y_i = y\}$, and $n_y$ denotes the number of calibration points labeled with $y$.

Now, we introduce the $\scp$-based method that ensures the label-conditional coverage guarantee. The idea follows the same logic as the standard method: We run $\scp$ within each class-specific partition of the real and synthetic calibration sets and then combine the results. 
For a class that does not appear in the synthetic dataset, 
we run the procedure with the entire synthetic dataset.

To formalize this idea, define 
\[\Y_1 = \{y \in \Y : \tilde{Y}_j = y \text{ for some } j \in [N]\} \text{ and } \Y_0 = \{y \in \Y : \tilde{Y}_j \neq y \text{ for all } j \in [N]\}.\]
Let $I_y = \{i \in [m] : Y_i = y\}$ for each $y \in \Y$, and $J_y = \{j \in [N] : \tilde{Y}_j = y\}$ for each $y \in \Y_1$. Then for each $y \in \Y_1$, we define the function $T^y(\cdot) = T(\cdot ; (S_i)_{i \in I_y}, (\tilde{S}_j)_{j \in J_y})$, following the definition in~\eqref{eqn:score_transporter}. For $y \in \Y_0$, we let $T^y(\cdot) = T(\cdot ; (S_i)_{i \in I_y}, (\tilde{S}_j)_{j \in [N]})$. Then we define $\tilde{Q}_{1-\alpha}^y$ for each $y$ as follows:

\begin{align*}
    \tilde{Q}_{1-\alpha}^y := 
    \begin{cases}
        \lceil (1-\alpha)(|J_y|+1)\rceil\text{-th smallest element in } \{\tilde{S}_j : j \in J_y\},\qquad &\text{ if } y \in \Y_1,\\
        \lceil (1-\alpha)(N+1)\rceil\text{-th smallest element in } \{\tilde{S}_j : j \in [N]\},\qquad &\text{ if } y \in \Y_0.
    \end{cases}
\end{align*}

Then we construct the prediction set as
\begin{equation}\label{eqn:scp_class}
    \ch(X_{m+1}) = \left\{y\in \Y : T^y(s(X_{m+1}, y)) \leq \tilde{Q}_{1-\alpha}^y\right\}.
\end{equation}

As a direct consequence of Theorem~\ref{thm:worst_case}, the prediction set~\eqref{eqn:scp_class} attains the following label-conditional coverage control: 
\begin{multline*}
\frac{|\{ j \in[m+1] : R_j^+ \le \left\lceil (1 - \alpha)(N_y + 1) \right\rceil \}|}{m+1} \le \PPst{Y_{m+1} \in \ch(X_{m+1})}{Y_{m+1} = y}\\
\le \frac{|\{ j \in[m+1] : R_j^- \le \left\lceil (1 - \alpha)(N_y + 1) \right\rceil \}|}{m+1}, \qquad \text{ for all $y \in \Y$},
\end{multline*} 
where we let $N_y = |J_y|$ for $y \in \Y_1$ and $N_y = N$ for $y \in \Y_0$.

\section{Mathematical proofs}\label{app-sec:proofs}

\subsection{Proof of Lemma~\ref{lem:window}}
The result follows directly from the work of~\citet{lee2024batch}, but we provide the proof for completeness. Define
\[R_r = \min\{\tau \in [N+1] : \tilde{S}_{(\tau)} \ge S_{(r)}\}\]
for each $r \in [m+1]$, where we let $R_{r} = N+1$ if $S_{(r)} \ge \tilde{S}_{(N)}$. Note that $R_r$ is random, whereas $R_r^-$ and $R_r^+$ are not. Then by the exchangeability of $(S_r)_{r \in [m+1]}$ and $(\tilde{S}_j)_{j \in [N]}$, the distribution of the vector $(R_1,R_2,\cdots,R_{m+1})$ is given by
\[(R_1,R_2,\cdots,R_{m+1}) \sim \textnormal{Unif}\big(\{(\zeta_1,\cdots,\zeta_{m+1}) : 1 \le \zeta_1 \le \cdots \le \zeta_{m+1} \le N+1\}\big).\]
Therefore, for each $k \in [N+1]$, we have
\begin{align*}
    &\PP{R_r = k} = \frac{|\{(\zeta_1,\cdots,\zeta_{m+1}) : 1 \le \zeta_1 \le \cdots \le \zeta_{m+1} \le N+1 \text{ and } \zeta_r = k\}|}{|\{(\zeta_1,\cdots,\zeta_{m+1}) : 1 \le \zeta_1 \le \cdots \le \zeta_{m+1} \le N+1\}|}\\
    &= \frac{|\{\zeta_{1:(r-1)} : 1 \le \zeta_1 \le \cdots \le \zeta_{r-1} \le k\}| \cdot |\{ \zeta_{(r+1):(m+1)} : k \le \zeta_{r+1} \le \cdots \le \zeta_{m+1} \le N+1 \}|}{|\{(\zeta_1,\cdots,\zeta_{m+1}) : 1 \le \zeta_1 \le \cdots \le \zeta_{m+1} \le N+1\}|}\\
    &= \frac{_{k}\mathrm{H}_{r-1} \cdot _{N-k+2}\mathrm{H}_{m-r+1}}{_{N+1}\mathrm{H}_{m+1}} = \frac{\binom{k+r-2}{r-1}\cdot\binom{N+m-k-r+2}{m-r+1}}{\binom{N+m+1}{m+1}},
\end{align*} 
where we use the notation $_{n}\mathrm{H}_r$ to denote the number of ways to select $r$ items with replacement from $n$ items. Therefore,
\begin{multline*}
    \PP{S_{(r)} \in I_m(r)} = \PP{\tilde{S}_{(R_r^-)} \le S_{(r)} \le \tilde{S}_{(R_r^+)}} = \PP{\tilde{S}_{(R_r^-)} \le \tilde{S}_{(R_r)} \le \tilde{S}_{(R_r^+)}}\\
    = \PP{R_r^- \le R_r \le R_r^+} = F(R_r^+) - F(R_r^--1) \geq 1-\beta,
\end{multline*}
where the inequality follows from the definition of $R_r^-$ and $R_r^+$.

\subsection{Proof of Theorem~\ref{thm:coverage_bound}}

Let $r_{m+1} = \sum_{i=1}^m \One{S_i < S_{m+1}} + 1$ denote the rank of $S_{m+1}$ in the increasing order among ${S_1, \dots, S_m, S_{m+1}}$. Observe that $T(S_{m+1}) \le S_{m+1}$ holds if 
$L_{m}(r_{m+1}) \le S_{m+1}$, by the construction of the mapping $T$. 
Therefore, writing $L_r = L_{m}(r)$ and $U_r = U_{m}(r)$ for simplicity, we have
\begin{align*}
&\PP{Y_{m+1} \in \ch(X_{m+1})} = \PP{T(S_{m+1}) \le \tilde{Q}_{1 - \alpha}}\\
&= \PP{T(S_{m+1}) \le \tilde{Q}_{1 - \alpha}, S_{m+1} \in [L_{r_{m+1}}, U_{r_{m+1}}]} \\
&\hspace{60mm}
+ \PP{T(S_{m+1}) \le \tilde{Q}_{1 - \alpha}, S_{m+1} \notin [L_{r_{m+1}}, U_{r_{m+1}}]} \\
&\ge \PP{S_{m+1} \le \tilde{Q}_{1 - \alpha}, S_{m+1} \in [L_{r_{m+1}}, U_{r_{m+1}}]}.
\end{align*}
We can condition on $r_{m+1}$ to write that this equals
\begin{align*}
\EE{\PPst{S_{m+1} \le \tilde{Q}_{1 - \alpha}, S_{m+1} \in [L_{r_{m+1}}, U_{r_{m+1}}]}{r_{m+1}}}.
\end{align*}
Further, since $r_{m+1} \sim \text{Unif}({1,2,\dots,m+1})$ by the exchangeability of $(S_i)_{i \in [m+1]}$, and since $r_{m+1}$ is independent of the order statistics $S_{(1)}, \cdots, S_{(m+1)}$, the expression further simplifies to
\begin{multline*}
\frac{1}{m+1}\sum_{i=1}^{m+1} \PPst{S_{(r)} \le \tilde{Q}_{1 - \alpha}, L_r \le S_{(r)} \le U_r}{r_{m+1}=r}\\
=\frac{1}{m+1}\sum_{i=1}^{m+1} \PP{S_{(r)} \le \tilde{Q}_{1 - \alpha}, L_r \le S_{(r)} \le U_r}.
\end{multline*}

Now we fix $r \in [m+1]$ and examine the probability in the summation. The event inside the probability is a function of $S_{(r)} \sim P_{(r)}^{m+1}$ and $\tilde{S}_1, \cdots, \tilde{S}_N \iidsim Q$. Thus, we have 
\begin{align*}
    &\Pp{S_{(r)} \sim P_{(r)}^{m+1}, \tilde{S}_{1:N} \sim Q^{N}}{S_{(r)} \le \tilde{Q}_{1 - \alpha}, L_r \le S_{(r)} \le U_r}\\
    &\ge \Pp{S_{(r)} \sim Q_{(r)}^{m+1}, \tilde{S}_{1:N} \sim Q^{N}}{S_{(r)} \le \tilde{Q}_{1 - \alpha}, L_r \le S_{(r)} \le U_r} \\    
    &\hspace{75mm}- \dtv(P_{(r)}^{m+1} \times Q^{N}, Q_{(r)}^{m+1} \times Q^{N})\\
    &=  \Pp{S_{(r)} \sim Q_{(r)}^{m+1}, \tilde{S}_{1:N} \sim Q^{N}}{S_{(r)} \le \tilde{Q}_{1 - \alpha}, L_r \le S_{(r)} \le U_r} - \dtv(P_{(r)}^{m+1}, Q_{(r)}^{m+1}).
\end{align*}
Therefore, putting everything together, we have
\begin{align*}
&\PP{Y_{m+1} \in \ch(X_{m+1})}\\
&\ge \frac{1}{m+1}\sum_{i=1}^{m+1}\Pp{S_{(r)} \sim Q_{(r)}^{m+1}, \tilde{S}_{1:N} \sim Q^{N}}{S_{(r)} \le \tilde{Q}_{1 - \alpha}, L_r \le S_{(r)} \le U_r}
 \\    
&\hspace{75mm}
- \frac{1}{m+1}\sum_{i=1}^{m+1}\dtv(P_{(r)}^{m+1}, Q_{(r)}^{m+1})\\
&=\Pp{S_{(r)} \sim Q_{(r)}^{m+1}, \tilde{S}_{1:N} \sim Q^{N}}{S_{m+1} \le \tilde{Q}_{1 - \alpha}, S_{m+1} \in [L_{r_{m+1}}, U_{r_{m+1}}]} - \ep_{P,Q}^{m+1}.
\end{align*}
The probability in the last term is equivalently taken with respect to $S_1,\cdots, S_{m+1}, \tilde{S}_1, \cdots, \tilde{S}_N \iidsim Q$, and thus we have
\[\Pp{S_{(r)} \sim Q_{(r)}^{m+1}, \tilde{S}_{1:N} \sim Q^{N}}{S_{m+1} \le \tilde{Q}_{1 - \alpha}} \ge 1-\alpha,\]
by the standard conformal prediction coverage guarantee~\eqref{eqn:bound_split}, and
\[\Pp{S_{(r)} \sim Q_{(r)}^{m+1}, \tilde{S}_{1:N} \sim Q^{N}}{S_{m+1} \in [L_{r_{m+1}}, U_{r_{m+1}}]} = \frac{1}{m+1}\sum_{i=1}^{m+1} \PP{S_r \in [L_r,U_r]} \ge 1-\beta,\]
by Lemma~\ref{lem:window}. Therefore, by the union bound, we have
\[\PP{Y_{m+1} \in \ch(X_{m+1})} \ge 1-\alpha-\beta-\ep_{P,Q}^{m+1}.\]
Next, defining $\tilde{Q}_{1 - \alpha}'$ as in Section \ref{simp}, 
the events $S_{m+1} \in [L_{r_{m+1}}, U_{r_{m+1}}]$ and $S_{m+1} \le \tilde{Q}_{1 - \alpha}'$ 
together imply $T(S_{m+1}) \le \tilde{Q}_{1 - \alpha}$, by the construction of $T$, and thus we have
\begin{align*}
\PP{T(S_{m+1}) \le \tilde{Q}_{1 - \alpha}}  \le \PP{S_{m+1} \le \tilde{Q}_{1 - \alpha}' \text{ or } S_{m+1} \notin [L_{r_{m+1}}, U_{r_{m+1}}]}.
\end{align*}
Therefore, applying arguments analogous to the ones above, we have
\begin{align*}
    &\PP{Y_{m+1} \in \ch(X_{m+1})}\\
    &\le \Pp{S_{(r)} \sim Q_{(r)}^{m+1}, \tilde{S}_{1:N} \sim Q^{N}}{S_{m+1} \le \tilde{Q}_{1 - \alpha}' \text{ or } S_{m+1} \notin [L_{r_{m+1}}, U_{r_{m+1}}]} + \ep_{P,Q}^{m+1}\\
    &\le 1-\alpha+\beta+ \ep_{P,Q}^{m+1}+\frac{1}{N+1},
\end{align*}
since $\PP{S_{m+1} \le \tilde{Q}_{1 - \alpha}'} \le 1-\alpha+\frac{1}{N+1}$ under exchangeability due to the standard conformal prediction coverage guarantee~\eqref{eqn:bound_split}.

\subsection{Proof of Theorem~\ref{thm:worst_case}}

Let us define $r_{m+1}$ as in the proof of Theorem~\ref{thm:coverage_bound}. By the continuity assumption on $Q$, the synthetic scores are almost surely all distinct, and their order is well-defined. Now observe the deterministic relation

\begin{equation}\label{eqn:sub}
\begin{split}
    &\{R_{r_{m+1}}^+ \le \lceil {(1 - \alpha)(N+1)}\rceil\} = \{U_{m}(r_{m+1}) \le \tilde{Q}_{1-\alpha}\} \subset \{Y_{m+1} \in \ch(X_{m+1})\}\\
    &\hspace{40mm}\subset \{L_{m}(r_{m+1}) \le \tilde{Q}_{1-\alpha}\} = \{R_{r_{m+1}}^- \le \lceil {(1 - \alpha)(N+1)}\rceil\},
\end{split}
\end{equation}
which holds by the construction of $\smash{\ch(X_{m+1})}$ and the definition of the interval $[L_{m}(r_{m+1}), U_{m}(r_{m+1})]$. Therefore, the desired inequalities directly follow from the fact that $r_{m+1} \sim \text{Unif}([m+1])$ due to the exchangeability of the scores $(S_i)_{i \in [m+1]}$.

\subsection{Proof of Proposition~\ref{prop:pred_set}}
We show that for any $x \in \mathcal{X}$, the following relation holds:
\[\ch(x) \,\triangle\, \ch^{\mathrm{fast}}(x) \subset \{y \in \Y : s(x,y) \in \{\tilde{S}_j : j \in [N]\}\}.\]
The claim then follows directly from the continuity of $Q$.
    
    Fix any $x\in\mathcal{X}$. It is sufficient to prove that for any $y$ in the set  $\Lambda := \{y' : s(x,y') \notin \{\tilde{S}_j : j \in [N]\}\}$, we have 
    $y \in \ch(x)$ if and only if $y \in \ch^{\mathrm{fast}}(x)$ holds.
    
    Let us first take any $y \in \ch^{\mathrm{fast}}(x) \cap \Lambda$, and define $r_{m+1}^{(x,y)} = \sum_{i=1}^m \One{S_i < s(x,y)} + 1$. Then we have the following:
    \begin{align*}
        &\left\{y \in \ch^{\mathrm{fast}}(x)\right\}
        = \left(\left\{s(x,y) \le \tilde{Q}_{1-\alpha}'\right\} \cap \left\{s(x,y) \le S_{(\tilde{R}^-)}\right\}\right) \cup \left\{s(x,y) \le S_{(\tilde{R}^+)}\right\}\\
        &= \left(\left\{s(x,y) \le \tilde{Q}_{1-\alpha}'\right\} \cap \left\{r_{m+1}^{(x,y)} \le \tilde{R}^-\right\}\right) \cup \left\{r_{m+1}^{(x,y)} \le \tilde{R}^+\right\} \quad \text{ since $y \in \Lambda$}\\
        &= \left(\left\{s(x,y) \le \tilde{Q}_{1-\alpha}'\right\} \cap \left\{L_{m}(r_{m+1}^{(x,y)}) \le \tilde{Q}_{1-\alpha}\right\}\right) \cup \left\{U_{m}(r_{m+1}^{(x,y)}) \le \tilde{Q}_{1-\alpha}\right\},
    \end{align*}
and the final set can be expressed as a disjoint union of two events:
\[\text{(i) } s(x,y) \le \tilde{Q}_{1-\alpha}' \text{ and } L_{m}(r_{m+1}^{(x,y)}) \le \tilde{Q}_{1-\alpha} < U_{m}(r_{m+1}^{(x,y)}), \quad \text{(ii) } U_{m}(r_{m+1}^{(x,y)}) \le \tilde{Q}_{1-\alpha}.\]
Note that in the case (ii), $T(s(x,y)) \le \tilde{Q}_{1-\alpha}$ directly follows, since $T(s(x,y)) \le U_{m}(r_{m+1}^{(x,y)})$ holds deterministically. In the case (i), we have $s(x,y) \le \tilde{Q}_{1-\alpha}' \le U_{m}(r_{m+1}^{(x,y)})$, and thus $T(s(x,y))$ is equal to either $L_{m}(r_{m+1}^{(x,y)})$ or $\mathrm{NN}_{m}^-(r_{m+1}^{(x,y)}, s(x,y))$, which are both less than or equal to $\tilde{Q}_{1-\alpha}$:
the first by the condition (i), 
the second by the definition of $\mathrm{NN}_{m}^-$. 
Therefore, in either case, we have $y \in \ch(x)$.

Next, to prove the contrapositive, let $y \notin \ch^{\mathrm{fast}}(x)$---more precisely, $y \in \ch^{\mathrm{fast}}(x)^c \cap \Lambda$. From the observations above, we have
\begin{align*}
    &\left\{y \notin \ch^{\mathrm{fast}}(x)\right\}
    = \left(\left\{s(x,y) > \tilde{Q}_{1-\alpha}'\right\} \cup \left\{L_{m}(r_{m+1}^{(x,y)}) > \tilde{Q}_{1-\alpha}\right\}\right) \cap \left\{U_{m}(r_{m+1}^{(x,y)}) > \tilde{Q}_{1-\alpha}\right\}\\
    &= \left(\left\{s(x,y) > \tilde{Q}_{1-\alpha}'\right\} \cap \left\{U_{m}(r_{m+1}^{(x,y)}) > \tilde{Q}_{1-\alpha}\right\}\right) \cup \left\{L_{m}(r_{m+1}^{(x,y)}) > \tilde{Q}_{1-\alpha}\right\},
\end{align*}
where the second equality applies De Morgan's law. The final set is a disjoint union of the following two events:
\[\text{(i) } s(x,y) > \tilde{Q}_{1-\alpha}' \text{ and } L_{m}(r_{m+1}^{(x,y)}) \le \tilde{Q}_{1-\alpha} < U_{m}(r_{m+1}^{(x,y)}), \quad \text{(ii) } L_{m}(r_{m+1}^{(x,y)}) > \tilde{Q}_{1-\alpha}.\]
In case (ii), we have $T(s(x,y)) > \tilde{Q}_{1-\alpha}$, since $T(s(x,y)) \ge L_{m}(r_{m+1}^{(x,y)})$ holds deterministically. 
In case (i), $T(s(x,y))$ is equal to either $U_{m}(r_{m+1}^{(x,y)})$ or $\mathrm{NN}_{m}^-(r_{m+1}^{(x,y)}, s(x,y))$, which are both larger than $\tilde{Q}_{1-\alpha}$. This can be concluded as follows: 
In this case, we have $ s(x,y) > \tilde{Q}_{1-\alpha}' \ge \tilde{Q}_{1-\alpha}$.
If $L_{m}(r_{m+1}^{(x,y)}) \le s(x,y) < U_{m}(r_{m+1}^{(x,y)})$, 
by the construction of $T(s(x,y))$,
we have $T(s(x,y)) = \mathrm{NN}_{m}^-(r_{m+1}^{(x,y)}, s(x,y)) \ge \tilde{Q}_{1-\alpha}' \ge \tilde{Q}_{1-\alpha}$, and moreover we cannot have equality, since $y \in \Lambda$. Otherwise, $s(x,y) \ge U_{m}(r_{m+1}^{(x,y)})$, 
and therefore, $T(s(x,y)) = U_{m}(r_{m+1}^{(x,y)}) > \tilde{Q}_{1-\alpha}$. 
Therefore, in both cases, we have $y \notin \ch(x)$, as desired.~

\section{Experimental details}
\label{app-sec:experimental_details}

\subsection{Setup and environment}
\label{app-sec:setup_and_env}
The experiments were conducted on a system running Ubuntu 20.04.6 LTS, with 192 CPU cores of Intel(R) Xeon(R) Gold CPUs at 2.40 GHz, 1 TB of RAM, and 16 NVIDIA A40 GPUs. The software environment used Python 3.11.5, PyTorch 2.6, and CUDA 12.2.

\subsection{Datasets}
\label{app-sec:datasets}
Our experiments involve two datasets: ImageNet for image classification tasks and the Medical Expenditure Panel Survey (MEPS) for regression tasks.

\begin{itemize}
\item \textbf{ImageNet} \citep{deng2009imagenet}: We use the training split of ImageNet, focusing on 30 selected classes, which are listed in \Cref{app-tab:acc_per_class}.
\item \textbf{MEPS}: The MEPS dataset is a medical survey used for regression tasks, with the goal of predicting healthcare expenditures. For the regression experiments: MEPS-19~\citep{MEPS19}, MEPS-20~\citep{MEPS20}, and MEPS-21~\citep{MEPS21}. Each survey includes 139 features, such as demographic information (e.g., age, gender), and clinical data (e.g., chronic conditions, medical history).
\end{itemize}

\subsection{Model details}
\label{app-sec:models}
We have applied the following models to compute the nonconformity scores:
\begin{itemize}
\item \textbf{ImageNet experiments}: We employed a CLIP model based ViT-B/32 backbone, pre-trained on the LAION-2B dataset  \citep{radford2021learning,ilharco_gabriel_2021_5143773,schuhmann2022laion}, using the HuggingFace API. \Cref{app-tab:acc_per_class} reports the top-1 and top-2 accuracies of this model on the ImageNet training set for the subset of classes used in our experiments.
\item \textbf{MEPS experiments}: 
The dataset was filtered to include only non-Hispanic White and non-White individuals. Panel-specific variables were renamed for consistency across panels 19-21, and rows with missing or invalid values were removed. A healthcare utilization variable was computed as the sum of expenditures across outpatient, office-based, emergency room, inpatient, and home health services, serving as the regression target. Preprocessing steps included retaining common features across panels, standardizing covariates, and applying a log transformation to the target variable to reduce skewness.

A deep neural network was trained to estimate the lower and upper quantile bounds of healthcare utilization using a quantile regression approach, with different $\alpha$ levels used for each experiment. The network architecture consisted of four hidden layers (256, 128, 64, and 32 units) with LeakyReLU activations, dropout regularization (rate = 0.3), and optional batch normalization. The model was optimized using the pinball loss function and trained on 2019 data with early stopping based on validation loss (up to 50 epochs, batch size = 128, learning rate = 1e-4). 
\end{itemize}

\subsection{Data generation}
\label{app-sec:data_generation}
\subsubsection{Stable Diffusion}
\begin{figure}[h]
    \centering
    \setlength{\tabcolsep}{2pt} 
    \renewcommand{\arraystretch}{0.8} 

    \begin{tabular}{c!{\vrule width 2pt}cccccc}
            \includegraphics[height=0.1\linewidth]{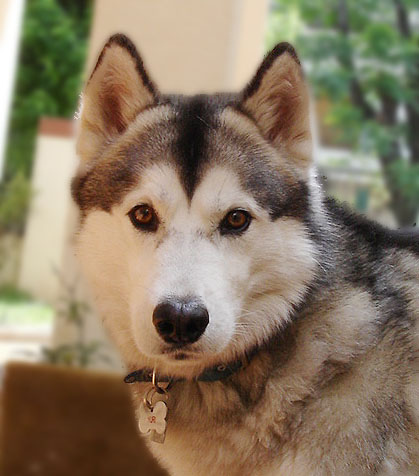} &
            \includegraphics[height=0.1\linewidth]{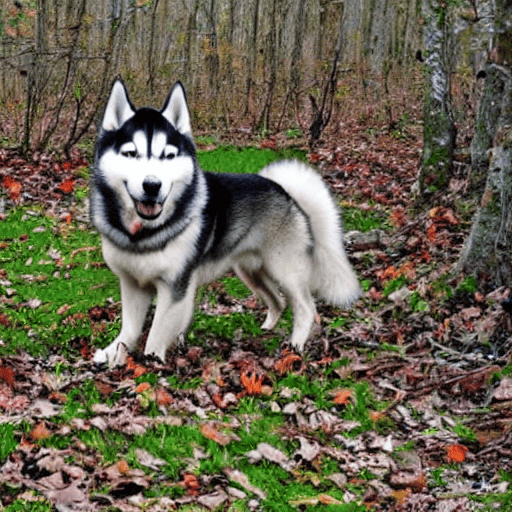} &
            \includegraphics[height=0.1\linewidth]{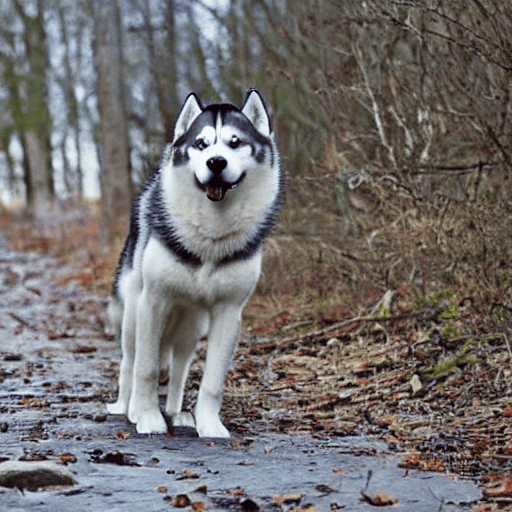} &
            \includegraphics[height=0.1\linewidth]{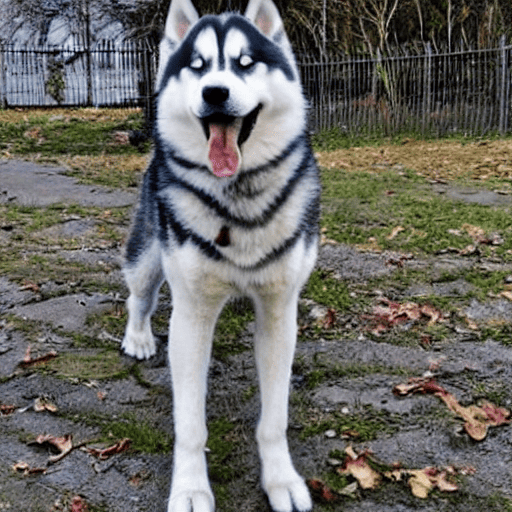} &
            \includegraphics[height=0.1\linewidth]{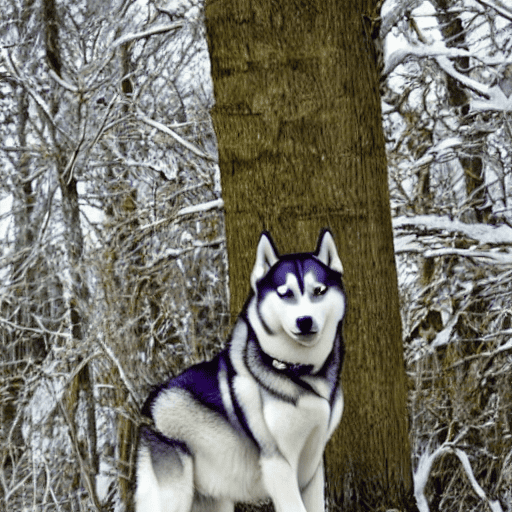} &
            \includegraphics[height=0.1\linewidth]{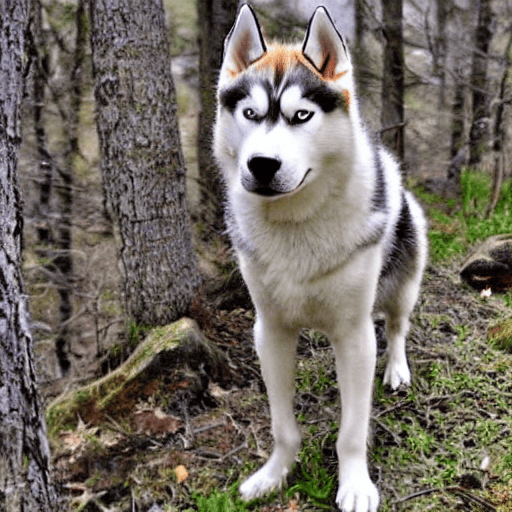} &
            \includegraphics[height=0.1\linewidth]{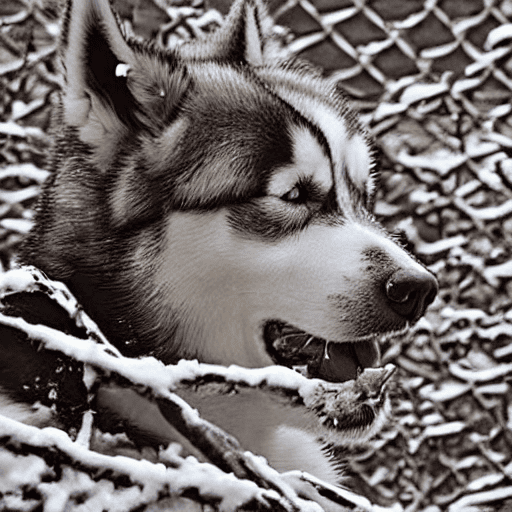} \\
        
            \includegraphics[height=0.1\linewidth]{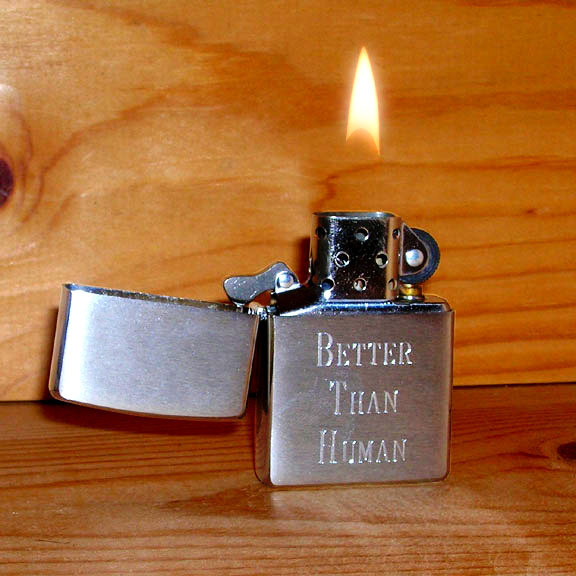} &
            \includegraphics[height=0.1\linewidth]{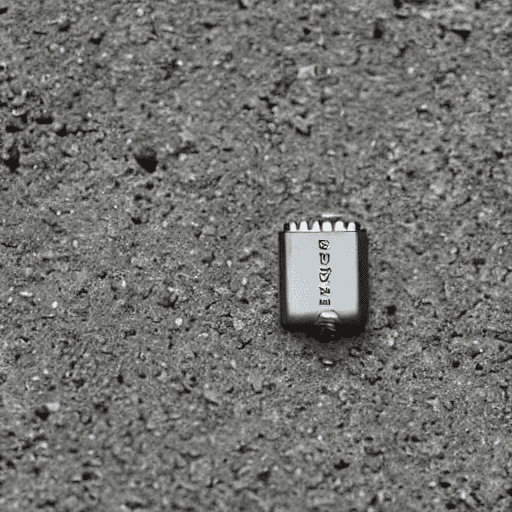} &
            \includegraphics[height=0.1\linewidth]{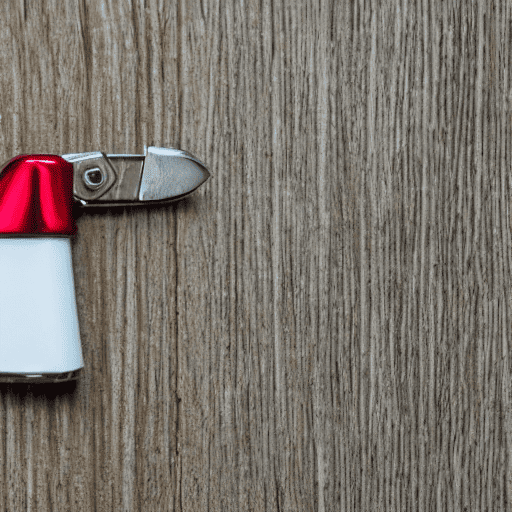} &
            \includegraphics[height=0.1\linewidth]{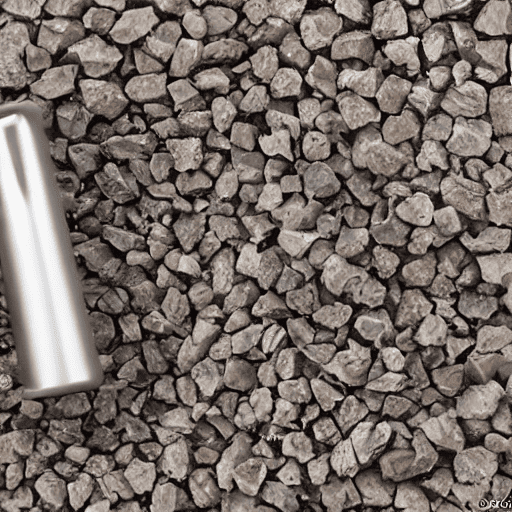} &
            \includegraphics[height=0.1\linewidth]{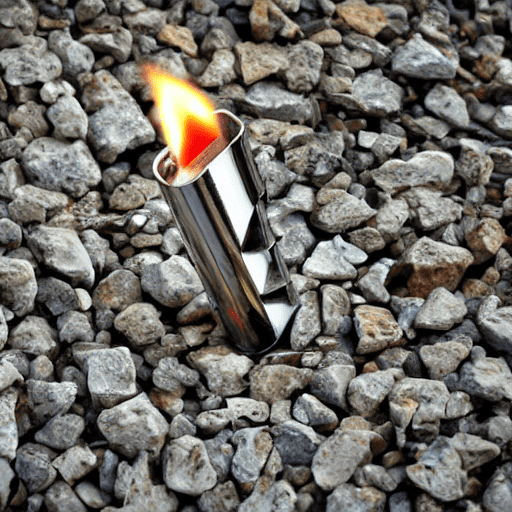} &
            \includegraphics[height=0.1\linewidth]{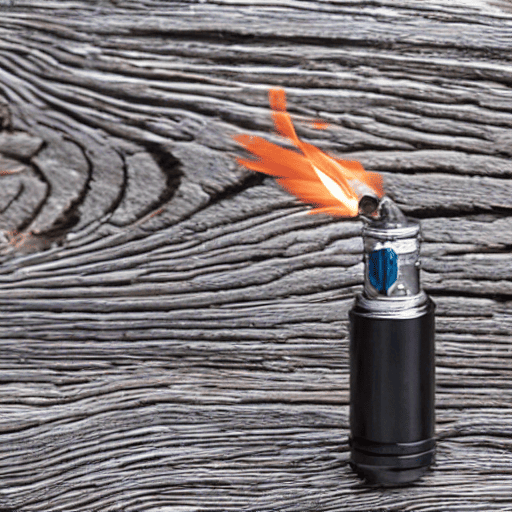} &
            \includegraphics[height=0.1\linewidth]{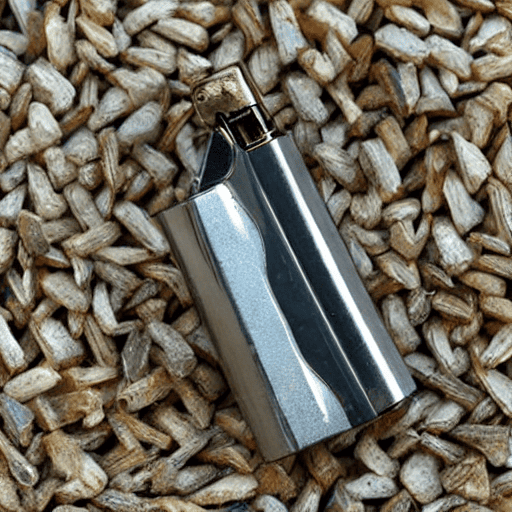} \\
			
            \includegraphics[height=0.1\linewidth]{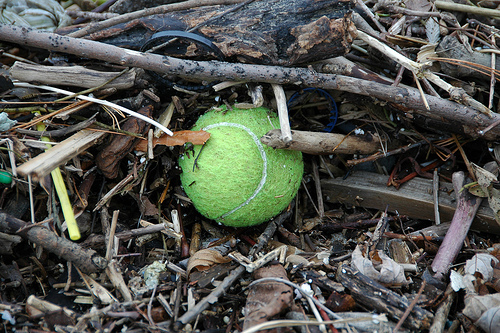} &
            \includegraphics[height=0.1\linewidth]{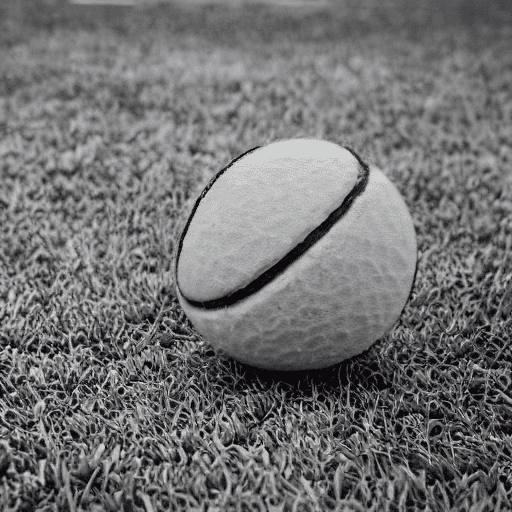} &
            \includegraphics[height=0.1\linewidth]{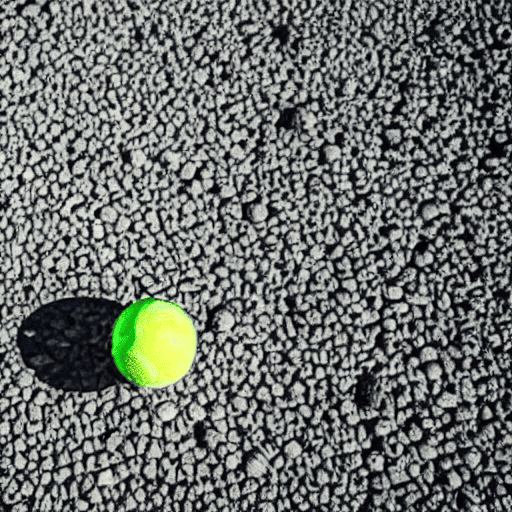} &
            \includegraphics[height=0.1\linewidth]{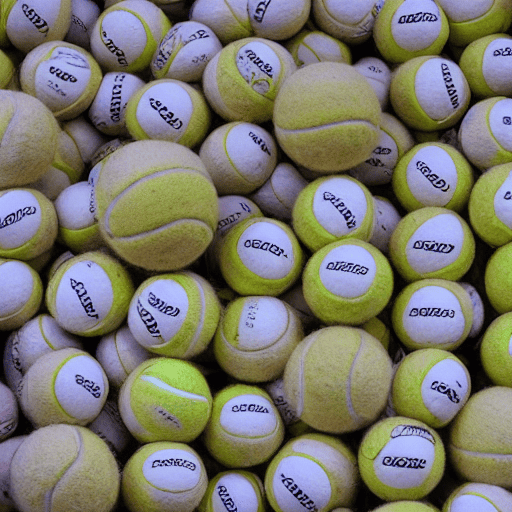} &
            \includegraphics[height=0.1\linewidth]{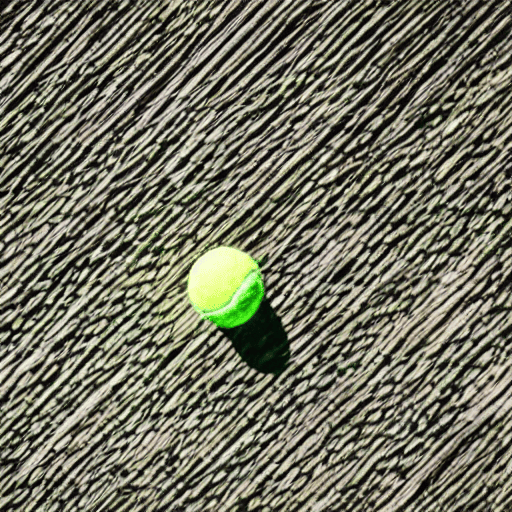} &
            \includegraphics[height=0.1\linewidth]{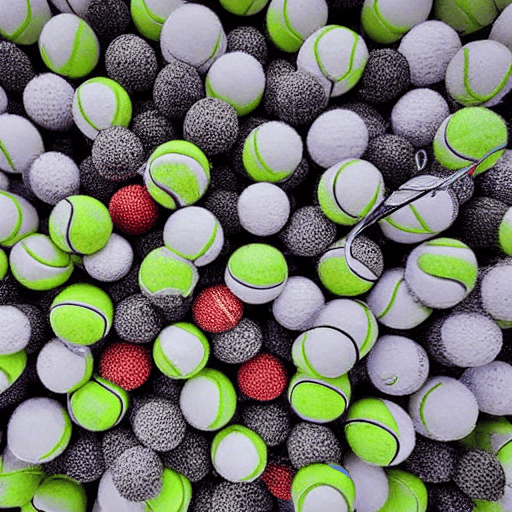} &
            \includegraphics[height=0.1\linewidth]{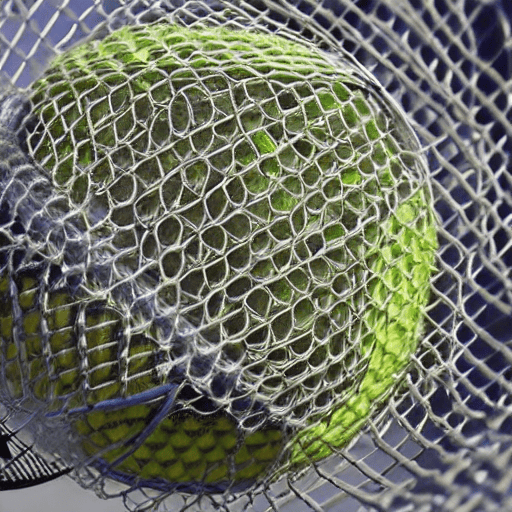} \\

            \includegraphics[height=0.1\linewidth]{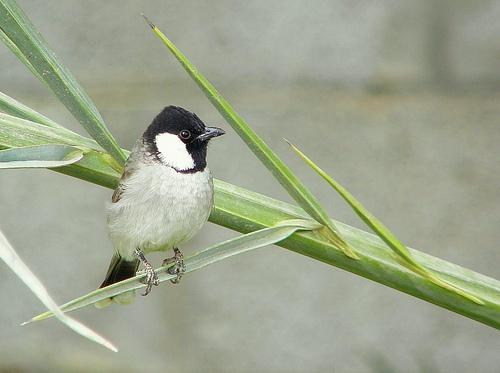} &
            \includegraphics[height=0.1\linewidth]{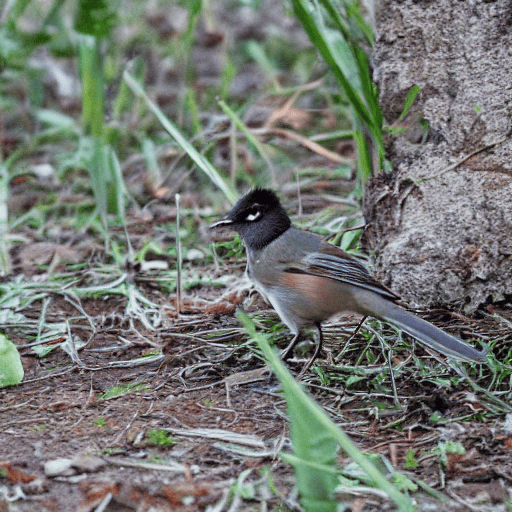} &
            \includegraphics[height=0.1\linewidth]{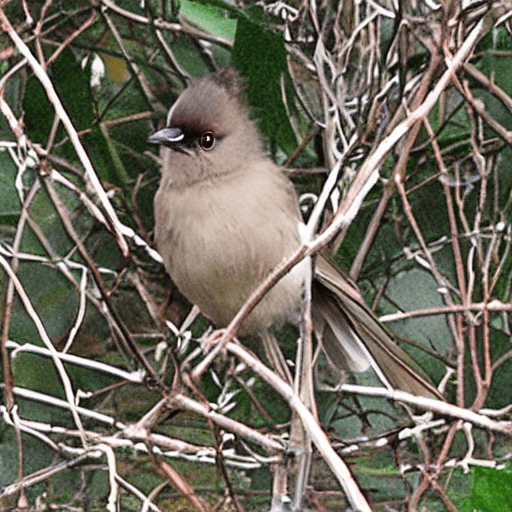} &
            \includegraphics[height=0.1\linewidth]{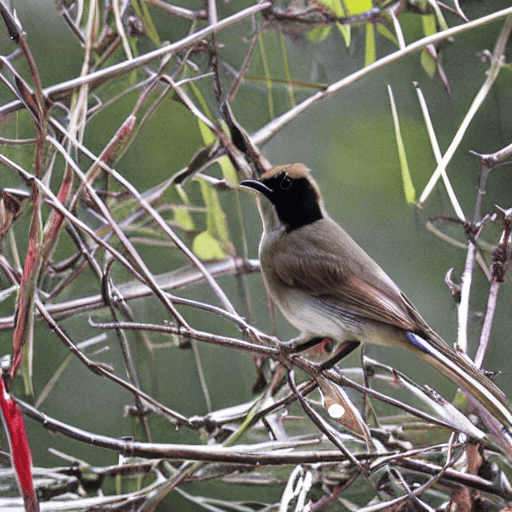} &
            \includegraphics[height=0.1\linewidth]{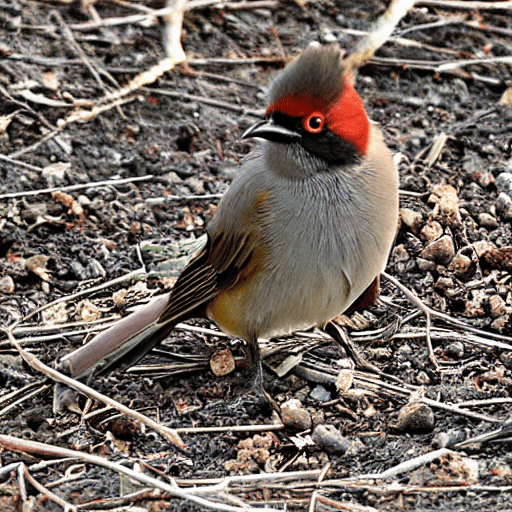} &
            \includegraphics[height=0.1\linewidth]{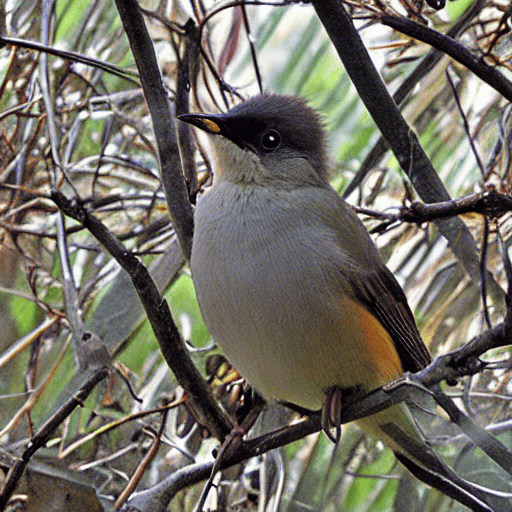} &
            \includegraphics[height=0.1\linewidth]{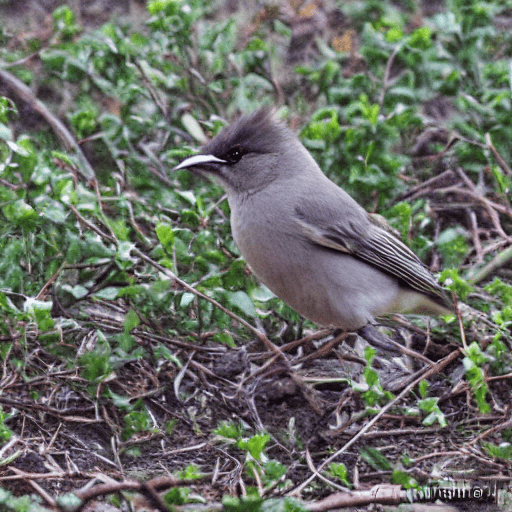} \\
			
    \end{tabular}

    \caption{Comparison between real and Stable Diffusion-generated images for selected ImageNet classes. Each row corresponds to a class, with the first column showing a real ImageNet image and the remaining columns showing generated datapoints.}
    \label{app-fig:imagenet-images}
\end{figure}

We generated synthetic images using the Stable Diffusion v1.5 model~\citep{Rombach_2022_CVPR}. For each class listed in \Cref{app-tab:acc_per_class}, we generated 2,000 images using the following configuration:
\begin{itemize}
\item \underline{Prompt}: ``A photo of a \{class name\}'', where \{class name\} refers to the corresponding ImageNet label, as shown to be effective in \citep{radford2021learning}.
\item \underline{Inference steps}: 260
\item \underline{Guidance scale}: 7.5
\end{itemize}
\Cref{app-fig:imagenet-images} presents examples of generated images alongside real ImageNet training images from the same class.

\subsubsection{FLUX}
\label{app-sec:data_generation_flux}
\begin{figure}[h]
    \centering
    \setlength{\tabcolsep}{2pt} 
    \renewcommand{\arraystretch}{0.8} 

    \begin{tabular}{c!{\vrule width 2pt}cccccc}
        \includegraphics[height=0.1\linewidth]{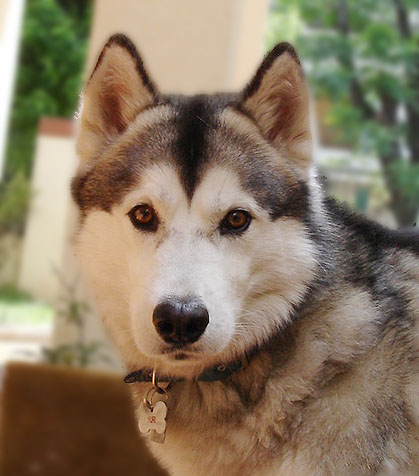} &
        \includegraphics[height=0.1\linewidth]{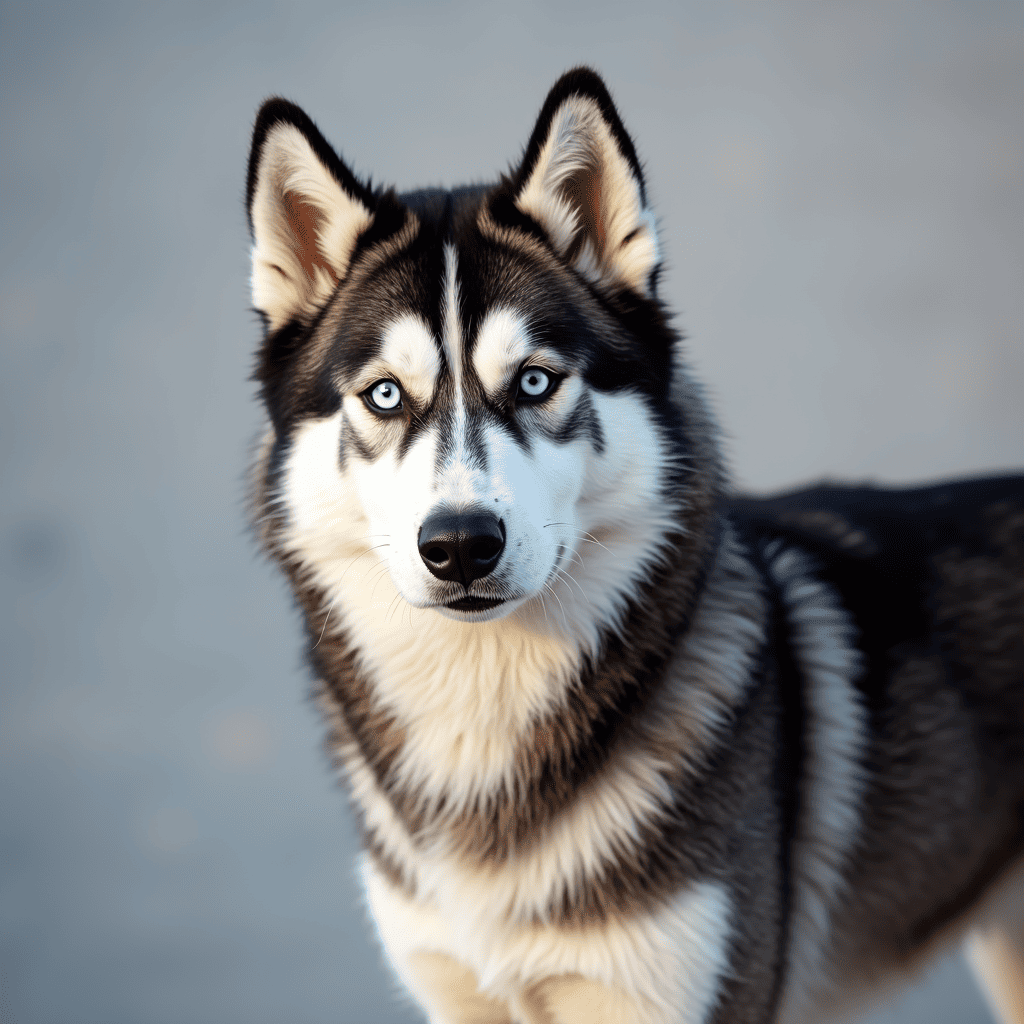} &
        \includegraphics[height=0.1\linewidth]{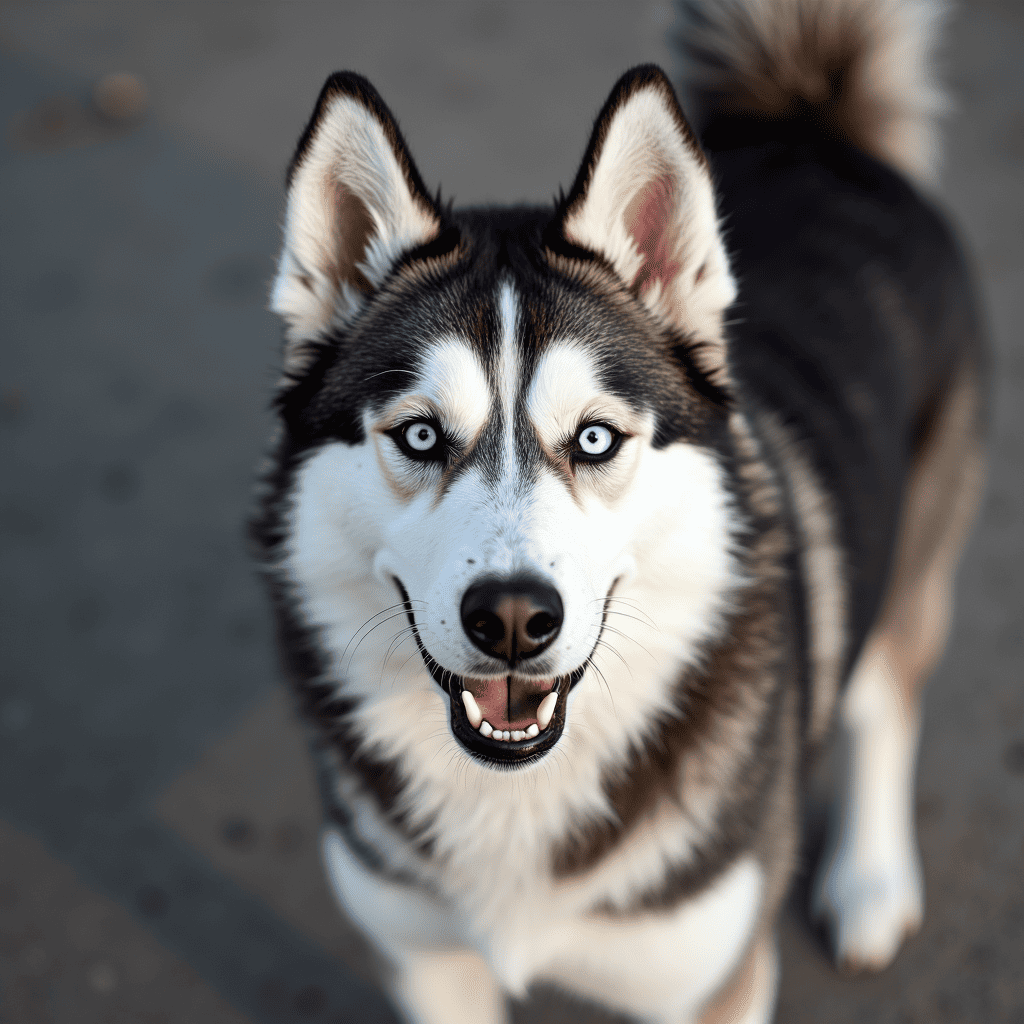} &
        \includegraphics[height=0.1\linewidth]{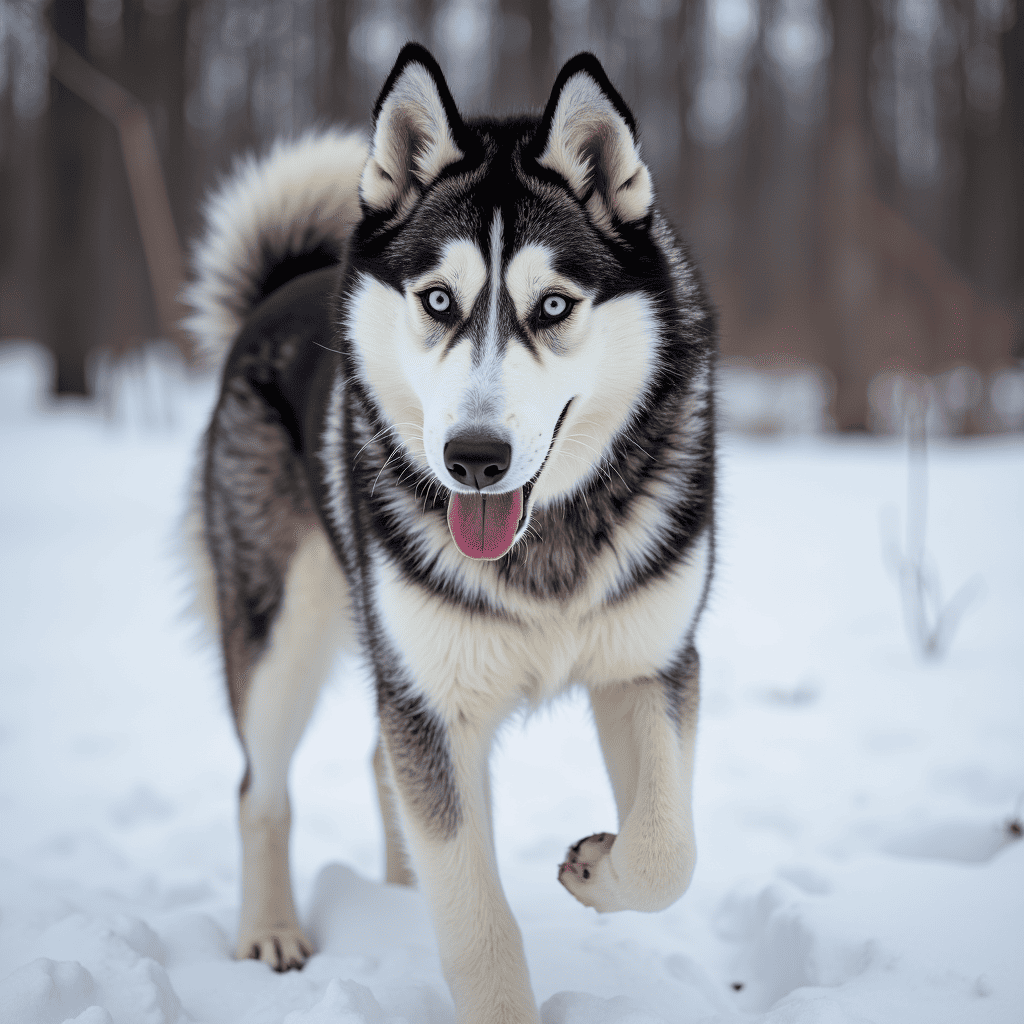} &
        \includegraphics[height=0.1\linewidth]{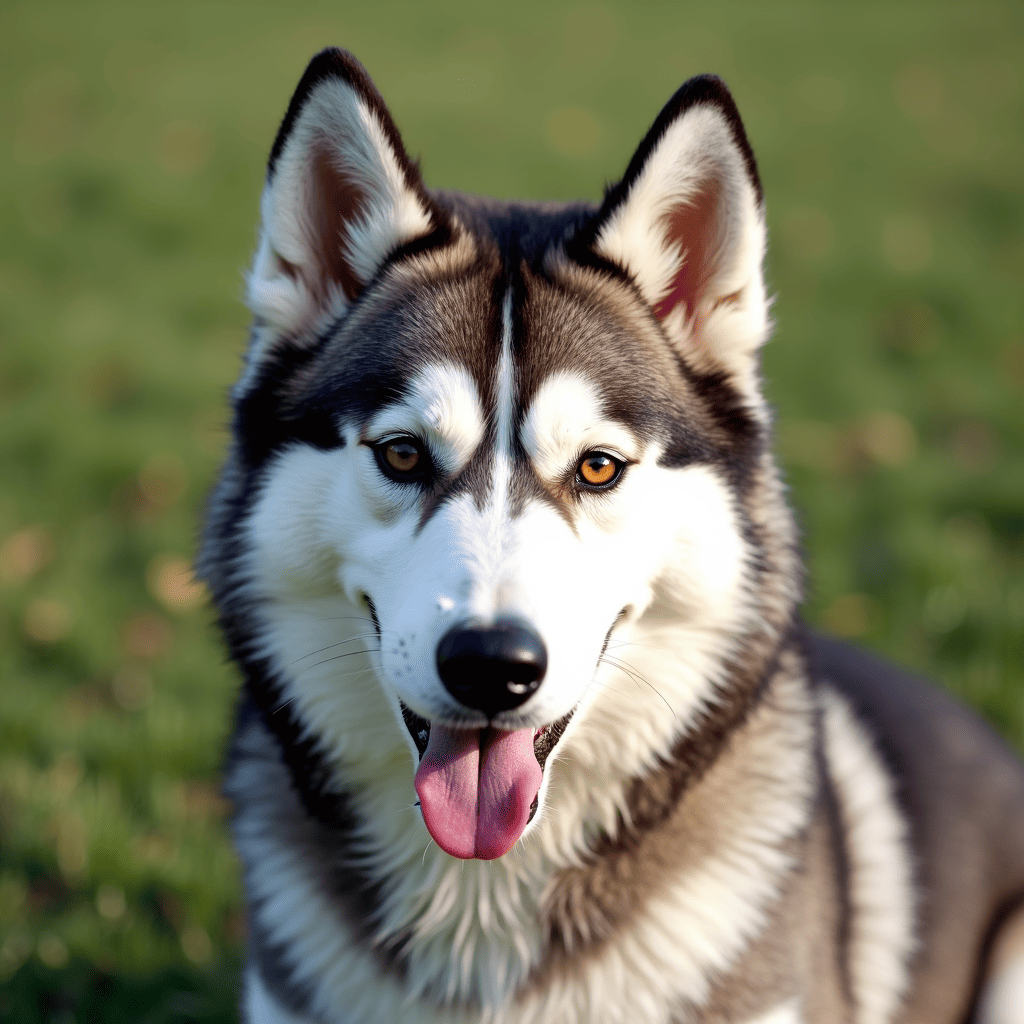} &
        \includegraphics[height=0.1\linewidth]{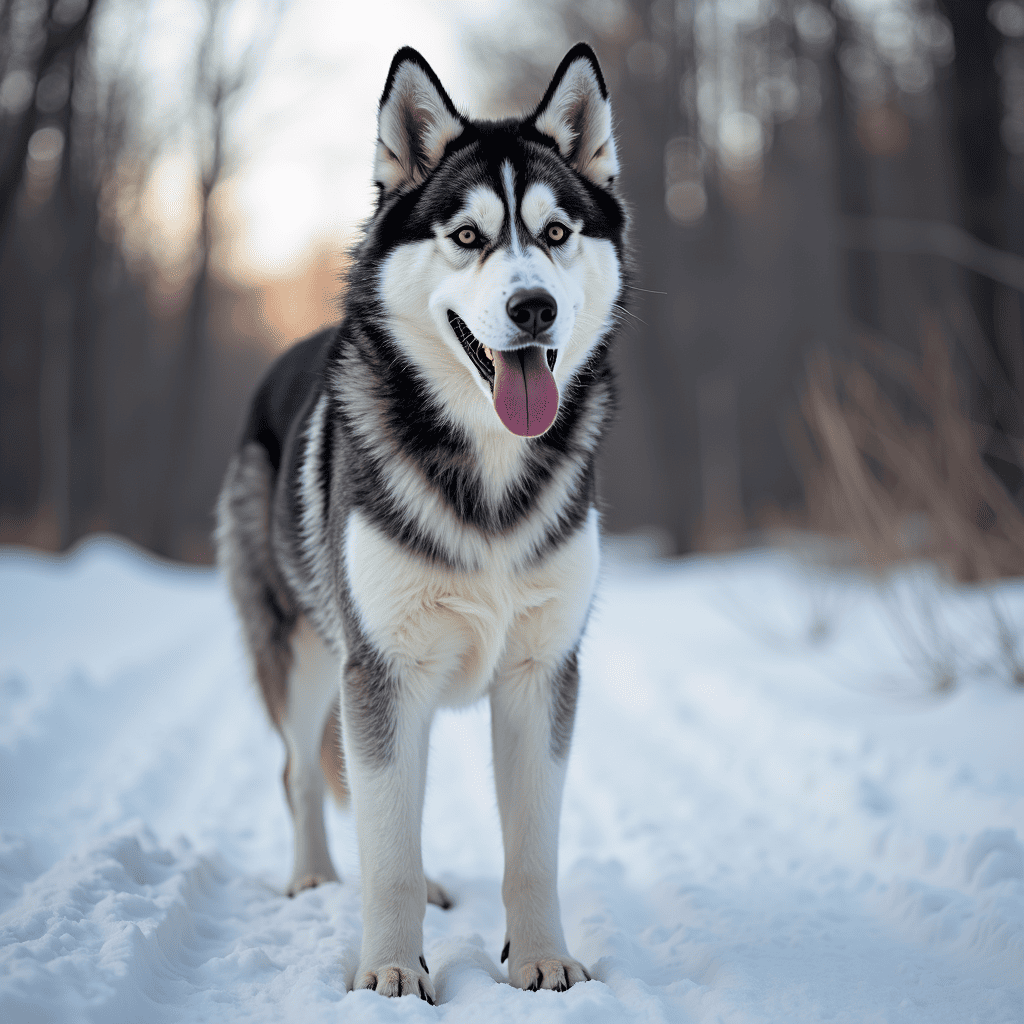} &
        \includegraphics[height=0.1\linewidth]{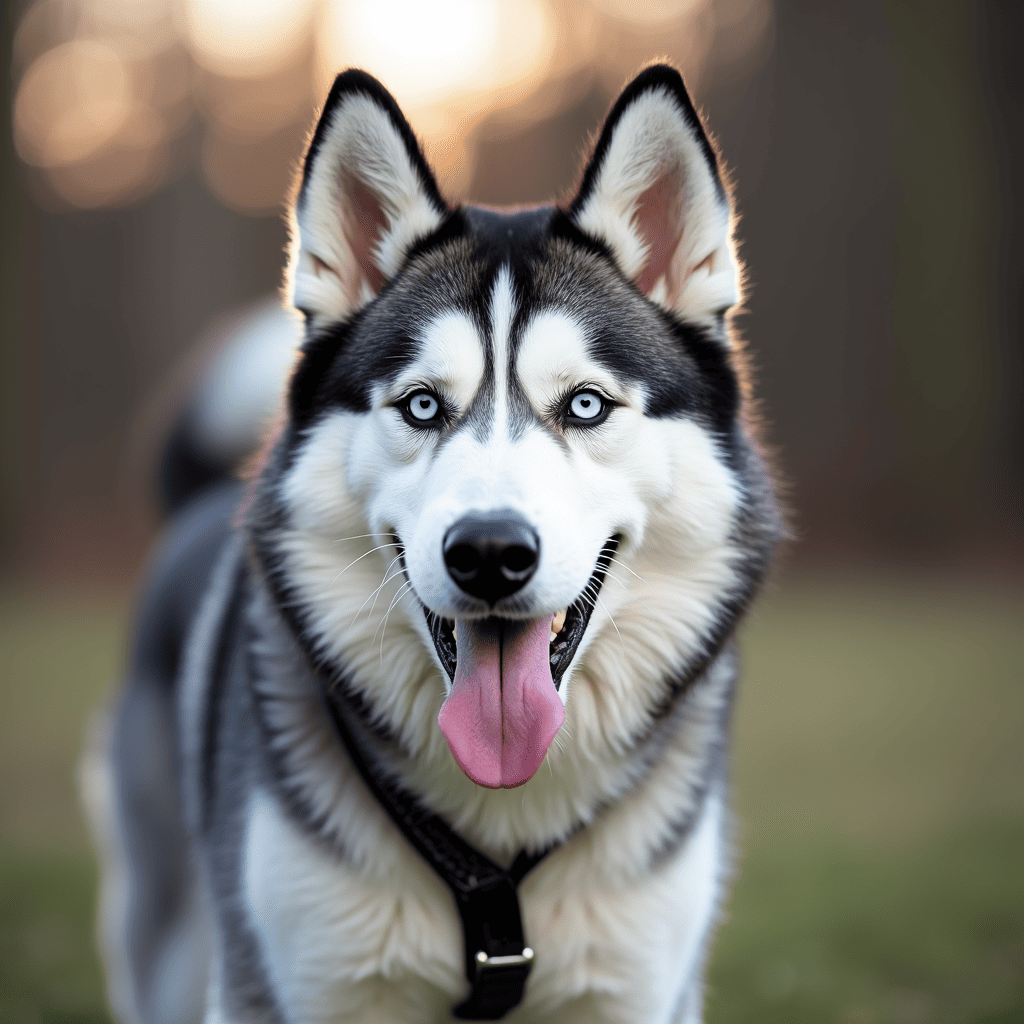} \\
        
        \includegraphics[height=0.1\linewidth]{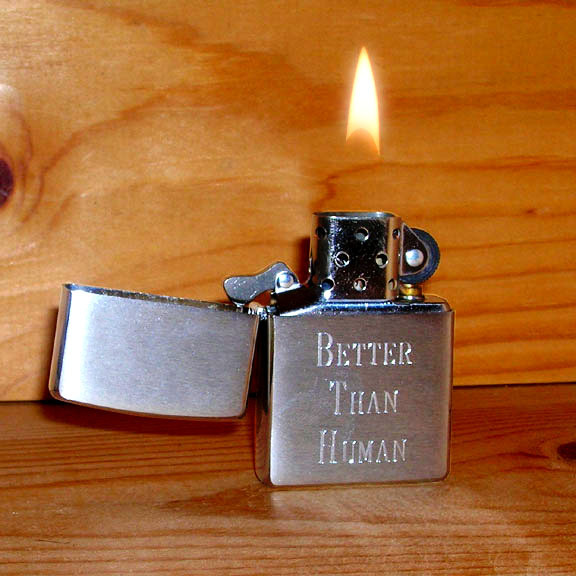} &
        \includegraphics[height=0.1\linewidth]{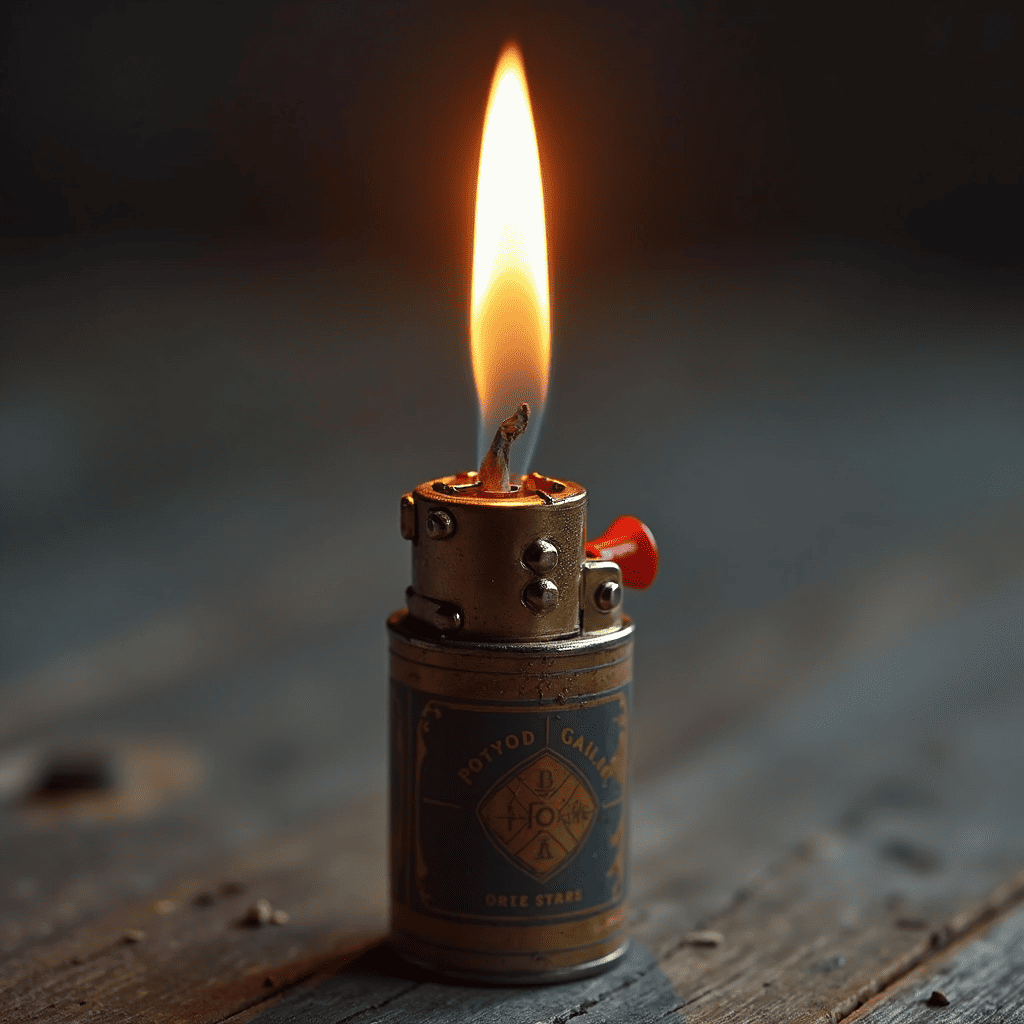} &
        \includegraphics[height=0.1\linewidth]{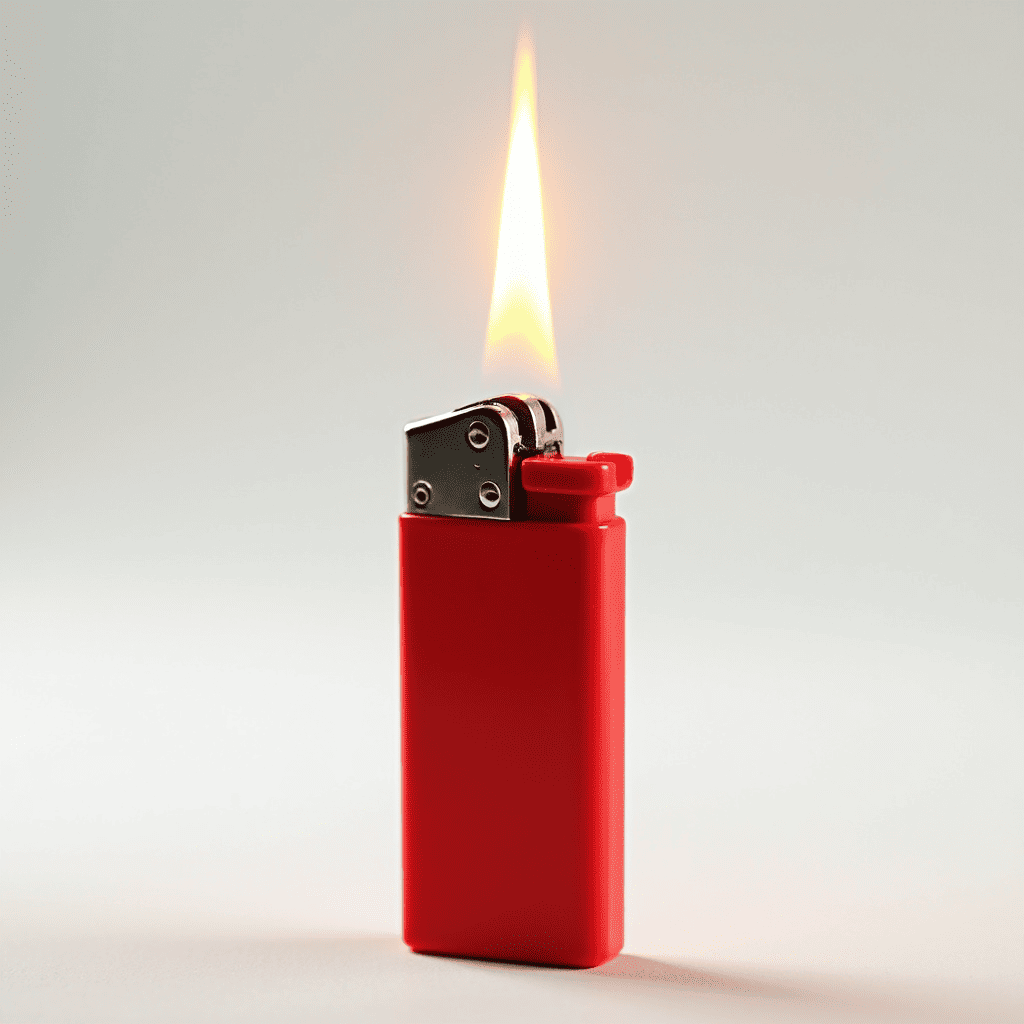} &
        \includegraphics[height=0.1\linewidth]{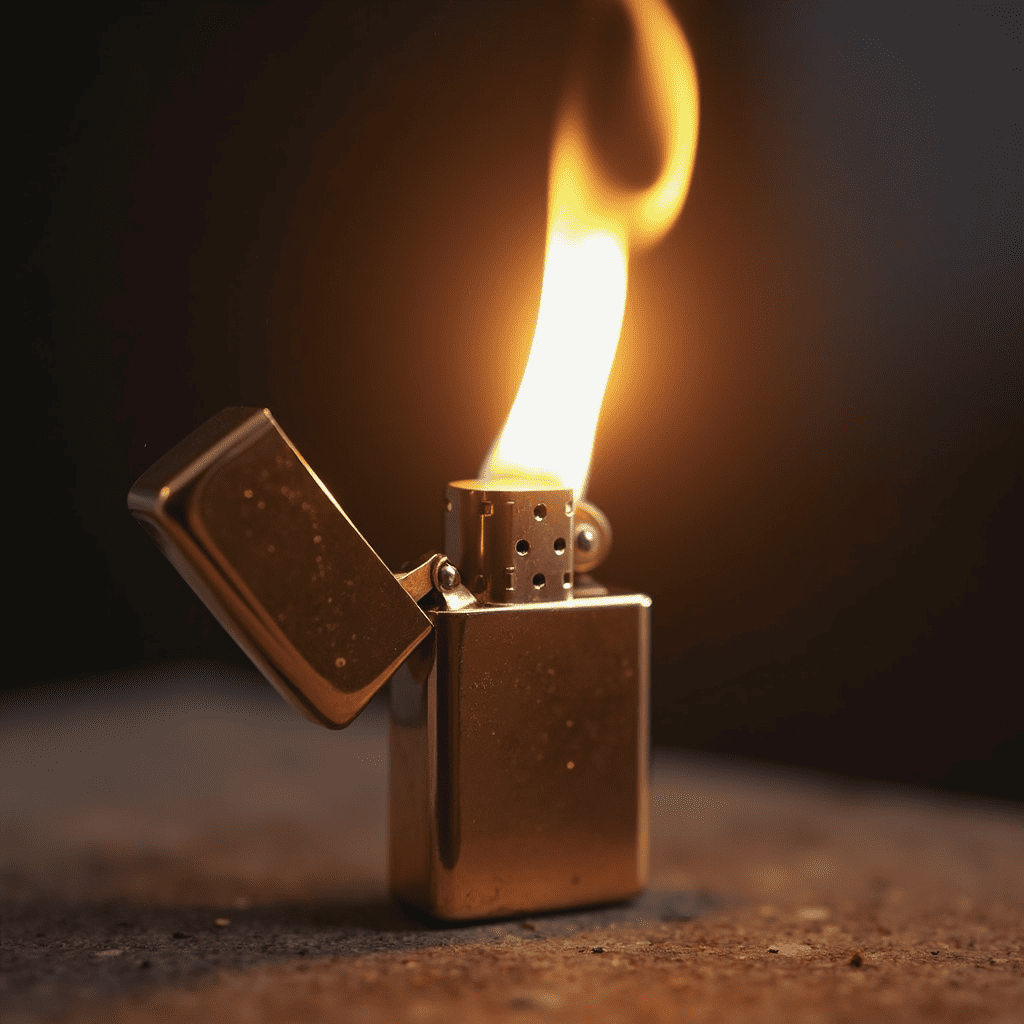} &
        \includegraphics[height=0.1\linewidth]{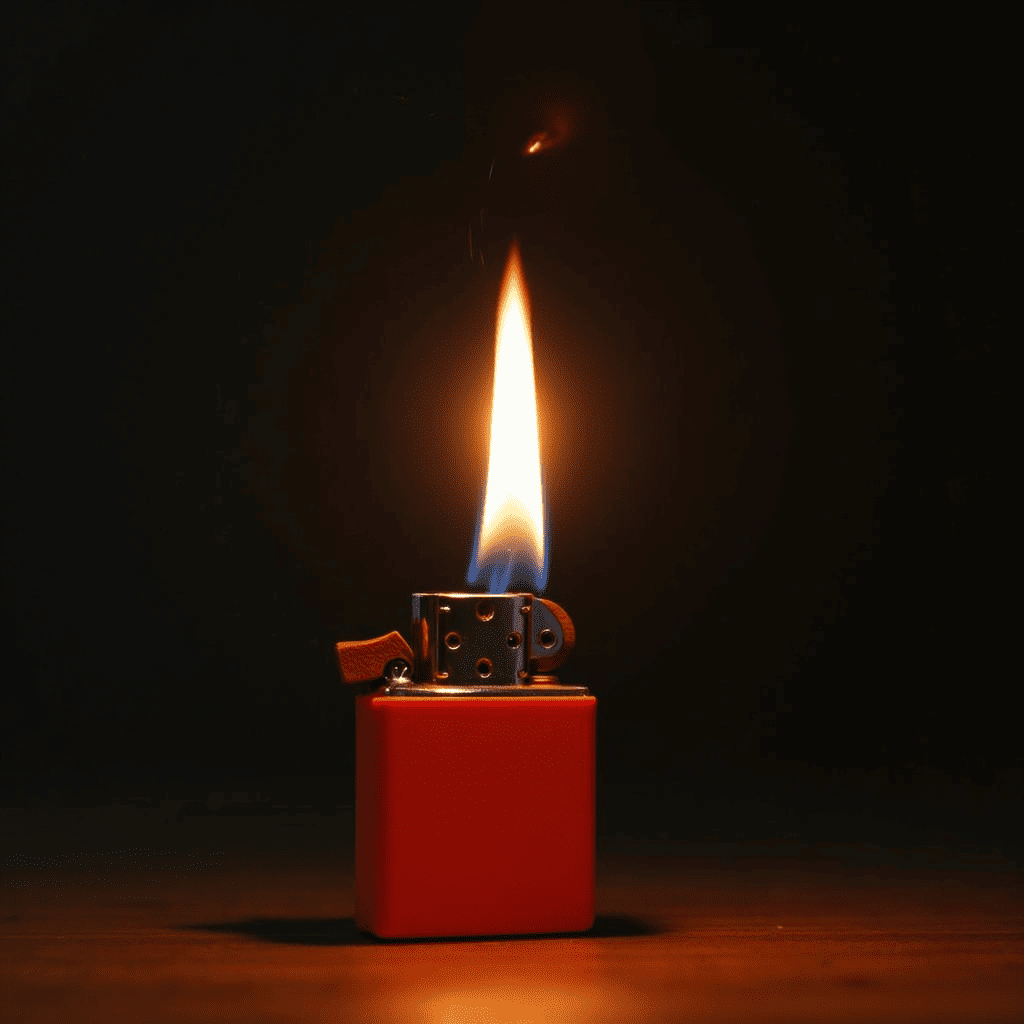} &
        \includegraphics[height=0.1\linewidth]{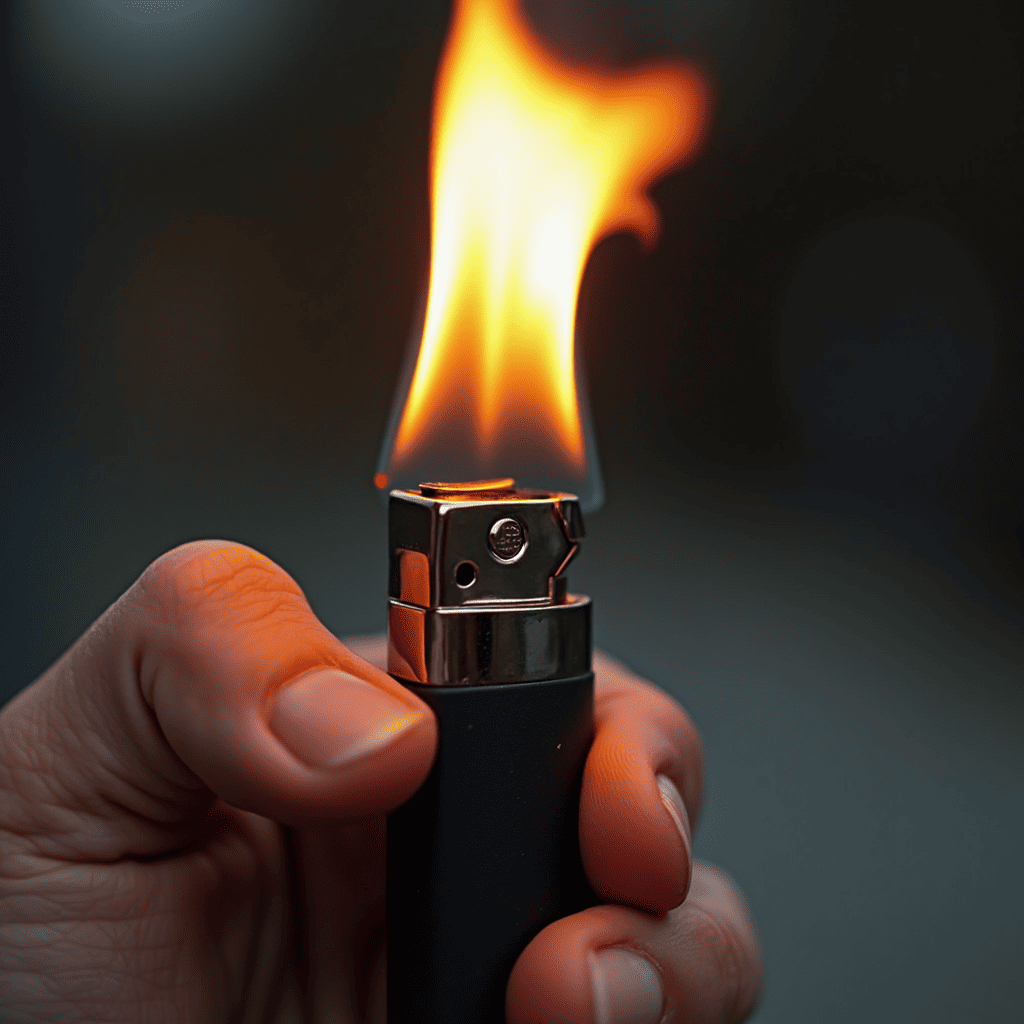} &
        \includegraphics[height=0.1\linewidth]{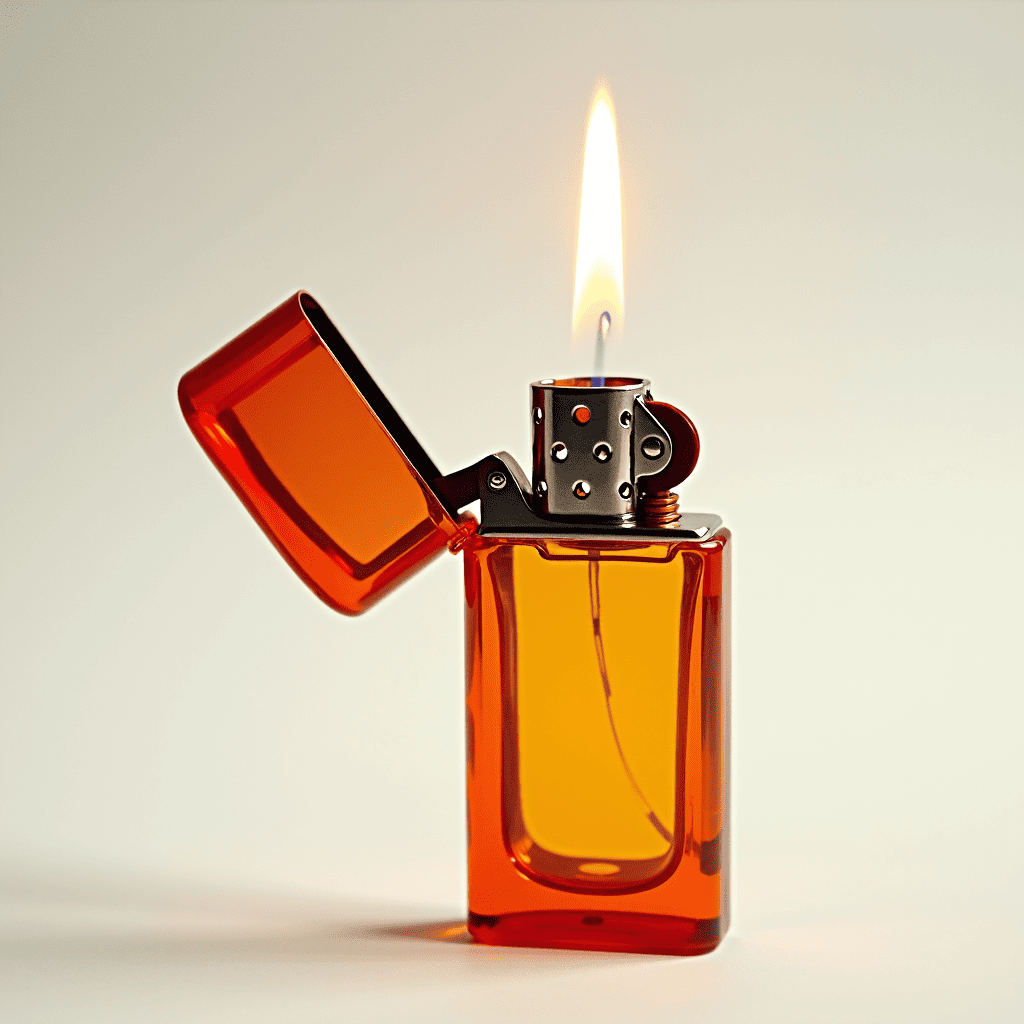} \\

        \includegraphics[height=0.1\linewidth]{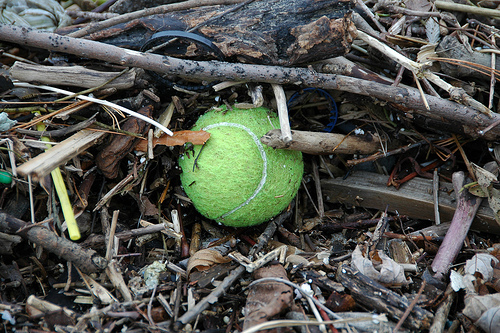} &
        \includegraphics[height=0.1\linewidth]{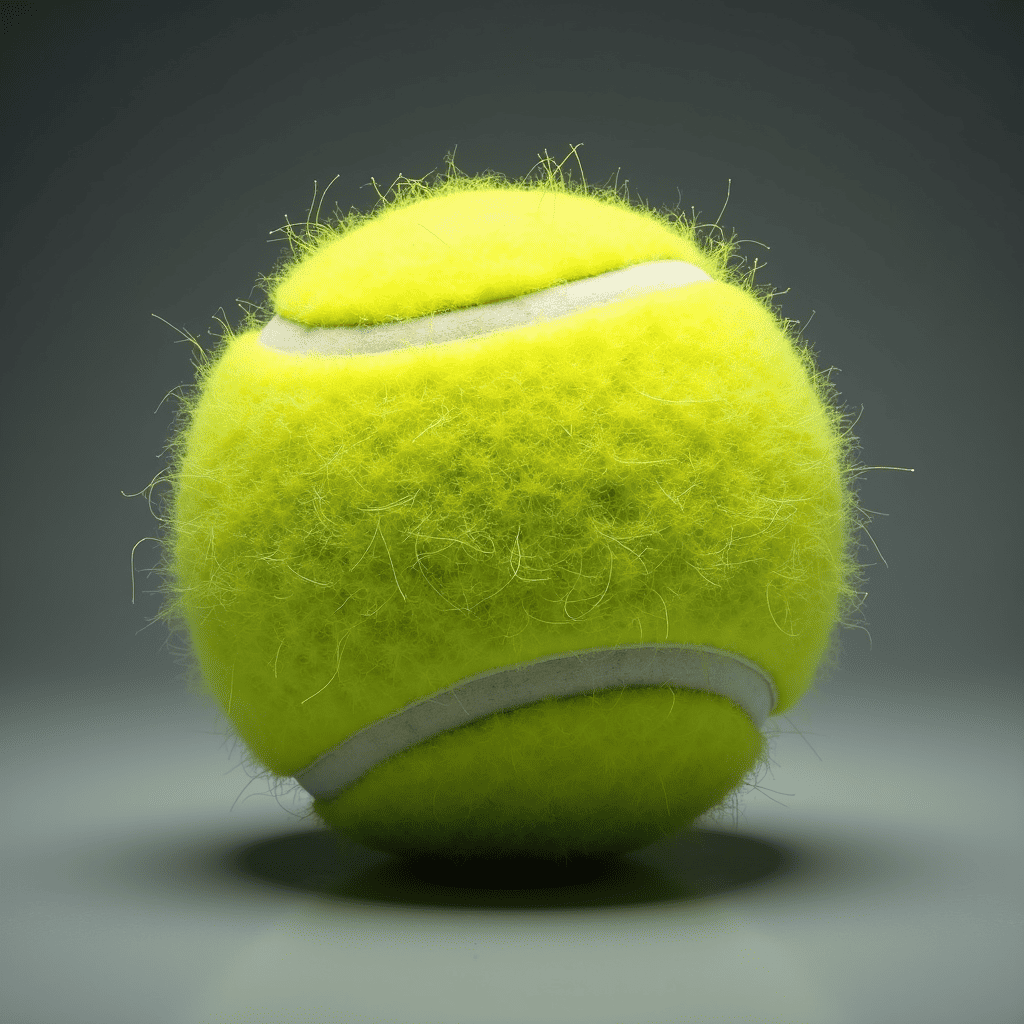} &
        \includegraphics[height=0.1\linewidth]{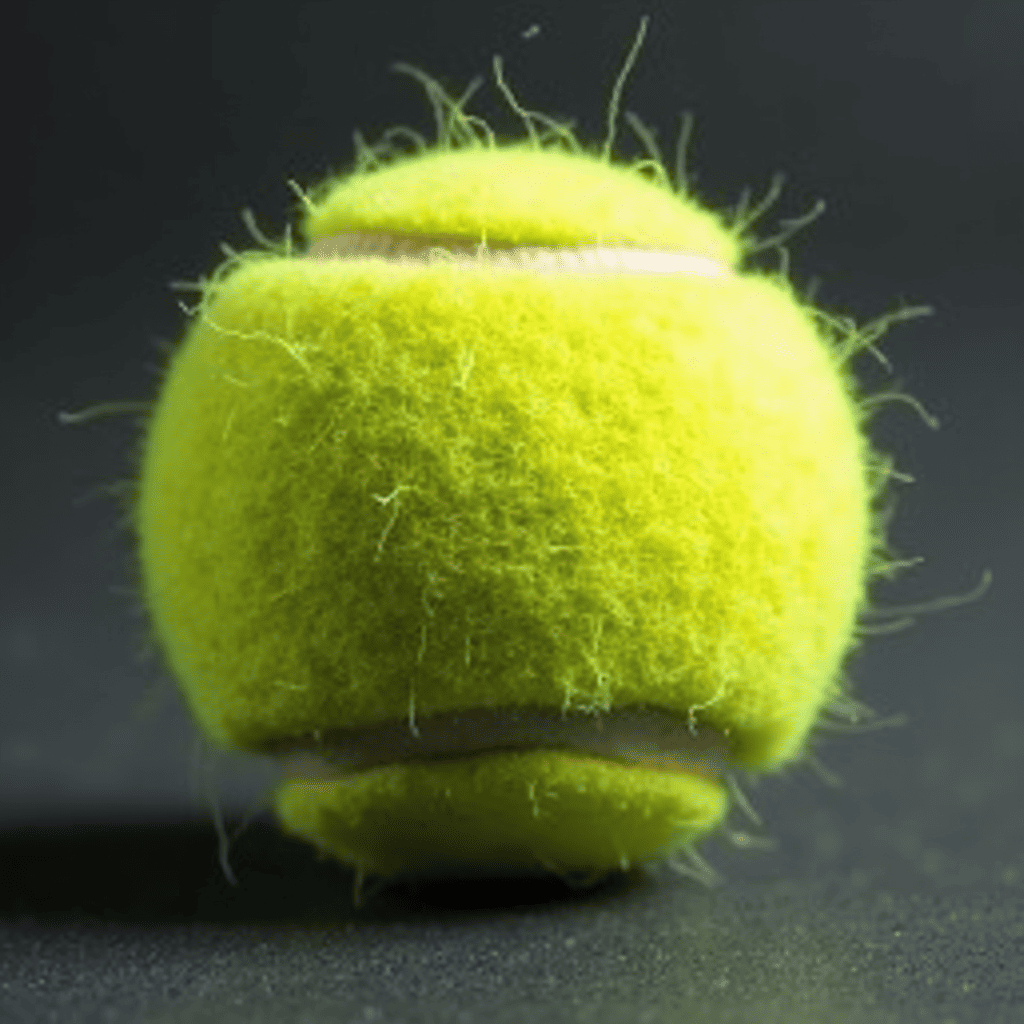} &
        \includegraphics[height=0.1\linewidth]{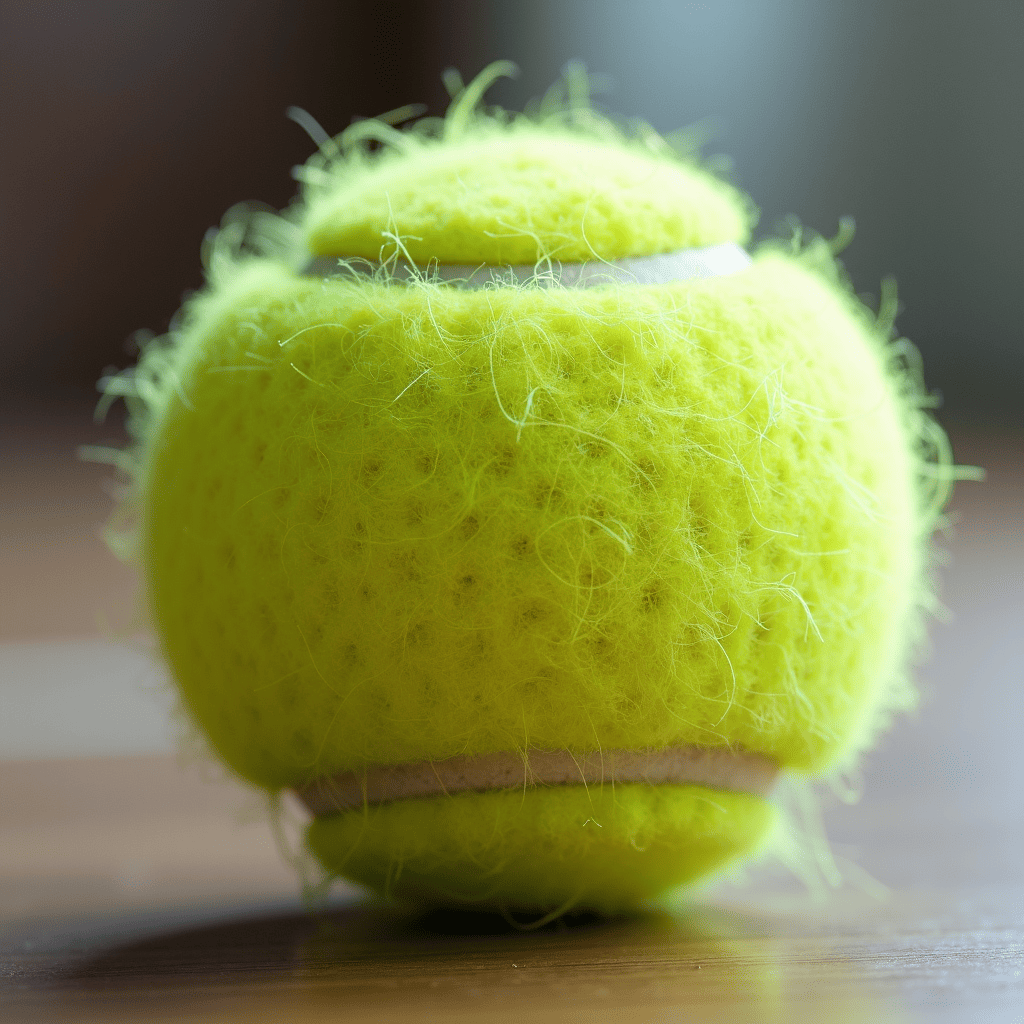} &
        \includegraphics[height=0.1\linewidth]{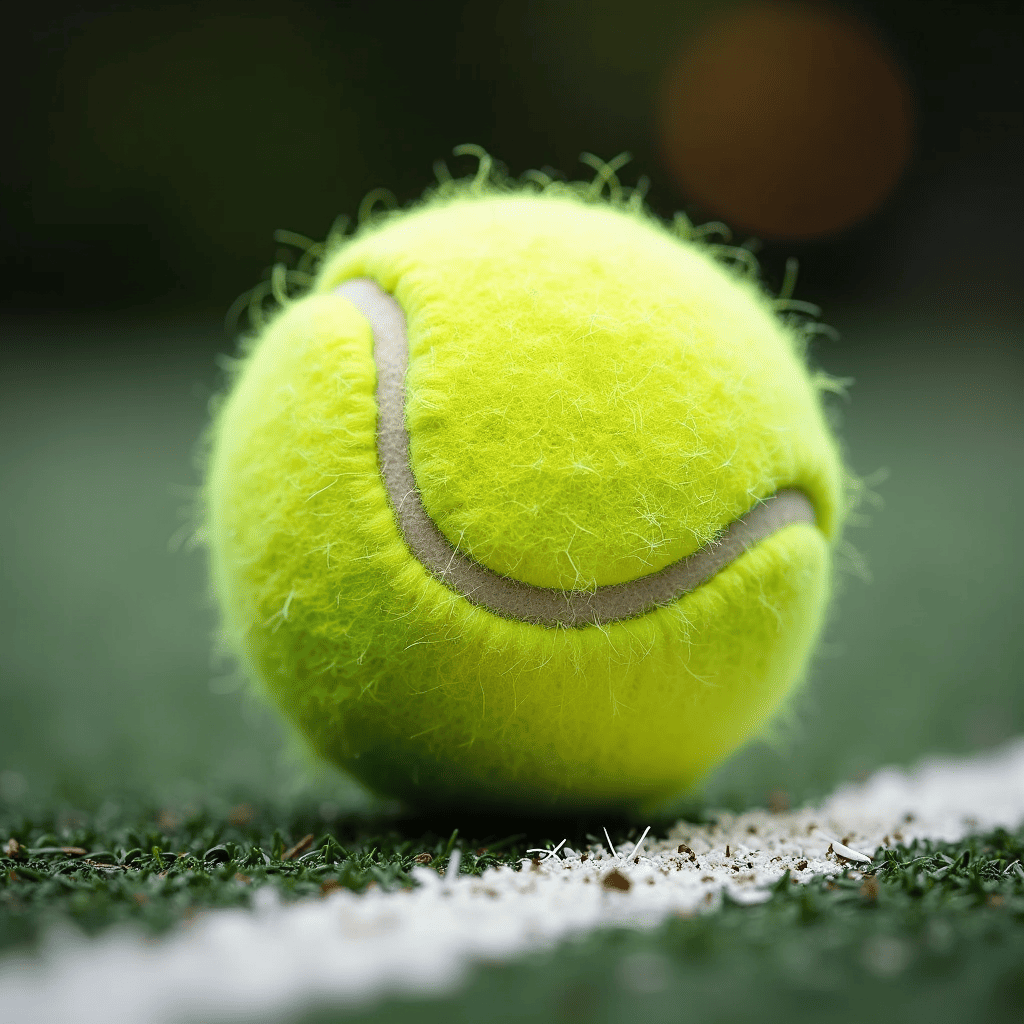} &
        \includegraphics[height=0.1\linewidth]{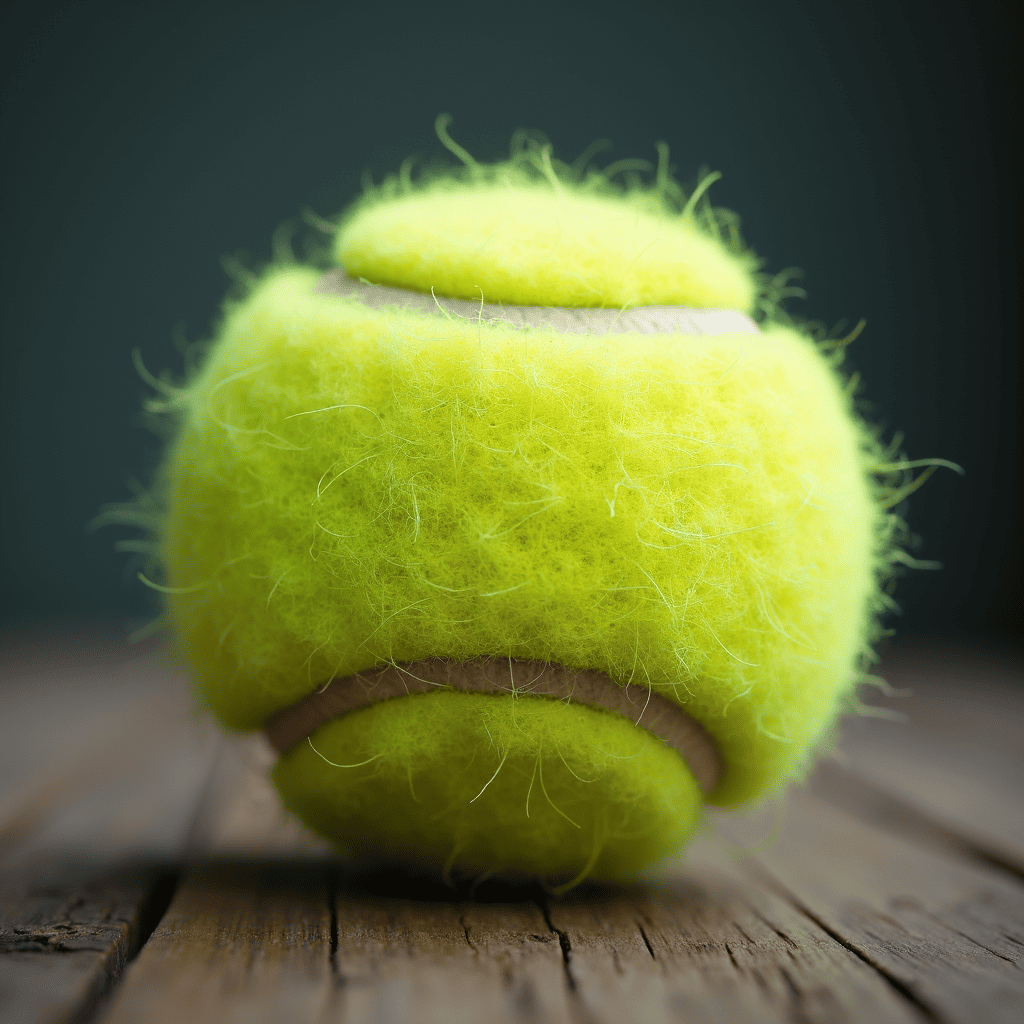} &
        \includegraphics[height=0.1\linewidth]{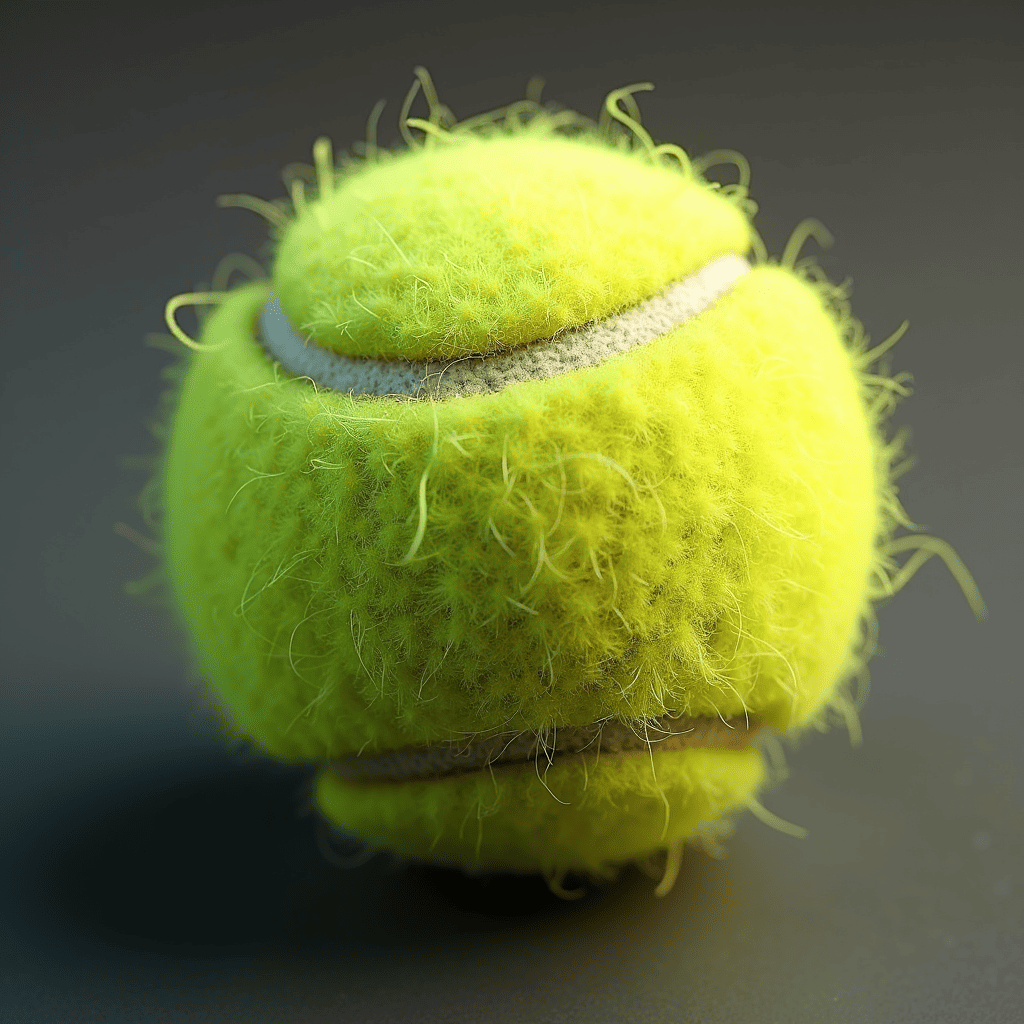} \\

        \includegraphics[height=0.1\linewidth]{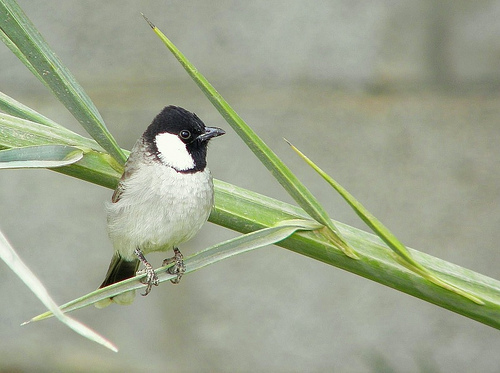} &
        \includegraphics[height=0.1\linewidth]{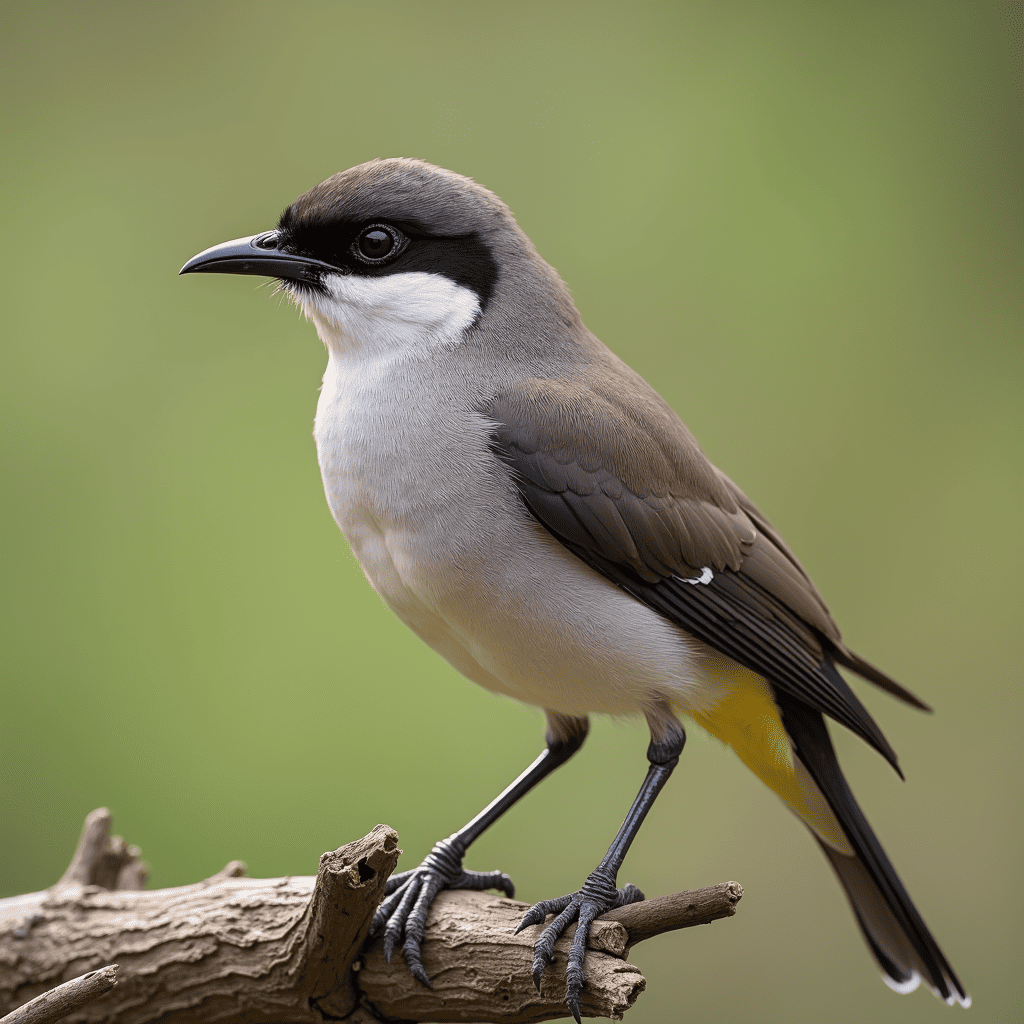} &
        \includegraphics[height=0.1\linewidth]{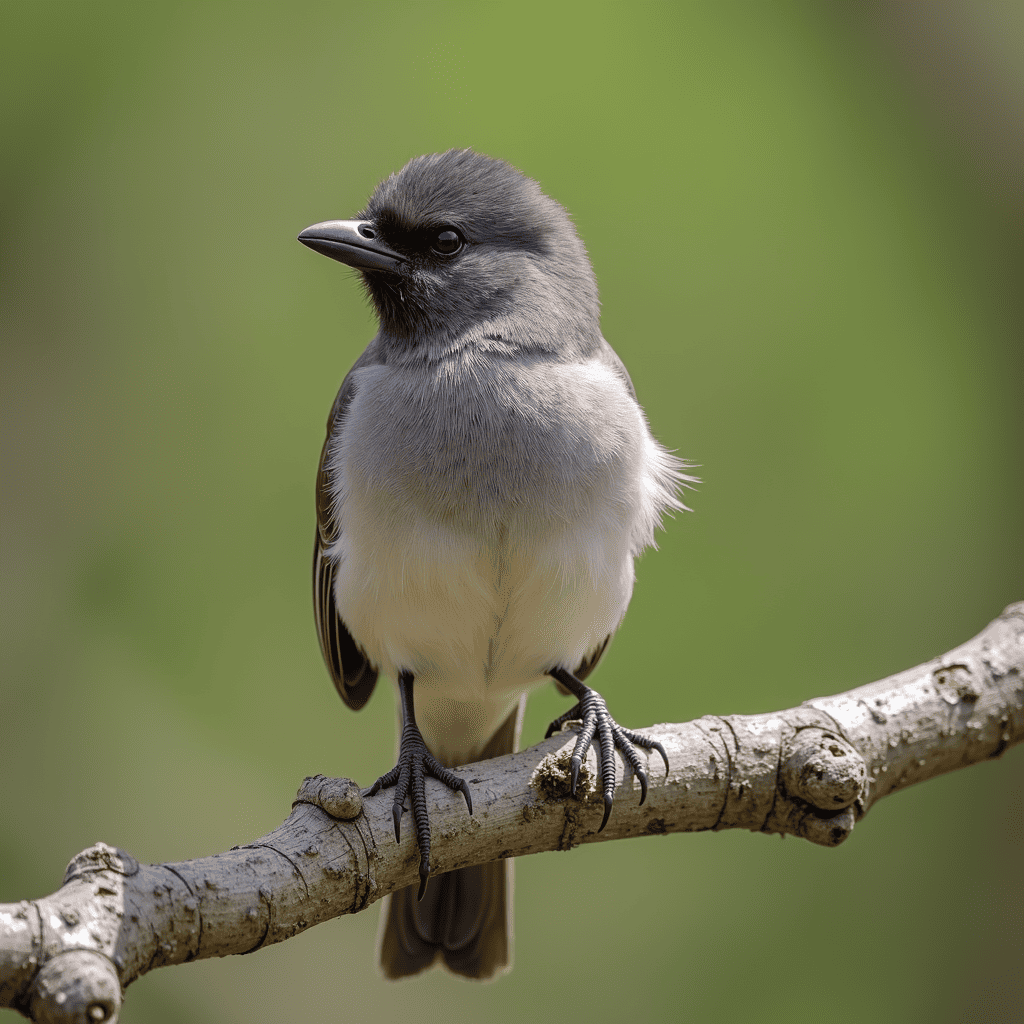} &
        \includegraphics[height=0.1\linewidth]{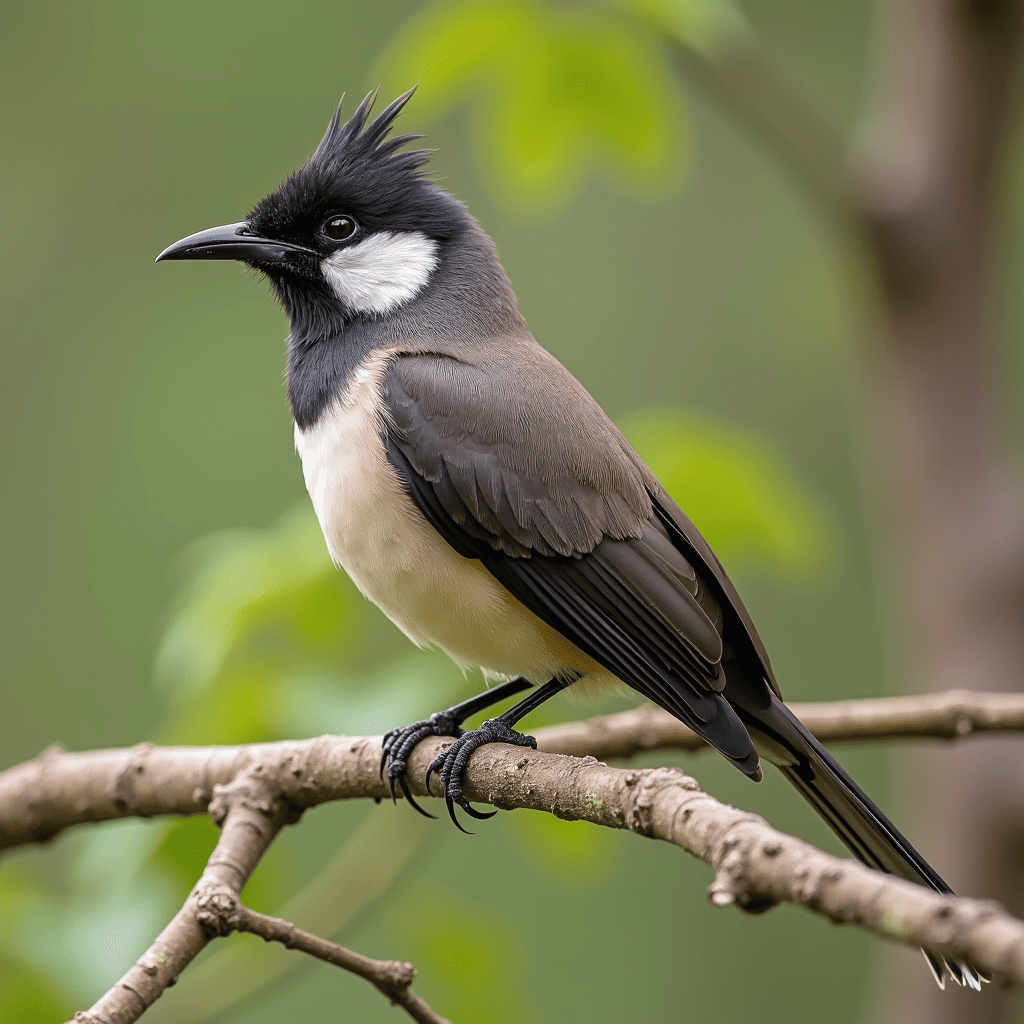} &
        \includegraphics[height=0.1\linewidth]{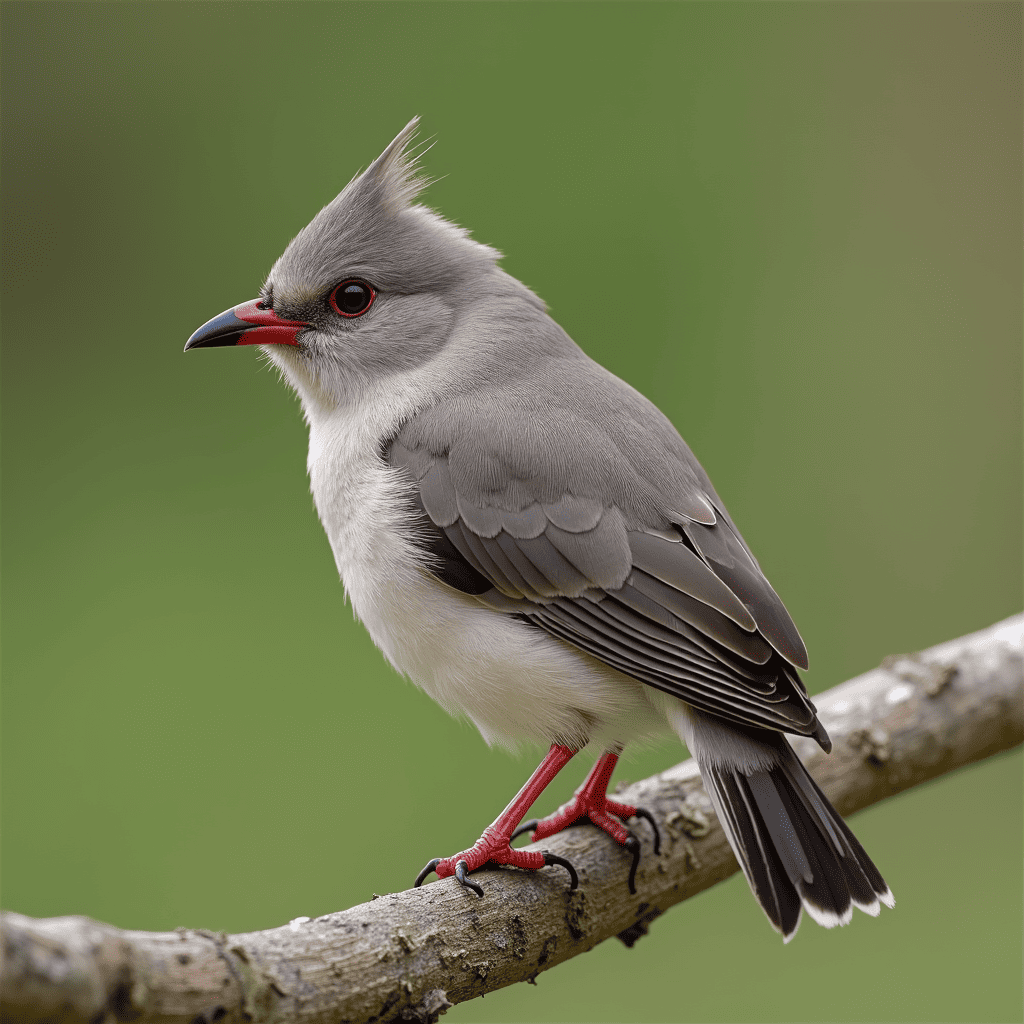} &
        \includegraphics[height=0.1\linewidth]{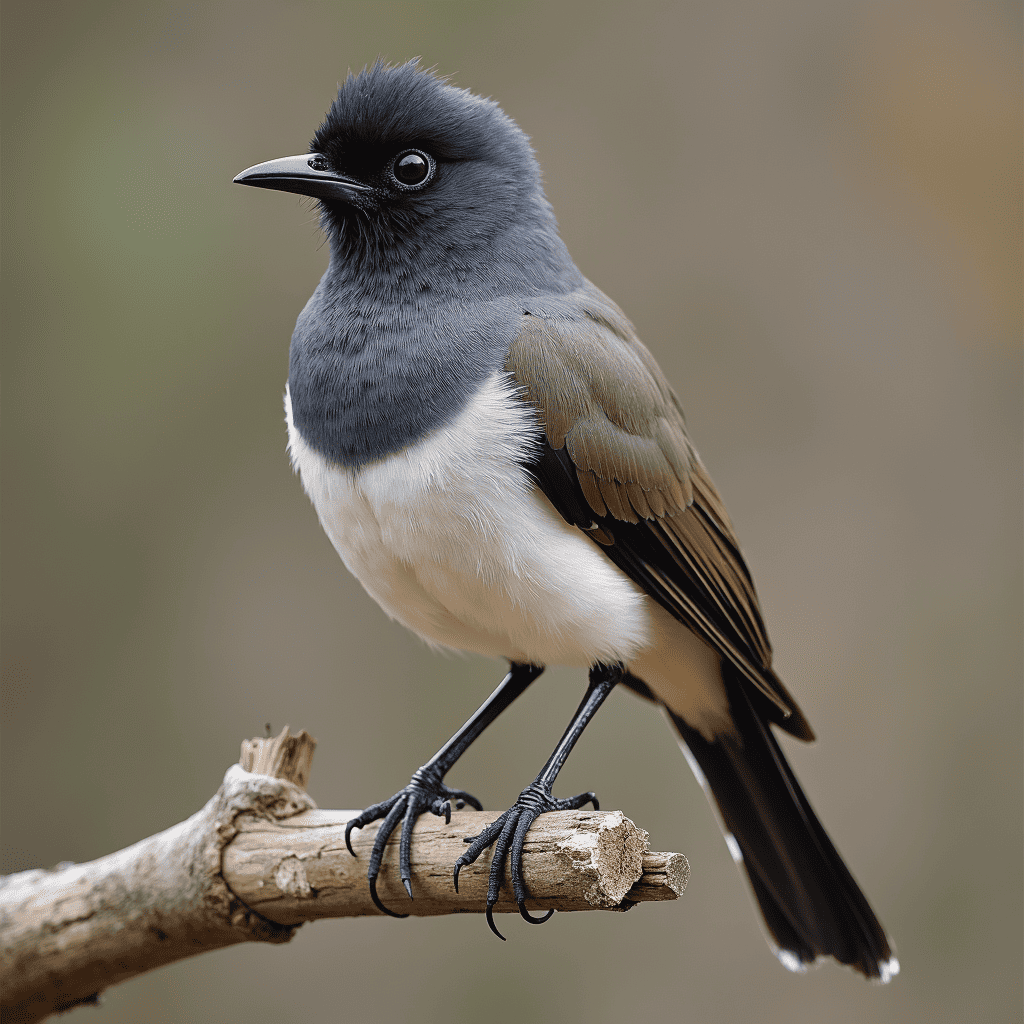} &
        \includegraphics[height=0.1\linewidth]{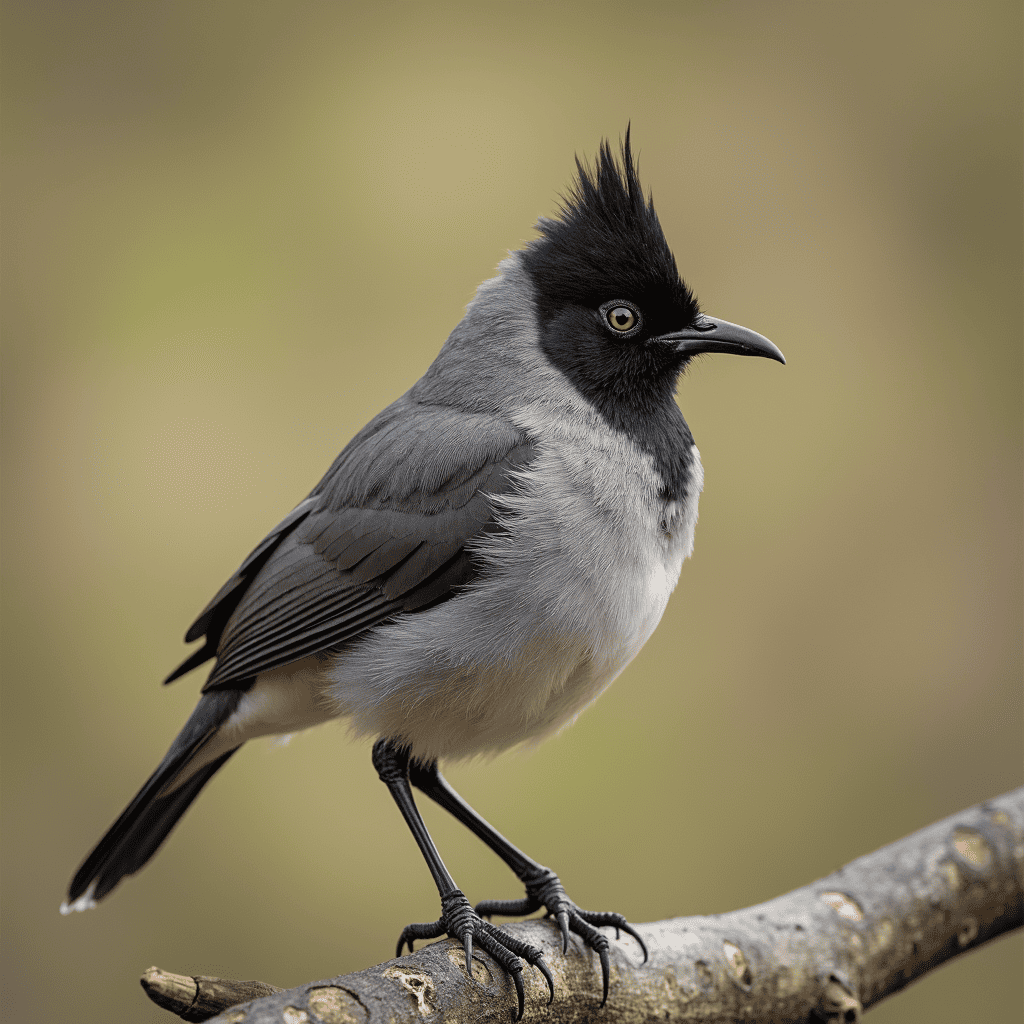} \\
    \end{tabular}

    \caption{Comparison between real and FLUX-generated images for selected ImageNet classes. Each row corresponds to a class, with the first column showing a real ImageNet image and the remaining columns showing generated datapoints.}
    \label{app-fig:flux-imagenet-images}
\end{figure}

We generated synthetic images using the FLUX.1 model~\citep{flux2024} from Black Forest Labs. For each class listed in \Cref{app-tab:acc_per_class}, we generated 2,000 images using the following configuration:
\begin{itemize}
\item \underline{Prompt}: ``A photo of a {class name}'', where {class name} refers to the corresponding ImageNet label, as shown to be effective in \citep{radford2021learning}.
\item \underline{Inference steps}: 50
\end{itemize}
Images were generated using the \texttt{FluxPipeline} from the \texttt{diffusers} library, utilizing NVIDIA GPUs for acceleration with mixed-precision (\texttt{float16}) computation. The generation process was parallelized across multiple GPUs. 
\Cref{app-fig:flux-imagenet-images} presents examples of generated images alongside real ImageNet training images from the same class.

\begin{table}[!h]
\caption{Per-class accuracies of the pre-trained CLIP model with a ViT backbone on ImageNet. The first two columns (Top-1 and Top-2 accuracy) are computed over all ImageNet classes, while the last two columns are computed only over the subset of classes shown in this table.}
\label{app-tab:acc_per_class}
\centering
\begin{tabular}{lllll}
\toprule
Class & Top-1 (\%) & Top-2 (\%) & Top-1 (\%) & Top-2 (\%) \\
\midrule
Junco, snowbird & 91.8 & 95.1 & 94.7 & 98.3 \\
Bulbul & 89.8 & 96.2 & 96 & 99.5 \\
Jay & 9.6 & 22 & 29 & 58.3 \\
Magpie & 88.2 & 93.2 & 94.5 & 97.8 \\
Golden retriever & 66.5 & 78.4 & 83.9 & 95.5 \\
Labrador retriever & 53.8 & 66.3 & 83.9 & 94.2 \\
English springer & 58.7 & 79.9 & 96.2 & 97.7 \\
Kuvasz & 65.5 & 83.5 & 93 & 97.9 \\
Siberian husky & 13.8 & 40 & 87.3 & 94.2 \\
Marmot & 47.5 & 66  & 75.4 & 98.5 \\
Beaver & 59.6 & 73.5 & 93.6 & 98.5 \\
Bicycle & 91.5 & 96.2 & 97.8 & 99.9 \\
Lighter, Light & 35.3 & 44.3 & 72.2 & 84.8 \\
Muzzle & 52.5 & 62.1 & 89.2 & 94.5 \\
Tennis ball & 65.4 & 76.8 & 88.4 & 93.8 \\
Torch & 44.8 & 60.7 & 85.2 & 95.2 \\
Unicycle & 66.3 & 80.5 & 83.2 & 96.5 \\
White wolf & 63.9 & 79.5 & 87.3 & 95.4 \\
Water ouzel & 88.9 & 93.5 & 93.1 & 96.1 \\
American robin & 87.2 & 92.3 & 94.5 & 98.2 \\
Admiral & 0.1 & 0.1 & 0.1 & 0.1 \\
Rock beauty & 4.6 & 28.9 & 66.6 & 89.5 \\
Papillon & 59.8 & 73 & 93.8 & 97.3 \\
Lycaenid butterfly & 70.1 & 92.8 & 95.3 & 99.4 \\
Gyromitra & 0.1 & 0.1 & 0.1 & 0.1 \\
Coral fungus & 78.9 & 91 & 87.6 & 98.8 \\
Stinkhorn & 65.3 & 79.2 & 87.1 & 97.7 \\
Barracouta & 1.4 & 7.1 & 4.4 & 41.8 \\
Garfish & 48.2 & 64.8 & 85.8 & 94.3 \\
Tinca tinca & 89.4 & 94.2 & 96.5 & 98.8 \\
\bottomrule
\end{tabular}
\end{table}

\FloatBarrier
\clearpage

\section{Additional ImageNet experiments}
\label{app-sec:imagenet-exp}
This section provides supplementary results that complement those in \Cref{sec:imagenet-exp}, including additional experiments on the ImageNet dataset. The following two subsections---\Cref{app-sec:exp-imagenet-gen,app-sec:exp-imagenet-clusters}---correspond to \Cref{sec:exp-imagenet-gen,sec:exp-imagenet-clusters} of the main manuscript, respectively, and follow the same experimental settings.

\subsection{Experiments with generated synthetic data}
\label{app-sec:exp-imagenet-gen}

\begin{figure}[!h]
    \begin{subfigure}[t]{\linewidth}
    \includegraphics[height=0.22\textwidth, valign=t]{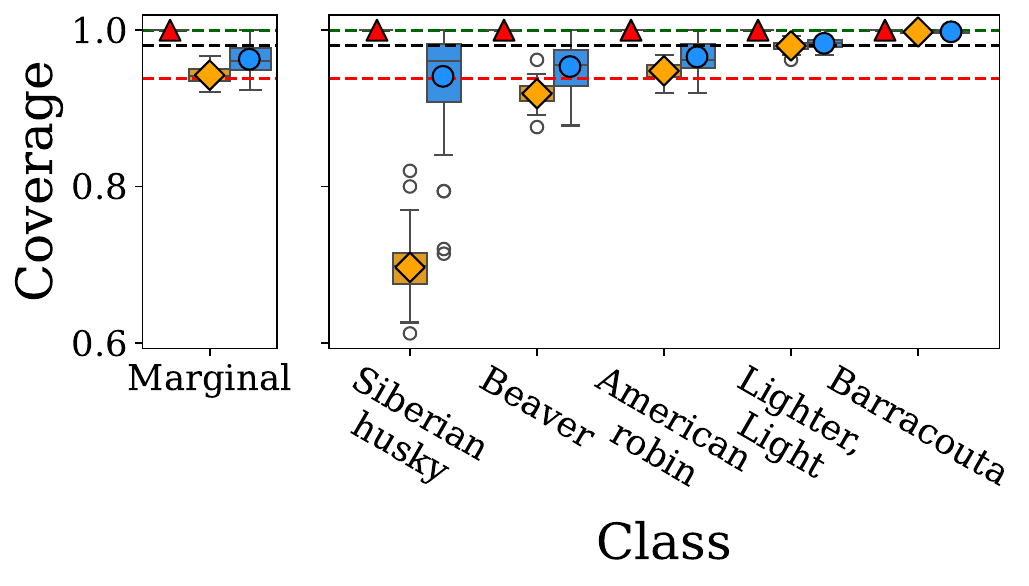}
    \includegraphics[height=0.22\textwidth, valign=t]{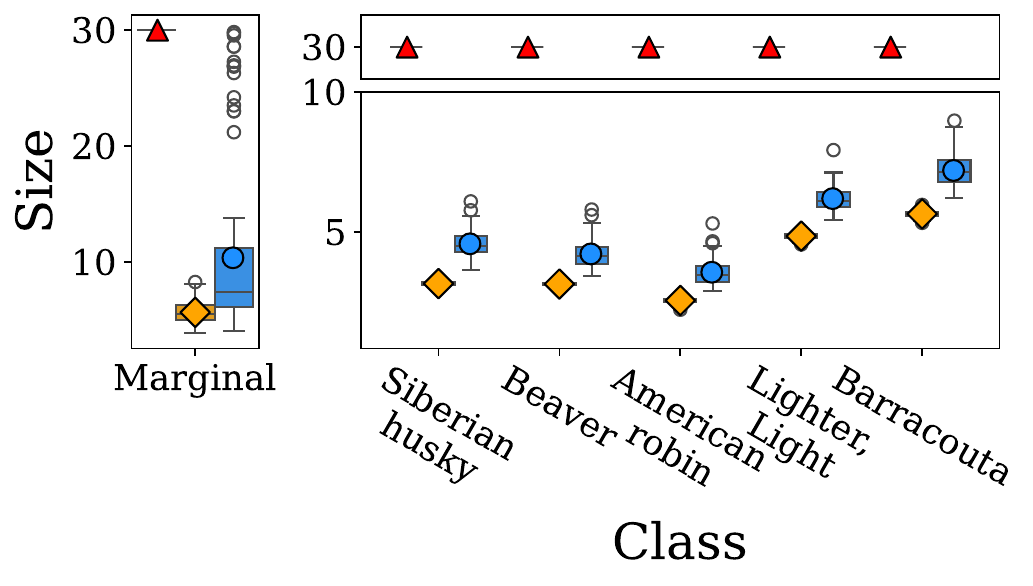}
    \caption{$\alpha=0.02$}
    \label{app-fig:imagenet-gen_0.02}
    \end{subfigure}
    \begin{subfigure}[t]{\linewidth}
    \includegraphics[height=0.22\textwidth, valign=t]{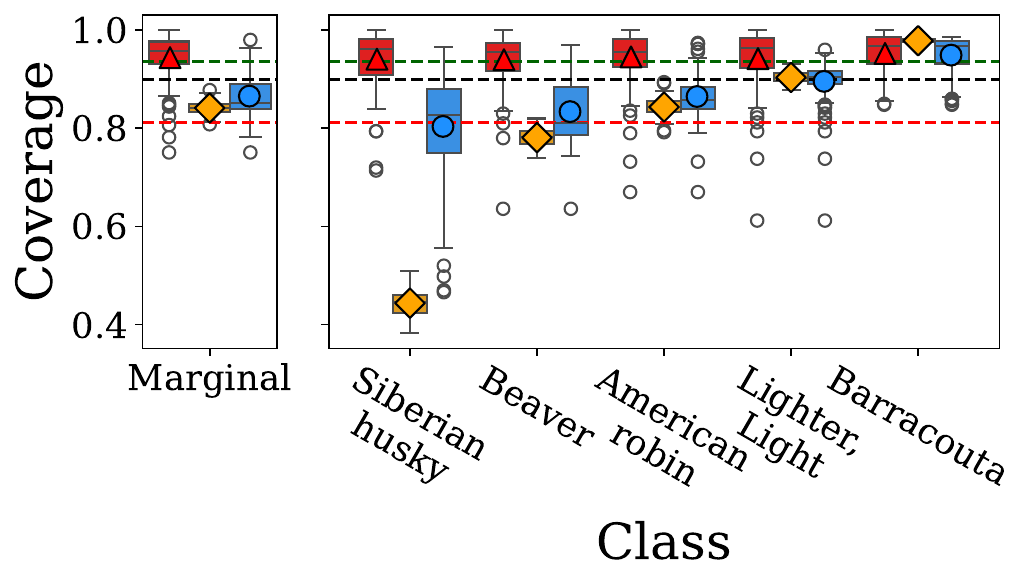}
    \includegraphics[height=0.22\textwidth, valign=t]{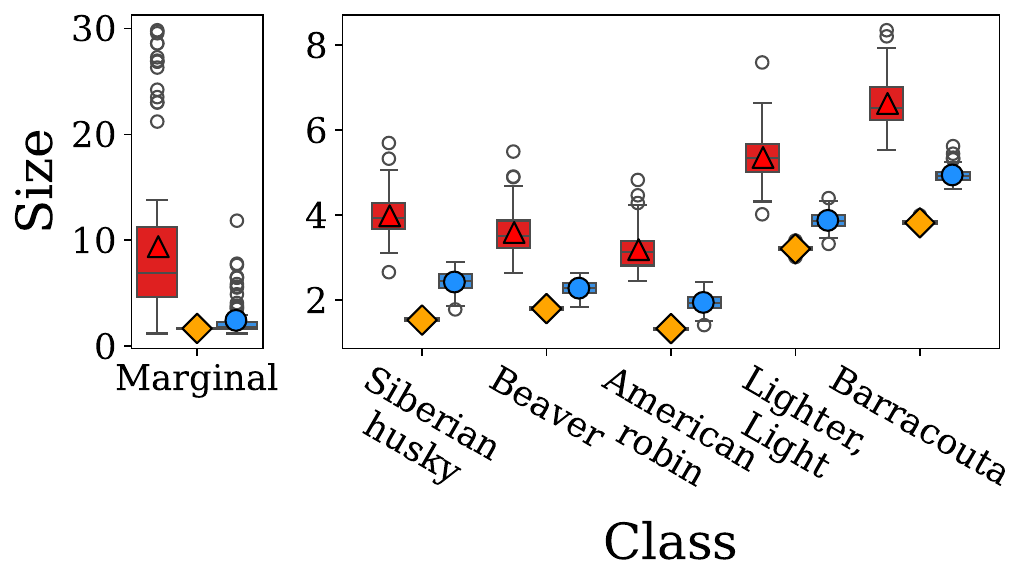}
    \includegraphics[height=0.22\textwidth, valign=t]{figures/results/legend_spi.pdf}
    \caption{$\alpha=0.1$}
    \label{app-fig:imagenet-gen_0.1}
    \end{subfigure}
    \caption{Results for the ImageNet data:
     Coverage rates of $\cp$, $\cpsynt$, and $\scp$ run at level $\alpha = 0.02$ (a) and $0.1$ (b), averaged over 100 trials. Left: Average coverage. Right: Average prediction set size, both under marginal (leftmost box in each group) and label-conditional coverage settings.  Label-conditional results are shown for selected classes; see~\Cref{app-tab:imagenet-gen-all-classes_0.02,app-tab:imagenet-gen-all-classes_0.1} for results across all classes.}
    \label{app-fig:imagenet-gen}
\end{figure}

\Cref{app-fig:imagenet-gen} presents the performance under both marginal and label-conditional guarantees at levels $\alpha=0.02$ and $0.1$. We observe a similar trend to that seen in~\Cref{fig:imagenet-gen_0.05}.
Following that figure, we can see that the standard conformal prediction method, $\cp$, controls the coverage at the $ 1-\alpha$ level as expected, but it produces overly conservative prediction sets.
$\cpsynt$ method fails to achieve the target coverage level of $1 - \alpha$, under-covering some classes, while in others, it becomes overly conservative, depending on the unknown distribution shift between the real and synthetic data.
In contrast, the proposed method, $\scp$, stays within the theoretical bounds and produces informative prediction sets.

\Cref{app-tab:imagenet-gen-all-classes_0.05,app-tab:imagenet-gen-all-classes_0.02,app-tab:imagenet-gen-all-classes_0.1} present results for all 30 classes in the real calibration set corresponding to \Cref{fig:imagenet-gen_0.05} in the main manuscript and \Cref{app-fig:imagenet-gen_0.02,app-fig:imagenet-gen_0.1} above.

\begin{table}[!h]
\caption{Per-class conditional coverage (in \%) and prediction set size for each method, computed over 100 trials. Standard errors are shown in parentheses. The target coverage level is $1-\alpha = 0.98$. The theoretical coverage guarantees for $\scp$ are in the range  $[93.7, 100]$. Other details are as in~\Cref{app-fig:imagenet-gen}.}
\label{app-tab:imagenet-gen-all-classes_0.02}
\centering
\begin{tabular}{lp{1.3cm}p{1.6cm}p{1.6cm}p{1.1cm}p{1.2cm}p{1.5cm}}
\toprule
& \multicolumn{3}{c}{Coverage (\%)} & \multicolumn{3}{c}{Size} \\
\toprule
Class & \texttt{Only Real} & \texttt{Only Synth} & \texttt{SPI} & \texttt{Only Real} & \texttt{Only Synth} & \texttt{SPI} \\
\midrule
Admiral & 100 (± 0) & 6.9 (± 0.3) & 93.6 (± 0.6) & 30 (± 0) & 4.5 (± 0) & 6.1 (± 0) \\
American robin & 100 (± 0) & 94.8 (± 0.1) & 96.6 (± 0.2) & 30 (± 0) & 2.6 (± 0) & 3.6 (± 0) \\
Barracouta & 100 (± 0) & 99.8 (± 0) & 99.8 (± 0) & 30 (± 0) & 5.6 (± 0) & 7.2 (± 0.1) \\
Beaver & 100 (± 0) & 91.9 (± 0.1) & 95.3 (± 0.3) & 30 (± 0) & 3.1 (± 0) & 4.2 (± 0) \\
Bicycle & 100 (± 0) & 96.3 (± 0.1) & 97.3 (± 0.1) & 30 (± 0) & 2.8 (± 0) & 3.5 (± 0) \\
Bulbul & 100 (± 0) & 99.4 (± 0) & 99.5 (± 0) & 30 (± 0) & 2.6 (± 0) & 3.7 (± 0) \\
Coral fungus & 100 (± 0) & 99.4 (± 0) & 99.4 (± 0) & 30 (± 0) & 2.7 (± 0) & 3.3 (± 0) \\
English springer & 100 (± 0) & 94.2 (± 0.1) & 96.5 (± 0.2) & 30 (± 0) & 2.9 (± 0) & 4.1 (± 0) \\
Garfish & 100 (± 0) & 92.1 (± 0.1) & 95.3 (± 0.3) & 30 (± 0) & 4.6 (± 0) & 5.9 (± 0) \\
Golden retriever & 100 (± 0) & 94.3 (± 0.1) & 96.4 (± 0.2) & 30 (± 0) & 4.3 (± 0) & 5.8 (± 0.1) \\
Gyromitra & 100 (± 0) & 92.4 (± 0.2) & 96.4 (± 0.3) & 30 (± 0) & 3.1 (± 0) & 3.7 (± 0) \\
Jay & 100 (± 0) & 92.0 (± 0.2) & 95.6 (± 0.3) & 30 (± 0) & 6.4 (± 0) & 8.0 (± 0.1) \\
Junco, snowbird & 100 (± 0) & 97.5 (± 0.1) & 98.0 (± 0.1) & 30 (± 0) & 2.3 (± 0) & 3.1 (± 0) \\
Kuvasz & 100 (± 0) & 95.1 (± 0.1) & 96.5 (± 0.2) & 30 (± 0) & 2.8 (± 0) & 4.0 (± 0) \\
Labrador retriever & 100 (± 0) & 96.2 (± 0.1) & 97.2 (± 0.1) & 30 (± 0) & 4.6 (± 0) & 6.3 (± 0.1) \\
Lighter, Light & 100 (± 0) & 98.0 (± 0.1) & 98.3 (± 0.1) & 30 (± 0) & 4.9 (± 0) & 6.2 (± 0) \\
Lycaenid butterfly & 100 (± 0) & 93.5 (± 0.1) & 95.7 (± 0.2) & 30 (± 0) & 2.9 (± 0) & 4.1 (± 0.1) \\
Magpie & 100 (± 0) & 96.7 (± 0.1) & 97.4 (± 0.1) & 30 (± 0) & 2.5 (± 0) & 3.4 (± 0) \\
Marmot & 100 (± 0) & 95.5 (± 0.1) & 97.0 (± 0.2) & 30 (± 0) & 4.0 (± 0) & 5.6 (± 0.1) \\
Muzzle & 100 (± 0) & 97.5 (± 0) & 98.0 (± 0.1) & 30 (± 0) & 3.5 (± 0) & 4.9 (± 0) \\
Papillon & 100 (± 0) & 89.7 (± 0.2) & 94.9 (± 0.4) & 30 (± 0) & 3.1 (± 0) & 4.3 (± 0) \\
Rock beauty & 100 (± 0) & 90.6 (± 0.2) & 95.3 (± 0.3) & 30 (± 0) & 4.8 (± 0) & 6.0 (± 0) \\
Siberian husky & 100 (± 0) & 69.7 (± 0.3) & 94.1 (± 0.6) & 30 (± 0) & 3.2 (± 0) & 4.6 (± 0) \\
Stinkhorn & 100 (± 0) & 97.9 (± 0.1) & 98.2 (± 0.1) & 30 (± 0) & 4.6 (± 0) & 5.5 (± 0) \\
Tennis ball & 100 (± 0) & 97.5 (± 0.1) & 98.0 (± 0.1) & 30 (± 0) & 3.1 (± 0) & 4.0 (± 0) \\
Tinca tinca & 100 (± 0) & 98.5 (± 0) & 98.6 (± 0.1) & 30 (± 0) & 3.3 (± 0) & 4.2 (± 0) \\
Torch & 100 (± 0) & 98.3 (± 0) & 98.7 (± 0.1) & 30 (± 0) & 5.6 (± 0) & 7.1 (± 0.1) \\
Unicycle & 100 (± 0) & 96.3 (± 0.1) & 97.2 (± 0.1) & 30 (± 0) & 4.2 (± 0) & 5.5 (± 0) \\
Water ouzel & 100 (± 0) & 95.7 (± 0.1) & 97.0 (± 0.2) & 30 (± 0) & 2.5 (± 0) & 3.3 (± 0) \\
White wolf & 100 (± 0) & 85.5 (± 0.2) & 94.3 (± 0.5) & 30 (± 0) & 2.9 (± 0) & 4.0 (± 0) \\
\bottomrule
\end{tabular}
\end{table}

\begin{table}[!h]
\caption{Per-class conditional coverage (in \%) and prediction set size for each method, computed over 100 trials. Standard errors are shown in parentheses.~The target coverage level is $1-\alpha = 0.95$. The theoretical coverage guarantees for $\scp$ are in the range $[93.7, 100]$. Other details are as in~\Cref{fig:imagenet-gen_0.05}.}
\label{app-tab:imagenet-gen-all-classes_0.05}
\centering
\begin{tabular}{lp{1.3cm}p{1.6cm}p{1.6cm}p{1.1cm}p{1.2cm}p{1.5cm}}
\toprule
& \multicolumn{3}{c}{Coverage (\%)} & \multicolumn{3}{c}{Size} \\
\toprule
Class & \texttt{Only Real} & \texttt{Only Synth} & \texttt{SPI} & \texttt{Only Real} & \texttt{Only Synth} & \texttt{SPI} \\
\midrule
Admiral & 100 (± 0) & 0.6 (± 0) & 93.6 (± 0.6) & 30 (± 0) & 3.6 (± 0) & 5.8 (± 0) \\
American robin & 100 (± 0) & 90.5 (± 0.2) & 95.4 (± 0.3) & 30 (± 0) & 1.8 (± 0) & 3.3 (± 0) \\
Barracouta & 100 (± 0) & 99.3 (± 0) & 99.4 (± 0) & 30 (± 0) & 4.5 (± 0) & 6.8 (± 0.1) \\
Beaver & 100 (± 0) & 85.1 (± 0.2) & 94.3 (± 0.5) & 30 (± 0) & 2.5 (± 0) & 3.9 (± 0) \\
Bicycle & 100 (± 0) & 92.5 (± 0.1) & 95.6 (± 0.3) & 30 (± 0) & 2.3 (± 0) & 3.3 (± 0) \\
Bulbul & 100 (± 0) & 98.4 (± 0.1) & 98.6 (± 0.1) & 30 (± 0) & 2.0 (± 0) & 3.4 (± 0) \\
Coral fungus & 100 (± 0) & 98.7 (± 0) & 98.8 (± 0.1) & 30 (± 0) & 2.1 (± 0) & 3.1 (± 0) \\
English springer & 100 (± 0) & 90.1 (± 0.1) & 95.2 (± 0.4) & 30 (± 0) & 2.2 (± 0) & 3.8 (± 0) \\
Garfish & 100 (± 0) & 86.4 (± 0.1) & 94.2 (± 0.4) & 30 (± 0) & 3.7 (± 0) & 5.6 (± 0) \\
Golden retriever & 100 (± 0) & 88.8 (± 0.2) & 94.9 (± 0.4) & 30 (± 0) & 3.3 (± 0) & 5.3 (± 0.1) \\
Gyromitra & 100 (± 0) & 80.3 (± 0.3) & 95.1 (± 0.5) & 30 (± 0) & 2.7 (± 0) & 3.6 (± 0) \\
Jay & 100 (± 0) & 80.4 (± 0.2) & 93.7 (± 0.6) & 30 (± 0) & 4.8 (± 0) & 7.4 (± 0.1) \\
Junco, snowbird & 100 (± 0) & 94.7 (± 0.1) & 96.4 (± 0.2) & 30 (± 0) & 1.7 (± 0) & 2.9 (± 0) \\
Kuvasz & 100 (± 0) & 91.6 (± 0.1) & 95.1 (± 0.3) & 30 (± 0) & 2.1 (± 0) & 3.7 (± 0) \\
Labrador retriever & 100 (± 0) & 91.9 (± 0.1) & 95.6 (± 0.3) & 30 (± 0) & 3.6 (± 0) & 5.8 (± 0.1) \\
Lighter, Light & 100 (± 0) & 95.2 (± 0.1) & 96.8 (± 0.2) & 30 (± 0) & 3.9 (± 0) & 5.8 (± 0) \\
Lycaenid butterfly & 100 (± 0) & 88.2 (± 0.1) & 94.5 (± 0.4) & 30 (± 0) & 2.3 (± 0) & 4.0 (± 0.1) \\
Magpie & 100 (± 0) & 93.6 (± 0.1) & 95.7 (± 0.2) & 30 (± 0) & 1.9 (± 0) & 3.1 (± 0) \\
Marmot & 100 (± 0) & 93.0 (± 0.1) & 96.2 (± 0.3) & 30 (± 0) & 3.2 (± 0) & 5.2 (± 0.1) \\
Muzzle & 100 (± 0) & 95.7 (± 0.1) & 96.9 (± 0.2) & 30 (± 0) & 2.8 (± 0) & 4.6 (± 0) \\
Papillon & 100 (± 0) & 83.6 (± 0.2) & 94.1 (± 0.5) & 30 (± 0) & 2.3 (± 0) & 4.0 (± 0) \\
Rock beauty & 100 (± 0) & 68.5 (± 0.4) & 94.2 (± 0.5) & 30 (± 0) & 3.5 (± 0) & 5.5 (± 0) \\
Siberian husky & 100 (± 0) & 56.7 (± 0.3) & 94.1 (± 0.6) & 30 (± 0) & 2.3 (± 0) & 4.2 (± 0) \\
Stinkhorn & 100 (± 0) & 95.6 (± 0.1) & 96.6 (± 0.2) & 30 (± 0) & 3.6 (± 0) & 5.1 (± 0) \\
Tennis ball & 100 (± 0) & 94.3 (± 0.1) & 96.0 (± 0.2) & 30 (± 0) & 2.5 (± 0) & 3.7 (± 0) \\
Tinca tinca & 100 (± 0) & 96.5 (± 0.1) & 97.2 (± 0.1) & 30 (± 0) & 2.6 (± 0) & 4.0 (± 0) \\
Torch & 100 (± 0) & 96.7 (± 0.1) & 97.6 (± 0.1) & 30 (± 0) & 4.5 (± 0) & 6.6 (± 0.1) \\
Unicycle & 100 (± 0) & 92.9 (± 0.1) & 95.7 (± 0.3) & 30 (± 0) & 3.3 (± 0) & 5.1 (± 0) \\
Water ouzel & 100 (± 0) & 92.6 (± 0.1) & 95.7 (± 0.3) & 30 (± 0) & 1.8 (± 0) & 3.1 (± 0) \\
White wolf & 100 (± 0) & 80.5 (± 0.2) & 93.9 (± 0.6) & 30 (± 0) & 2.1 (± 0) & 3.7 (± 0) \\
\bottomrule
\end{tabular}
\end{table}

\begin{table}[!h]
\caption{Per-class conditional coverage (in \%) and prediction set size for each method, computed over 100 trials. Standard errors are shown in parentheses.~The target coverage level is $1-\alpha = 0.9$. The theoretical coverage guarantees for $\scp$ are in the range $[81.2, 93.7]$. Other details as in~\Cref{app-fig:imagenet-gen}.}
\label{app-tab:imagenet-gen-all-classes_0.1}
\centering
\begin{tabular}{lp{1.6cm}p{1.6cm}p{1.6cm}p{1.5cm}p{1.2cm}p{1.2cm}}
\toprule
& \multicolumn{3}{c}{Coverage (\%)} & \multicolumn{3}{c}{Size} \\
\toprule
Class & \texttt{Only Real} & \texttt{Only Synth} & \texttt{SPI} & \texttt{Only Real} & \texttt{Only Synth} & \texttt{SPI} \\

\midrule
Admiral & 93.6 (± 0.6) & 0.2 (± 0) & 81.3 (± 0.9) & 5.6 (± 0) & 3.1 (± 0) & 4.3 (± 0) \\
American robin & 94.5 (± 0.5) & 84.3 (± 0.2) & 86.5 (± 0.5) & 3.2 (± 0) & 1.3 (± 0) & 1.9 (± 0) \\
Barracouta & 95.3 (± 0.4) & 97.8 (± 0) & 94.9 (± 0.4) & 6.6 (± 0.1) & 3.8 (± 0) & 4.9 (± 0) \\
Beaver & 94.0 (± 0.6) & 78.1 (± 0.2) & 83.4 (± 0.6) & 3.6 (± 0.1) & 1.8 (± 0) & 2.3 (± 0) \\
Bicycle & 94.2 (± 0.5) & 87.1 (± 0.2) & 88.2 (± 0.4) & 3.2 (± 0) & 2.0 (± 0) & 2.3 (± 0) \\
Bulbul & 93.8 (± 0.6) & 95.8 (± 0.1) & 92.8 (± 0.6) & 3.3 (± 0) & 1.6 (± 0) & 2.1 (± 0) \\
Coral fungus & 93.5 (± 0.6) & 97.7 (± 0.1) & 93.2 (± 0.6) & 2.9 (± 0) & 1.7 (± 0) & 2.1 (± 0) \\
English springer & 93.9 (± 0.6) & 83.1 (± 0.2) & 85.3 (± 0.4) & 3.6 (± 0) & 1.5 (± 0) & 2.2 (± 0) \\
Garfish & 93.5 (± 0.6) & 79.2 (± 0.2) & 83.4 (± 0.5) & 5.4 (± 0.1) & 3.2 (± 0) & 4.0 (± 0) \\
Golden retriever & 94.1 (± 0.6) & 82.4 (± 0.2) & 85.8 (± 0.5) & 5.0 (± 0.1) & 2.3 (± 0) & 3.1 (± 0) \\
Gyromitra & 95.0 (± 0.6) & 64.1 (± 0.3) & 85.2 (± 0.9) & 3.3 (± 0) & 2.3 (± 0) & 2.5 (± 0) \\
Jay & 93.3 (± 0.7) & 67.1 (± 0.3) & 80.9 (± 0.9) & 6.8 (± 0.1) & 3.7 (± 0) & 4.8 (± 0) \\
Junco, snowbird & 94.2 (± 0.5) & 90.0 (± 0.1) & 89.9 (± 0.4) & 2.7 (± 0) & 1.3 (± 0) & 1.7 (± 0) \\
Kuvasz & 93.4 (± 0.6) & 85.0 (± 0.2) & 86.3 (± 0.4) & 3.5 (± 0) & 1.5 (± 0) & 2.0 (± 0) \\
Labrador retriever & 93.5 (± 0.7) & 84.6 (± 0.2) & 85.9 (± 0.5) & 5.4 (± 0.1) & 2.7 (± 0) & 3.5 (± 0) \\
Lighter, Light & 94.2 (± 0.6) & 90.4 (± 0.1) & 89.6 (± 0.4) & 5.4 (± 0.1) & 3.2 (± 0) & 3.9 (± 0) \\
Lycaenid butterfly & 94.0 (± 0.5) & 81.3 (± 0.2) & 85.7 (± 0.5) & 3.9 (± 0.1) & 1.9 (± 0) & 2.5 (± 0) \\
Magpie & 93.5 (± 0.6) & 88.3 (± 0.2) & 88.3 (± 0.5) & 3.0 (± 0) & 1.4 (± 0) & 1.9 (± 0) \\
Marmot & 94.0 (± 0.6) & 89.8 (± 0.1) & 89.1 (± 0.4) & 4.9 (± 0.1) & 2.6 (± 0) & 3.1 (± 0) \\
Muzzle & 93.5 (± 0.6) & 92.1 (± 0.1) & 90.8 (± 0.5) & 4.3 (± 0) & 2.3 (± 0) & 3.0 (± 0) \\
Papillon & 93.8 (± 0.6) & 75.8 (± 0.2) & 82.4 (± 0.6) & 3.8 (± 0.1) & 1.7 (± 0) & 2.4 (± 0) \\
Rock beauty & 94.2 (± 0.5) & 44.6 (± 0.3) & 80.7 (± 1.0) & 5.1 (± 0.1) & 2.5 (± 0) & 3.8 (± 0) \\
Siberian husky & 94.1 (± 0.6) & 44.4 (± 0.3) & 80.4 (± 1.1) & 4.0 (± 0) & 1.5 (± 0) & 2.4 (± 0) \\
Stinkhorn & 93.4 (± 0.6) & 92.2 (± 0.1) & 90.7 (± 0.4) & 4.5 (± 0.1) & 2.8 (± 0) & 3.2 (± 0) \\
Tennis ball & 93.5 (± 0.6) & 88.6 (± 0.1) & 88.6 (± 0.3) & 3.5 (± 0) & 2.0 (± 0) & 2.5 (± 0) \\
Tinca tinca & 93.2 (± 0.6) & 93.2 (± 0.1) & 91.2 (± 0.5) & 3.8 (± 0) & 2.1 (± 0) & 2.7 (± 0) \\
Torch & 94.8 (± 0.5) & 93.9 (± 0.1) & 92.6 (± 0.4) & 6.2 (± 0.1) & 3.8 (± 0) & 4.6 (± 0) \\
Unicycle & 93.3 (± 0.6) & 86.5 (± 0.2) & 87.1 (± 0.4) & 4.8 (± 0) & 2.8 (± 0) & 3.3 (± 0) \\
Water ouzel & 94.3 (± 0.5) & 87.7 (± 0.1) & 88.3 (± 0.3) & 3.0 (± 0) & 1.4 (± 0) & 1.9 (± 0) \\
White wolf & 93.9 (± 0.6) & 74.1 (± 0.2) & 82.4 (± 0.7) & 3.4 (± 0) & 1.5 (± 0) & 2.1 (± 0) \\
\bottomrule
\end{tabular}
\end{table}

 \FloatBarrier
\subsubsection{The effect of the real calibration set size}
\label{app-sec:imagenet-gen-calib-size}
Here, we evaluate the performance of different methods as a function of the real calibration set size $m$, following the same setup described in~\Cref{sec:exp-imagenet-gen}. This parameter directly affects the performance of both the standard conformal prediction method, $\cp$, and our proposed method, $\scp$, including the theoretical bounds established in~\Cref{thm:worst_case}. In contrast, $\cpsynt$, which relies solely on the synthetic calibration set, is unaffected by changes in $m$. As such, it serves as a useful baseline for assessing how well the synthetic calibration set aligns with the real one.

\begin{figure}[!h]
    \centering
    \begin{subfigure}[t]{\linewidth}
    \includegraphics[height=0.22\textwidth]{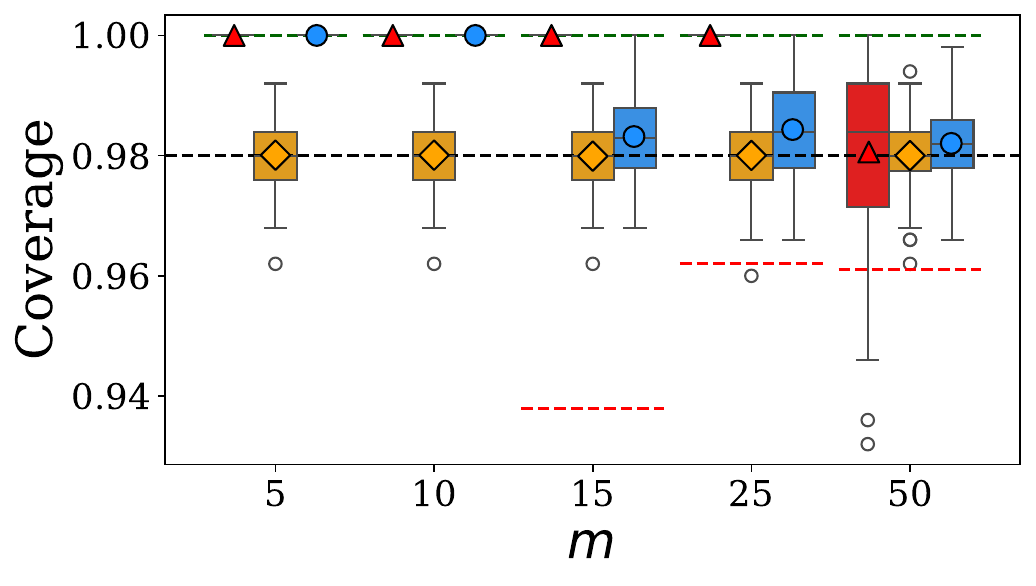}
    \includegraphics[height=0.22\textwidth]{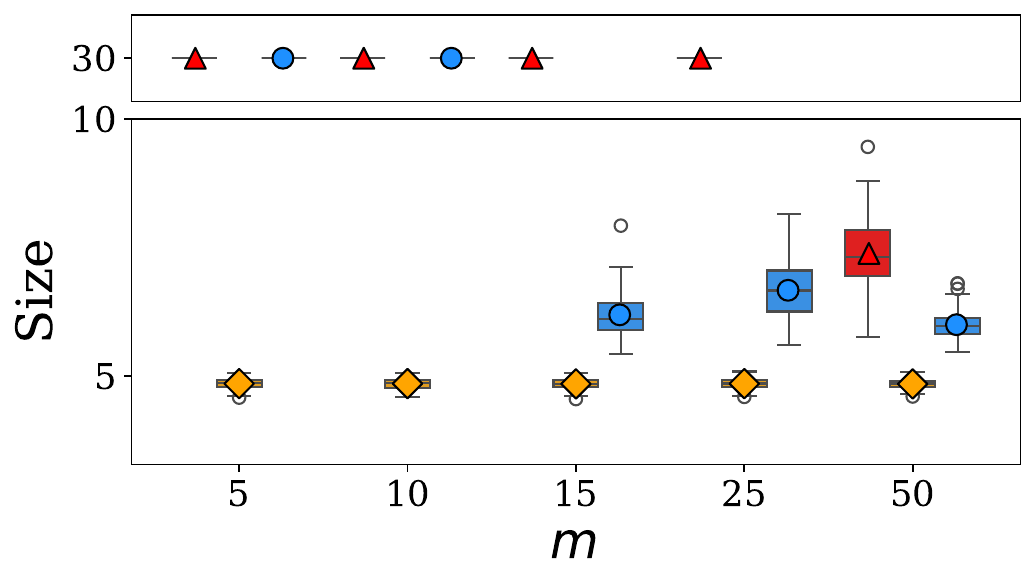}
    \caption{$\alpha=0.02$}
    \label{app-fig:imagenet-l-m-0.02}
    \end{subfigure}
    \begin{subfigure}[t]{\linewidth}
    \includegraphics[height=0.22\textwidth]{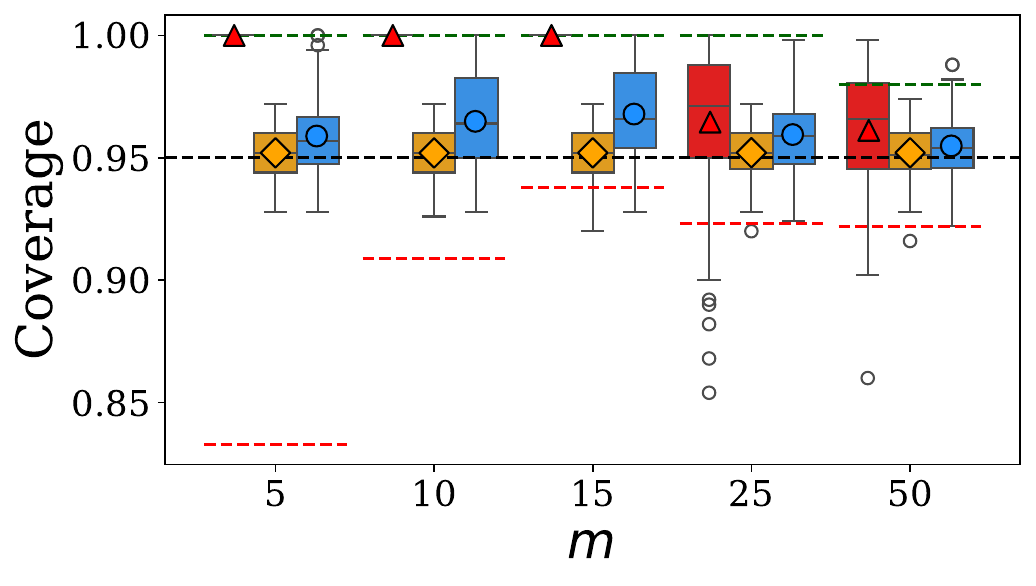}
    \includegraphics[height=0.22\textwidth]{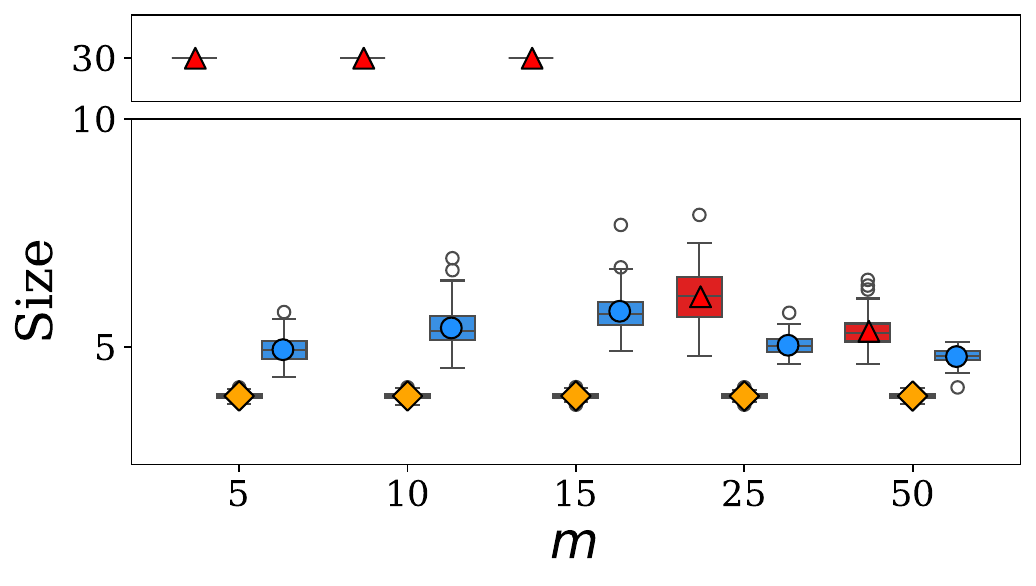}
    \caption{$\alpha=0.05$}
    \label{app-fig:imagenet-l-m-0.05}
    \end{subfigure}
    
    \begin{subfigure}[t]{\linewidth}
    \includegraphics[height=0.22\textwidth]{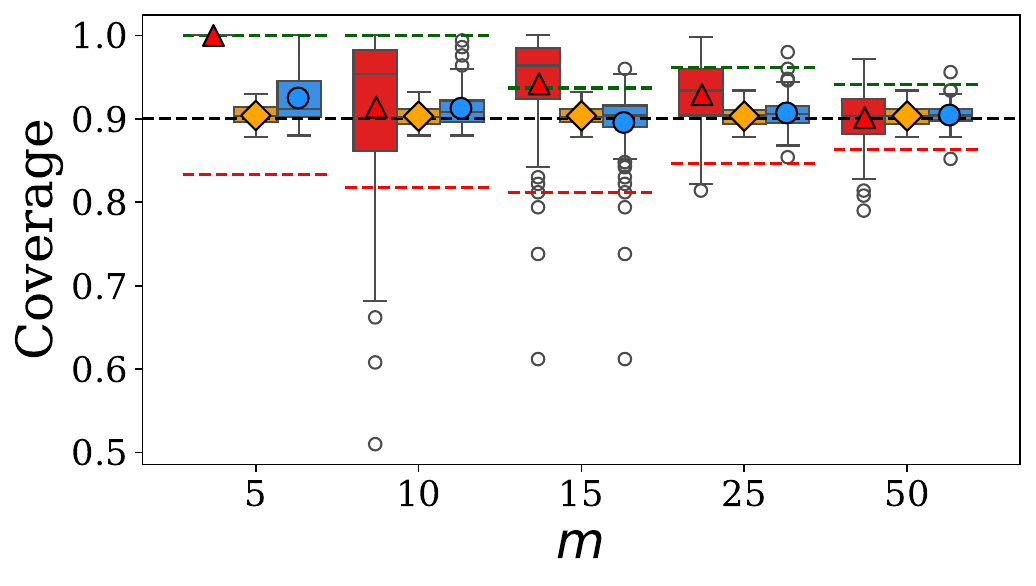}
    \includegraphics[height=0.22\textwidth]{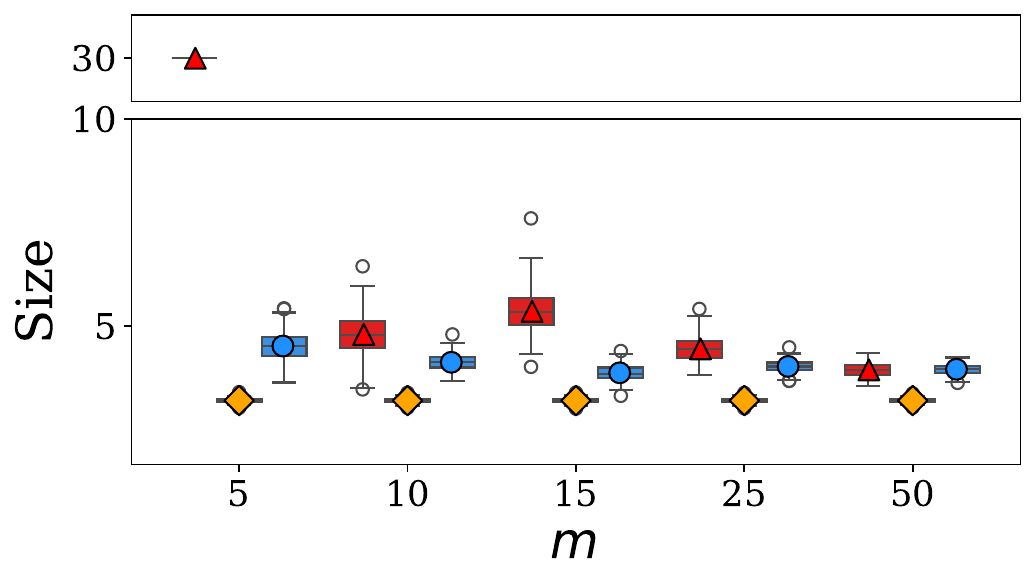}
    \includegraphics[height=0.22\linewidth]{figures/results/legend_spi.pdf}
    \caption{$\alpha=0.1$}
    \label{app-fig:imagenet-l-m-0.1}
    \end{subfigure}
    \caption{Results for the ImageNet data: Coverage rate for $\cp$, $\cpsynt$, and $\scp$ on the ``Lighter'' class as a function of the real calibration set size $m$, for levels $\alpha=0.02$ (a), $\alpha=0.05$ (b), and $\alpha=0.1$ (c).}
    \label{app-fig:imagenet-l-m}
\end{figure}

\Cref{app-fig:imagenet-l-m} presents the performance of all methods for the ``Lighter'' class across varying values of $m$ and $\alpha$ levels. Notably, although $\cpsynt$ does not have formal coverage guarantees, its empirical coverage closely matches the target level $1-\alpha$. This alignment suggests that the synthetic calibration data approximate the real distribution well.

\Cref{app-fig:imagenet-l-m-0.02} presents results for $\alpha=0.02$. At this low level, the standard conformal prediction, $\cp$, controls the coverage at level $1-\alpha$, but---as expected---produces trivial prediction sets when $m < 50$.

In contrast, our proposed method, $\scp$, achieves coverage within the theoretical bounds even for small $m$, with reduced variance in coverage and smaller prediction sets for $m\ge 15$. 
Interestingly, for $\alpha=0.02$ and $m=5$ or $10$, the theoretical lower and upper coverage bounds are both equal to unity, indicating that we know \emph{a priori} 
that the proposed method yields trivial prediction sets for this window construction.

For $\alpha=0.05$ and $\alpha=0.1$ (\Cref{app-fig:imagenet-l-m-0.05,app-fig:imagenet-l-m-0.1}, respectively), we observe similar trends. Our method, $\scp$, consistently achieves coverage within the theoretical bounds, remaining close to the target coverage level $1-\alpha$, while also exhibiting reduced variance in coverage and producing smaller, more informative prediction sets compared to the baseline, $\cp$.

\begin{figure}[!h]
    \centering
    \begin{subfigure}[t]{\linewidth}
    \includegraphics[height=0.22\textwidth]{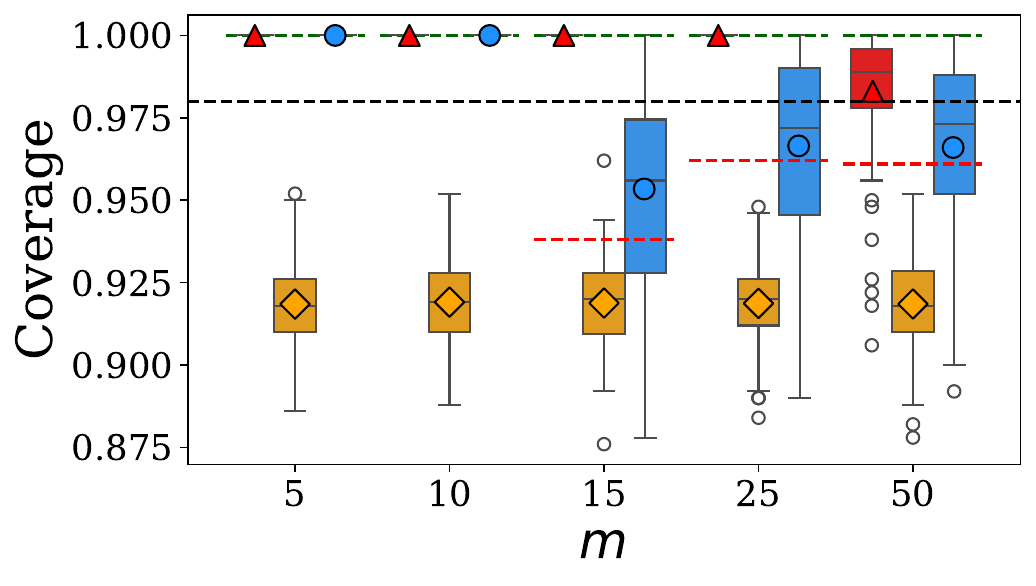}
    \includegraphics[height=0.22\textwidth]{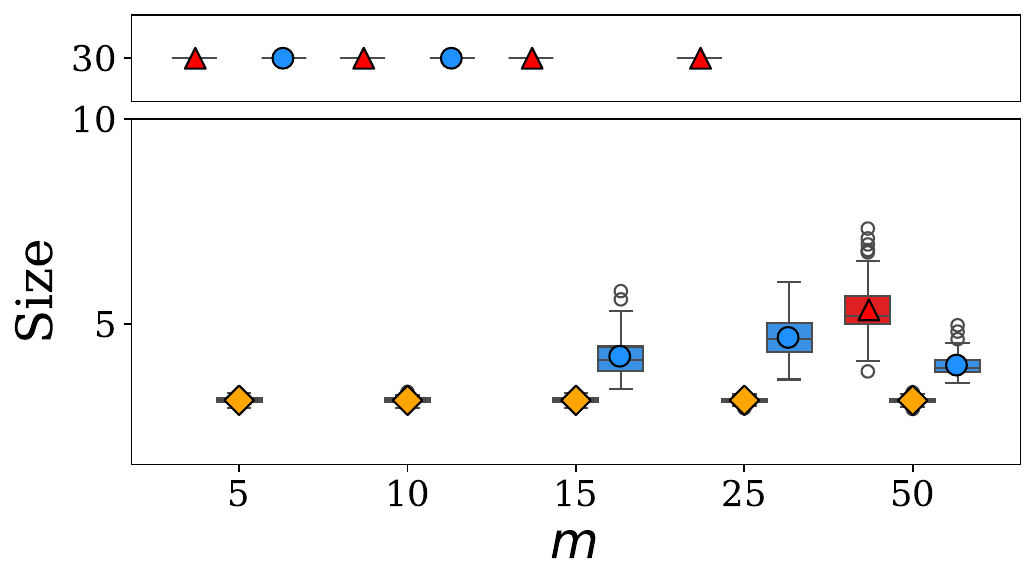}
    \caption{$\alpha=0.02$}
    \end{subfigure}
    \begin{subfigure}[t]{\linewidth}
    \includegraphics[height=0.22\textwidth]{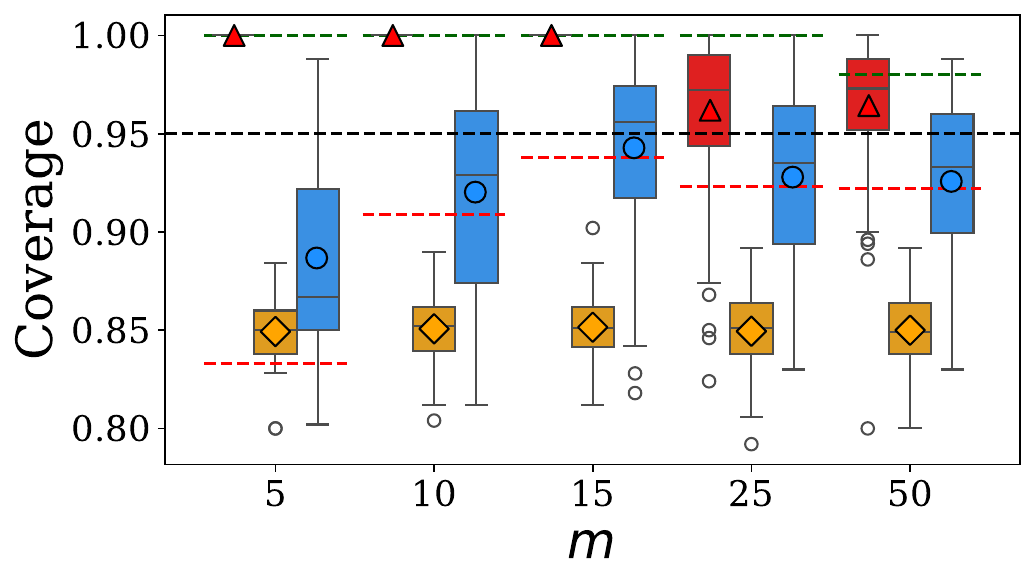}
    \includegraphics[height=0.22\textwidth]{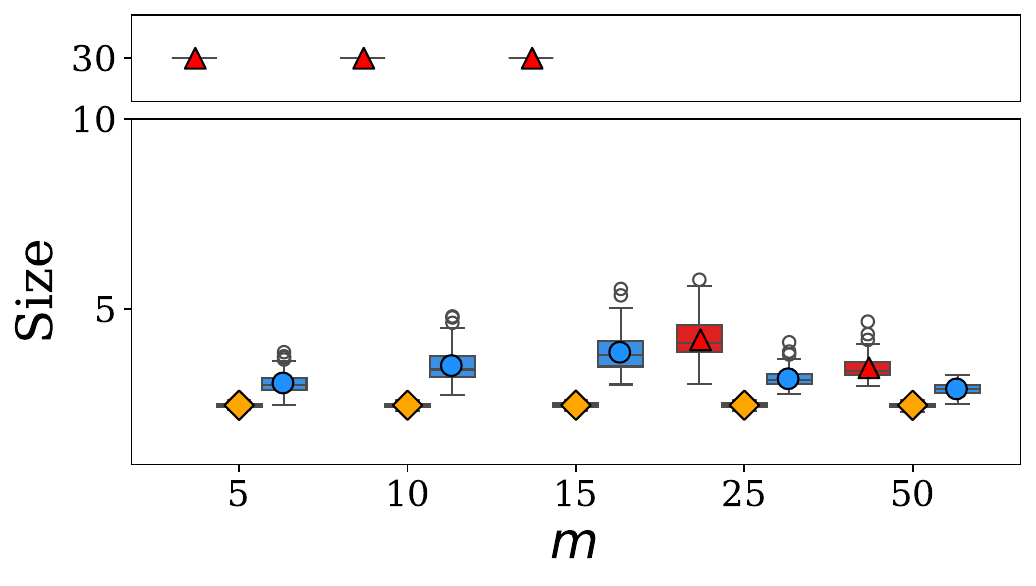}
    \caption{$\alpha=0.05$}
    \end{subfigure}
    \begin{subfigure}[t]{\linewidth}
    \includegraphics[height=0.22\textwidth]{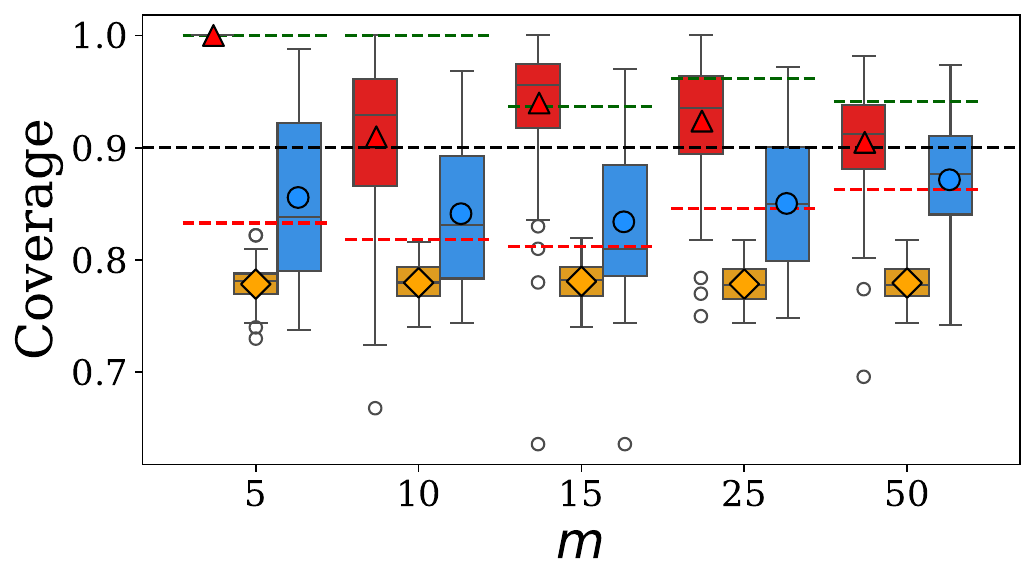}
    \includegraphics[height=0.22\textwidth]{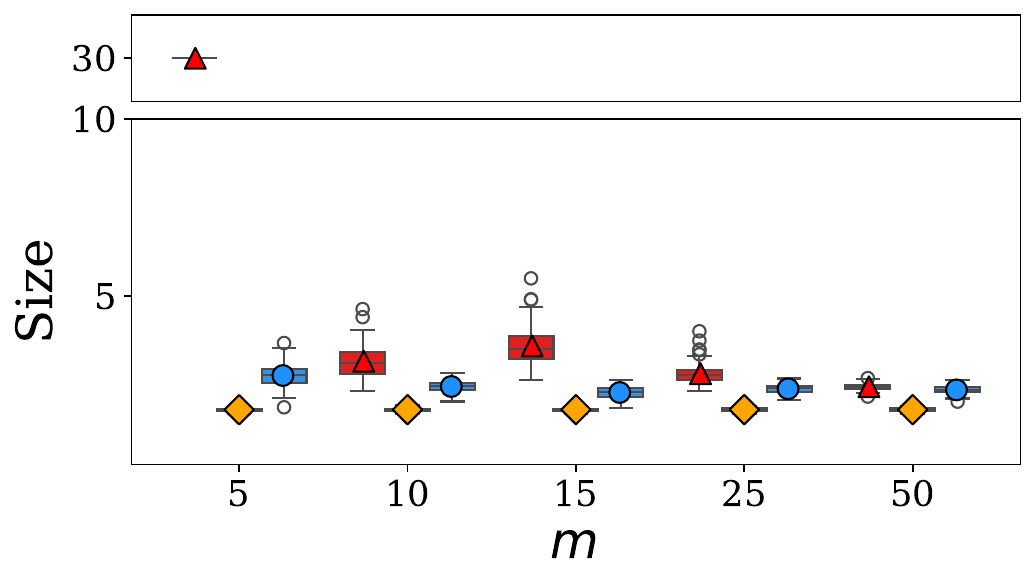}
    \includegraphics[height=0.22\linewidth]{figures/results/legend_spi.pdf}
    \caption{$\alpha=0.1$}
    \end{subfigure}
    \caption{Results for the ImageNet data: Coverage rate for $\cp$, $\cpsynt$, and $\scp$ on the Beaver class as a function of the real calibration set size $m$, for levels $\alpha=0.02$ (a), $\alpha=0.05$ (b), and $\alpha=0.1$ (c).}
    \label{app-fig:imagenet-b-m}
\end{figure}

Additionally, \Cref{app-fig:imagenet-b-m} presents the same experiment as in \Cref{app-fig:imagenet-l-m}, but for the ``Beaver'' class. In this case, the $\cpsynt$ method yields coverage that falls significantly below the target level $1-\alpha$, indicating that the synthetic calibration set differs substantially from the real data. Nevertheless, our proposed method, 
$\scp$, 
achieves coverage within the theoretical bounds across all $\alpha$ levels and calibration set sizes, while also producing informative prediction sets.

\FloatBarrier
\clearpage
\subsubsection{Results for \texorpdfstring{$\scp$}{SPI} with FLUX-generated synthetic data}
\label{app-sec:exp-imagenet-flux}

In this section, we evaluate the performance of the proposed method, $\scp$, using synthetic images generated by the FLUX.1 model~\citep{flux2024}. The experimental setup follows the same procedure described in~\Cref{sec:exp-imagenet-gen} of the main manuscript. As before, we aim for both marginal and label-conditional coverage guarantees.

\begin{figure}[!h]
    \centering
    \begin{subfigure}[t]{\linewidth}
    \includegraphics[height=0.22\textwidth]{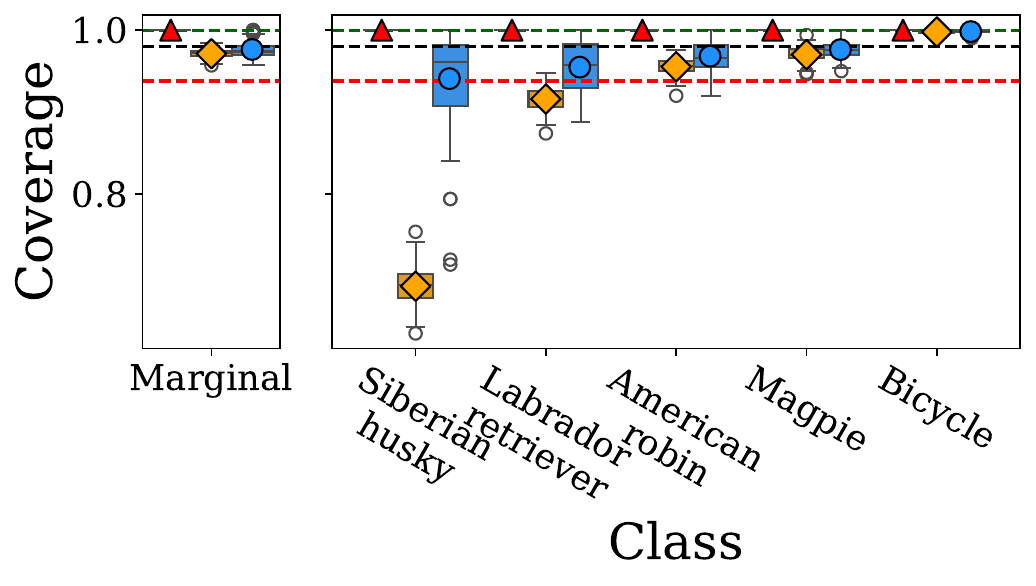}
    \includegraphics[height=0.22\textwidth]{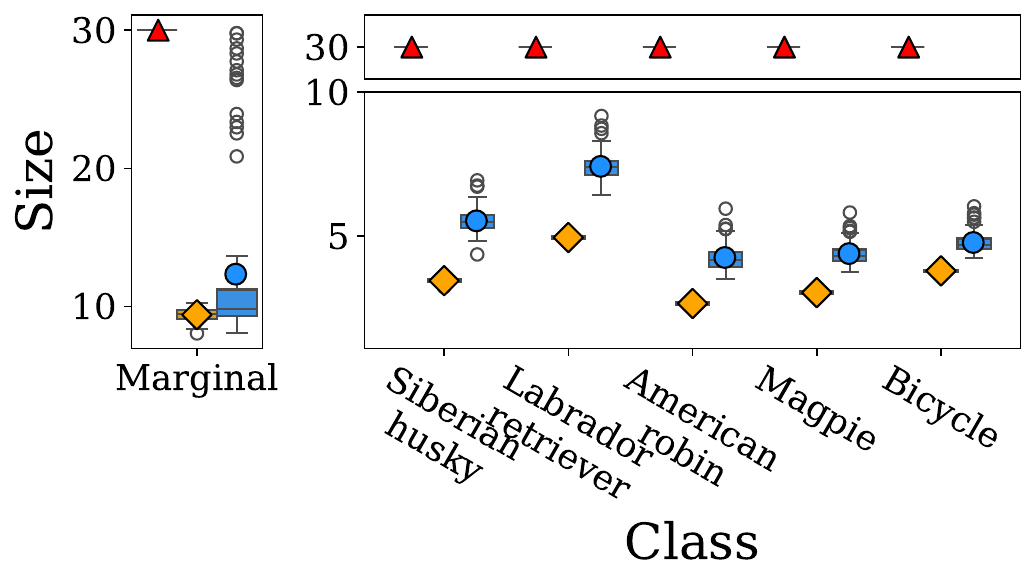}
    \caption{$\alpha=0.02$}
    \end{subfigure}
    \begin{subfigure}[t]{\linewidth}
    \includegraphics[height=0.22\textwidth]{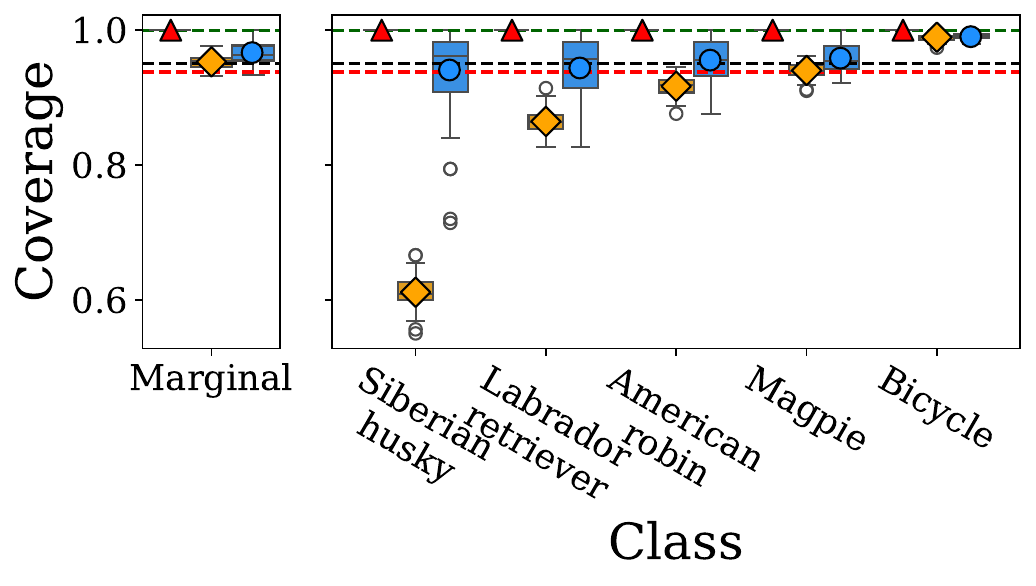}
    \includegraphics[height=0.22\textwidth]{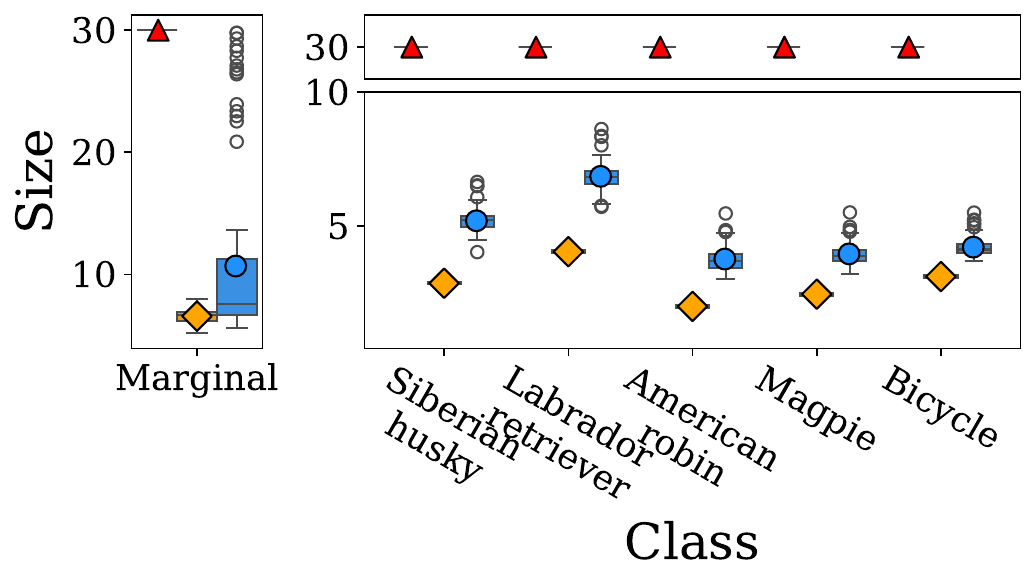}
    \caption{$\alpha=0.05$}
    \end{subfigure}
    \begin{subfigure}[t]{\linewidth}
    \includegraphics[height=0.22\textwidth]{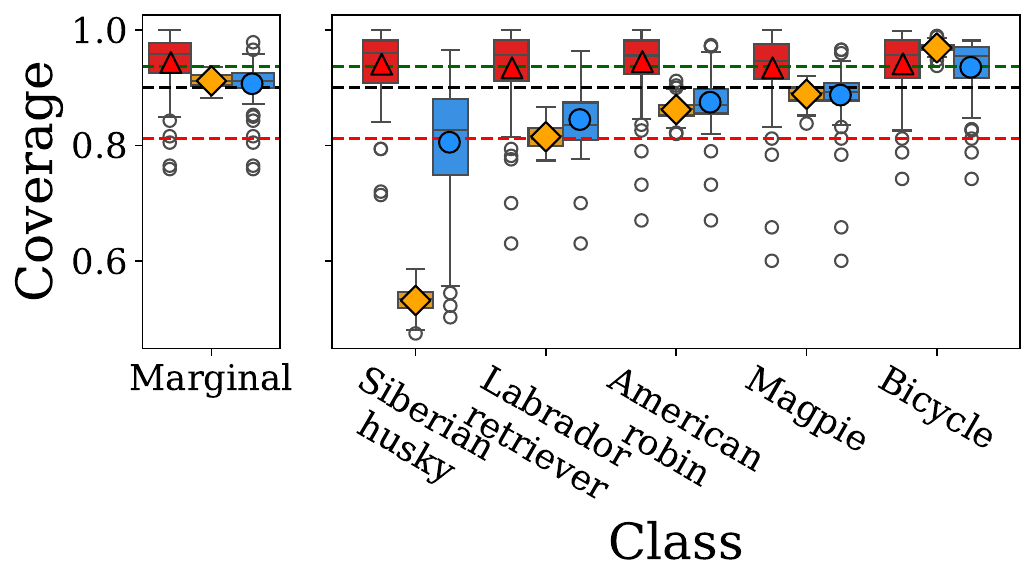}
    \includegraphics[height=0.22\textwidth]{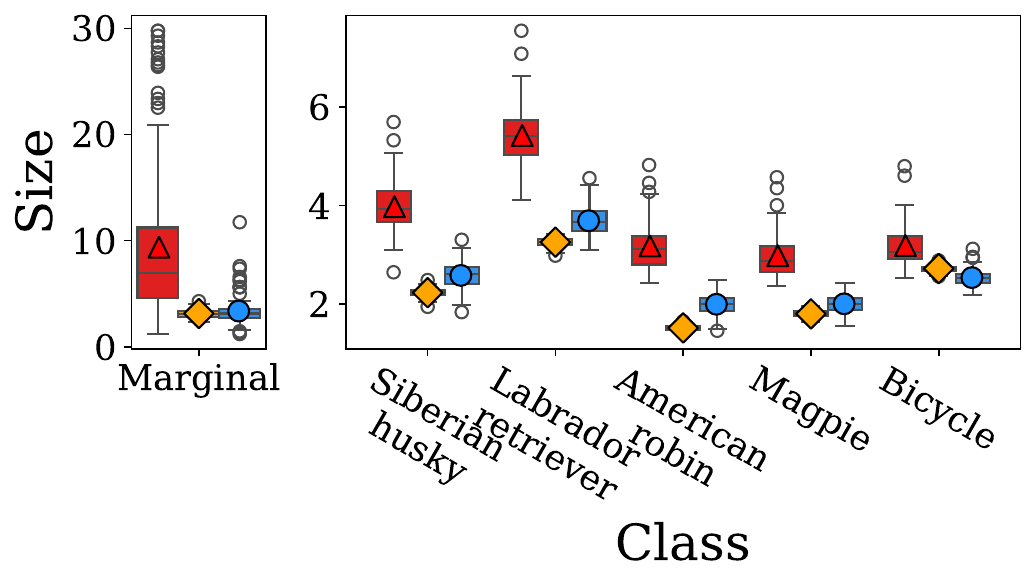}
    \includegraphics[height=0.22\linewidth]{figures/results/legend_spi.pdf}
    \caption{$\alpha=0.1$}
    \end{subfigure}
    \caption{Results for the ImageNet data using FLUX-generated synthetic images:
     Coverage rates of $\cp$, $\cpsynt$, and $\scp$ run at level $\alpha = 0.02$ (a), $0.05$ (b), and $0.1$ (b), averaged over 100 trials. Left: Average coverage. Right: Average prediction set size, both under marginal (leftmost box in each group) and label-conditional coverage settings.  Label-conditional results are shown for selected classes; see~\Cref{app-tab:imagenet-flux_0.02,app-tab:imagenet-flux_0.05,app-tab:imagenet-flux_0.1} for results across all classes.}
    \label{app-fig:imagenet-flux}
\end{figure}

\Cref{app-fig:imagenet-flux} presents the
marginal and label-conditional coverage of various methods
at levels $\alpha=0.02$, $0.05$, and $0.1$. 
The results for label-conditional guarantees are presented for representative classes; results for all classes in the real population are detailed in \Cref{app-tab:imagenet-flux_0.02,app-tab:imagenet-flux_0.05,app-tab:imagenet-flux_0.1}. 
We observe similar trends to those observed using synthetic images generated by Stable Diffusion. The standard conformal method, $\cp$, controls the coverage at the $1-\alpha$ level;
however, it yields overly conservative prediction sets due to the small sample size. 
$\cpsynt$ fails to control the coverage at the desired level, exhibiting under-coverage for some classes and over-coverage for others. 
In contrast, the proposed method, $\scp$, achieves coverage within the theoretical bounds while providing smaller, more informative prediction sets.

\begin{table}[!h]
\caption{Per-class conditional coverage (in \%) and prediction set size for each method, computed over 100 trials using FLUX-generated synthetic data. The target coverage level is $1 - \alpha = 0.98$. The theoretical coverage guarantees for $\scp$ are in the range $[93.7, 100]$. Standard errors are shown in parentheses. Other experimental details follow~\Cref{app-fig:imagenet-flux}.}
\label{app-tab:imagenet-flux_0.02}
\centering
\begin{tabular}{lp{1.3cm}p{1.6cm}p{1.6cm}p{1.1cm}p{1.6cm}p{1.5cm}}
\toprule
& \multicolumn{3}{c}{Coverage (\%)} & \multicolumn{3}{c}{Size} \\
\toprule
Class & \texttt{Only Real} & \texttt{Only Synth} & \texttt{SPI} & \texttt{Only Real} & \texttt{Only Synth} & \texttt{SPI} \\
\midrule
Admiral & 100 (± 0) & 0.2 (± 0) & 93.6 (± 0.6) & 30 (± 0) & 4.7 (± 0) & 6.5 (± 0) \\
American robin & 100 (± 0) & 95.5 (± 0.1) & 96.8 (± 0.2) & 30 (± 0) & 2.7 (± 0) & 4.3 (± 0) \\
Barracouta & 100 (± 0) & 99.9 (± 0) & 99.9 (± 0) & 30 (± 0) & 5.6 (± 0) & 8.2 (± 0.1) \\
Beaver & 100 (± 0) & 86.7 (± 0.2) & 94.5 (± 0.4) & 30 (± 0) & 4.1 (± 0) & 5.6 (± 0) \\
Bicycle & 100 (± 0) & 99.8 (± 0) & 99.8 (± 0) & 30 (± 0) & 3.8 (± 0) & 4.8 (± 0) \\
Bulbul & 100 (± 0) & 97.3 (± 0.1) & 97.9 (± 0.1) & 30 (± 0) & 3.2 (± 0) & 4.7 (± 0) \\
Coral fungus & 100 (± 0) & 99.6 (± 0) & 99.6 (± 0) & 30 (± 0) & 2.9 (± 0) & 3.9 (± 0) \\
English springer & 100 (± 0) & 96.2 (± 0.1) & 97.3 (± 0.1) & 30 (± 0) & 3.3 (± 0) & 5.1 (± 0) \\
Garfish & 100 (± 0) & 90.4 (± 0.1) & 95.0 (± 0.3) & 30 (± 0) & 4.9 (± 0) & 6.9 (± 0) \\
Golden retriever & 100 (± 0) & 93.6 (± 0.1) & 96.0 (± 0.2) & 30 (± 0) & 4.6 (± 0) & 6.9 (± 0) \\
Gyromitra & 100 (± 0) & 57.2 (± 0.3) & 95.0 (± 0.6) & 30 (± 0) & 3.6 (± 0) & 4.5 (± 0) \\
Jay & 100 (± 0) & 50.1 (± 0.5) & 93.3 (± 0.7) & 30 (± 0) & 6.9 (± 0) & 9.1 (± 0) \\
Junco, snowbird & 100 (± 0) & 99.2 (± 0) & 99.3 (± 0) & 30 (± 0) & 2.4 (± 0) & 3.6 (± 0) \\
Kuvasz & 100 (± 0) & 99.4 (± 0) & 99.4 (± 0) & 30 (± 0) & 2.8 (± 0) & 4.6 (± 0) \\
Labrador retriever & 100 (± 0) & 91.6 (± 0.2) & 95.5 (± 0.3) & 30 (± 0) & 4.9 (± 0) & 7.4 (± 0) \\
Lighter, Light & 100 (± 0) & 75.6 (± 0.2) & 94.4 (± 0.6) & 30 (± 0) & 4.7 (± 0) & 7.0 (± 0) \\
Lycaenid butterfly & 100 (± 0) & 93.4 (± 0.1) & 95.7 (± 0.2) & 30 (± 0) & 3.8 (± 0) & 4.7 (± 0) \\
Magpie & 100 (± 0) & 97.0 (± 0.1) & 97.6 (± 0.1) & 30 (± 0) & 3.0 (± 0) & 4.4 (± 0) \\
Marmot & 100 (± 0) & 98.0 (± 0.1) & 98.3 (± 0.1) & 30 (± 0) & 4.9 (± 0) & 6.9 (± 0.1) \\
Muzzle & 100 (± 0) & 96.2 (± 0.1) & 97.2 (± 0.1) & 30 (± 0) & 3.7 (± 0) & 5.8 (± 0) \\
Papillon & 100 (± 0) & 99.9 (± 0) & 99.9 (± 0) & 30 (± 0) & 2.5 (± 0) & 4.4 (± 0) \\
Rock beauty & 100 (± 0) & 93.6 (± 0.1) & 96.0 (± 0.2) & 30 (± 0) & 4.9 (± 0) & 7.1 (± 0) \\
Siberian husky & 100 (± 0) & 68.8 (± 0.2) & 94.1 (± 0.6) & 30 (± 0) & 3.4 (± 0) & 5.5 (± 0) \\
Stinkhorn & 100 (± 0) & 98.3 (± 0) & 98.5 (± 0.1) & 30 (± 0) & 5.1 (± 0) & 6.5 (± 0) \\
Tennis ball & 100 (± 0) & 89.5 (± 0.1) & 94.5 (± 0.4) & 30 (± 0) & 3.4 (± 0) & 4.8 (± 0) \\
Tinca tinca & 100 (± 0) & 99.4 (± 0) & 99.4 (± 0) & 30 (± 0) & 3.3 (± 0) & 5.0 (± 0) \\
Torch & 100 (± 0) & 91.5 (± 0.1) & 95.8 (± 0.3) & 30 (± 0) & 6.0 (± 0) & 8.5 (± 0) \\
Unicycle & 100 (± 0) & 99.8 (± 0) & 99.8 (± 0) & 30 (± 0) & 4.8 (± 0) & 6.6 (± 0) \\
Water ouzel & 100 (± 0) & 99.1 (± 0) & 99.2 (± 0) & 30 (± 0) & 2.6 (± 0) & 3.9 (± 0) \\
White wolf & 100 (± 0) & 83.9 (± 0.2) & 94.2 (± 0.5) & 30 (± 0) & 3.4 (± 0) & 5.1 (± 0) \\
\bottomrule
\end{tabular}
\end{table}

\begin{table}[!h]
\caption{Per-class conditional coverage (in \%) and prediction set size for each method, computed over 100 trials using FLUX-generated synthetic data. The target coverage level is $1 - \alpha = 0.95$. The theoretical coverage guarantees for $\scp$ are in the range $[93.7, 100]$. Standard errors are shown in parentheses. Other experimental details follow~\Cref{app-fig:imagenet-flux}.}

\label{app-tab:imagenet-flux_0.05}
\centering
\begin{tabular}{lp{1.3cm}p{1.6cm}p{1.6cm}p{1.1cm}p{1.6cm}p{1.5cm}}
\toprule
& \multicolumn{3}{c}{Coverage (\%)} & \multicolumn{3}{c}{Size} \\
\toprule
Class & \texttt{Only Real} & \texttt{Only Synth} & \texttt{SPI} & \texttt{Only Real} & \texttt{Only Synth} & \texttt{SPI} \\
\midrule

Admiral & 100 (± 0) & 0.2 (± 0) & 93.6 (± 0.6) & 30 (± 0) & 4.4 (± 0) & 6.3 (± 0) \\
American robin & 100 (± 0) & 91.7 (± 0.1) & 95.6 (± 0.3) & 30 (± 0) & 2.0 (± 0) & 3.8 (± 0) \\
Barracouta & 100 (± 0) & 99.9 (± 0) & 99.9 (± 0) & 30 (± 0) & 4.8 (± 0) & 7.7 (± 0.1) \\
Beaver & 100 (± 0) & 81.0 (± 0.2) & 94.1 (± 0.5) & 30 (± 0) & 3.1 (± 0) & 4.8 (± 0) \\
Bicycle & 100 (± 0) & 98.9 (± 0) & 99.0 (± 0) & 30 (± 0) & 3.1 (± 0) & 4.2 (± 0) \\
Bulbul & 100 (± 0) & 93.8 (± 0.1) & 96.2 (± 0.2) & 30 (± 0) & 2.6 (± 0) & 4.3 (± 0) \\
Coral fungus & 100 (± 0) & 99.4 (± 0) & 99.4 (± 0) & 30 (± 0) & 2.4 (± 0) & 3.6 (± 0) \\
English springer & 100 (± 0) & 95.1 (± 0.1) & 96.8 (± 0.2) & 30 (± 0) & 3.0 (± 0) & 4.9 (± 0) \\
Garfish & 100 (± 0) & 84.7 (± 0.2) & 93.9 (± 0.5) & 30 (± 0) & 4.0 (± 0) & 6.3 (± 0) \\
Golden retriever & 100 (± 0) & 89.9 (± 0.1) & 95.0 (± 0.3) & 30 (± 0) & 3.8 (± 0) & 6.5 (± 0.1) \\
Gyromitra & 100 (± 0) & 46.6 (± 0.2) & 95.0 (± 0.6) & 30 (± 0) & 3.0 (± 0) & 4.1 (± 0) \\
Jay & 100 (± 0) & 31.7 (± 0.2) & 93.3 (± 0.7) & 30 (± 0) & 5.7 (± 0) & 8.5 (± 0.1) \\
Junco, snowbird & 100 (± 0) & 97.9 (± 0.1) & 98.3 (± 0.1) & 30 (± 0) & 2.0 (± 0) & 3.3 (± 0) \\
Kuvasz & 100 (± 0) & 99.2 (± 0) & 99.3 (± 0) & 30 (± 0) & 2.3 (± 0) & 4.3 (± 0) \\
Labrador retriever & 100 (± 0) & 86.5 (± 0.2) & 94.4 (± 0.5) & 30 (± 0) & 4.1 (± 0) & 6.9 (± 0.1) \\
Lighter, Light & 100 (± 0) & 67.0 (± 0.2) & 94.2 (± 0.6) & 30 (± 0) & 3.9 (± 0) & 6.7 (± 0) \\
Lycaenid butterfly & 100 (± 0) & 88.5 (± 0.1) & 94.6 (± 0.4) & 30 (± 0) & 3.3 (± 0) & 4.6 (± 0) \\
Magpie & 100 (± 0) & 94.0 (± 0.1) & 95.8 (± 0.2) & 30 (± 0) & 2.5 (± 0) & 4.0 (± 0) \\
Marmot & 100 (± 0) & 96.9 (± 0.1) & 97.7 (± 0.1) & 30 (± 0) & 3.9 (± 0) & 6.2 (± 0.1) \\
Muzzle & 100 (± 0) & 94.2 (± 0.1) & 96.2 (± 0.2) & 30 (± 0) & 3.2 (± 0) & 5.5 (± 0) \\
Papillon & 100 (± 0) & 99.9 (± 0) & 99.9 (± 0) & 30 (± 0) & 2.2 (± 0) & 4.3 (± 0) \\
Rock beauty & 100 (± 0) & 87.9 (± 0.1) & 94.7 (± 0.4) & 30 (± 0) & 4.4 (± 0) & 6.8 (± 0) \\
Siberian husky & 100 (± 0) & 61.1 (± 0.2) & 94.1 (± 0.6) & 30 (± 0) & 2.9 (± 0) & 5.2 (± 0) \\
Stinkhorn & 100 (± 0) & 97.3 (± 0.1) & 97.7 (± 0.1) & 30 (± 0) & 4.4 (± 0) & 6.0 (± 0) \\
Tennis ball & 100 (± 0) & 84.4 (± 0.2) & 93.7 (± 0.5) & 30 (± 0) & 2.8 (± 0) & 4.4 (± 0) \\
Tinca tinca & 100 (± 0) & 98.7 (± 0) & 98.8 (± 0.1) & 30 (± 0) & 2.7 (± 0) & 4.6 (± 0) \\
Torch & 100 (± 0) & 86.9 (± 0.2) & 95.2 (± 0.4) & 30 (± 0) & 5.2 (± 0) & 8.0 (± 0) \\
Unicycle & 100 (± 0) & 99.8 (± 0) & 99.8 (± 0) & 30 (± 0) & 4.0 (± 0) & 6.0 (± 0) \\
Water ouzel & 100 (± 0) & 98.9 (± 0) & 99.0 (± 0) & 30 (± 0) & 2.0 (± 0) & 3.5 (± 0) \\
White wolf & 100 (± 0) & 78.5 (± 0.2) & 93.9 (± 0.6) & 30 (± 0) & 2.8 (± 0) & 4.7 (± 0) \\
\bottomrule
\end{tabular}
\end{table}

\begin{table}[!h]
\caption{Per-class conditional coverage (in \%) and prediction set size for each method, computed over 100 trials using FLUX-generated synthetic data. The target coverage level is $1 - \alpha = 0.9$. The theoretical coverage guarantees for $\scp$ are in the range $[81.2, 93.7]$. Standard errors are shown in parentheses. Other experimental details follow~\Cref{app-fig:imagenet-flux}.}
\label{app-tab:imagenet-flux_0.1}
\centering
\begin{tabular}{lp{1.6cm}p{1.6cm}p{1.6cm}p{1.5cm}p{1.2cm}p{1.2cm}}
\toprule
& \multicolumn{3}{c}{Coverage (\%)} & \multicolumn{3}{c}{Size} \\
\toprule
Class & \texttt{Only Real} & \texttt{Only Synth} & \texttt{SPI} & \texttt{Only Real} & \texttt{Only Synth} & \texttt{SPI} \\
\midrule
Admiral & 93.6 (± 0.6) & 0.1 (± 0) & 81.3 (± 0.9) & 5.6 (± 0) & 4.1 (± 0) & 4.6 (± 0) \\
American robin & 94.5 (± 0.5) & 86.2 (± 0.2) & 87.5 (± 0.4) & 3.2 (± 0) & 1.5 (± 0) & 2.0 (± 0) \\
Barracouta & 95.3 (± 0.4) & 99.9 (± 0) & 95.3 (± 0.4) & 6.6 (± 0.1) & 3.9 (± 0) & 5.1 (± 0) \\
Beaver & 94.0 (± 0.6) & 74.5 (± 0.2) & 82.5 (± 0.7) & 3.6 (± 0.1) & 2.4 (± 0) & 2.5 (± 0) \\
Bicycle & 94.2 (± 0.5) & 96.9 (± 0.1) & 93.6 (± 0.5) & 3.2 (± 0) & 2.7 (± 0) & 2.5 (± 0) \\
Bulbul & 93.8 (± 0.6) & 88.3 (± 0.2) & 88.4 (± 0.4) & 3.3 (± 0) & 1.9 (± 0) & 2.2 (± 0) \\
Coral fungus & 93.5 (± 0.6) & 98.4 (± 0) & 93.4 (± 0.6) & 2.9 (± 0) & 2.0 (± 0) & 2.2 (± 0) \\
English springer & 93.9 (± 0.6) & 93.3 (± 0.1) & 91.3 (± 0.4) & 3.6 (± 0) & 2.5 (± 0) & 2.5 (± 0) \\
Garfish & 93.5 (± 0.6) & 77.2 (± 0.2) & 82.7 (± 0.6) & 5.4 (± 0.1) & 3.3 (± 0) & 4.1 (± 0) \\
Golden retriever & 94.1 (± 0.6) & 84.0 (± 0.2) & 86.5 (± 0.5) & 5.0 (± 0.1) & 3.0 (± 0) & 3.3 (± 0) \\
Gyromitra & 95.0 (± 0.6) & 38.8 (± 0.2) & 84.8 (± 1.0) & 3.3 (± 0) & 2.4 (± 0) & 2.6 (± 0) \\
Jay & 93.3 (± 0.7) & 23.5 (± 0.2) & 80.5 (± 1.0) & 6.8 (± 0.1) & 4.6 (± 0) & 5.1 (± 0) \\
Junco, snowbird & 94.2 (± 0.5) & 95.3 (± 0.1) & 92.9 (± 0.4) & 2.7 (± 0) & 1.6 (± 0) & 1.8 (± 0) \\
Kuvasz & 93.4 (± 0.6) & 99.0 (± 0) & 93.3 (± 0.6) & 3.5 (± 0) & 1.8 (± 0) & 2.1 (± 0) \\
Labrador retriever & 93.5 (± 0.7) & 81.5 (± 0.2) & 84.5 (± 0.5) & 5.4 (± 0.1) & 3.3 (± 0) & 3.7 (± 0) \\
Lighter, Light & 94.2 (± 0.6) & 57.8 (± 0.2) & 81.5 (± 0.9) & 5.4 (± 0.1) & 3.1 (± 0) & 3.8 (± 0) \\
Lycaenid butterfly & 94.0 (± 0.5) & 82.2 (± 0.2) & 86.2 (± 0.5) & 3.9 (± 0.1) & 3.0 (± 0) & 3.0 (± 0) \\
Magpie & 93.5 (± 0.6) & 88.9 (± 0.2) & 88.7 (± 0.5) & 3.0 (± 0) & 1.8 (± 0) & 2.0 (± 0) \\
Marmot & 94.0 (± 0.6) & 96.0 (± 0.1) & 92.9 (± 0.6) & 4.9 (± 0.1) & 3.1 (± 0) & 3.3 (± 0) \\
Muzzle & 93.5 (± 0.6) & 90.5 (± 0.1) & 90.0 (± 0.5) & 4.3 (± 0) & 2.6 (± 0) & 3.1 (± 0) \\
Papillon & 93.8 (± 0.6) & 99.7 (± 0) & 93.8 (± 0.6) & 3.8 (± 0.1) & 1.9 (± 0) & 2.5 (± 0) \\
Rock beauty & 94.2 (± 0.5) & 80.3 (± 0.2) & 84.4 (± 0.5) & 5.1 (± 0.1) & 3.7 (± 0) & 4.1 (± 0) \\
Siberian husky & 94.1 (± 0.6) & 53.1 (± 0.2) & 80.6 (± 1.0) & 4.0 (± 0) & 2.2 (± 0) & 2.6 (± 0) \\
Stinkhorn & 93.4 (± 0.6) & 96.0 (± 0.1) & 92.7 (± 0.5) & 4.5 (± 0.1) & 3.5 (± 0) & 3.4 (± 0) \\
Tennis ball & 93.5 (± 0.6) & 77.2 (± 0.2) & 82.7 (± 0.6) & 3.5 (± 0) & 2.1 (± 0) & 2.5 (± 0) \\
Tinca tinca & 93.2 (± 0.6) & 97.5 (± 0.1) & 93.0 (± 0.6) & 3.8 (± 0) & 2.2 (± 0) & 2.8 (± 0) \\
Torch & 94.8 (± 0.5) & 79.5 (± 0.2) & 84.5 (± 0.5) & 6.2 (± 0.1) & 4.4 (± 0) & 4.8 (± 0) \\
Unicycle & 93.3 (± 0.6) & 99.7 (± 0) & 93.3 (± 0.6) & 4.8 (± 0) & 3.4 (± 0) & 3.5 (± 0) \\
Water ouzel & 94.3 (± 0.5) & 98.8 (± 0) & 94.2 (± 0.5) & 3.0 (± 0) & 1.6 (± 0) & 2.0 (± 0) \\
White wolf & 93.9 (± 0.6) & 72.8 (± 0.2) & 82.2 (± 0.8) & 3.4 (± 0) & 2.2 (± 0) & 2.3 (± 0) \\
\bottomrule
\end{tabular}
\end{table}
\FloatBarrier

\subsection{Experiments with auxiliary labeled data}
\label{app-sec:exp-imagenet-clusters}
In this section, we follow the experimental setup described in~\Cref{sec:exp-imagenet-clusters}, where the synthetic data comprise 100 classes, none of which are included in the real calibration set.

\begin{figure}[!h]
    \centering
    \begin{subfigure}[t]{\linewidth}
    \includegraphics[height=0.22\linewidth]{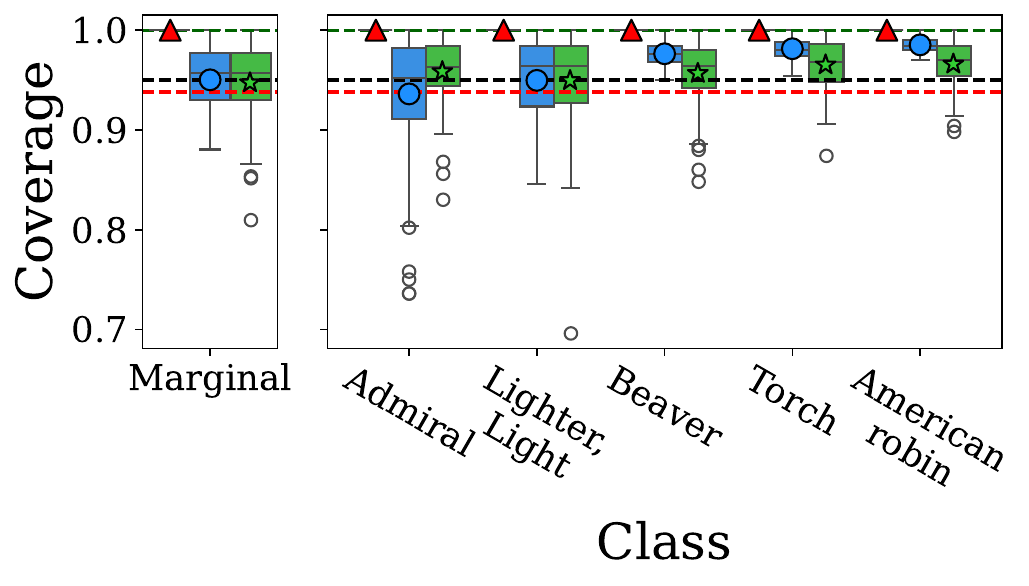}
    \includegraphics[height=0.22\linewidth]{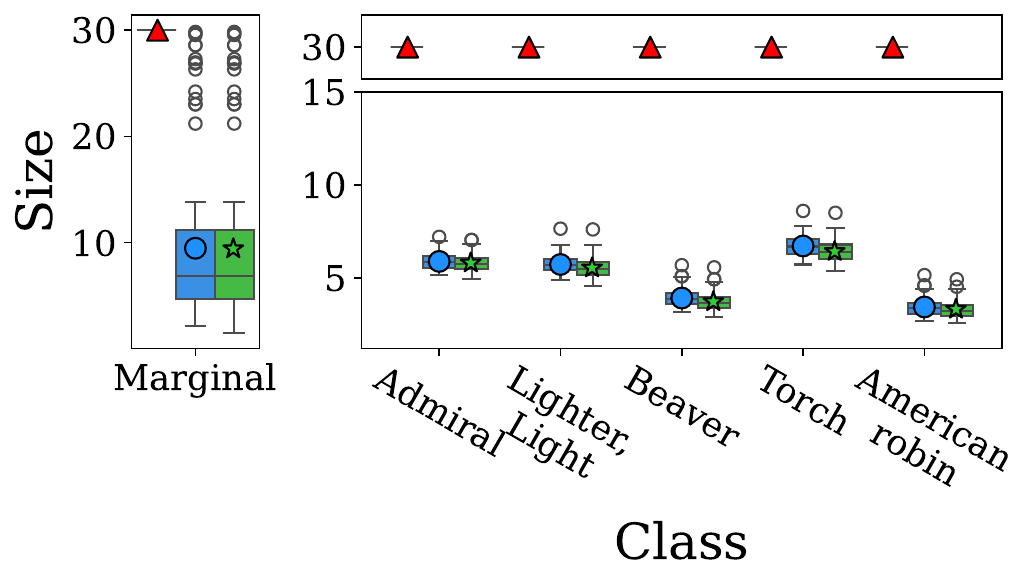}
    \caption{$\alpha=0.05$}
    \label{app-fig:imagenet-clusters_0.05}
    \end{subfigure}
    \begin{subfigure}[t]{\linewidth}
    \includegraphics[height=0.22\linewidth]{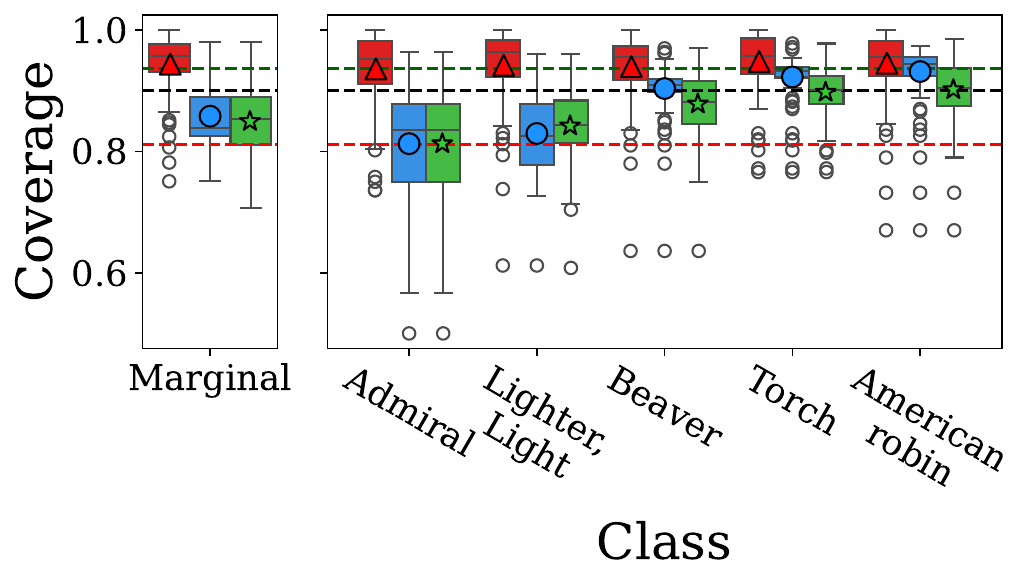}
    \includegraphics[height=0.22\linewidth]{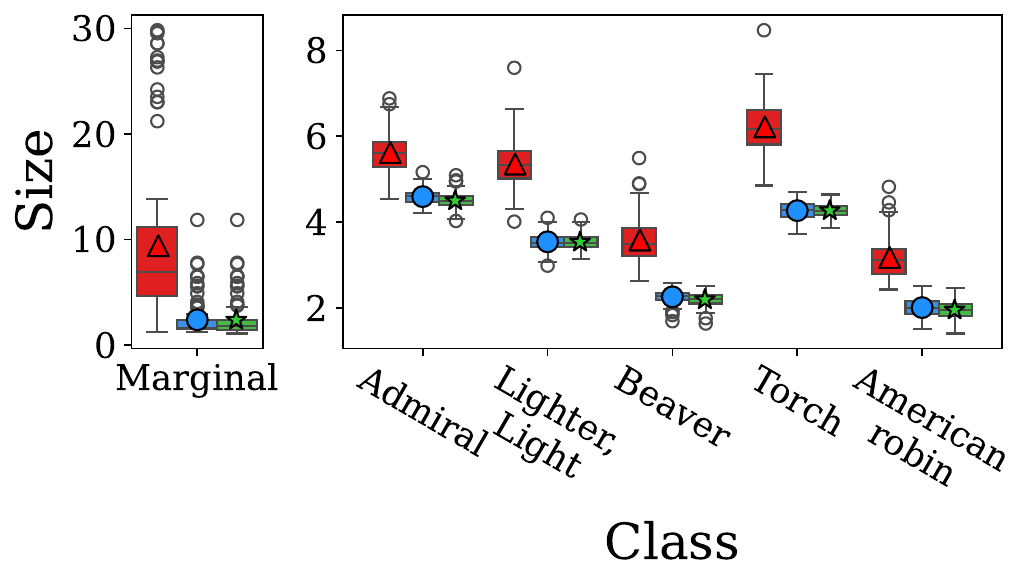}
    \includegraphics[height=0.2\linewidth]{figures/results/legend_clusters_wo_more_spi.pdf}
    \caption{$\alpha=0.1$}
    \label{app-fig:imagenet-clusters_0.1}
    \end{subfigure}
    \caption{
    Results for the ImageNet data:
    Coverage rates of $\cp$, $\scph$, and $\scpc$ run at level $\alpha = 0.05$ (a) and $0.1$ (b), averaged over 100 trials. Left: Average coverage. Right: Average prediction set size, both under marginal (leftmost box in each group) and label-conditional coverage settings.  Label-conditional results are shown for selected classes; see~\Cref{app-tab:imagenet-clusters-all-classes_0.05,app-tab:imagenet-clusters-all-classes_0.1} for results across all classes.}
    \label{app-fig:imagenet-clusters}
\end{figure}

\Cref{app-fig:imagenet-clusters} presents the results
for both marginal and label-conditional guarantees at levels $\alpha=0.05$ and $0.1$, demonstrating 
trends similar to those observed in~\Cref{fig:imagenet-clusters_0.02}. 
The standard conformal prediction, $\cp$, conservatively controls coverage at the target level $1 - \alpha$, but results in larger and noisier prediction sets due to the limited sample size. In contrast, both $\scph$ and $\scpc$ substantially reduce the size and variance of the prediction sets and, as expected, achieve coverage within the theoretical bounds.

Notably, for the ``American robin'' and ``Torch'' classes, the $\scpc$ variant achieves coverage more tightly aligned with the target level $1 - \alpha$, outperforming the standard $\scph$ method.

We include the results for all real classes in \Cref{app-tab:imagenet-clusters-all-classes_0.02,app-tab:imagenet-clusters-all-classes_0.05,app-tab:imagenet-clusters-all-classes_0.1}, corresponding to \Cref{fig:imagenet-clusters_0.02,app-fig:imagenet-clusters_0.05,app-fig:imagenet-clusters_0.1}, respectively.

\begin{table}[!h]
\caption{Per-class conditional coverage (in \%) and prediction set size for each method, computed over 100 trials. Standard errors are shown in parentheses.~The target coverage level is $1-\alpha = 0.98$. The theoretical coverage guarantees for both $\scph$ and $\scpc$ are in the range  $[93.7, 100]$. Other details are as in~\Cref{fig:imagenet-clusters_0.02}.}
\label{app-tab:imagenet-clusters-all-classes_0.02}
\centering
\begin{tabular}{lp{1.3cm}p{1.6cm}p{1.6cm}p{1.1cm}p{1.6cm}p{1.5cm}}
\toprule
& \multicolumn{3}{c}{Coverage (\%)} & \multicolumn{3}{c}{Size} \\
\toprule
Class & \texttt{Only Real} & \texttt{SPI Whole} & \texttt{SPI Subset} & \texttt{Only Real} & \texttt{SPI Whole} & \texttt{SPI Subset} \\
\midrule
Admiral & 100 (± 0) & 93.6 (± 0.6) & 99.8 (± 0) & 30 (± 0) & 7.6 (± 0.1) & 6.5 (± 0) \\
American robin & 100 (± 0) & 99.9 (± 0) & 98.7 (± 0.1) & 30 (± 0) & 5.0 (± 0.1) & 4.4 (± 0) \\
Barracouta & 100 (± 0) & 98.4 (± 0.1) & 98.0 (± 0.2) & 30 (± 0) & 9.8 (± 0.1) & 7.8 (± 0.1) \\
Beaver & 100 (± 0) & 99.5 (± 0) & 98.1 (± 0.2) & 30 (± 0) & 6.1 (± 0.1) & 4.9 (± 0) \\
Bicycle & 100 (± 0) & 100 (± 0) & 99.2 (± 0.1) & 30 (± 0) & 4.7 (± 0) & 4.0 (± 0) \\
Bulbul & 100 (± 0) & 99.9 (± 0) & 99.0 (± 0.1) & 30 (± 0) & 5.2 (± 0) & 4.3 (± 0) \\
Coral fungus & 100 (± 0) & 99.7 (± 0) & 98.2 (± 0.1) & 30 (± 0) & 4.1 (± 0) & 3.7 (± 0) \\
English springer & 100 (± 0) & 99.8 (± 0) & 98.8 (± 0.1) & 30 (± 0) & 5.7 (± 0.1) & 4.7 (± 0) \\
Garfish & 100 (± 0) & 99.4 (± 0) & 97.8 (± 0.2) & 30 (± 0) & 8.0 (± 0.1) & 6.6 (± 0) \\
Golden retriever & 100 (± 0) & 99.6 (± 0) & 98.0 (± 0.2) & 30 (± 0) & 7.9 (± 0.1) & 6.5 (± 0.1) \\
Gyromitra & 100 (± 0) & 95.0 (± 0.6) & 95.0 (± 0.6) & 30 (± 0) & 4.6 (± 0) & 4.3 (± 0) \\
Jay & 100 (± 0) & 98.3 (± 0.1) & 98.1 (± 0.1) & 30 (± 0) & 11.4 (± 0.1) & 8.7 (± 0.1) \\
Junco, snowbird & 100 (± 0) & 99.9 (± 0) & 98.7 (± 0.1) & 30 (± 0) & 4.3 (± 0) & 3.5 (± 0) \\
Kuvasz & 100 (± 0) & 99.5 (± 0) & 98.5 (± 0.1) & 30 (± 0) & 5.7 (± 0) & 4.7 (± 0.1) \\
Labrador retriever & 100 (± 0) & 99.3 (± 0) & 97.6 (± 0.2) & 30 (± 0) & 8.7 (± 0.1) & 7.0 (± 0.1) \\
Lighter, Light & 100 (± 0) & 97.8 (± 0.1) & 96.9 (± 0.2) & 30 (± 0) & 8.4 (± 0.1) & 6.6 (± 0.1) \\
Lycaenid butterfly & 100 (± 0) & 99.8 (± 0) & 99.1 (± 0.1) & 30 (± 0) & 5.2 (± 0) & 5.2 (± 0) \\
Magpie & 100 (± 0) & 99.7 (± 0) & 98.6 (± 0.1) & 30 (± 0) & 4.7 (± 0) & 3.9 (± 0) \\
Marmot & 100 (± 0) & 99.7 (± 0) & 98.5 (± 0.1) & 30 (± 0) & 7.5 (± 0.1) & 6.4 (± 0.1) \\
Muzzle & 100 (± 0) & 98.8 (± 0.1) & 96.6 (± 0.3) & 30 (± 0) & 6.7 (± 0.1) & 5.4 (± 0) \\
Papillon & 100 (± 0) & 99.2 (± 0) & 97.6 (± 0.2) & 30 (± 0) & 5.7 (± 0.1) & 5.1 (± 0) \\
Rock beauty & 100 (± 0) & 98.9 (± 0.1) & 98.5 (± 0.1) & 30 (± 0) & 8.3 (± 0.1) & 6.2 (± 0.1) \\
Siberian husky & 100 (± 0) & 99.3 (± 0) & 98.8 (± 0.1) & 30 (± 0) & 6.3 (± 0.1) & 5.2 (± 0) \\
Stinkhorn & 100 (± 0) & 99.8 (± 0) & 98.3 (± 0.1) & 30 (± 0) & 7.2 (± 0.1) & 6.0 (± 0) \\
Tennis ball & 100 (± 0) & 98.6 (± 0.1) & 96.6 (± 0.3) & 30 (± 0) & 5.3 (± 0) & 4.5 (± 0) \\
Tinca tinca & 100 (± 0) & 99.8 (± 0) & 98.8 (± 0.1) & 30 (± 0) & 5.3 (± 0) & 4.7 (± 0) \\
Torch & 100 (± 0) & 99.6 (± 0) & 98.4 (± 0.1) & 30 (± 0) & 10.3 (± 0.1) & 7.6 (± 0.1) \\
Unicycle & 100 (± 0) & 99.7 (± 0) & 98.2 (± 0.1) & 30 (± 0) & 8.1 (± 0.1) & 6.1 (± 0.1) \\
Water ouzel & 100 (± 0) & 99.2 (± 0) & 98.0 (± 0.1) & 30 (± 0) & 5.0 (± 0) & 4.0 (± 0) \\
White wolf & 100 (± 0) & 99.1 (± 0) & 97.1 (± 0.2) & 30 (± 0) & 5.5 (± 0) & 4.6 (± 0.1) \\
\bottomrule
\end{tabular}
\end{table}

\begin{table}[!h]
\caption{Per-class conditional coverage (in \%) and prediction set size for each method, computed over 100 trials. Standard errors are shown in parentheses.~The target coverage level is $1-\alpha = 0.95$. The theoretical coverage guarantees for both $\scph$ and $\scpc$ are in the range  $[93.7, 100]$. Other details are as in~\Cref{app-fig:imagenet-clusters}.}
\label{app-tab:imagenet-clusters-all-classes_0.05}
\centering
\begin{tabular}{lp{1.3cm}p{1.6cm}p{1.6cm}p{1.1cm}p{1.5cm}p{1.5cm}}
\toprule
& \multicolumn{3}{c}{Coverage (\%)} & \multicolumn{3}{c}{Size} \\
\toprule
Class & \texttt{Only Real} & \texttt{SPI Whole} & \texttt{SPI Subset} & \texttt{Only Real} & \texttt{SPI Whole} & \texttt{SPI Subset} \\
\midrule
Admiral & 100 (± 0) & 93.6 (± 0.6) & 95.9 (± 0.3) & 30 (± 0) & 5.9 (± 0) & 5.8 (± 0) \\
American robin & 100 (± 0) & 98.5 (± 0.1) & 96.6 (± 0.2) & 30 (± 0) & 3.4 (± 0) & 3.3 (± 0) \\
Barracouta & 100 (± 0) & 95.3 (± 0.4) & 95.8 (± 0.3) & 30 (± 0) & 7.1 (± 0.1) & 6.9 (± 0.1) \\
Beaver & 100 (± 0) & 97.6 (± 0.1) & 95.7 (± 0.3) & 30 (± 0) & 3.9 (± 0) & 3.7 (± 0) \\
Bicycle & 100 (± 0) & 99.3 (± 0) & 97.2 (± 0.2) & 30 (± 0) & 3.4 (± 0) & 3.3 (± 0) \\
Bulbul & 100 (± 0) & 99.0 (± 0) & 96.8 (± 0.2) & 30 (± 0) & 3.5 (± 0) & 3.4 (± 0) \\
Coral fungus & 100 (± 0) & 98.1 (± 0.1) & 96.0 (± 0.3) & 30 (± 0) & 3.1 (± 0) & 3.0 (± 0) \\
English springer & 100 (± 0) & 98.9 (± 0.1) & 97.3 (± 0.2) & 30 (± 0) & 3.9 (± 0) & 3.8 (± 0) \\
Garfish & 100 (± 0) & 96.9 (± 0.2) & 95.2 (± 0.4) & 30 (± 0) & 5.8 (± 0) & 5.6 (± 0) \\
Golden retriever & 100 (± 0) & 97.7 (± 0.1) & 95.8 (± 0.3) & 30 (± 0) & 5.5 (± 0.1) & 5.3 (± 0.1) \\
Gyromitra & 100 (± 0) & 95.0 (± 0.6) & 95.0 (± 0.6) & 30 (± 0) & 3.6 (± 0) & 3.5 (± 0) \\
Jay & 100 (± 0) & 94.0 (± 0.5) & 95.3 (± 0.3) & 30 (± 0) & 7.9 (± 0.1) & 7.2 (± 0.1) \\
Junco, snowbird & 100 (± 0) & 98.8 (± 0.1) & 96.6 (± 0.2) & 30 (± 0) & 3.0 (± 0) & 2.8 (± 0) \\
Kuvasz & 100 (± 0) & 98.5 (± 0.1) & 96.5 (± 0.2) & 30 (± 0) & 3.9 (± 0) & 3.7 (± 0) \\
Labrador retriever & 100 (± 0) & 97.1 (± 0.1) & 95.3 (± 0.4) & 30 (± 0) & 6.0 (± 0.1) & 5.7 (± 0.1) \\
Lighter, Light & 100 (± 0) & 95.0 (± 0.4) & 95.0 (± 0.5) & 30 (± 0) & 5.7 (± 0) & 5.5 (± 0.1) \\
Lycaenid butterfly & 100 (± 0) & 99.2 (± 0) & 96.8 (± 0.2) & 30 (± 0) & 4.2 (± 0) & 4.1 (± 0.1) \\
Magpie & 100 (± 0) & 98.6 (± 0.1) & 96.4 (± 0.2) & 30 (± 0) & 3.2 (± 0) & 3.1 (± 0) \\
Marmot & 100 (± 0) & 98.1 (± 0.1) & 96.4 (± 0.3) & 30 (± 0) & 5.1 (± 0.1) & 5.1 (± 0.1) \\
Muzzle & 100 (± 0) & 95.0 (± 0.3) & 94.3 (± 0.5) & 30 (± 0) & 4.6 (± 0) & 4.5 (± 0) \\
Papillon & 100 (± 0) & 97.4 (± 0.1) & 95.3 (± 0.3) & 30 (± 0) & 4.0 (± 0) & 4.0 (± 0) \\
Rock beauty & 100 (± 0) & 94.5 (± 0.4) & 95.6 (± 0.3) & 30 (± 0) & 5.6 (± 0) & 5.3 (± 0) \\
Siberian husky & 100 (± 0) & 96.5 (± 0.2) & 97.0 (± 0.2) & 30 (± 0) & 4.3 (± 0) & 4.1 (± 0) \\
Stinkhorn & 100 (± 0) & 98.1 (± 0.1) & 95.9 (± 0.2) & 30 (± 0) & 5.0 (± 0) & 4.8 (± 0) \\
Tennis ball & 100 (± 0) & 95.7 (± 0.3) & 94.6 (± 0.4) & 30 (± 0) & 3.8 (± 0) & 3.6 (± 0) \\
Tinca tinca & 100 (± 0) & 98.8 (± 0.1) & 96.6 (± 0.2) & 30 (± 0) & 4.0 (± 0) & 3.9 (± 0) \\
Torch & 100 (± 0) & 98.1 (± 0.1) & 96.5 (± 0.3) & 30 (± 0) & 6.7 (± 0.1) & 6.4 (± 0.1) \\
Unicycle & 100 (± 0) & 97.9 (± 0.1) & 95.6 (± 0.4) & 30 (± 0) & 5.3 (± 0) & 5.0 (± 0) \\
Water ouzel & 100 (± 0) & 98.0 (± 0.1) & 96.2 (± 0.3) & 30 (± 0) & 3.3 (± 0) & 3.1 (± 0) \\
White wolf & 100 (± 0) & 96.6 (± 0.2) & 95.1 (± 0.4) & 30 (± 0) & 3.7 (± 0) & 3.6 (± 0) \\
\bottomrule
\end{tabular}
\end{table}

\begin{table}[!h]
\caption{Per-class conditional coverage (in \%) and prediction set size for each method, computed over 100 trials. Standard errors are shown in parentheses.~The target coverage level is $1-\alpha = 0.9$. The theoretical coverage guarantee for both $\scph$ and $\scpc$ are in the range $[81.2, 93.7]$. Other details are as in~\Cref{app-fig:imagenet-clusters}.}
\label{app-tab:imagenet-clusters-all-classes_0.1}
    \centering
\begin{tabular}{lp{1.6cm}p{1.6cm}p{1.6cm}p{1.5cm}p{1.2cm}p{1.2cm}}
\toprule
& \multicolumn{3}{c}{Coverage (\%)} & \multicolumn{3}{c}{Size} \\
\toprule
Class & \texttt{Only Real} & \texttt{SPI Whole} & \texttt{SPI Subset} & \texttt{Only Real} & \texttt{SPI Whole} & \texttt{SPI Subset} \\
\midrule
Admiral & 93.6 (± 0.6) & 81.3 (± 0.9) & 81.3 (± 0.9) & 5.6 (± 0) & 4.6 (± 0) & 4.5 (± 0) \\
American robin & 94.5 (± 0.5) & 93.2 (± 0.5) & 90.2 (± 0.5) & 3.2 (± 0) & 2.0 (± 0) & 2.0 (± 0) \\
Barracouta & 95.3 (± 0.4) & 83.4 (± 0.8) & 85.1 (± 0.6) & 6.6 (± 0.1) & 5.0 (± 0) & 4.9 (± 0) \\
Beaver & 94.0 (± 0.6) & 90.4 (± 0.4) & 87.8 (± 0.5) & 3.6 (± 0.1) & 2.3 (± 0) & 2.2 (± 0) \\
Bicycle & 94.2 (± 0.5) & 93.5 (± 0.5) & 90.9 (± 0.4) & 3.2 (± 0) & 2.3 (± 0) & 2.3 (± 0) \\
Bulbul & 93.8 (± 0.6) & 93.0 (± 0.6) & 90.0 (± 0.6) & 3.3 (± 0) & 2.1 (± 0) & 2.0 (± 0) \\
Coral fungus & 93.5 (± 0.6) & 91.7 (± 0.5) & 88.8 (± 0.5) & 2.9 (± 0) & 2.1 (± 0) & 2.0 (± 0) \\
English springer & 93.9 (± 0.6) & 93.0 (± 0.5) & 90.4 (± 0.5) & 3.6 (± 0) & 2.2 (± 0) & 2.2 (± 0) \\
Garfish & 93.5 (± 0.6) & 88.5 (± 0.4) & 86.8 (± 0.5) & 5.4 (± 0.1) & 3.9 (± 0) & 3.9 (± 0) \\
Golden retriever & 94.1 (± 0.6) & 91.7 (± 0.4) & 88.8 (± 0.5) & 5.0 (± 0.1) & 3.2 (± 0) & 3.1 (± 0) \\
Gyromitra & 95.0 (± 0.6) & 84.8 (± 1.0) & 84.8 (± 1.0) & 3.3 (± 0) & 2.5 (± 0) & 2.4 (± 0) \\
Jay & 93.3 (± 0.7) & 80.5 (± 1.0) & 83.7 (± 0.6) & 6.8 (± 0.1) & 4.8 (± 0) & 4.6 (± 0) \\
Junco, snowbird & 94.2 (± 0.5) & 93.0 (± 0.4) & 90.6 (± 0.5) & 2.7 (± 0) & 1.8 (± 0) & 1.7 (± 0) \\
Kuvasz & 93.4 (± 0.6) & 92.3 (± 0.5) & 90.1 (± 0.5) & 3.5 (± 0) & 2.1 (± 0) & 2.1 (± 0) \\
Labrador retriever & 93.5 (± 0.7) & 88.5 (± 0.5) & 87.0 (± 0.6) & 5.4 (± 0.1) & 3.4 (± 0) & 3.4 (± 0) \\
Lighter, Light & 94.2 (± 0.6) & 83.0 (± 0.7) & 84.2 (± 0.6) & 5.4 (± 0.1) & 3.5 (± 0) & 3.5 (± 0) \\
Lycaenid butterfly & 94.0 (± 0.5) & 93.3 (± 0.4) & 90.7 (± 0.5) & 3.9 (± 0.1) & 2.7 (± 0) & 2.7 (± 0) \\
Magpie & 93.5 (± 0.6) & 92.4 (± 0.6) & 90.1 (± 0.5) & 3.0 (± 0) & 1.9 (± 0) & 1.9 (± 0) \\
Marmot & 94.0 (± 0.6) & 90.9 (± 0.5) & 88.3 (± 0.6) & 4.9 (± 0.1) & 3.1 (± 0) & 3.0 (± 0) \\
Muzzle & 93.5 (± 0.6) & 86.1 (± 0.6) & 85.9 (± 0.6) & 4.3 (± 0) & 2.8 (± 0) & 2.8 (± 0) \\
Papillon & 93.8 (± 0.6) & 90.4 (± 0.4) & 87.9 (± 0.5) & 3.8 (± 0.1) & 2.4 (± 0) & 2.4 (± 0) \\
Rock beauty & 94.2 (± 0.5) & 80.9 (± 0.9) & 84.6 (± 0.5) & 5.1 (± 0.1) & 3.7 (± 0) & 3.7 (± 0) \\
Siberian husky & 94.1 (± 0.6) & 84.3 (± 0.5) & 89.1 (± 0.5) & 4.0 (± 0) & 2.3 (± 0) & 2.4 (± 0) \\
Stinkhorn & 93.4 (± 0.6) & 91.9 (± 0.5) & 89.3 (± 0.4) & 4.5 (± 0.1) & 3.0 (± 0) & 3.0 (± 0) \\
Tennis ball & 93.5 (± 0.6) & 87.7 (± 0.3) & 85.8 (± 0.5) & 3.5 (± 0) & 2.5 (± 0) & 2.4 (± 0) \\
Tinca tinca & 93.2 (± 0.6) & 92.6 (± 0.5) & 90.3 (± 0.5) & 3.8 (± 0) & 2.6 (± 0) & 2.6 (± 0) \\
Torch & 94.8 (± 0.5) & 92.3 (± 0.4) & 89.8 (± 0.4) & 6.2 (± 0.1) & 4.3 (± 0) & 4.3 (± 0) \\
Unicycle & 93.3 (± 0.6) & 90.3 (± 0.5) & 88.2 (± 0.6) & 4.8 (± 0) & 3.3 (± 0) & 3.2 (± 0) \\
Water ouzel & 94.3 (± 0.5) & 92.4 (± 0.4) & 89.9 (± 0.4) & 3.0 (± 0) & 2.0 (± 0) & 1.9 (± 0) \\
White wolf & 93.9 (± 0.6) & 89.2 (± 0.4) & 87.3 (± 0.5) & 3.4 (± 0) & 2.1 (± 0) & 2.1 (± 0) \\
\bottomrule
\end{tabular}
\end{table}
\FloatBarrier

\subsubsection{Results for \texorpdfstring{$\scpc$}{SPI-Subset} with different hyperparameter values}
\label{app-sec:imagenet-clusters-k}
In this section, we present results for the $\scpc$ procedure across different values of $k$, the number of subsets selected to construct the synthetic calibration set. We compare the performance of $\scpc$ with $\scph$---which uses all 100 synthetic classes---and the standard conformal prediction, $\cp$.

\begin{figure}[!h]
    \centering
    \begin{subfigure}[t]{\linewidth}
    \includegraphics[height=0.22\textwidth]{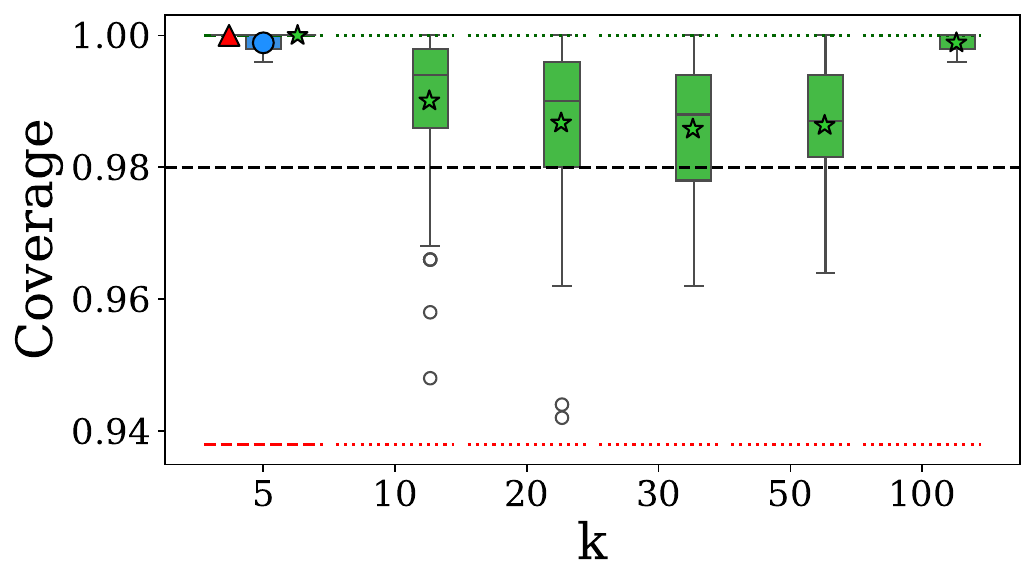}
    \includegraphics[height=0.22\textwidth]{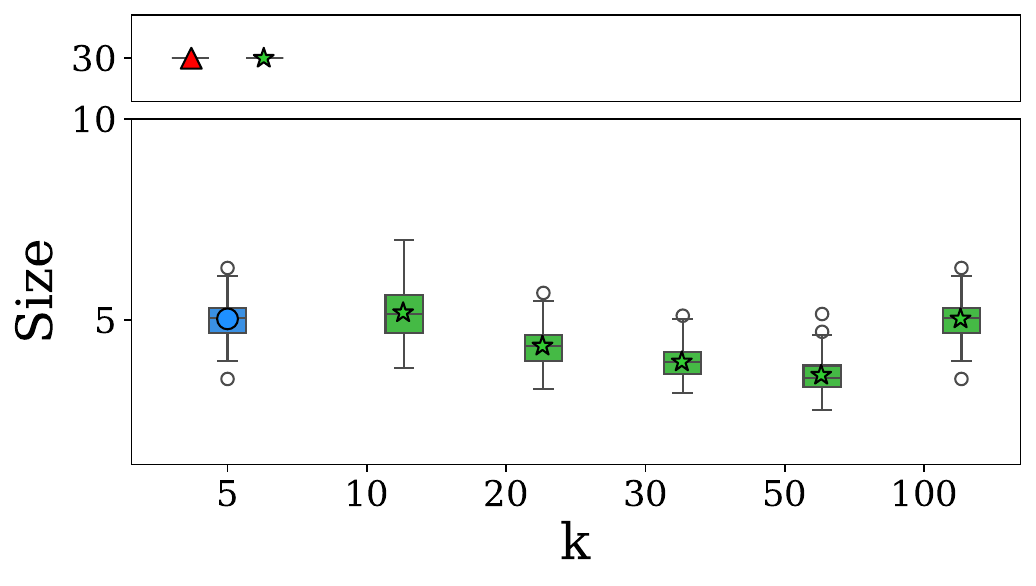}
    \caption{$\alpha=0.02$}
    \end{subfigure}
    \begin{subfigure}[t]{\linewidth}
    \includegraphics[height=0.22\textwidth]{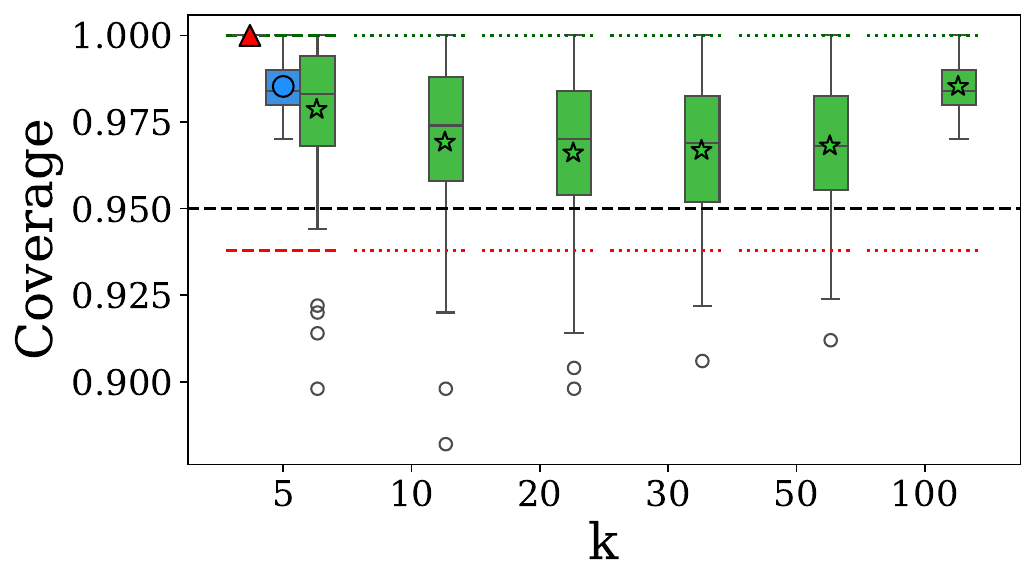}
    \includegraphics[height=0.22\textwidth]{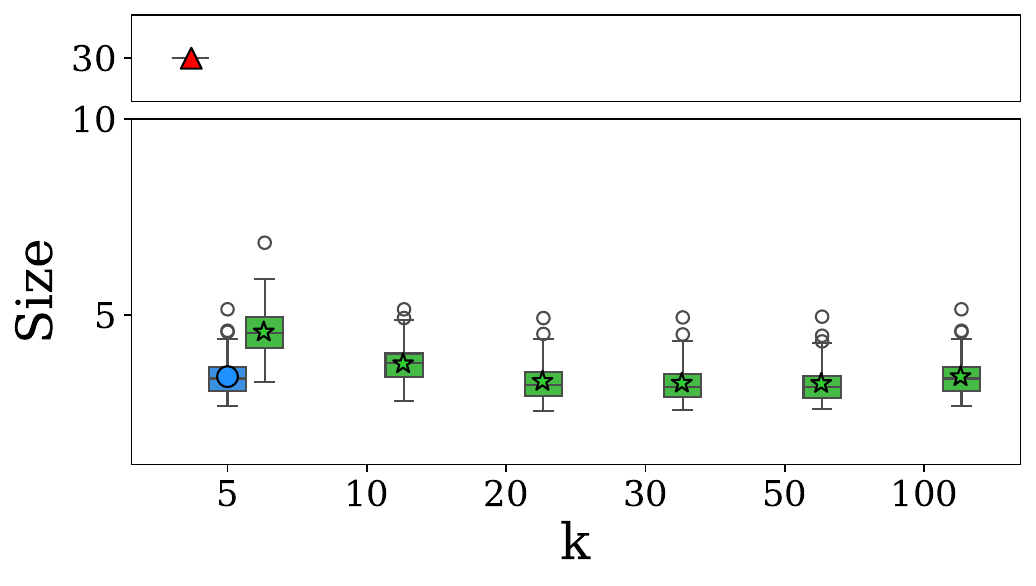}
    \caption{$\alpha=0.05$}
    \end{subfigure}
    \begin{subfigure}[t]{\linewidth}\includegraphics[height=0.22\textwidth]{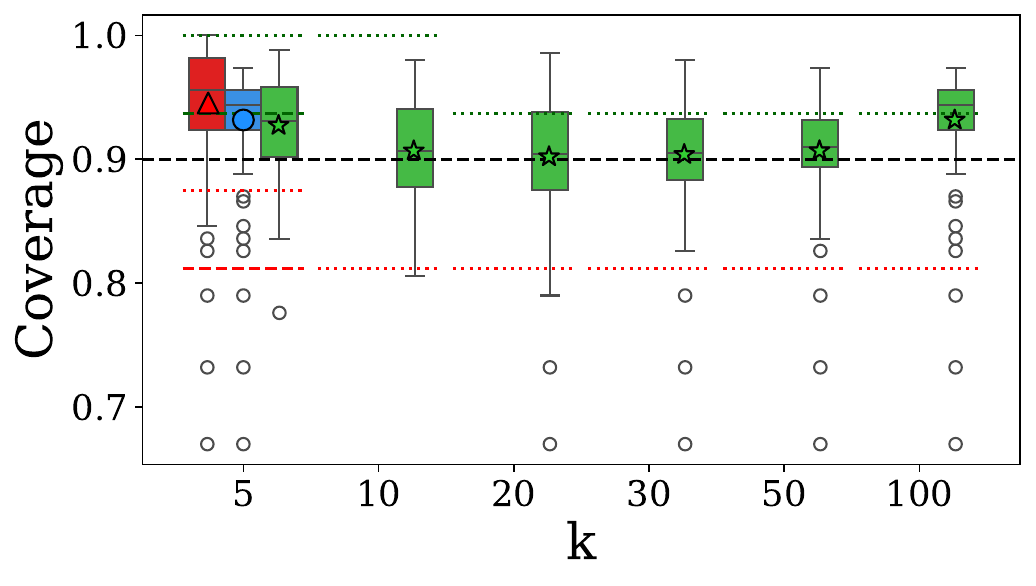}
    \includegraphics[height=0.22\textwidth]{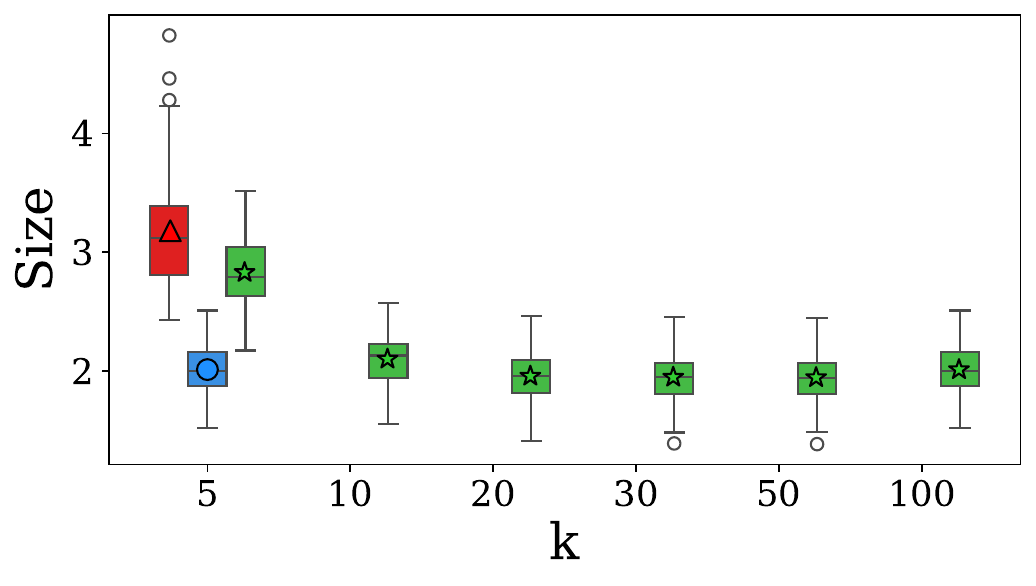}
    \includegraphics[height=0.22\linewidth]{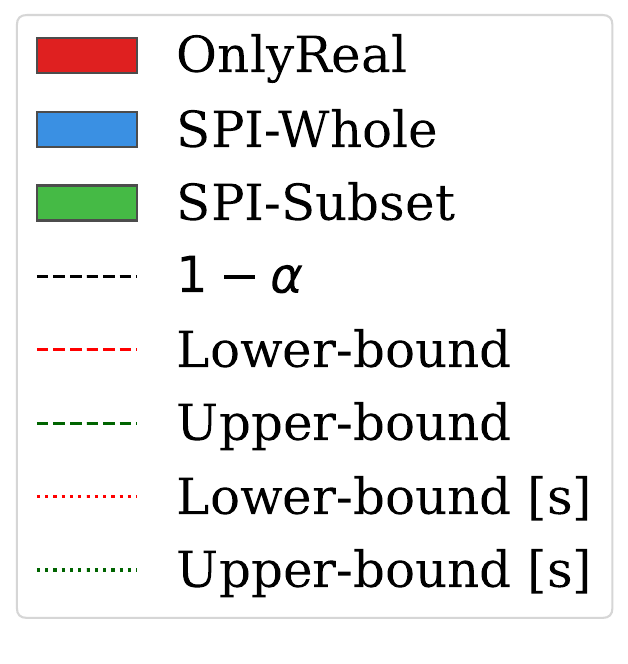}
    \caption{$\alpha=0.1$}
    \end{subfigure}
    \caption{Results for the ImageNet data: Coverage rate for $\cp$,$\scph$, and $\scpc$ on the American robin class as a function of the number of subsets $k$, for levels $\alpha=0.02$ (a), $\alpha=0.05$ (b), and $\alpha=0.1$ (c).}
    \label{app-fig:imagenet-clusters-ar-k}
\end{figure}

\Cref{app-fig:imagenet-clusters-ar-k} presents the performance of all methods for the ``American robin" class as a function of $k$, at different values of the level $\alpha$. Notably, for all values of $k$ and $\alpha$, $\scpc$ achieves coverage within the theoretical bounds.

The two methods, $\scpc$ and $\scph$, coincide when $k = 100$, as both use the full synthetic calibration set. However, for smaller values of $k$, the two methods exhibit significant differences. While $\scph$ tends to produce more conservative prediction sets, the $\scpc$ procedure more tightly achieves the target coverage level across different settings.

For the case $\alpha = 0.02$ and $k = 5$, both the theoretical lower and upper bounds on coverage are equal to unity, implying that $\scpc$ yields trivial prediction sets that include all possible classes.
This outcome is known a priori and can be avoided by selecting a different hyperparameter for window construction.

\begin{figure}[!h]
    \centering
    \begin{subfigure}[t]{\linewidth}
    \includegraphics[height=0.22\textwidth]{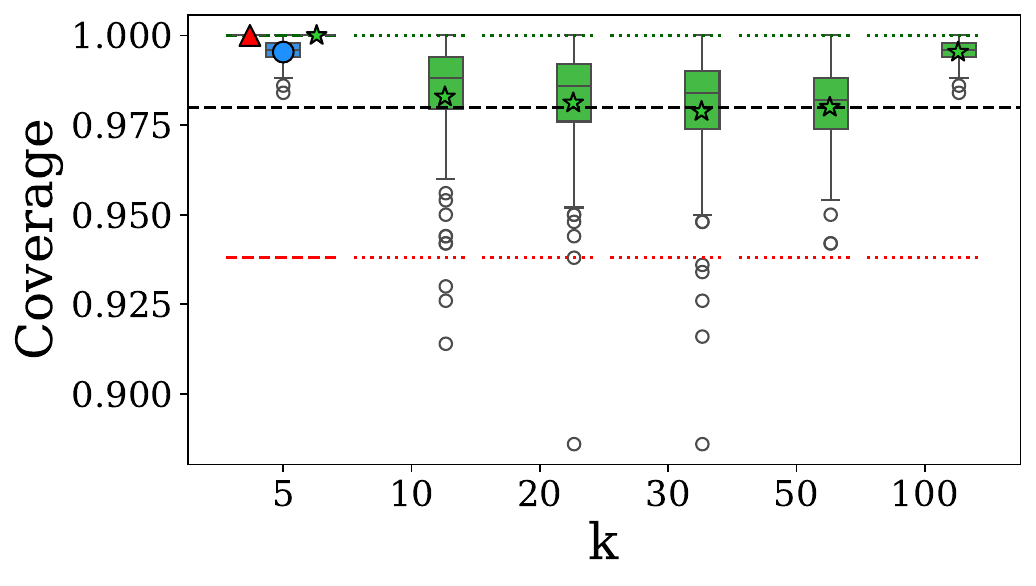}
    \includegraphics[height=0.22\textwidth]{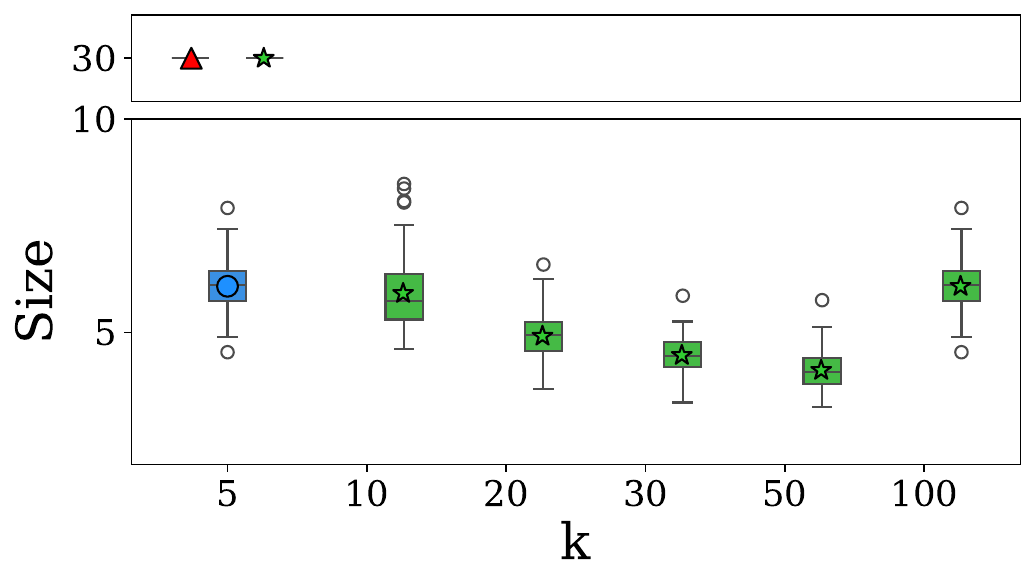}
    \caption{$\alpha=0.02$}
    \label{app-fig:imagenet-clusters-b-k-0.02}
    \end{subfigure}
    \begin{subfigure}[t]{\linewidth}
    \includegraphics[height=0.22\textwidth]{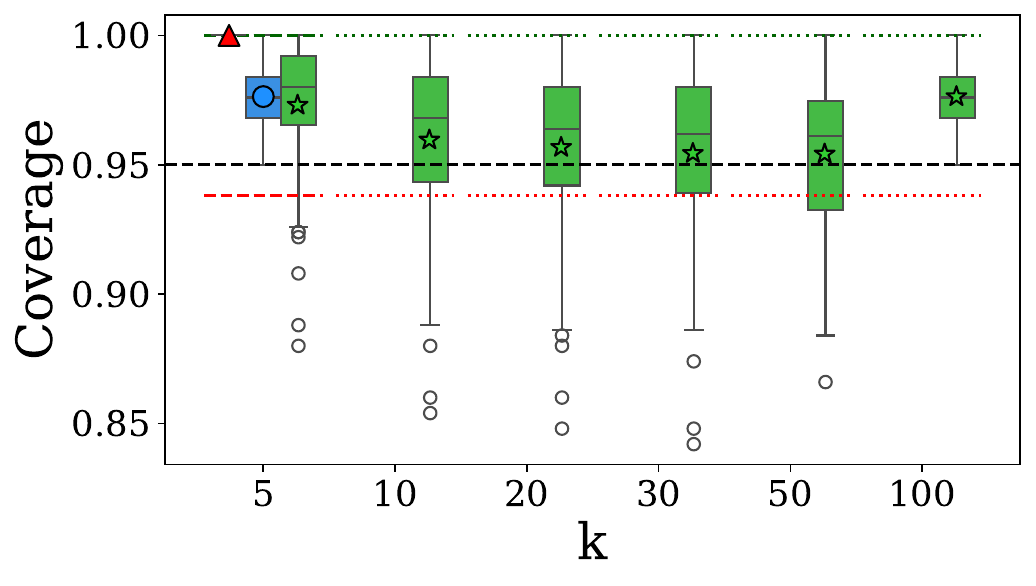}
    \includegraphics[height=0.22\textwidth]{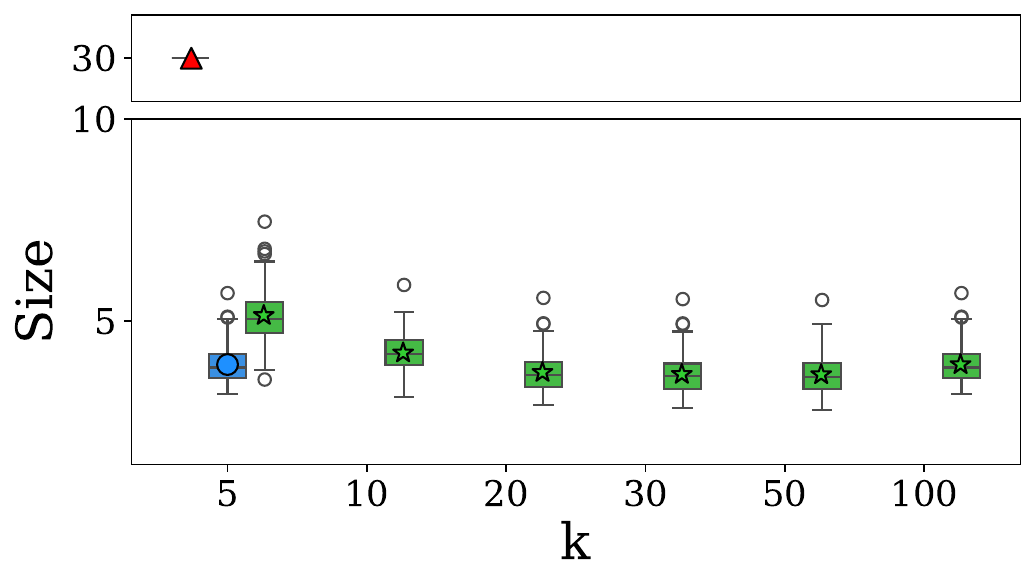}
    \caption{$\alpha=0.05$}
    \label{app-fig:imagenet-clusters-b-k-0.05}
    \end{subfigure}
    \begin{subfigure}[t]{\linewidth}
    \includegraphics[height=0.22\textwidth]{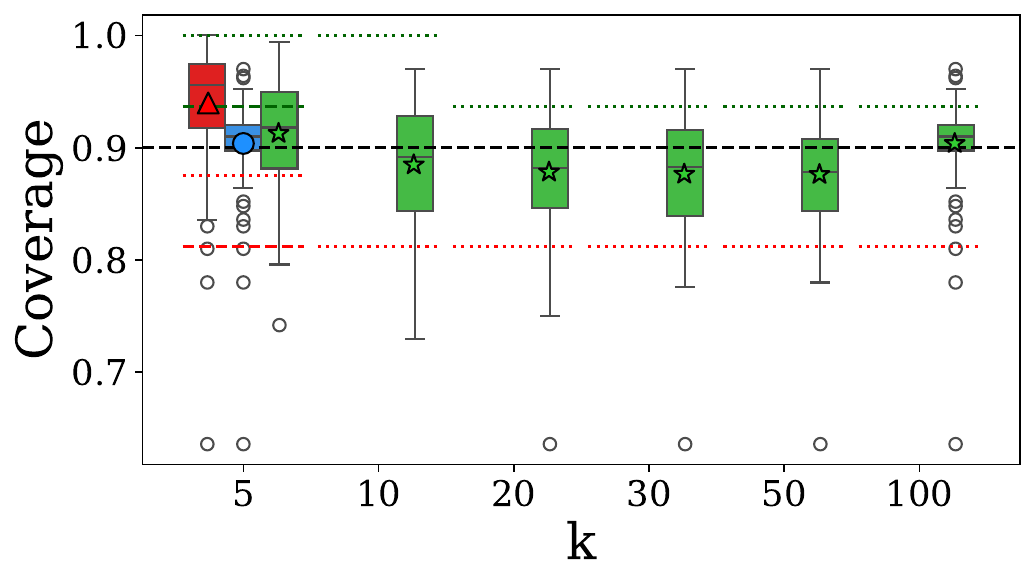}
    \includegraphics[height=0.22\textwidth]{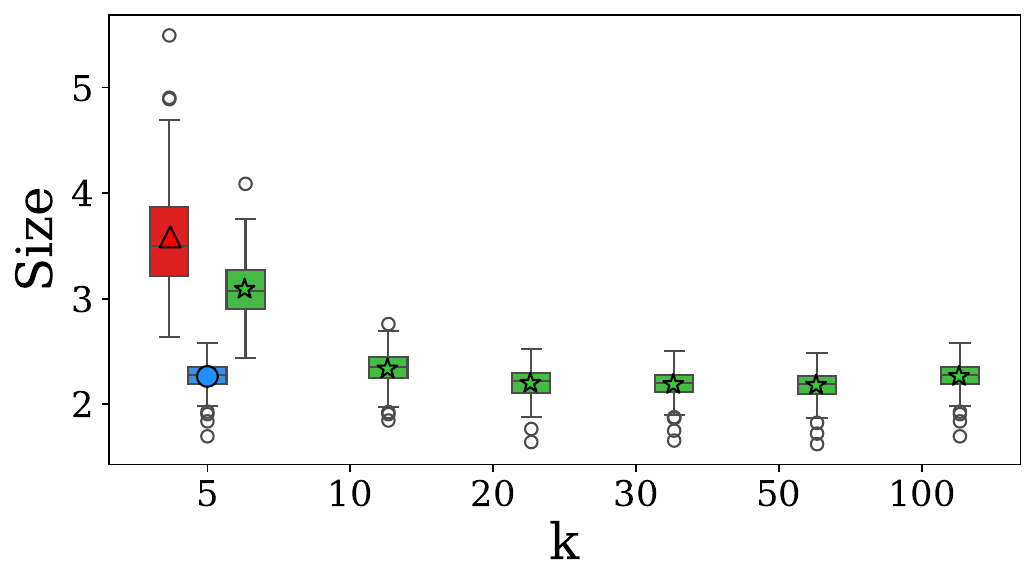}
    \includegraphics[height=0.22\linewidth]{figures/results/legend_clusters_spi.pdf}
    \caption{$\alpha=0.1$}
    \label{app-fig:imagenet-clusters-b-k-0.1}
    \end{subfigure}
    \caption{Results for the ImageNet data: Coverage rate for $\cp$, $\scph$, and $\scpc$ on the Beaver class as a function of the number of subsets $k$, for levels $\alpha=0.02$ (a), $\alpha=0.05$ (b), and $\alpha=0.1$ (c).}
    \label{app-fig:imagenet-clusters-b-k}
\end{figure}

\Cref{app-fig:imagenet-clusters-b-k} shows the results for the ``Beaver” class. For $\alpha = 0.02$ and $0.05$, we observe the same trend as in \Cref{app-fig:imagenet-clusters-ar-k}: $\scph$ yields relatively conservative coverage, while $\scpc$ with $k < 100$ achieves coverage closer to the nominal level $1 - \alpha$.

For $\alpha = 0.1$, $\scph$—which uses the full synthetic set—already achieves coverage close to the target level $1 - \alpha$, suggesting that the empirical $(1 - \alpha)$th quantile of the synthetic data closely matches that of the real data. Consequently, in this setting, using only a subset of the synthetic data results in an increase in the variance of the coverage rate.

\clearpage
\FloatBarrier
\section{Additional MEPS experiments}\label{app-sec:exp-meps}
In this section, we present additional results for the MEPS regression experiments, complementing those reported in \Cref{sec:exp-meps}. 

\begin{figure}[!h]
    \centering
    \begin{subfigure}[t]{\linewidth}
    \includegraphics[height=0.22\textwidth, valign=t]{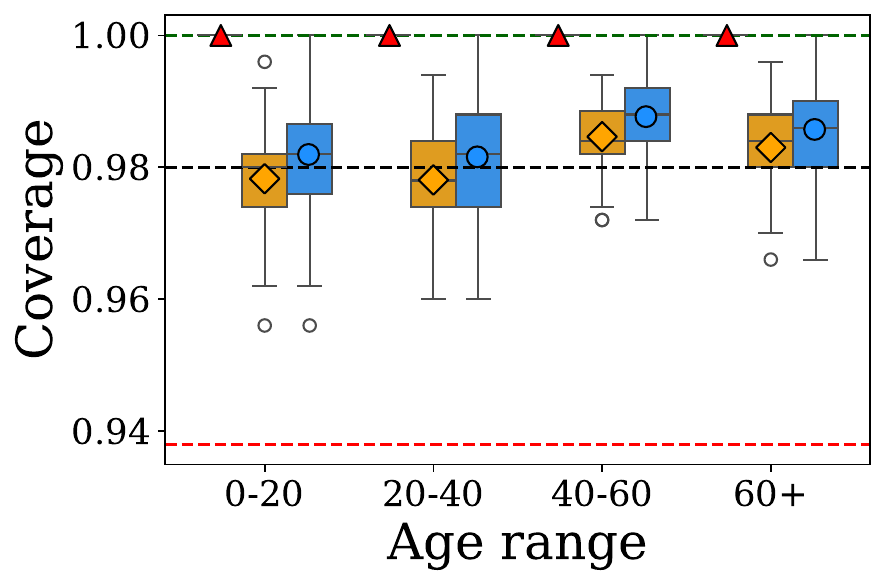}
    \includegraphics[height=0.22\textwidth, valign=t]{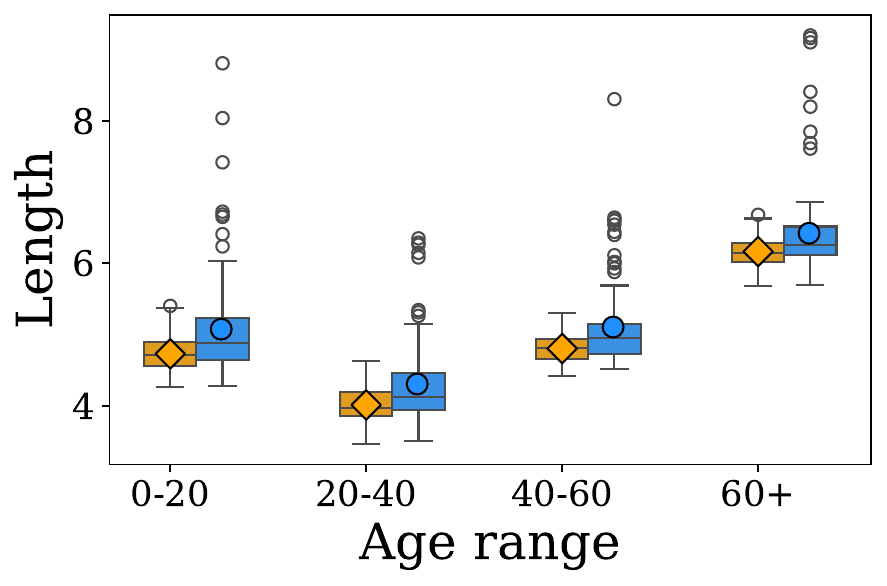}
    \caption{$\alpha=0.02$}
    \end{subfigure}
    \begin{subfigure}[t]{\linewidth}
    \includegraphics[height=0.22\textwidth, valign=t]{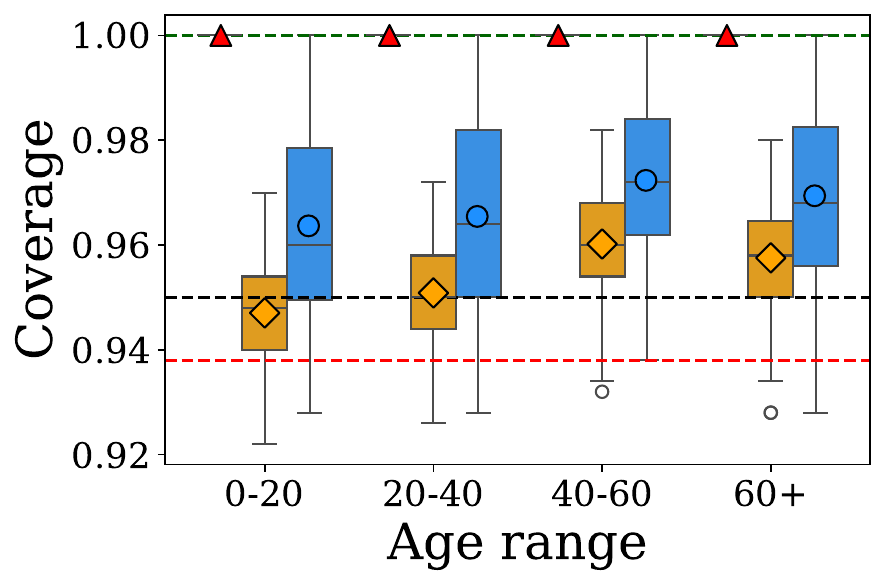}
    \includegraphics[height=0.22\textwidth, valign=t]{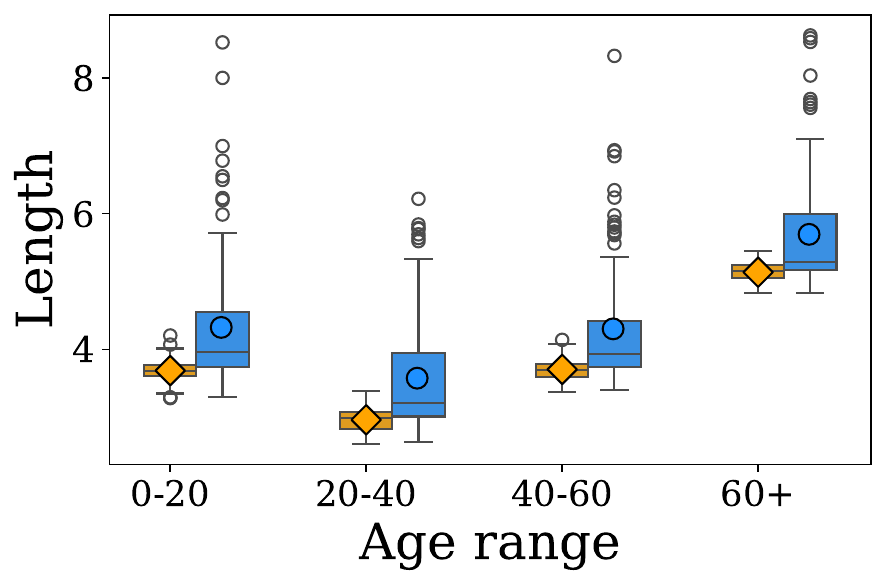}
    \includegraphics[height=0.22\textwidth, valign=t]{figures/results/legend_spi.pdf}
    \caption{$\alpha=0.05$}
    \end{subfigure}
    \caption{MEPS dataset results: coverage and interval length for each age group, obtained by $\cp$, $\cpsynt$, and $\scp$, at target coverage levels $1 - \alpha = 0.98$ (a), and $0.95$ (b). Experiments are repeated over 100 trials. $\cp$ produces trivial (infinite) prediction intervals; thus, its interval length is omitted.}
    \label{app-fig:meps}
\end{figure}

\Cref{app-fig:meps} reports the coverage rates and prediction interval lengths for all age groups, evaluated at $\alpha = 0.02$, and $0.05$. As in the main paper, we observe that $\scp$ achieves coverage rates that remain within the theoretical bounds, with lower variance compared to $\cp$. 

\subsection{The effect of the real calibration set size}
We replicate the experiments from~\Cref{app-sec:imagenet-gen-calib-size} on the MEPS dataset, evaluating the performance of different methods as a function of the real calibration set size, $m$.

\begin{figure}[!h]
    \centering
    \begin{subfigure}[t]{\linewidth}
    \includegraphics[height=0.22\textwidth, valign=t]{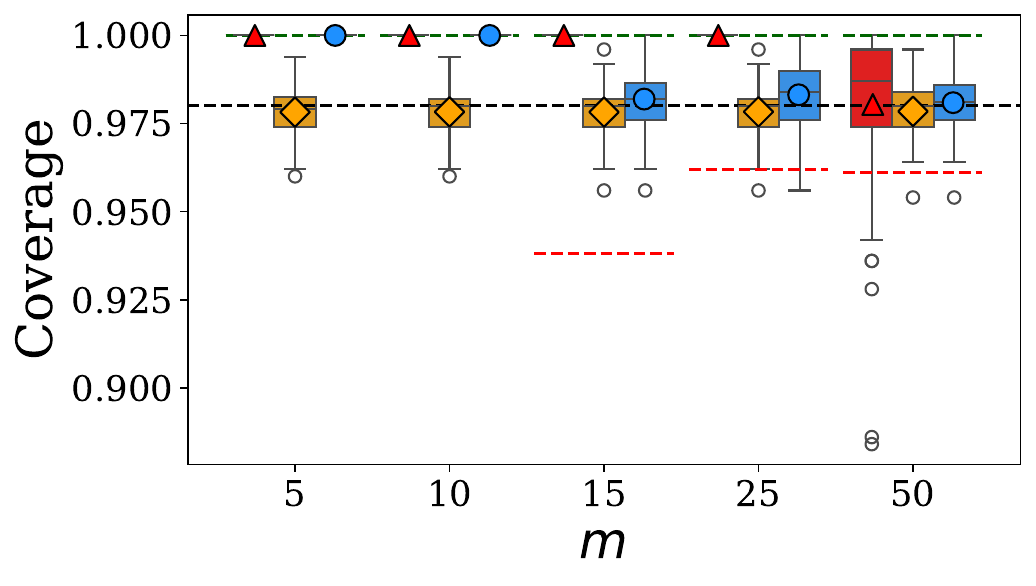}
    \includegraphics[height=0.22\textwidth, valign=t]{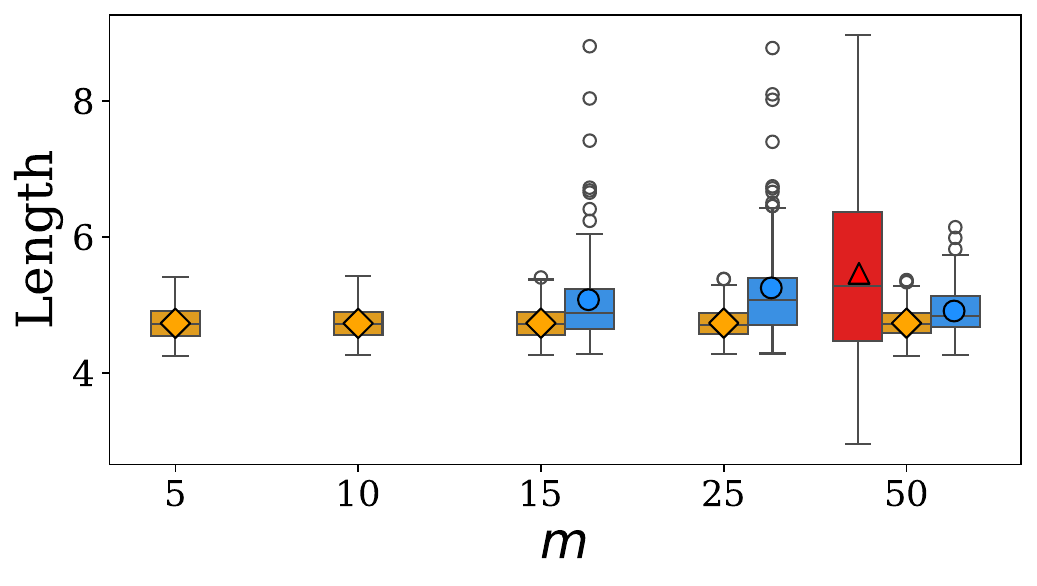}
    \caption{$\alpha=0.02$}
    \end{subfigure}
    \begin{subfigure}[t]{\linewidth}
    \includegraphics[height=0.22\textwidth, valign=t]{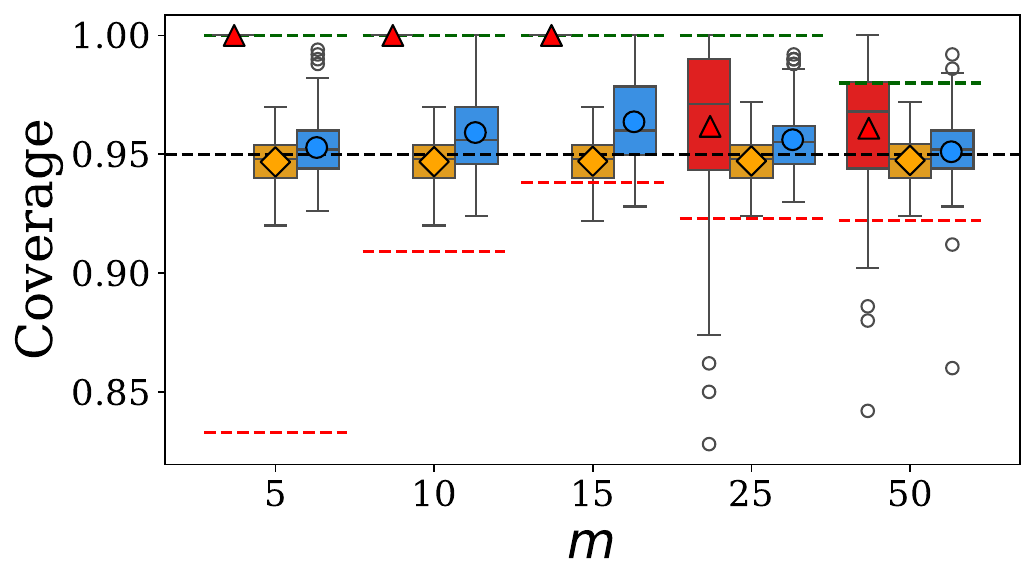}
    \includegraphics[height=0.22\textwidth, valign=t]{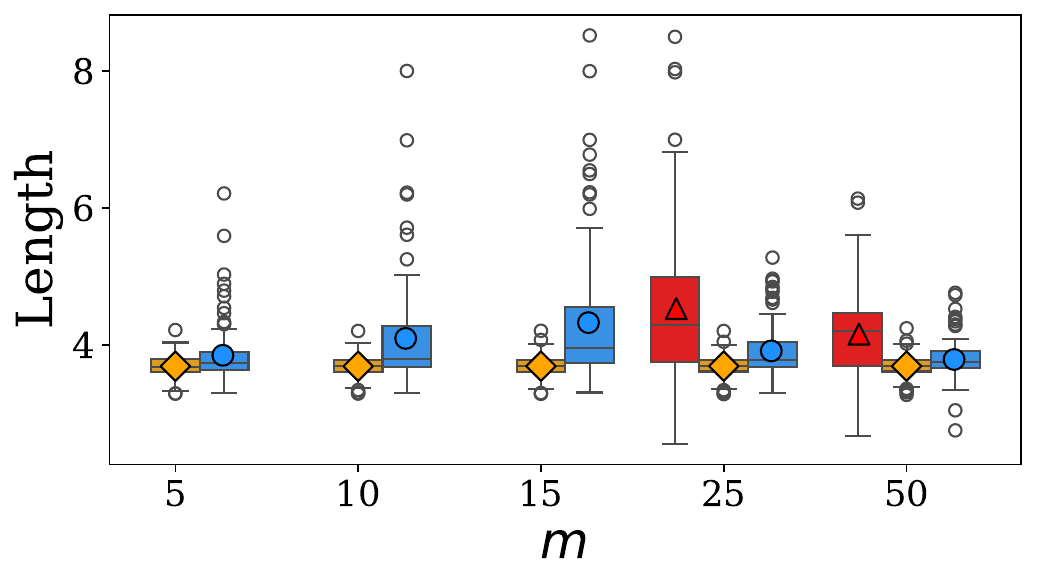}
    \caption{$\alpha=0.05$}
    \end{subfigure}
    \begin{subfigure}[t]{\linewidth}
    \includegraphics[height=0.22\textwidth, valign=t]{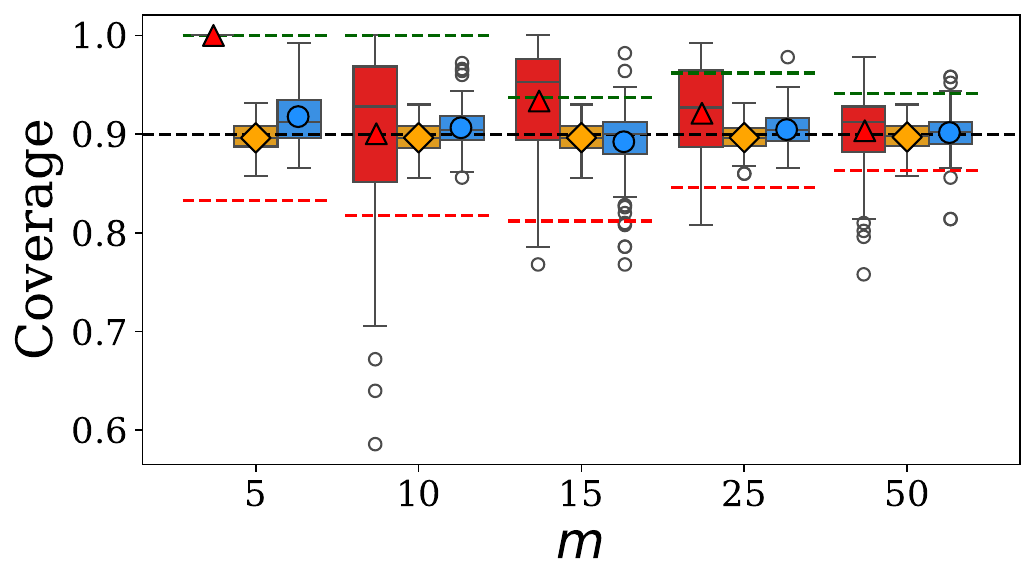}
    \includegraphics[height=0.22\textwidth, valign=t]{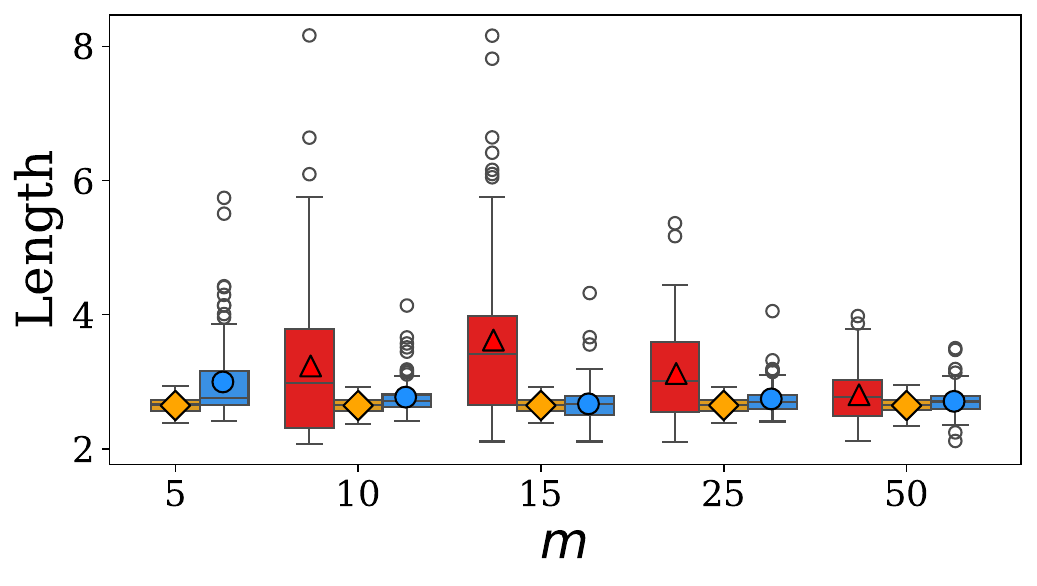}
    \includegraphics[height=0.22\textwidth, valign=t]{figures/results/legend_spi.pdf}
    \caption{$\alpha=0.1$}
    \end{subfigure}
    \caption{MEPS dataset results: coverage and interval length for the 0--20 age group, obtained by $\cp$, $\cpsynt$, and $\scp$, at target coverage levels $1 - \alpha = 0.98$ (a), $0.95$ (b), and $0.9$ (c). Experiments are repeated over 100 trials. For $\alpha=0.02$ and $0.05$, methods that produce trivial (infinite) prediction intervals are omitted from the interval length panel.}
    \label{app-fig:meps-n-cal-a-0-20}
\end{figure}

\begin{figure}[!h]
    \centering
    \begin{subfigure}[t]{\linewidth}
    \includegraphics[height=0.22\textwidth, valign=t]{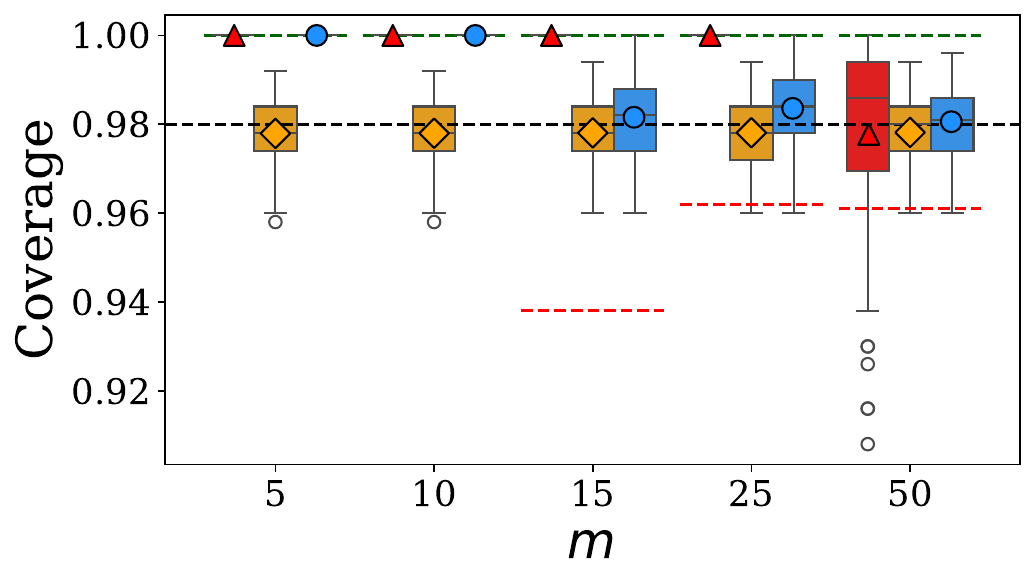}
    \includegraphics[height=0.22\textwidth, valign=t]{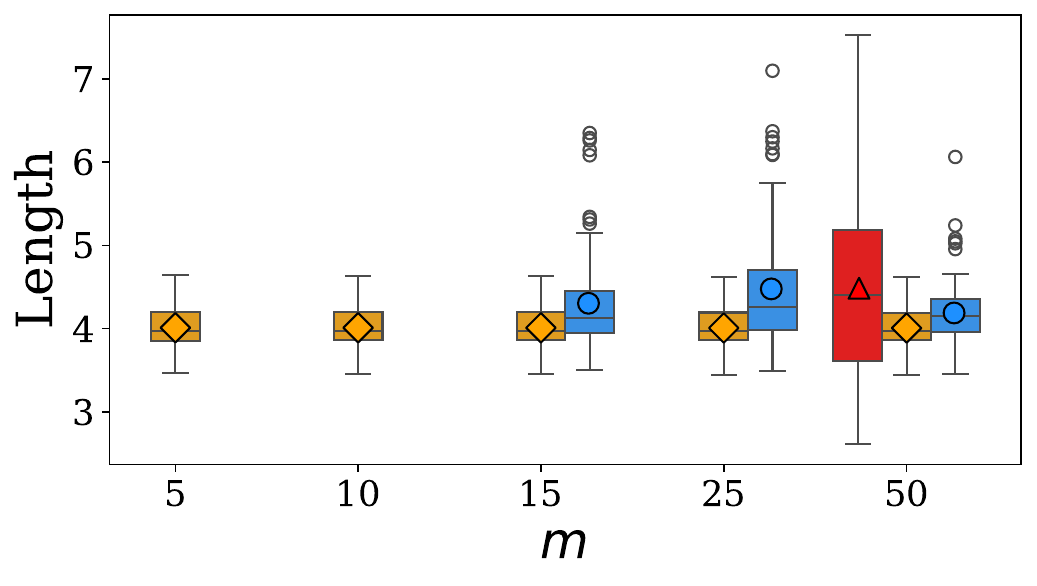}
    \caption{$\alpha=0.02$}
    \end{subfigure}
    \begin{subfigure}[t]{\linewidth}
    \includegraphics[height=0.22\textwidth, valign=t]{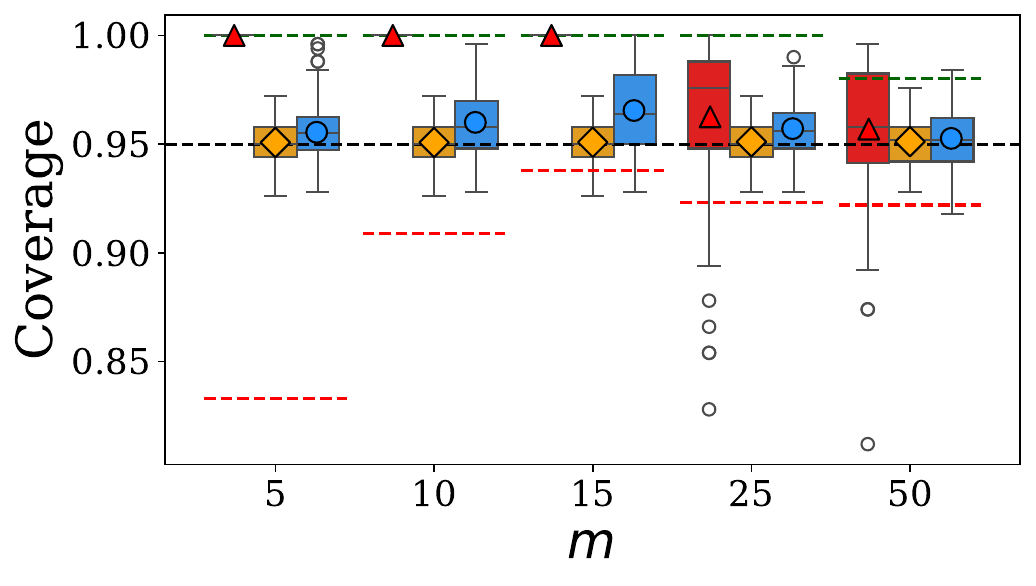}
    \includegraphics[height=0.22\textwidth, valign=t]{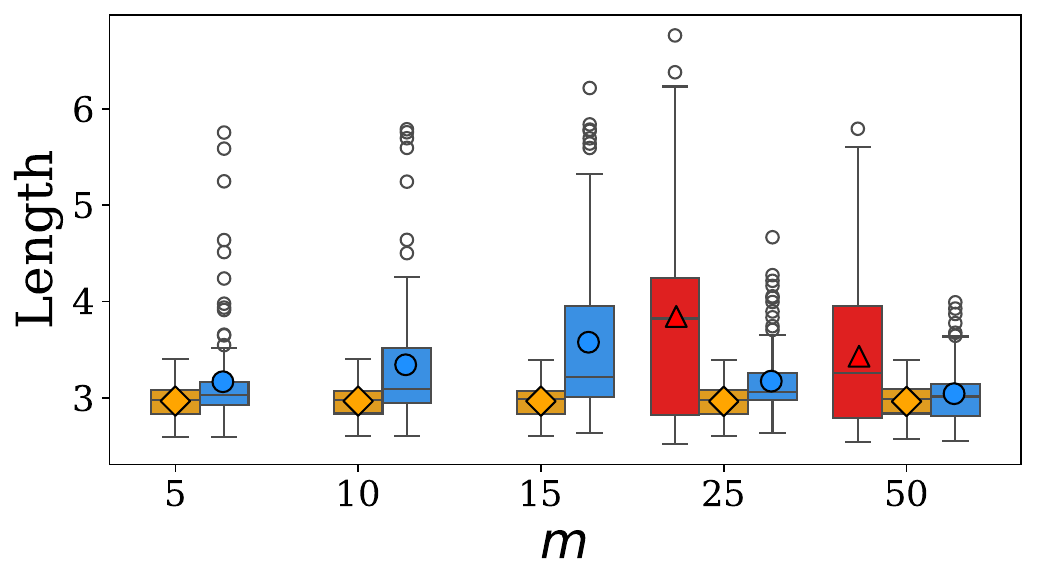}
    \caption{$\alpha=0.05$}
    \end{subfigure}
    \begin{subfigure}[t]{\linewidth}
    \includegraphics[height=0.22\textwidth, valign=t]{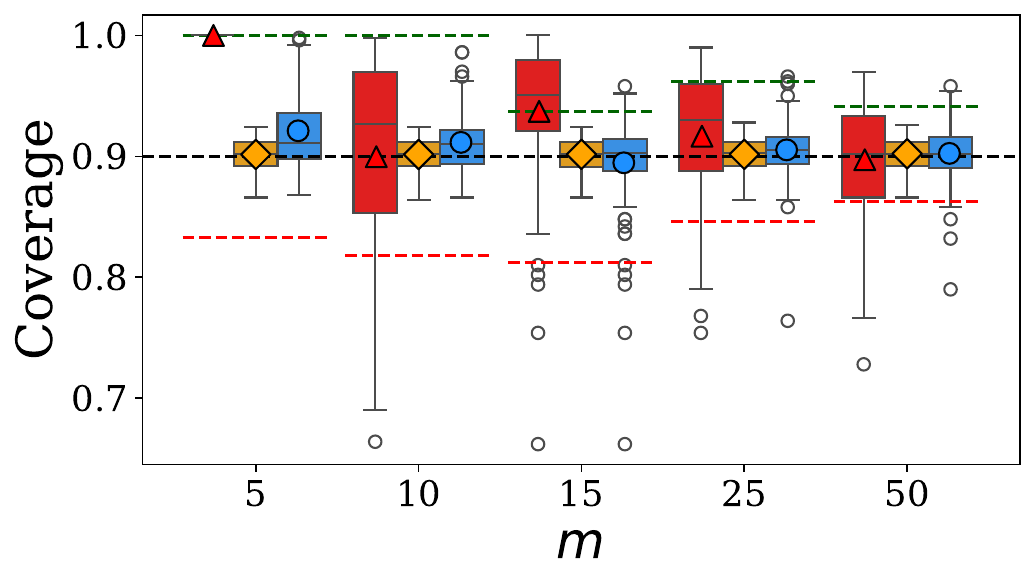}
    \includegraphics[height=0.22\textwidth, valign=t]{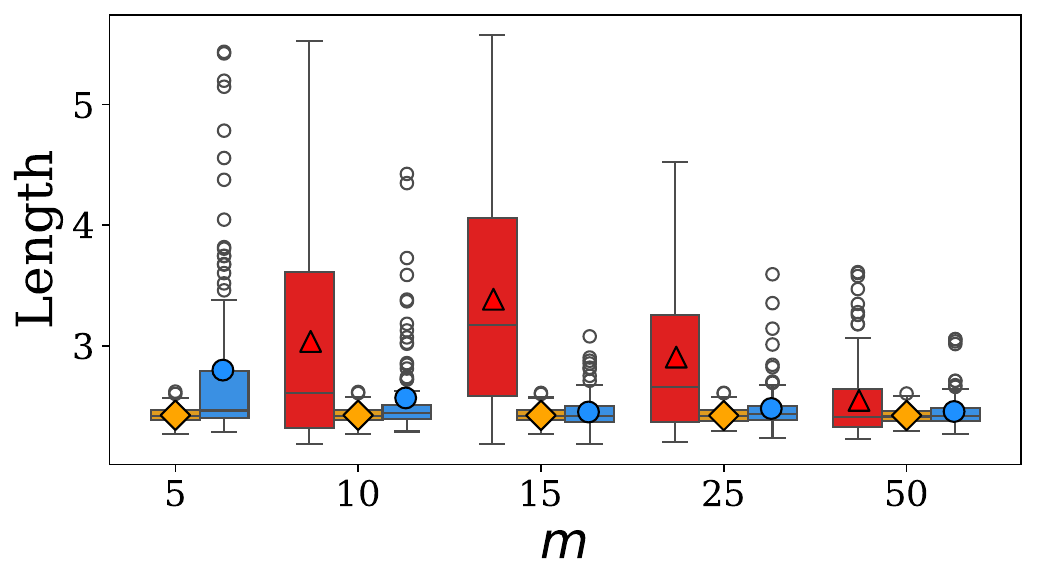}
    \includegraphics[height=0.22\textwidth, valign=t]{figures/results/legend_spi.pdf}
    \caption{$\alpha=0.1$}
    \end{subfigure}
    \caption{MEPS dataset results: coverage and interval length for the 20--40 age group, obtained by $\cp$, $\cpsynt$, and $\scp$, at target coverage levels $1 - \alpha = 0.98$ (a), $0.95$ (b), and $0.9$ (c). Experiments are repeated over 100 trials. For $\alpha=0.02$ and $0.05$, methods that produce trivial (infinite) prediction intervals are omitted from the interval length panel.}
    \label{app-fig:meps-n-cal-a-20-40}
\end{figure}

\Cref{app-fig:meps-n-cal-a-0-20} and \Cref{app-fig:meps-n-cal-a-20-40} present the performance of all methods for age groups 0--20 and 20--40, respectively, across different $\alpha$ levels and values of $m$. The standard conformal method, $\cp$, conservatively controls the coverage at the target level $1-\alpha$; however, it results in larger and noisier prediction intervals due to the small sample size.

Similar to the trends observed in the main manuscript, $\cpsynt$ achieves coverage close to the nominal $1-\alpha$ level, indicating that the synthetic data align well with the real one. However, this approach does not have coverage guarantees. 

In contrast, the proposed method, $\scp$, achieves coverage within the theoretical bounds, closely matching the target $1-\alpha$ level while reducing coverage variance and producing smaller, more informative prediction intervals.

For $\alpha=0.02$ with small calibration sizes ($m=5$ or $10$), the theoretical coverage bounds are equal to one under this window construction. This implies that the proposed method produces trivial prediction intervals. This behavior is known a priori and was also observed in the ImageNet experiment, where we used the same window construction parameters. Nevertheless, it can be avoided by employing a different window construction.

\FloatBarrier

\end{document}